%% file: arslan.tsiotras.arxiv.final.tex
\documentclass[11pt]{article}
\usepackage[margin=1in]{geometry}

\usepackage{tikz}
\usetikzlibrary{fit,calc}
\usepackage{relsize}
\newcommand{\tikzmark}[1]{\tikz[overlay,remember picture,baseline] \node [anchor=base] (#1) {};}

\usepackage[colorlinks]{hyperref}
\hypersetup{
   bookmarksnumbered,
   pdfstartview={FitH},
   citecolor={black},
   linkcolor={black},
   urlcolor={black},
   pdfpagemode={UseOutlines}
}

\usepackage[colorlinks]{hyperref}

\hypersetup{
   bookmarksnumbered,
   pdfstartview={FitH},
   citecolor={black},
   linkcolor={black},
   urlcolor={black},
   pdfpagemode={UseOutlines}
}

\usepackage{dsfont}
\usepackage{placeins}
\usepackage{color}
\usepackage{url}
\usepackage{exscale}
\usepackage{subfigure}
\usepackage{caption}
\usepackage{array}
\usepackage{epsfig}
\usepackage{xspace}
\usepackage{bm}
\usepackage{amsmath}
\usepackage{amssymb}
\usepackage{stmaryrd}
\usepackage{comment}
\usepackage{fancybox}
\usepackage{multirow}
\usepackage[vlined,linesnumbered,boxruled]{algorithm2e}
\usepackage{relsize}
\usepackage{rotating}

\definecolor{gray}{RGB}{190,190,190}
\definecolor{darkgreen}{rgb}{0,0.5,0} 
\definecolor{fullred}{rgb}{0.85,.0,.1} 
\definecolor{brown}{rgb}{0.65,0.16,0.16}

\newcommand{\AlgRRT}{\ensuremath{{\mathrm{RRT}}}}
\newcommand{\AlgPRM}{\ensuremath{{\mathrm{PRM}}}}
\newcommand{\AlgRRG}{\ensuremath{{\mathrm{RRG}}}}
\newcommand{\AlgRRTstar}{\ensuremath{{\mathrm{RRT}^*}}}
\newcommand{\AlgRRTsharp}{\ensuremath{{\mathrm{RRT}^\#}}}
\newcommand{\AlgRRTsharpNoBlackVertex}{\ensuremath{{\mathrm{RRT}^\#_{V1}}}}
\newcommand{\AlgRRTsharpPromisingParent}{\ensuremath{{\mathrm{RRT}^\#_{V2}}}}
\newcommand{\AlgRRTsharpPromisingNewVertex}{\ensuremath{{\mathrm{RRT}^\#_{V3}}}}

\newcommand{\reals}{\mathbb{R}}  
\newcommand{\naturals}{\mathbb{N}} 	
\newcommand{\X}{\mathcal{X}} 		
\newcommand{\Xfree}{\X_{\mathrm{free}}} 
\newcommand{\Xgoal}{\X_{\mathrm{goal}}} 
\newcommand{\Xobs}{\X_{\mathrm{obs}}} 	
\newcommand{\Xrel}{\X_{\mathrm{rel}}} 	
\newcommand{\xrand}{x_{\mathrm{rand}}}	
\newcommand{\xnew}{x_{\mathrm{new}}}	
\newcommand{\xinit}{x_{\mathrm{init}}}	
\newcommand{\xnearest}{x_{\mathrm{nearest}}}	
\newcommand{\xmingoal}{x^{*}_{\mathrm{goal}}}	

\newcommand{\tree}[1]{\mathcal{#1}} 		
\newcommand{\graph}[1]{\mathcal{#1}}		

\newcommand{\Vprom}{V_{\mathrm{prom}}}	
\newcommand{\vmingoal}{v^{*}_{\mathrm{goal}}}	

\newcommand{\True}{{\tt True}}
\newcommand{\False}{{\tt False}}

\newcommand{\lmcValue}{\mathtt{lmc}}		
\newcommand{\gValue}{\mathtt{g}}			
\newcommand{\gStarValue}{\mathtt{g^{*}}}	
\newcommand{\hValue}{\mathtt{h}}			
\newcommand{\costValue}{\mathtt{c}}			

\newcommand{\PrcSample}{\mathtt{Sample}} 
\newcommand{\PrcExtend}{\mathtt{Extend}} 
\newcommand{\PrcNearest}{\mathtt{Nearest}} 
\newcommand{\PrcNear}{\mathtt{Near}}      
\newcommand{\PrcSteer}{\mathtt{Steer}}   
\newcommand{\PrcObstacleFree}{\mathtt{ObstacleFree}} 
\newcommand{\PrcReduceInconsistency}{\mathtt{ReduceInconsistency}} 
\newcommand{\PrcUpdateState}{\mathtt{UpdateQueue}} 

\newcommand{\PrcSuccessor}{\mathtt{succ}} 
\newcommand{\PrcPredecessor}{\mathtt{pred}} 
\newcommand{\PrcParent}{\mathtt{parent}} 
\newcommand{\PrcKey}{\mathtt{Key}} 

\newcommand{\hideMaterial}[1]{}
\newcommand{\hideoldproofofnegresult}[1]{ }

\title{The Role of Vertex Consistency in Sampling-based Algorithms\\ for Optimal Motion Planning}

\date{}

\author{Oktay Arslan\thanks{Oktay Arslan is a graduate student with the D. Guggenheim School of Aerospace Engineering, Georgia Institute of Technology, Atlanta, GA 30332-0150, USA, Email: {oktay@gatech.edu}} \and Panagiotis Tsiotras\thanks{Professor Panagiotis Tsiotras is with the faculty of D. Guggenheim School of Aerospace Engineering, Georgia Institute of Technology, Atlanta, GA 30332-0150, USA, Email: {tsiotras@gatech.edu}}
}

\begin{document}

\maketitle

\begin{abstract}
Motion planning problems have been studied  by both the robotics and the controls research communities for a long time, and many algorithms have been developed for their solution. Among them, incremental sampling-based motion planning algorithms, such as the Rapidly-exploring Random Trees (\AlgRRT), and the Probabilistic Road Maps (\AlgPRM) have become very popular recently, owing to their implementation simplicity and their advantages in handling  high-dimensional problems. Although these algorithms work very well in practice, the quality of the computed solution is often not good, i.e., the solution can be far from the optimal one. A recent variation of \AlgRRT, namely the \AlgRRTstar{} algorithm, bypasses this drawback of the traditional \AlgRRT{} algorithm, by ensuring asymptotic optimality as the number of samples tends to infinity. Nonetheless, the convergence rate to the optimal solution may still be slow. This paper presents a new incremental sampling-based motion planning algorithm based on Rapidly-exploring Random Graphs (\AlgRRG), denoted \AlgRRTsharp{} (RRT ``sharp'') which also guarantees asymptotic optimality but, in addition, it also ensures that the constructed spanning tree of the geometric graph is consistent after each iteration. In consistent trees, the vertices which have the potential to be part of the optimal solution have the minimum cost-come-value. This implies that the best possible solution is readily computed if there are some vertices in the current graph that are already in the goal region. Numerical results compare with the \AlgRRTstar{} algorithm.
\end{abstract}

{\footnotesize {\bf Keywords}: optimal motion planning, \AlgRRT, \AlgRRG, \AlgRRTstar, \AlgRRTsharp, vertex consistency, consistent tree.}

\section{Introduction} \label{section:introduction}

Motion planning problems are crucial for the realization of truly autonomous vehicles and robots. Many approaches have been proposed in the literature (see for example, the excellent books by LaValle~\cite{LaValle2006} and Choset et al~\cite{Choset2005}). A bottleneck in most motion planning problems, especially those involving systems with high state dimensionality, is the computational overhead associated with discretizing (i.e., gridding) the state space. Hence, deterministic searches~\cite{Howard2007} are impractical for high dimensional state spaces. Probabilistic roadmap methods~\cite{Donald1993,Kavraki1993,Svestka1993,Kavraki1996}, \cite[Ch. 7]{Choset2005}, as well as methods that use rapidly exploring random trees (RRTs)~\cite{LaValle2001,lavalle2001randomized,Hsu2002,Frazzoli2002,Plaku2010}, are among the most popular. They can address the vehicle's kinematic and dynamic constraints during motion planning in high dimensional state spaces. In these methods, random samples of the obstacle-free space are connected to each other by feasible trajectories, and the resulting graph is searched for a sequence of connected samples from the initial state to the goal state. Sampling-based algorithms require efficient low-level collision detection and trajectory planning algorithms to find collision-free trajectories between different samples~\cite{lavalle2001randomized}.

Incremental sampling-based algorithms were first proposed by  Kavraki during the late 1990s. The so-called Probabilistic Road Map (\AlgPRM) was successfully implemented to solve multi-query motion planning problems and gained a lot of attention, both in industry and academia~\cite{kavraki1996probabilistic}. In \AlgPRM{} a graph of the environment is constructed by taking random samples from the configuration space of the robot and testing them to determine whether they belong to the free space. The \AlgPRM{} algorithm uses a local planner that attempts to find a feasible path between the sampled points. Once a reasonable graph is constructed, the initial and the goal states are added to the graph, and the optimal path is computed using a graph search algorithm.

Another important class of incremental sampled-based motion planning algorithm is the Rapidly-exploring Random Tree (\AlgRRT) and its numerous variants~\cite{lavalle2001randomized}. RRTs have achieved great success in solving single-query motion planning problems in many real-time applications. However, the quality of RRT-based algorithms is often poor (i.e., highly suboptimal). As a result, a lot of effort has been devoted to the development of heuristic techniques in order to refine the quality of the solution obtained from RRTs. However, it has been recently shown that the best path returned by RRTs when the algorithm converges is almost always (i.e., with probability one) far from optimal~\cite{KaramanFra2011}. This has renewed the interest to develop incremental sampled-based algorithms for motion-planning problems with optimality guarantees. In \cite{karaman2009sampling} the authors proposed the Rapidly-exploring Random Graphs (\AlgRRG) algorithm, which has asymptotic optimality properties, that is, it ensures that the optimal path will be found as the number of samples tends to infinity. Based on \AlgRRG, the  same authors later proposed a new algorithm, namely \AlgRRTstar{} that extracts a tree from the graph constructed by \AlgRRG~\cite{karaman2010incremental,KaramanFra2011}.

In this paper we present a new incremental sampling-based motion planning algorithm based on \AlgRRG, denoted \AlgRRTsharp (RRT ``sharp''), which also guarantees asymptotic optimality but, in addition, it also ensures that, at each step, the constructed spanning tree of the graph is consistent. Vertex consistency (see Section~\ref{section:formulation}) implies that the accumulated cost-to-come of each vertex equals to the optimal cost-to-come. This allows us classify the vertices according to their potential of being part of the optimal path, and thus to quickly identify the region where the optimal solution is more likely to be found. This information can be subsequently used to improve the  speed of convergence of the standard \AlgRRTstar{} algorithm, as well as in order to more efficiently explore the obstacle-free space. Three variants of the baseline \AlgRRTsharp{} algorithm are proposed that take advantage of this vertex classification to speed up convergence.

The organization of the paper is as follows: The problem formulation is given in the next section. In Section 3, an overview of the \AlgRRTsharp{} algorithm is introduced. The fundamental concepts and primitive functions used in the \AlgRRTsharp{} algorithm are explained. In Section 4, each step of the proposed approach is explained in detail, along with the pseudo-code of the algorithm and the main procedures used in the main algorithm. In Sections 5, simulation results are used to compare the solutions of the proposed approach with the well-known \AlgRRTstar{} algorithm. In Section 6, several variants of the baseline algorithm are presented by using simple vertex rejection techniques and improvements are demonstrated by doing extensive simulations in the subsequent section. We conclude  the paper with some possible extensions for future work.

\section{Problem Formulation} \label{section:formulation}

\subsection{Notation and Definitions}

Let $\X$ denote the state space, which is assumed to be an open subset of $\reals^{{d}}$, where $d \in \naturals$ with $d \geq 2$. Let the \emph{obstacle region} and the \emph{goal region} be denoted by $\Xobs$ and $\Xgoal$, respectively. The obstacle-free space is defined by $\Xfree = \X \setminus \Xobs$. Let the \emph{initial state} be denoted by $\xinit \in \Xfree$. The neighborhood of a state $x \in \X$ is defined as the open ball of radius $r \in \reals_{+}$ centered at $x$, that is, $B_{r}(x) = \{ x^{\prime} \in \X : \|x - x^{\prime} \| < r \}$.
Let $\graph{G} = (V, E)$ denote a graph, where $V$ and $E \subseteq V \times V$ are finite sets of vertices and edges, respectively. In the sequel, we will use graphs to represent the connections between a (finite) set of points selected randomly from $\Xfree$. With a slight abuse of notation, we will use $x$ to denote both the point in the space $\X$ and the corresponding vertex in the graph.

\emph{Geometric r-disc graph}: Let $V \subset \reals^{{d}}$ be a finite set, and $r \ge 0$. A geometric $r$-disc graph $\graph{G}(V;r) = (V, E)$ in $d$ dimensions is an undirected graph with vertex set $V$ and edge set $E = \{(u, v): u,v \in \mathcal{V} \text{ and } \|u - v \| < r\}$.

\emph{Successor vertices}:
Given a vertex $v \in V$, the set-valued function $\PrcSuccessor : (\graph{G}, v) \mapsto V^{\prime} \subseteq V$ returns the vertices in $V$ that can be reached from vertex $v$,

$$\PrcSuccessor(\graph{G}, v) := \left\{ u \in V: (v,u) \in E \right\}$$

\emph{Predecessor vertices}:
Given a vertex $v \in V$ in a directed graph $\graph{G}=(V,E)$, the function $\PrcPredecessor : (\graph{G}, v) \mapsto V^{\prime} \subseteq V$ returns the vertices in $V$ that are the tails of the edges going into $v$,

$$\PrcPredecessor(\graph{G}, v) := \left\{ u \in V: (u, v) \in E \right\}$$

\emph{Parent vertex}: Given a vertex $v \in V$, the function $\PrcParent : v \mapsto u$ returns the unique vertex $u \in V$ such that $(u,v) \in E$ and $u \in \PrcPredecessor(\graph{G}, v)$.

\emph{Spanning tree}: Given the graph $\mathcal{G} = (V,E)$, a spanning tree of $\mathcal{G}$ can be defined such that $\mathcal{T} = (V_{s}, E_{s})$, where $V_{s} = V$ and $E_{s}=\{(u,v): u,v \in V, (u,v) \in E \mathrm{~and~} \PrcParent(v) = u\}$.

\emph{Edge cost value}: Given an edge $e = (u,v) \in E$, the function $\costValue : e \mapsto r$ returns a non-negative real number. Then $\costValue(u,v)$ where $v \in \PrcSuccessor(\graph{G},u)$ is the cost incurred by moving from $u$ to $v$.
\emph{Cost-to-come value}: Given a vertex $v \in V$, the function $\gValue : v \mapsto r$ returns a non-negative real number $r$, which is the cost of the path to $v$ from a given initial state $\xinit \in \Xfree$. Let $\gStarValue(v)$ be the optimal cost-to-come value of the vertex $v$. The optimal cost-to-come satisfies the following relationship:

$$
\gStarValue(v) =
\begin{cases}
0, & \text{if } v = \xinit, \\
\min_{u \in \PrcPredecessor(\graph{G}, v)}( \gStarValue(u) + \costValue(u,v)), & \text{otherwise}.
\end{cases}
$$

Each vertex $v$ is associated with two estimates of the optimal cost-to-come value $\gStarValue(v)$, namely, $\gValue(v)$ (g-value) and $\lmcValue(v)$ (locally minimum cost-to-come estimate, or lmc-value). The $\lmcValue(v)$ is the best estimate of the cost-to-come of the vertex $v$, computed based on the g-value of the vertices in the predecessor set $\PrcPredecessor(v)$. The lmc-value (also called rhs-value in~\cite{koenig2004lifelong}) is a one-step ahead lookahead value based on the g-value and is thus potentially better informed than the g-value of the vetrex.
The lmc-value satisfies the following relationship
$$
\lmcValue(v) =
\begin{cases}
0, & \text{if } v = \xinit, \\
\min_{u \in \PrcPredecessor(\graph{G}, v)}( \gValue(u) + \costValue(u,v)), & \text{otherwise}.
\end{cases}
$$

\emph{Heuristic value}: Given a vertex $v \in V$, and a goal region $\Xgoal$, the function $\hValue : (v,\Xgoal) \mapsto r \in \reals$ returns an estimate of the optimal cost from $v$ to $\Xgoal$; it is 0 if $v \in \Xgoal$. It is an admissible heuristic if it never overestimates the actual cost of reaching $\Xgoal$. In this paper, we always assume an admissible heuristic. It is well known that inadmissible heuristics can be used to speed-up the algorithm, but they lead to suboptimal paths \cite{pearl1984heuristics}.

\emph{Relevant region}: Let $\xmingoal \in \Xgoal$ be the point in the goal region that has the lowest optimal cost-to-come value in $\Xgoal$, i.e., $\xmingoal = \mathrm{argmin}_{x \in \Xgoal}\gStarValue(x)$. The \emph{relevant region} of $\Xfree$ is the set of points $x$ for which the optimal cost-to-come value of $x$, plus the estimate of the optimal cost moving from $x$ to  $\Xgoal$ is less than the optimal cost-to-come value of $\xmingoal$, that is,

$$\Xrel = \{ x \in \Xfree : \gStarValue(x) + \hValue(x) < \gStarValue(\xmingoal) \}$$

Points that lie in the $\Xrel$ have the potential to be part of the optimal path starting at $\xinit$ and reaching $\Xgoal$.

\emph{Key value}: Given a vertex $v \in V$, the function $\PrcKey : v \mapsto k $ returns a real vector $k \in \reals^\mathrm{2}$, whose components are $k_{1}(v) = \min(\gValue(v),\lmcValue(v)) + \hValue(v)$ and $k_{2}(v) = \min(\gValue(v),\lmcValue(v))$, respectively. Components of the keys correspond to the f-values and g-values in the $\mathrm{A}^{*}$~algorithm, respectively~\cite{nilsson1971problem}.

\emph{Promising vertices}: Let $\vmingoal \in V$ be the vertex that has the lowest key value, i.e., $\vmingoal = \mathrm{argmin}_{v \in V \cap \Xgoal}\PrcKey(v)$. The \emph{promising vertices} $\Vprom \subseteq V$ is the set of vertices that have better key value than  $\vmingoal$, that is,

$$\Vprom = \{ v \in V : \PrcKey(v) \prec \PrcKey(\vmingoal) \}$$

\emph{Priority of vertices}: The priority of vertices in the queue is the same as the priority of their associated keys, and the precedence relation between keys is determined according to lexicographical ordering. Given two keys $k, k^{\prime} \in \reals^{\mathrm{2}}$, the Boolean function ${\preccurlyeq} : (k, k^{\prime}) \mapsto \{\False,\True\}$ returns $\True$  if and only if either $k_{1} < k^{\prime}_{1}$ or $\left( k_{1} = k^{\prime}_{1} \mathrm{~and~} k_{2} \leq k^{\prime}_{2} \right)$, and $\False$ otherwise.

\emph{Consistency}: A vertex $v \in V$ is called \emph{locally consistent} if and only if its g-value equals its lmc-value \cite{koenig2004lifelong}. Otherwise, it is an \emph{inconsistent} vertex. The notion of \emph{consistency} is very important because it allows one to update cost-to-come values of all vertices by propagating the effects of the changes in the topology of the graph. This way, an incremental search can reuse information from the previous searches, thus speeding up the whole algorithm.
The lmc-value always keeps the best up-to-date estimate of the cost-to-come value based on the current topology of the graph, whereas the g-value keeps an estimate of the cost-to-come value computed from a previous topology of the graph. Equality of the g- and lmc-values of a vertex implies that the changes in the topology of the graph will not effect the cost-to-come value of that vertex, that is, the topology of the graph is consistent with its previous configuration in the locality of the vertex.

A tree $\tree{T} = (V_{s},E_{s})$ is called a \emph{consistent tree} if and only if all of its promising vertices are consistent.

The g-value of all vertices equals to their respective optimal cost-to-come value if and only if all vertices are locally consistent~\cite{koenig2004lifelong}. The g-values have the following form when all vertices are locally consistent

$$
\gValue(v) =
\begin{cases}
0, & \text{if } v = \xinit, \\
\min_{u \in \PrcPredecessor(\graph{G}, v)}( \gValue(u) + \costValue(u,v)), & \text{otherwise}.
\end{cases}
$$

Then, the shortest path from $\xinit \in \Xfree$ to any vertex $v \in V$ can be found by starting at $v$ and traversing iteratively from the current vertex $u \in V$ to any of its predecessor $u^{\prime} \in \PrcPredecessor(\graph{G}, u)$ that minimizes $\gValue(u^{\prime}) + \costValue(u^{\prime},u)$ (ties can be broken arbitrarily), until $\xinit$ is reached.

\subsection{Problem Definition}

The proposed \AlgRRTsharp{} algorithm  solves the following motion planning problem:
Given a bounded and connected open set $\X \subset \reals^{{d}}$, and the sets $\Xfree$ and $\Xobs = \X \backslash \Xfree$, and given an initial point $\xinit \in \Xfree$ and a goal region $\Xgoal \subset \Xfree$, find the minimum-cost path connecting $\xinit$ to the goal region $\Xgoal$. If no such path exists, then report that no solution is possible.

\section{The \AlgRRTsharp{} Algorithm - Overview} \label{section:overview}

A brief description of each function used in the \AlgRRTsharp{} algorithm is given below.

\emph{Sampling}: $\PrcSample : \naturals \to \Xfree$ is a function that returns independent, identically distributed (i.i.d) samples from $\Xfree$.

\emph{Nearest neighbor}: $\PrcNearest$ is a function that returns a point from a given finite set $V$, which is the closest to a given point $x$ in terms of a given distance function.

\emph{Near vertices}: $\PrcNear$ is a function that returns $n$ number of points from a given finite set $V$, which is the closest to a given point $x$ in terms of a given distance function.

\emph{Steering}: $\PrcSteer$ is a function that returns the closest point in a ball centered around a given state $x$ to another given point $\xnew$.

\emph{Collision checking}: Given two points, the Boolean function $\PrcObstacleFree$ checks whether the minimum distance path connecting these two points belongs to $\Xfree$. It returns $\True$ if the line segment is a subset of the $\Xfree$.

\emph{Tree extension}: $\PrcExtend$ is a function that extends the nearest vertex of the tree $\tree{T}$ towards the randomly sampled point $\xrand$.

\emph{Reducing inconsistency}: Given a graph $\graph{G} = (V, E)$, a corresponding spanning tree $\tree{T} = (V_{s}, E_{s})$, where $V_{s} = V$ and $E_{s} \subset V \times V$ and a goal region $\Xgoal \subset \Xfree$, the function $\PrcReduceInconsistency : (\graph{G}, \tree{T}, \Xgoal) \mapsto (\graph{G}, \tree{T}^{\prime})$ operates on the inconsistent vertices of the tree $\tree{T}$ iteratively, and continues until the tree becomes consistent, that is, all vertices of the tree that are promising (see Section~4) are consistent. The $\PrcReduceInconsistency$ function is used to propagate the effects of the topological changes in the graph $\graph{G}$ as new vertices are added with each iteration.

A priority queue is used to sort all of the inconsistent vertices of the tree $\tree{T}$ based on their respective key values. The following functions are defined to manage the priority queue.

\emph{Update queue}: Given a vertex $v \in V$, the function $\PrcUpdateState$ changes the content of the queue based on the g- and lmc-values of the vertex $v$. If the vertex $v$ is inconsistent, then it is either inserted into the queue or its priority in the queue is updated based on its up-to-date key value if it is already inside the queue. Otherwise, the vertex is removed from the queue if it is a consistent vertex.

\emph{Find minimum}: The function $findmin()$ returns the vertex with the highest priority of all vertices in the queue, i.e., the vertex of minimum key value.

\emph{Remove a vertex}: Given a vertex $v \in V$, the function $remove()$ deletes the vertex $v$ from content of the queue.

\emph{Update priority}: Given a vertex $v \in V$, and a key value $k$, the function $update()$ changes the priority of the vertex $v$ in priority queue $q$, i.e., it reassigns the key value of the vertex $v$ with the new given key value $k$.

\emph{Inserting a vertex}: Given a vertex $v \in V$, and a key $k$, the function $insert()$ adds the vertex $v$ with the key value $k$ into queue.

\section{The \AlgRRTsharp{} Algorithm - Details} \label{section:details}

The body of the \AlgRRTsharp{} algorithm is given in Algorithm~\ref{alg:rrtsharp} and it is  similar to the other RRT-variants (\AlgRRT, \AlgRRG, \AlgRRTstar, etc) with the notable exception that it keeps track of vertex consistency using the key values of all current vertices in the graph.
One of the important difference between the \AlgRRTstar{} and \AlgRRTsharp{} algorithms is that all vertices in the tree computed by the \AlgRRTstar{} algorithm have a uniform type based on their finite cost-to-come value, whereas in the \AlgRRTsharp{} algorithm the vertices have different types based on their pair of estimates of the cost-to-come value. In the \AlgRRTsharp{} algorithm, each vertex $v$ can be classified in one of the following four categories based on the values of its $(\gValue(v), \lmcValue(v))$ pair.

\begin{itemize}
\item Consistent with finite key value: $\gValue(v) < \infty , \lmcValue(v) < \infty \mathrm{~and~} \gValue(v) = \lmcValue(v)$
\item Consistent with infinite key value: $\gValue(v) = \infty , \lmcValue(v) = \infty$
\item Inconsistent with finite key value: $\gValue(v) < \infty , \lmcValue(v) < \infty \mathrm{~and~} \gValue(v) \neq \lmcValue(v)$
\item Inconsistent with infinite g-value and finite lmc-value: $\gValue(v) = \infty , \lmcValue(v) < \infty$
\end{itemize}

Vertices in the second category are always non-promising, whereas vertices in the rest of categories can be either promising or non-promising. The promising vertices can be used to approximate the region $\Xrel \subseteq \Xfree$ of the free space that may contain the optimal path.


\input{rrtsharp.tex}

The algorithm starts by adding the initial point $\xinit$ into the vertex set of the underlying graph. Then, it incrementally grows the graph in $\Xfree$ by sampling a random point $\xrand$ from $\Xfree$ and extending some parts of the graph towards $\xrand$. Later, the $\PrcReduceInconsistency$ procedure, which is provided in Algorithm~\ref{alg:reduce_inconsistency}, propagates the new information due to the extension across the whole graph in order to improve the estimate of the cost-to-come value of the promising vertices in the graph. All computations due to the sampling and extension steps, followed by information propagation (Lines~\ref{line:rrtsharp_itbegin}-\ref{line:rrtsharp_itend} of Algorithm~\ref{alg:rrtsharp}), form a single \emph{iteration} of the algorithm. The process is repeated for a given fixed number of iterations, and the consistent spanning tree of the final graph is returned at the end.

The key difference between the \AlgRRTsharp{} algorithm and other RRT-variants is that a unique consistent spanning tree of the graph is maintained at the end of the each iteration of the algorithm. Since this tree is consistent, it contains information of the lowest-cost path, which can be achieved on the current graph, for each promising vertex of the graph. In addition, the g-value of the promising vertices equals to their respective optimal cost-to-come value that can be achieved through the edges of the tree. Therefore, each new vertex is initialized with the minimum possible estimate of its respective optimal cost-to-come value during extension (since all of its promising neighbor vertices have the lowest g-value), and this estimate  keeps improving to the best possible value whenever new information becomes available on any part of the graph. Hence, the g-value of each promising vertex of the graph converges to its optimal cost-to-come value very quickly.

\input{extend_rrtsharp.tex}

The $\PrcExtend$ procedure for the \AlgRRTsharp{} algorithm is given in Algorithm~\ref{alg:extend_rrtsharp}. During each iteration, the $\PrcExtend$ procedure tries to extend the graph towards the randomly sampled point $\xrand \in \Xfree$. First, the closest vertex in the graph $\xnearest$ is found in Line 3, then  $\xnearest$ is steered towards the randomly sampled point $\xrand$ in the next line. If the line segment connecting the steered point $\xnew$ and $\xnearest$ is feasible, then the new point $\xnew$ is prepared for inclusion to the vertex set of the graph. First, its cost-to-come estimate, i.e., the g-value and lmc-values, and the parent vertex are initialized by using information of the nearest vertex $\xnearest$. Then, a local search is performed in some neighborhood of $\xnew$, i.e., the set of vertices returned by the $\PrcNear$ procedure, in order to find the local minimum cost-to-come estimate value in Lines 10-15 and the corresponding parent vertex. The new vertex $\xnew$ and all extensions resulting in feasible trajectories are added to the vertex and edge set of the graph in Lines 16-17, respectively. In the end, the new vertex is decided to be inserted in the priority queue or not based on its consistency in the $\PrcUpdateState$ procedure.

\input{reduce_inconsistency.tex}

\input{auxiliary_procedures.tex}

Inclusion of each new vertex may result in an inconsistent vertex in the graph if a finite lmc-value is achieved. Therefore, consistency of the spanning tree needs to be checked, and appropriate operations must be performed in order to make it consistent, if necessary. The $\PrcReduceInconsistency$ procedure, which is provided in Algorithm~\ref{alg:reduce_inconsistency}, is called to make the spanning tree consistent by operating on the inconsistent and \textit{promising} vertices of the graph, iteratively. It simply pops the most promising inconsistent vertex from the priority queue, if there are any, and this inconsistent vertex is made consistent by assigning its lmc-value to its g-value. Then, its new g-value information is propagated among its neighbors in order to improve their lmc-values in Lines 7-11. However, this information propagation may also cause some vertices to be inconsistent; therefore, all resulting inconsistent vertices are inserted in the priority queue as well. This process continues until a consistent spanning tree is computed, that is, there is no inconsistent promising vertex left in the priority queue.

\FloatBarrier

\section{Numerical Simulations 1} \label{section:experiments}\label{section:simulations}

The \AlgRRTsharp{} algorithm was developed in C++ and run on a computer with a 2.40 GHz processor and 12GB RAM
running the Ubuntu 11.10 Linux operating system. A Fibonacci heap was implemented as priority queue to store inconsistent vertices during the search~\cite{fredman1987fibonacci}. Extensive simulations were run to compare the performance of the \AlgRRTsharp{} algorithm with the \AlgRRTstar{} algorithm, whose C implementation is available to download from the \AlgRRTstar{} authors' website (\url{http://sertac.scripts.mit.edu/rrtstar/}).

Both \AlgRRTsharp{} and \AlgRRTstar{} algorithms were run on three different problem types with the same sample sequence in order to demonstrate the difference in their behavior while growing the tree. All problems tested require finding an optimal path in a square environment minimizing the Euclidean path length. The heuristic value of a vertex is the Euclidean distance from the vertex to the goal. In the first problem type, there are no obstacles in the environment, whereas there are some box-like obstacles in the second and third problem types. In the third problem type, the environment is more cluttered than the one in the second problem type, containing many widely distributed small obstacles.

\begin{figure*}[htp]
  \begin{center}
	\mbox{
	
		\subfigure[]{\scalebox{0.28}{\includegraphics[trim = 4.0cm 6.937cm 3.587cm 7.0cm, clip =
          true]{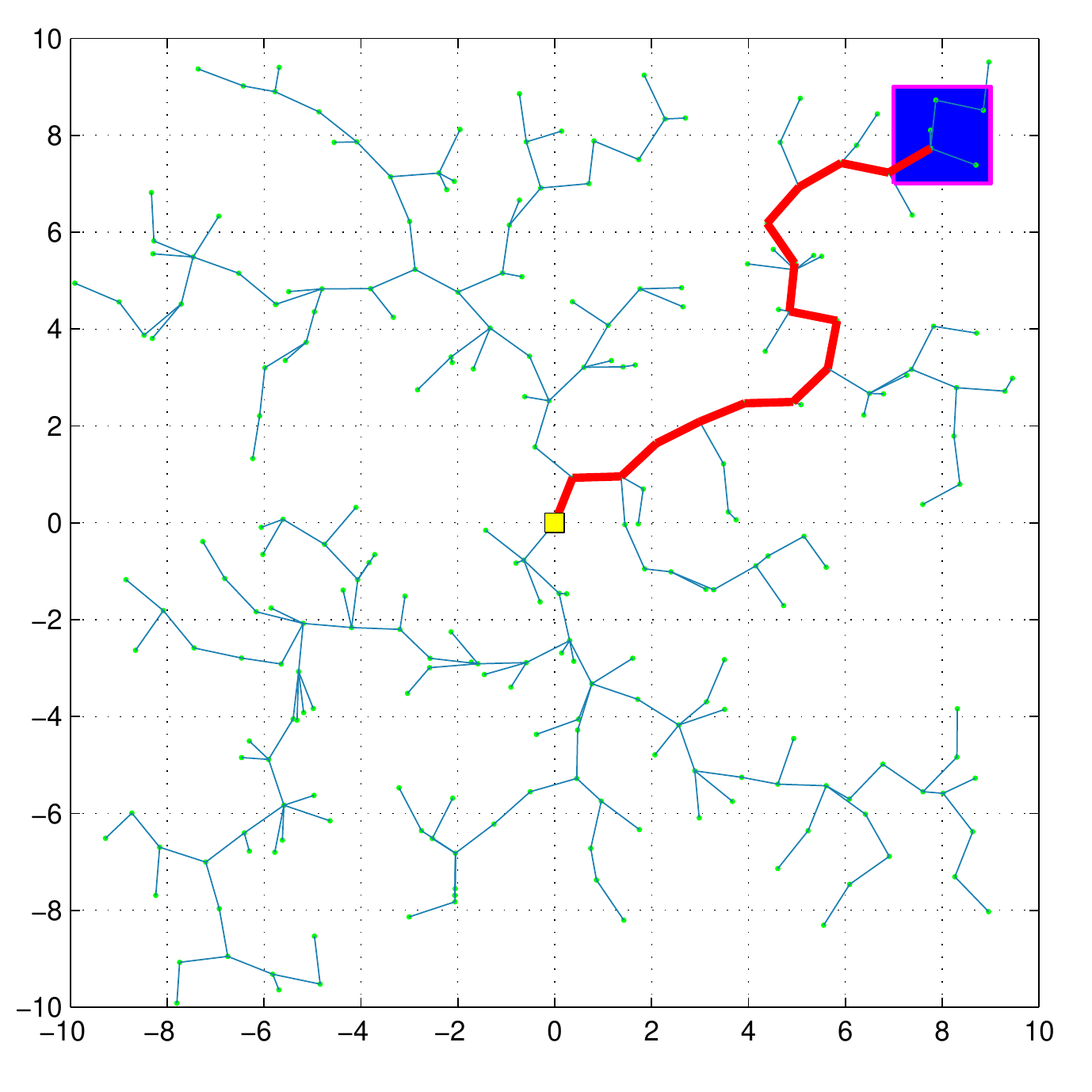}}\label{figure:pt1_rrtstar_it250}}
        \subfigure[]{\scalebox{0.28}{\includegraphics[trim = 4.0cm 6.937cm 3.587cm 7.0cm, clip =
          true]{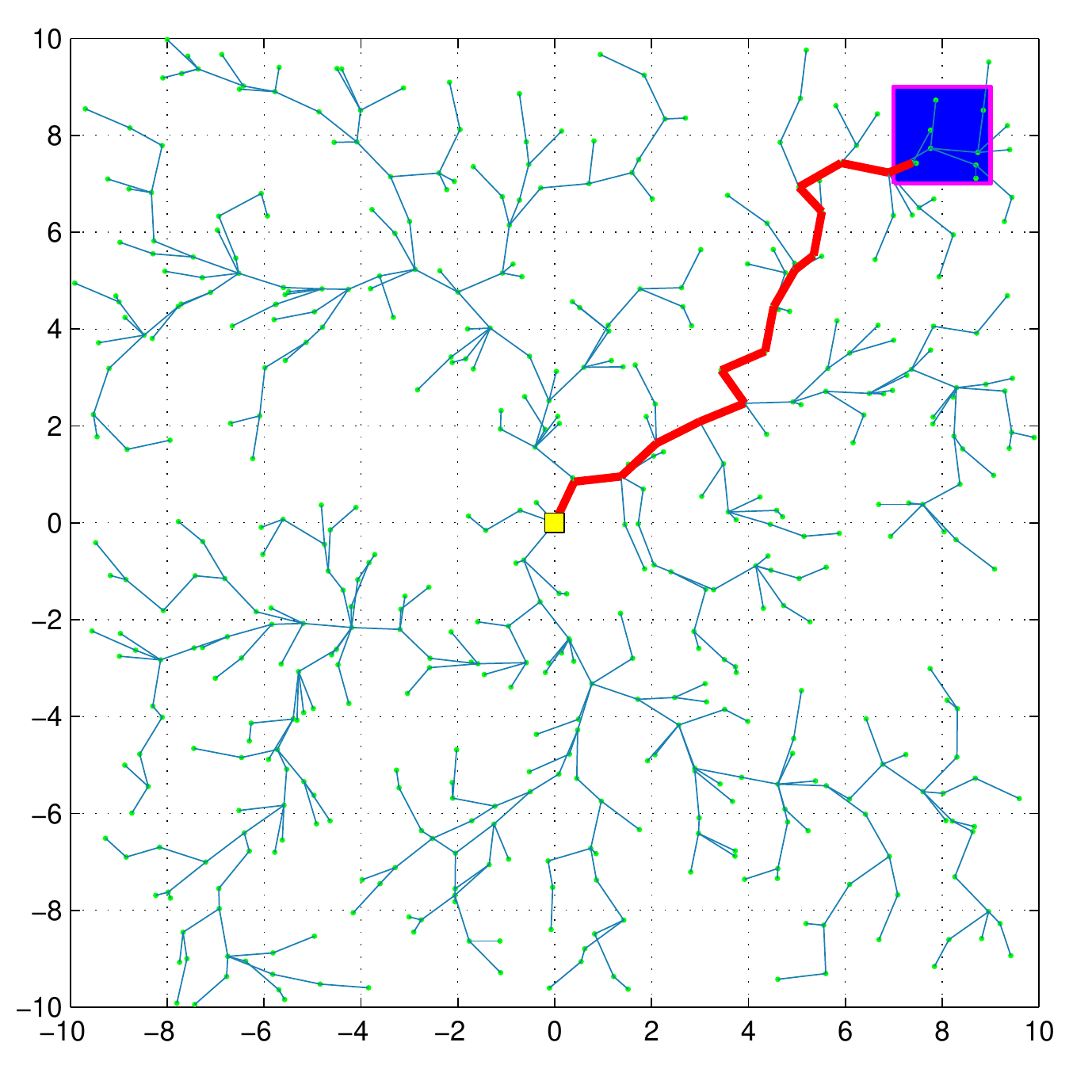}}\label{figure:pt1_rrtstar_it500}}
        \subfigure[]{\scalebox{0.28}{\includegraphics[trim = 4.0cm 6.937cm 3.587cm 7.0cm, clip =
          true]{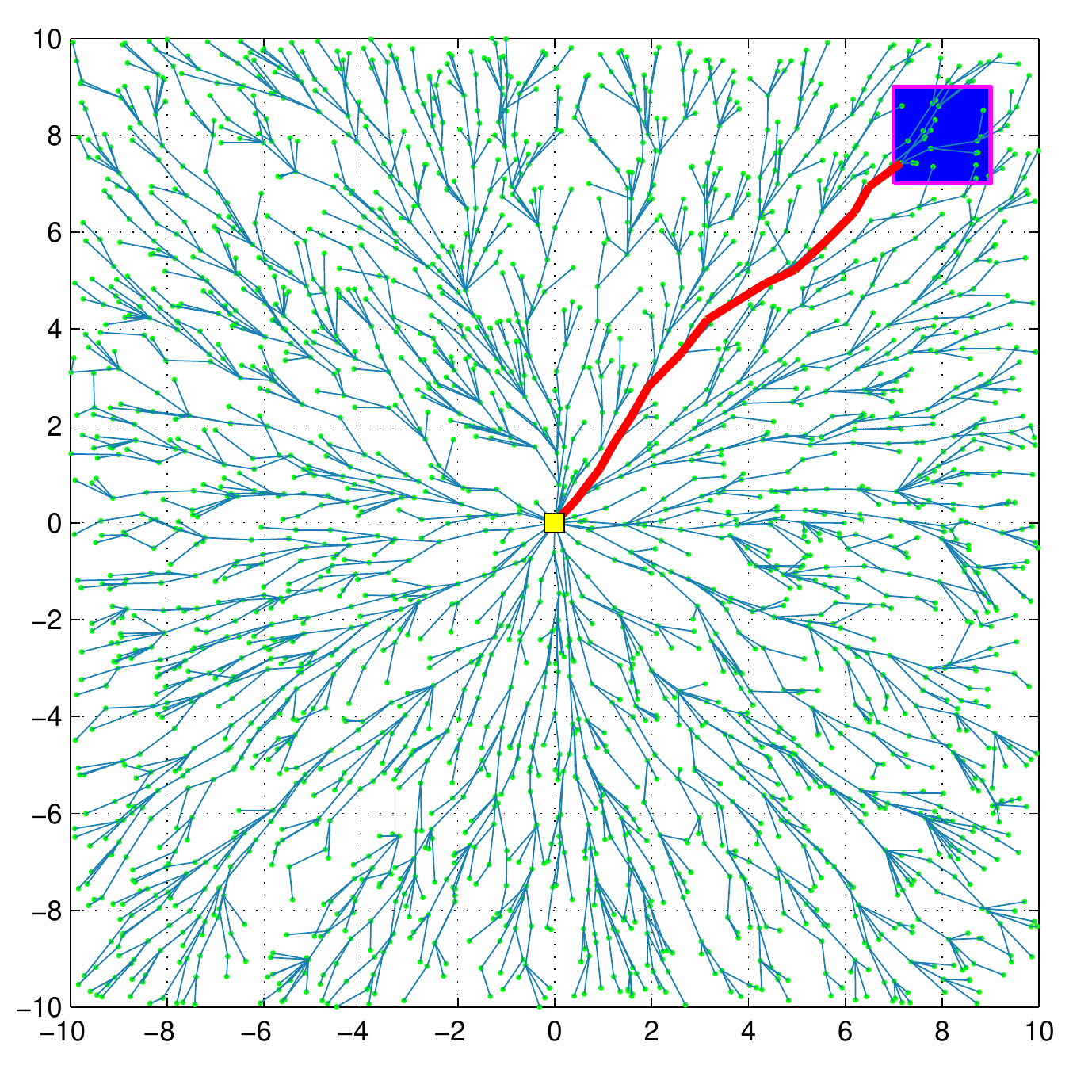}}\label{figure:pt1_rrtstar_it2500}}
        \subfigure[]{\scalebox{0.28}{\includegraphics[trim = 4.0cm 6.937cm 3.587cm 7.0cm, clip =
          true]{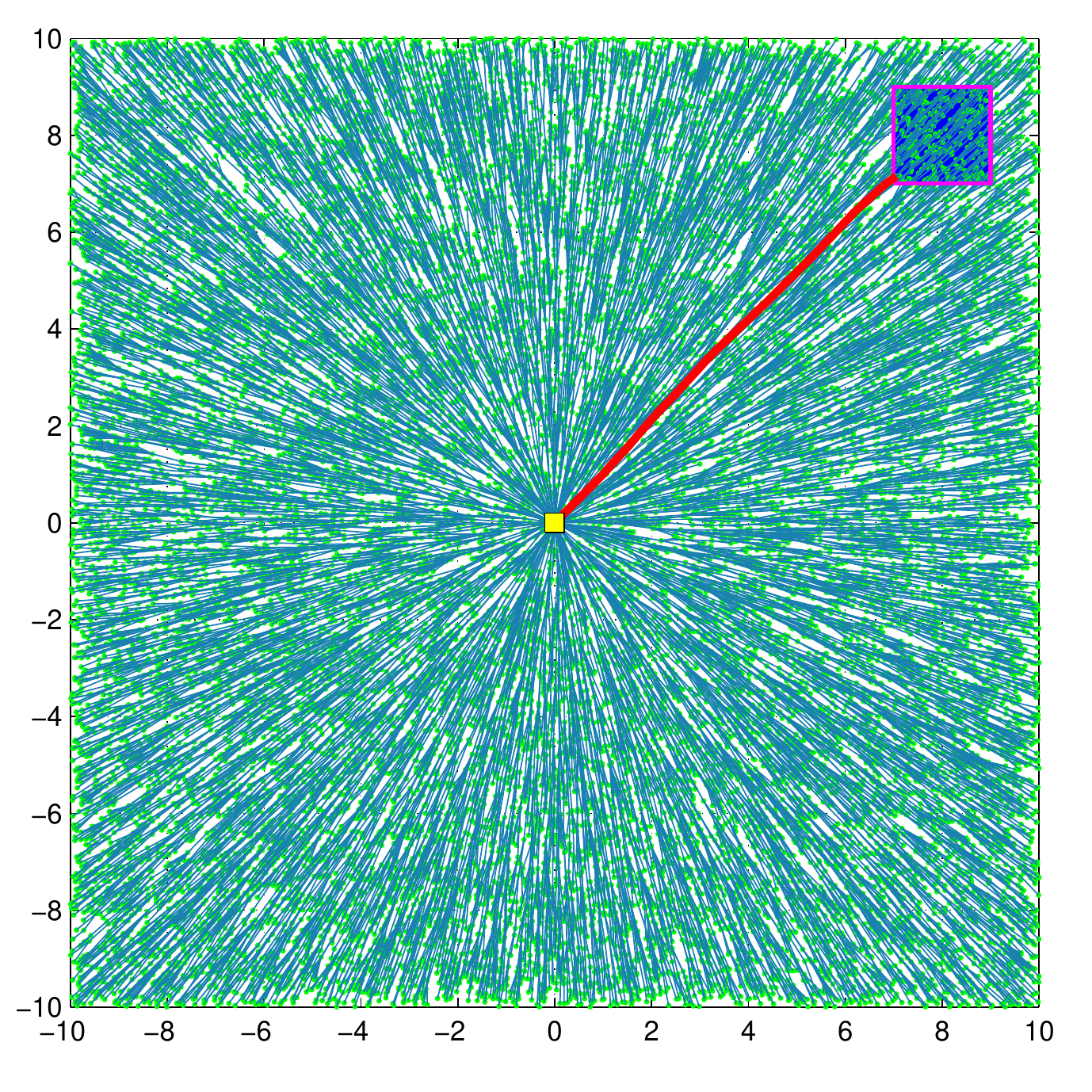}}\label{figure:pt1_rrtstar_it24999}}
    }
	\mbox{
		\subfigure[]{\scalebox{0.28}{\includegraphics[trim = 4.0cm 6.937cm 3.587cm 7.0cm, clip =
          true]{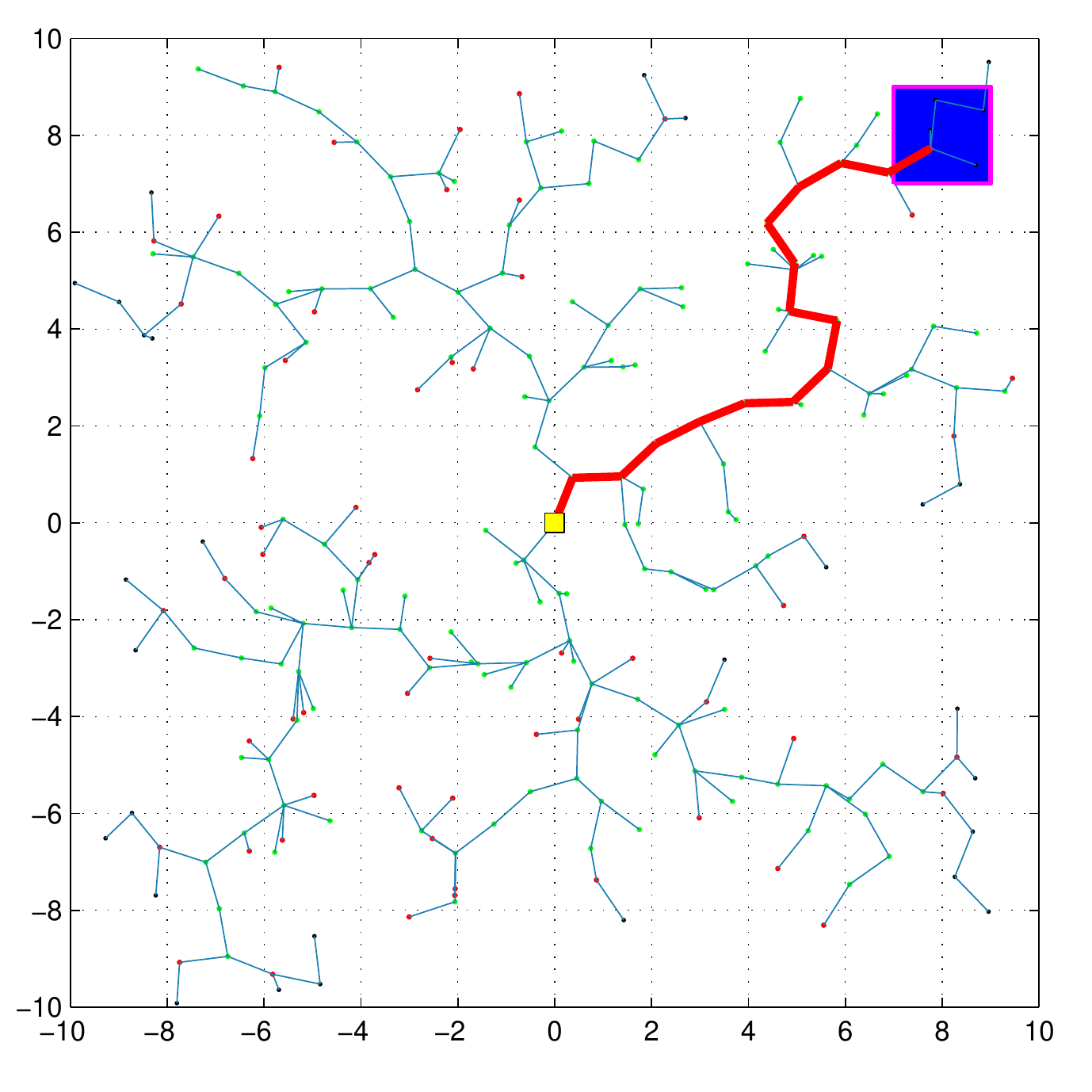}}\label{figure:pt1_rrtsharp_v0_it250}}
        \subfigure[]{\scalebox{0.28}{\includegraphics[trim = 4.0cm 6.937cm 3.587cm 7.0cm, clip =
          true]{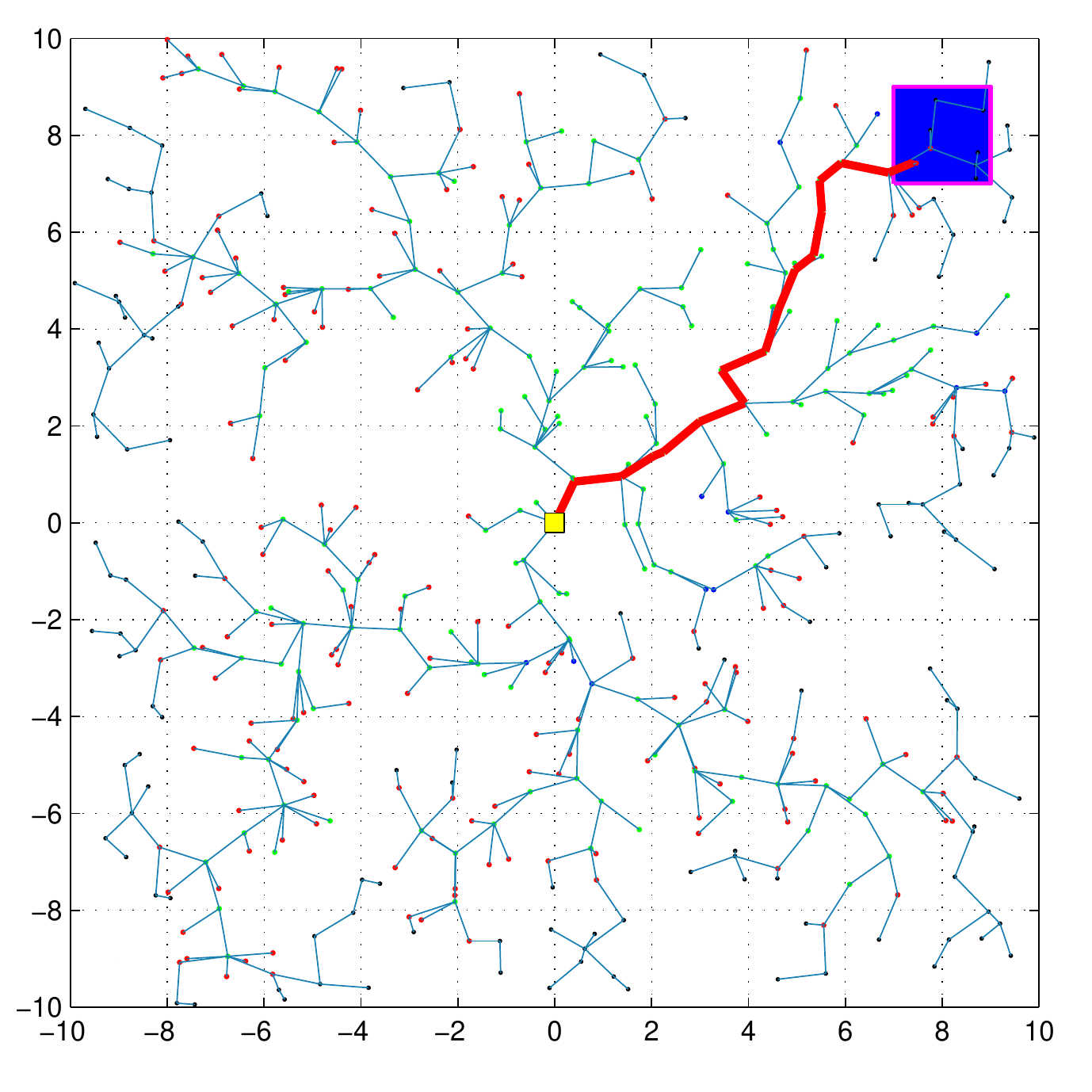}}\label{figure:pt1_rrtsharp_v0_it500}}
        \subfigure[]{\scalebox{0.28}{\includegraphics[trim = 4.0cm 6.937cm 3.587cm 7.0cm, clip =
          true]{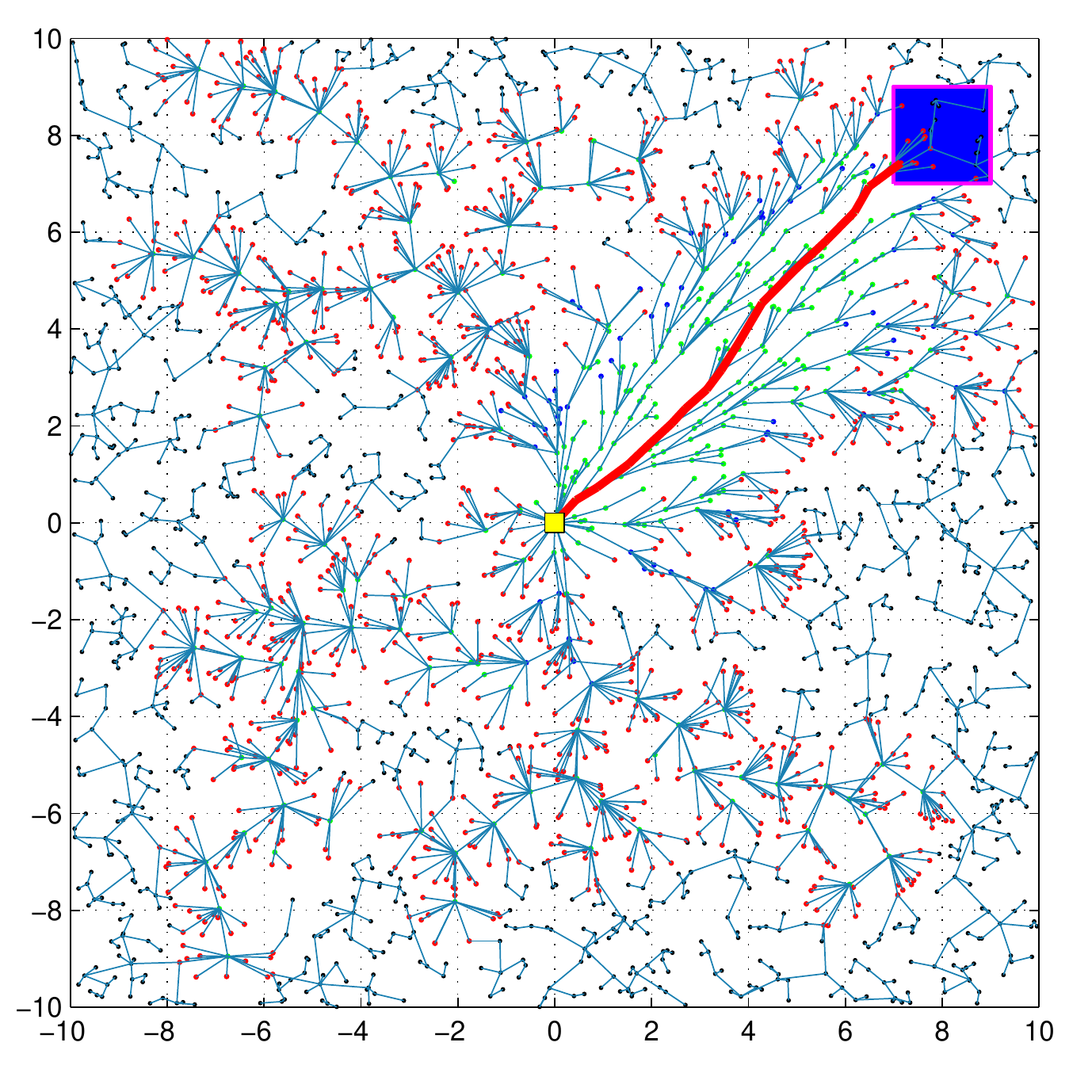}}\label{figure:pt1_rrtsharp_v0_it2500}}
        \subfigure[]{\scalebox{0.28}{\includegraphics[trim = 4.0cm 6.937cm 3.587cm 7.0cm, clip =
          true]{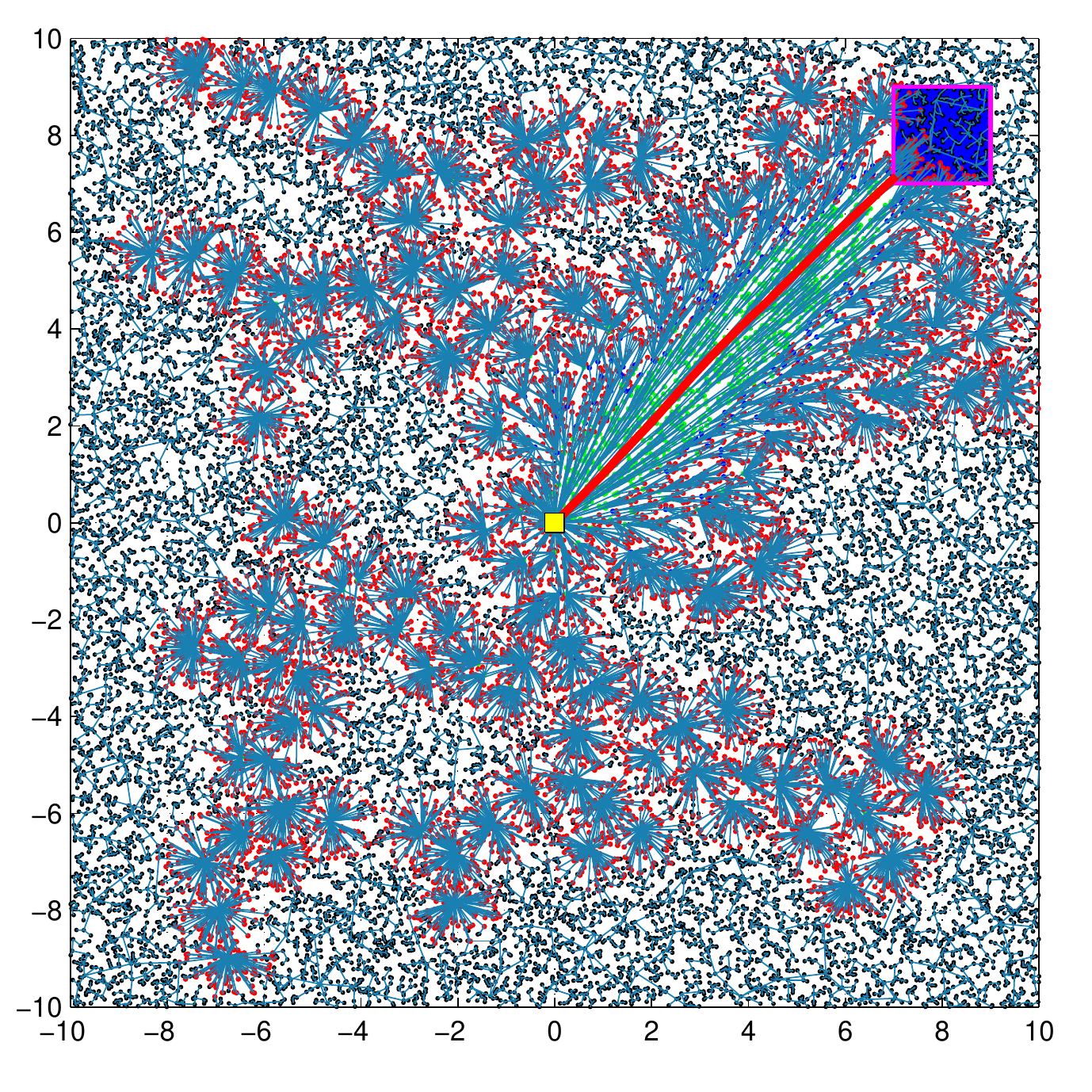}}\label{figure:pt1_rrtsharp_v0_it24999}}
    }
    \caption{The evolution of the tree computed by \AlgRRTstar{} and \AlgRRTsharp{} algorithms is shown in  \subref{figure:pt1_rrtstar_it250}-\subref{figure:pt1_rrtstar_it24999} and \subref{figure:pt1_rrtsharp_v0_it250}-\subref{figure:pt1_rrtsharp_v0_it24999}, respectively. The configuration of the trees \subref{figure:pt1_rrtstar_it250}, \subref{figure:pt1_rrtsharp_v0_it250} is at 250 iterations, \subref{figure:pt1_rrtstar_it500}, \subref{figure:pt1_rrtsharp_v0_it500} is at 500 iterations, \subref{figure:pt1_rrtstar_it2500}, \subref{figure:pt1_rrtsharp_v0_it2500} is at 2500 iterations, 
    and \subref{figure:pt1_rrtstar_it24999}, \subref{figure:pt1_rrtsharp_v0_it24999} is at 25000 iterations.}
    \label{figure:sim_d2_pt1_rrtstar_rrtsharp_v0_iterations}
  \end{center}
\end{figure*}

For the first problem type, the trees computed by both algorithms at different stages are shown in Figure~\ref{figure:sim_d2_pt1_rrtstar_rrtsharp_v0_iterations}. The initial state is plotted as a yellow square and the goal region is shown in blue with magenta border (upper right). The minimal-length path is shown in red. As shown in Figure~\ref{figure:sim_d2_pt1_rrtstar_rrtsharp_v0_iterations}, the best path computed by the \AlgRRTsharp{} algorithm converges to the optimal path. As mentioned earlier, one of the important differences between the \AlgRRTstar{} and \AlgRRTsharp{} algorithms is that the latter classifies the vertices in one of the following four categories based on the values of its $(\gValue(v), \lmcValue(v))$ pair: Consistent with finite key value (shown in green), consistent with infinite key value (shown in black), inconsistent with finite key value (shown in blue), and inconsistent with infinite g-value and finite lmc-value (shown in red).

Since only the points in the relevant region $\Xrel$ have the potential to be part of the optimal path, the \AlgRRTsharp{} algorithm tries to approximate $\Xrel$ with the set of promising vertices $\Vprom$ and tends to stop rewiring the parts of the tree which lie outside of the $\Xrel$ as iterations go to infinity. As seen in Figure~\ref{figure:sim_d2_pt1_rrtstar_rrtsharp_v0_iterations}, for this particular scenario, $\Xrel$ is an elliptic region, which is much smaller than the whole $\Xfree$. Therefore, uniform random sampling on $\Xfree$ results in too many vertices of different types (green, black, red, and blue vertices) outside of the relevant region during the search. The estimate of $\Xrel$ can be used to implement more intelligent sampling strategies, if needed, although this possibility was not pursued in this paper, where all sampling was uniform.

\begin{figure*}[htp]
  \begin{center}
	\mbox{
        \subfigure[]{\scalebox{0.26}{\includegraphics[trim = 4.0cm 3.0cm 4.0cm 3.0cm, clip =
          true]{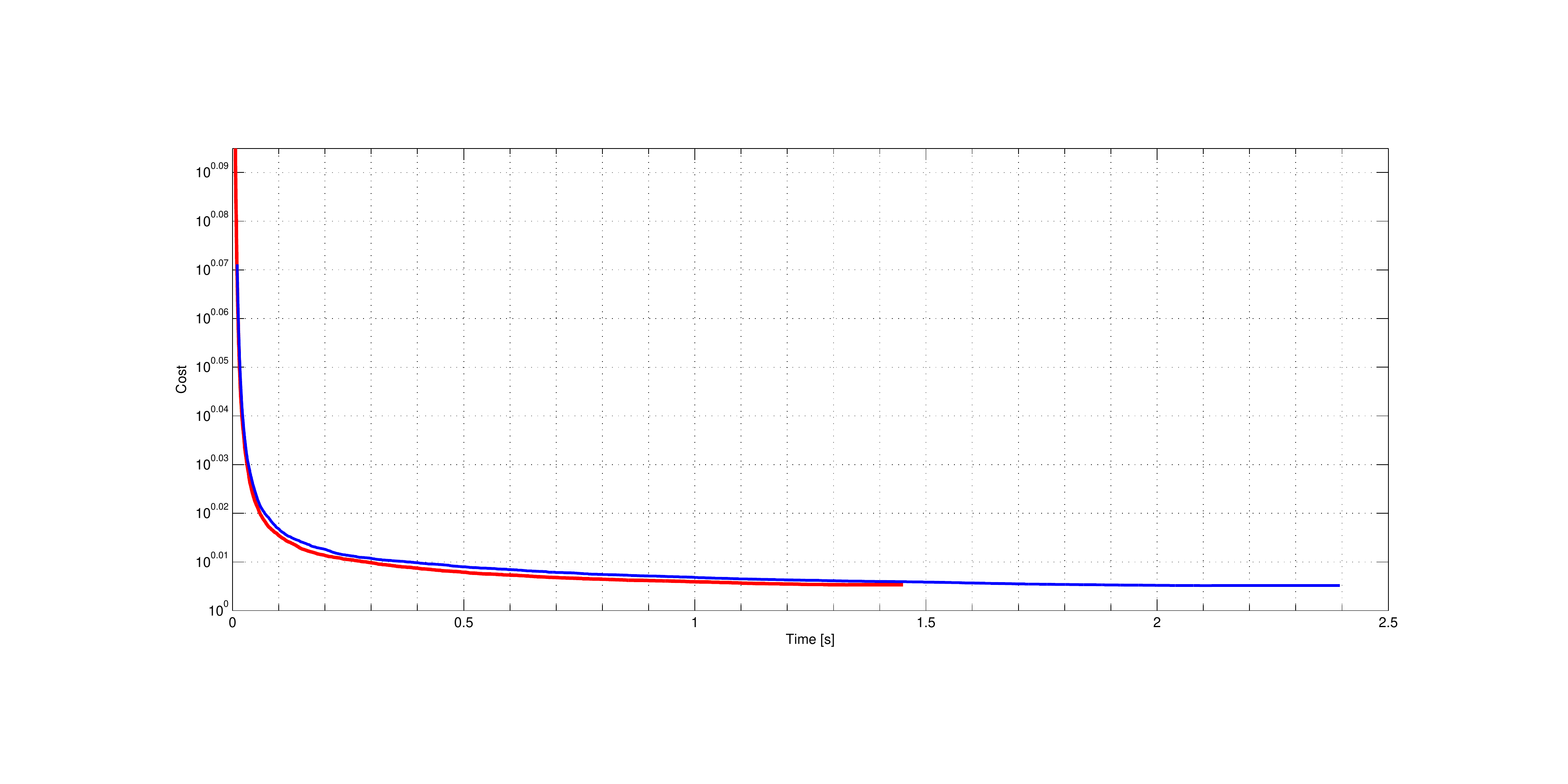}}\label{figure:time_cost_mean_d2_pt1_rrtstar_rrtsharp_v0}}
        \subfigure[]{\scalebox{0.26}{\includegraphics[trim = 4.0cm 3.0cm 4.0cm 3.0cm, clip =
          true]{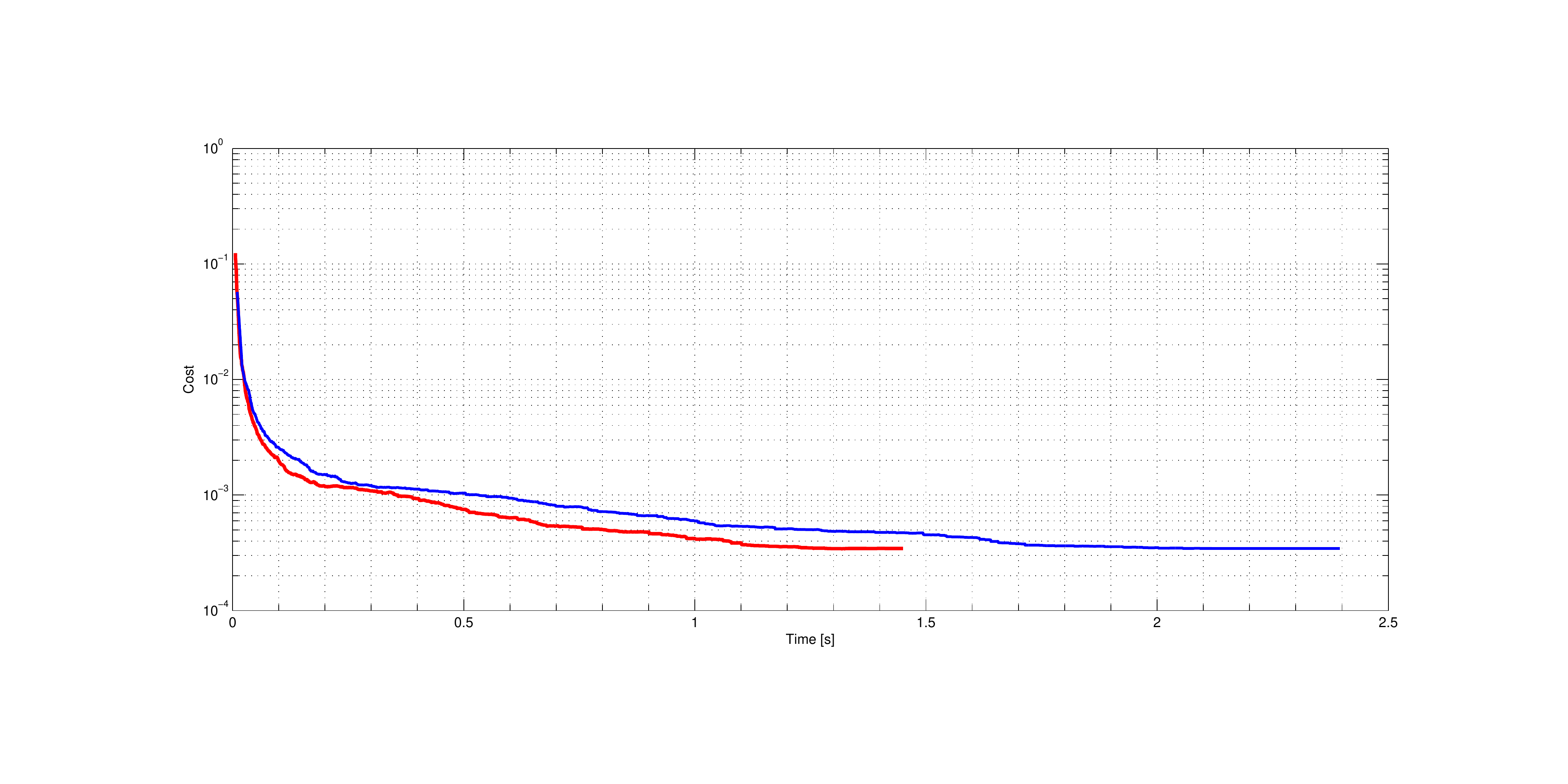}}\label{figure:time_cost_variance_d2_pt1_rrtstar_rrtsharp_v0}}
    }


    \caption{The change in the cost of the best paths computed by \AlgRRTstar{} and \AlgRRTsharp{} algorithms, and the variance in the trials are shown in \subref{figure:time_cost_mean_d2_pt1_rrtstar_rrtsharp_v0} and \subref{figure:time_cost_variance_d2_pt1_rrtstar_rrtsharp_v0}, respectively.}
    \label{figure:sim_d2_pt1_rrtstar_rrtsharp_v0_histories}
  \end{center}
\end{figure*}

\FloatBarrier

\begin{figure*}[htp]
  \begin{center}

	\mbox{
    \subfigure[]{\scalebox{0.28}{\includegraphics[trim = 4.0cm 6.937cm 3.587cm 7.0cm, clip =
          true]{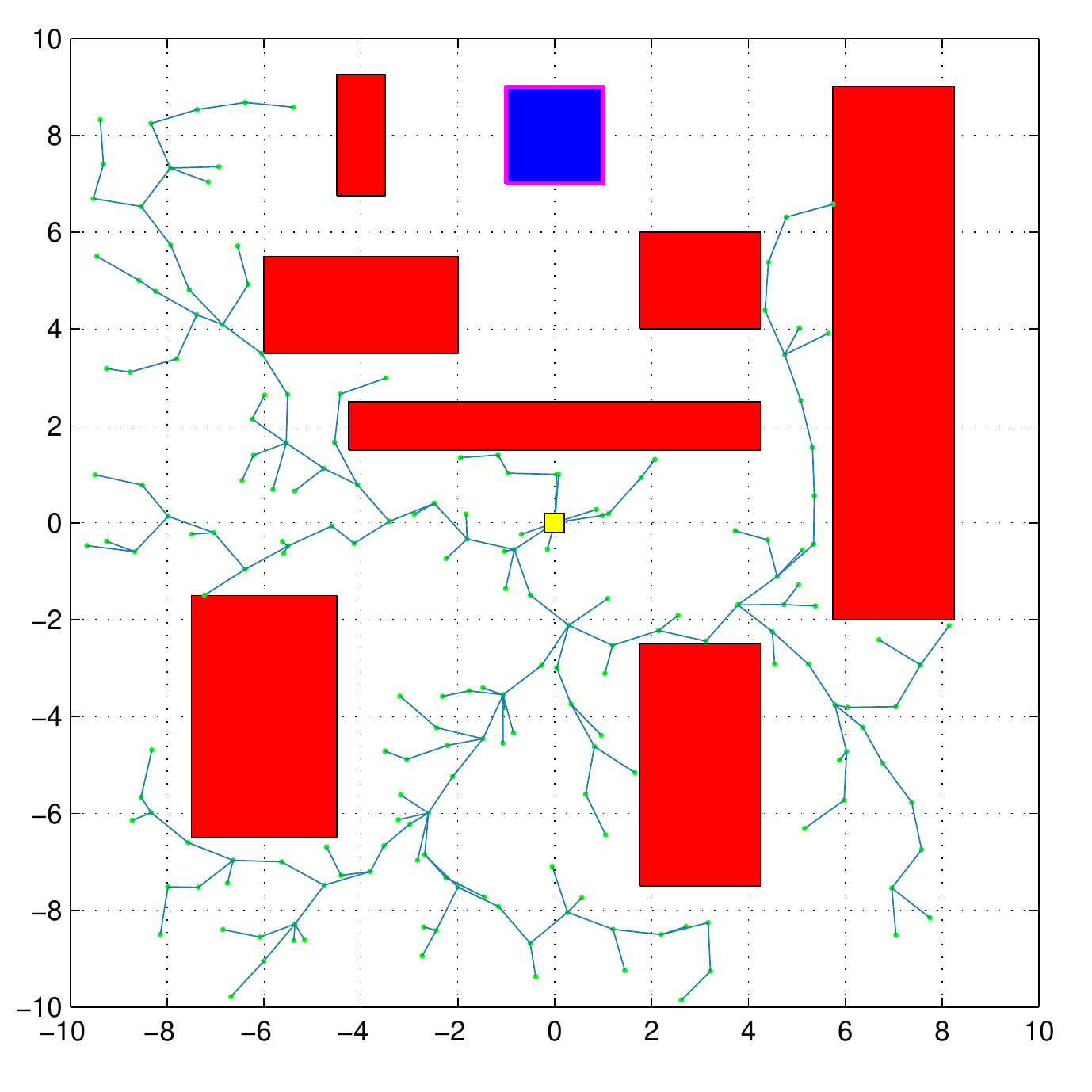}} \label{figure:pt2_rrtstar_it250}} 		
    \subfigure[]{\scalebox{0.28}{\includegraphics[trim = 4.0cm 6.937cm 3.587cm 7.0cm, clip =
          true]{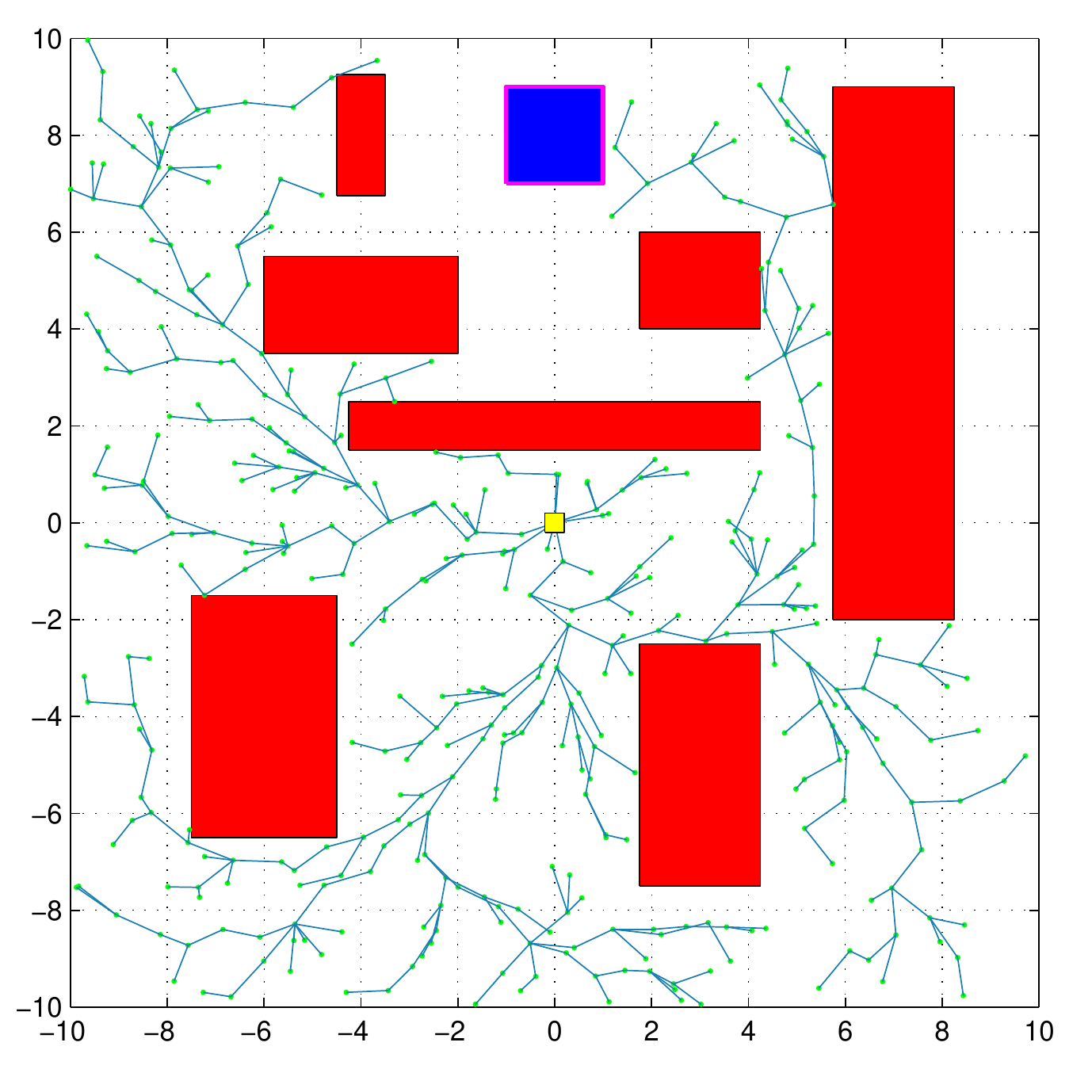}} \label{figure:pt2_rrtstar_it500}}
    \subfigure[]{\scalebox{0.28}{\includegraphics[trim = 4.0cm 6.937cm 3.587cm 7.0cm, clip =
          true]{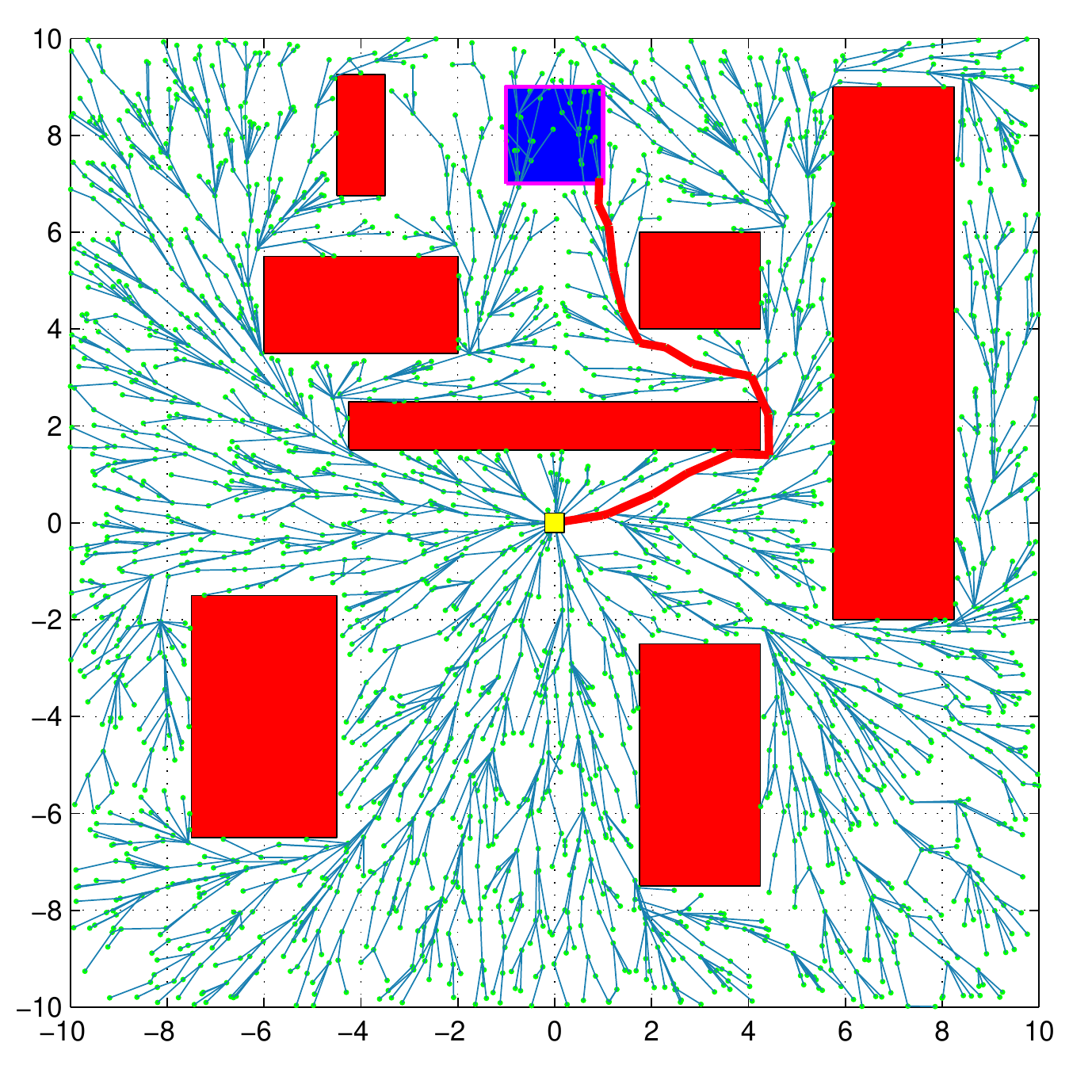}} \label{figure:pt2_rrtstar_it2500}}
    \subfigure[]{\scalebox{0.28}{\includegraphics[trim = 4.0cm 6.937cm 3.587cm 7.0cm, clip =
          true]{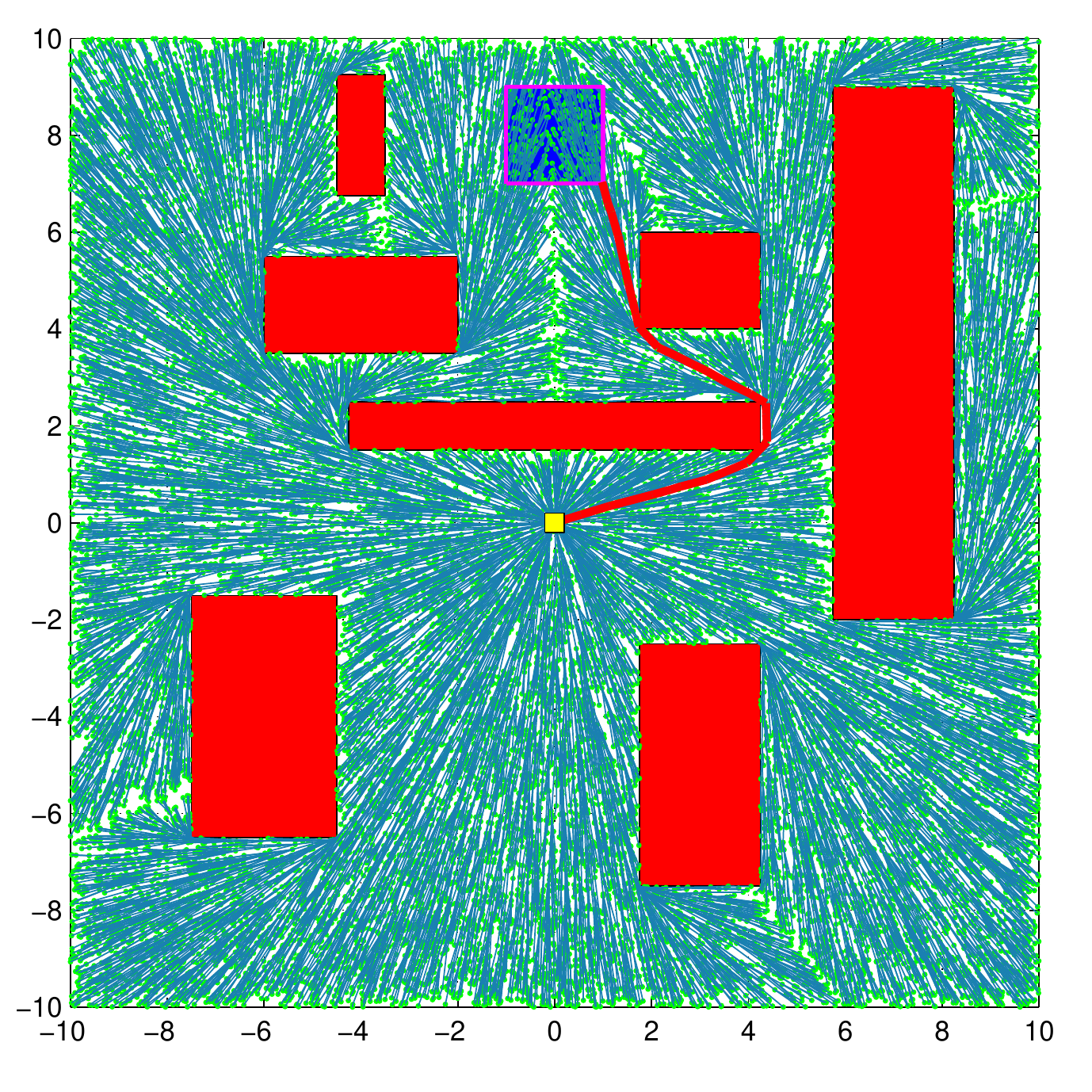}} \label{figure:pt2_rrtstar_it24999}}
    }

	\mbox{
    \subfigure[]{\scalebox{0.28}{\includegraphics[trim = 4.0cm 6.937cm 3.587cm 7.0cm, clip =
          true]{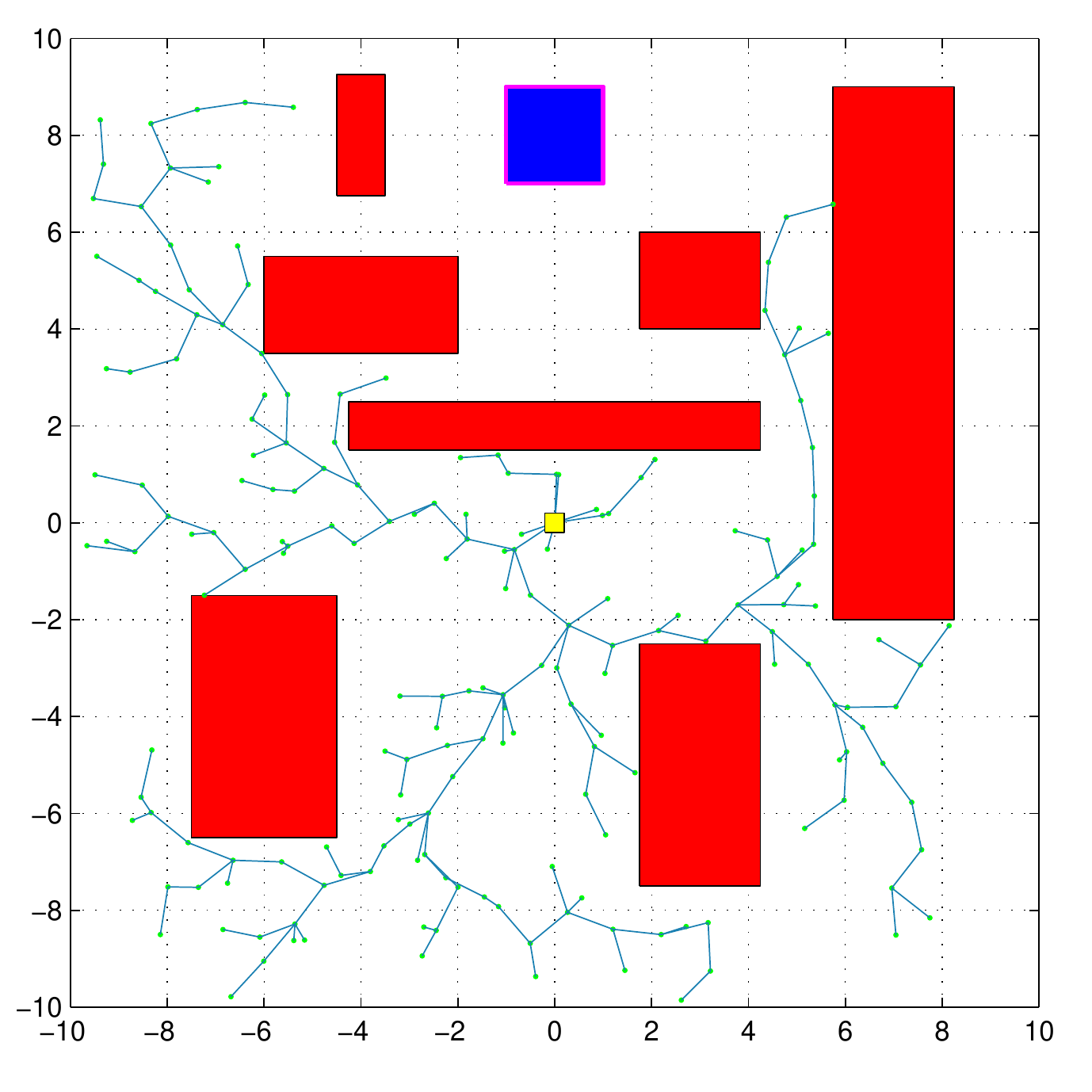}} \label{figure:pt2_rrtsharp_v0_it250}}
    \subfigure[]{\scalebox{0.28}{\includegraphics[trim = 4.0cm 6.937cm 3.587cm 7.0cm, clip =
          true]{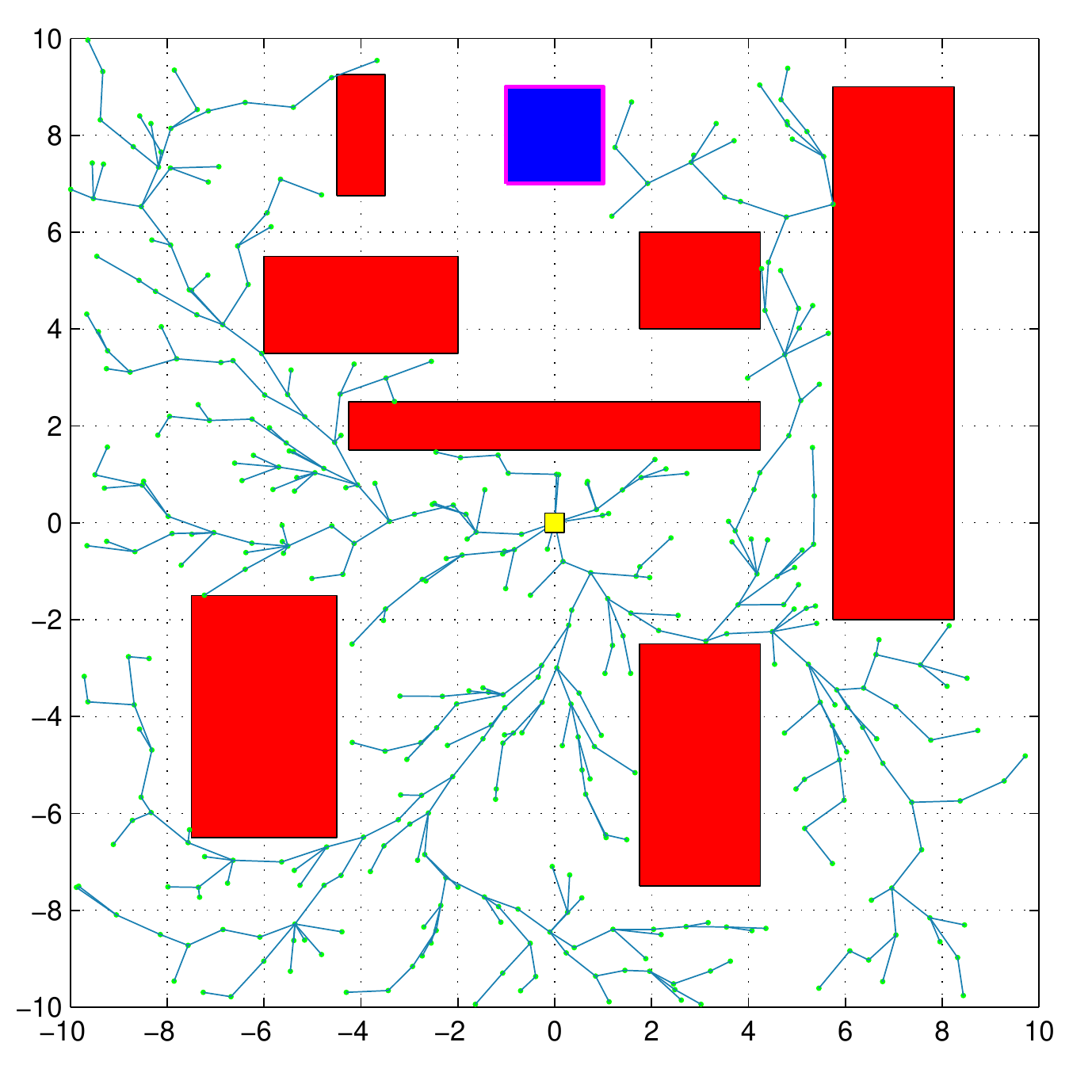}} \label{figure:pt2_rrtsharp_v0_it500}}
    \subfigure[]{\scalebox{0.28}{\includegraphics[trim = 4.0cm 6.937cm 3.587cm 7.0cm, clip =
          true]{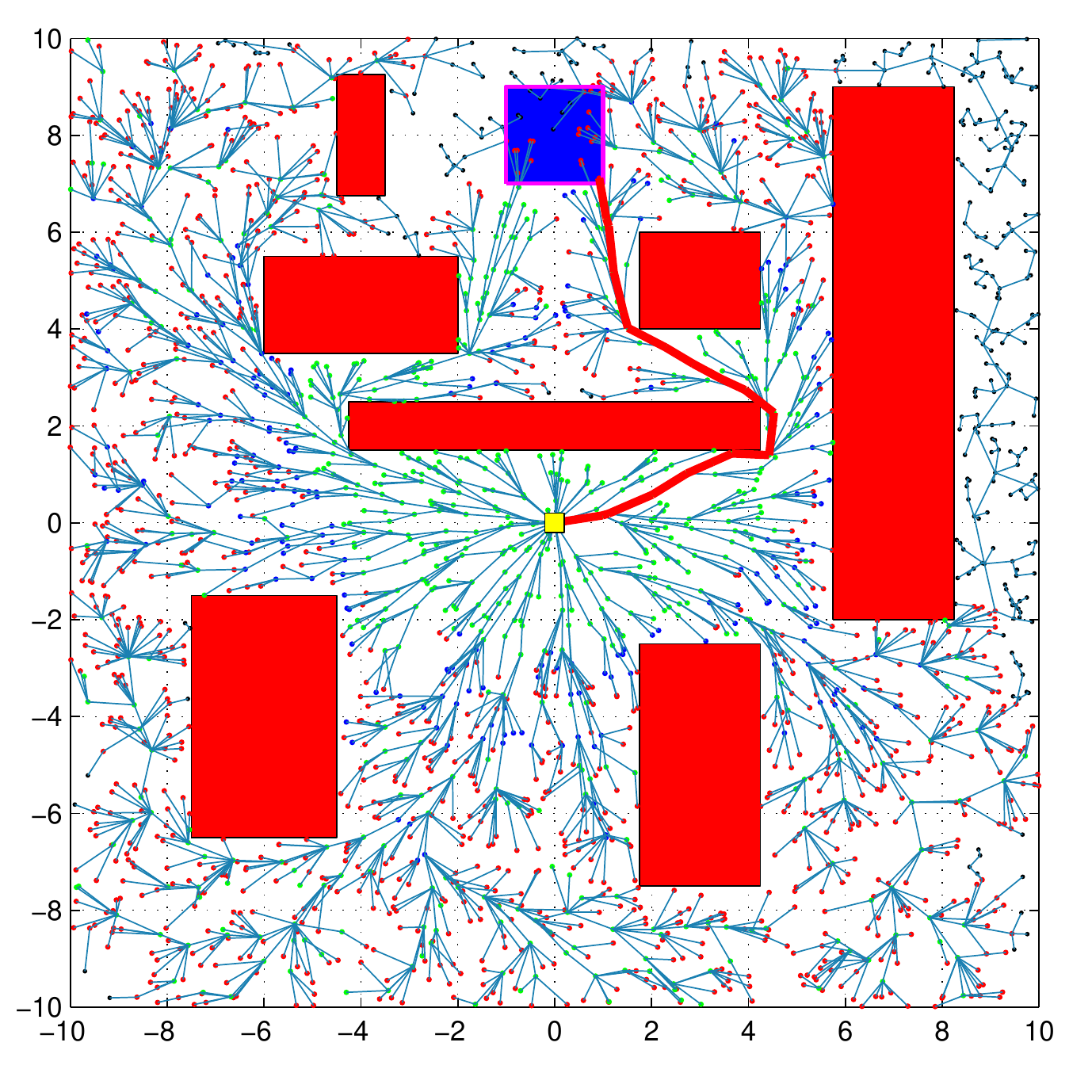}} \label{figure:pt2_rrtsharp_v0_it2500}}
    \subfigure[]{\scalebox{0.28}{\includegraphics[trim = 4.0cm 6.937cm 3.587cm 7.0cm, clip =
          true]{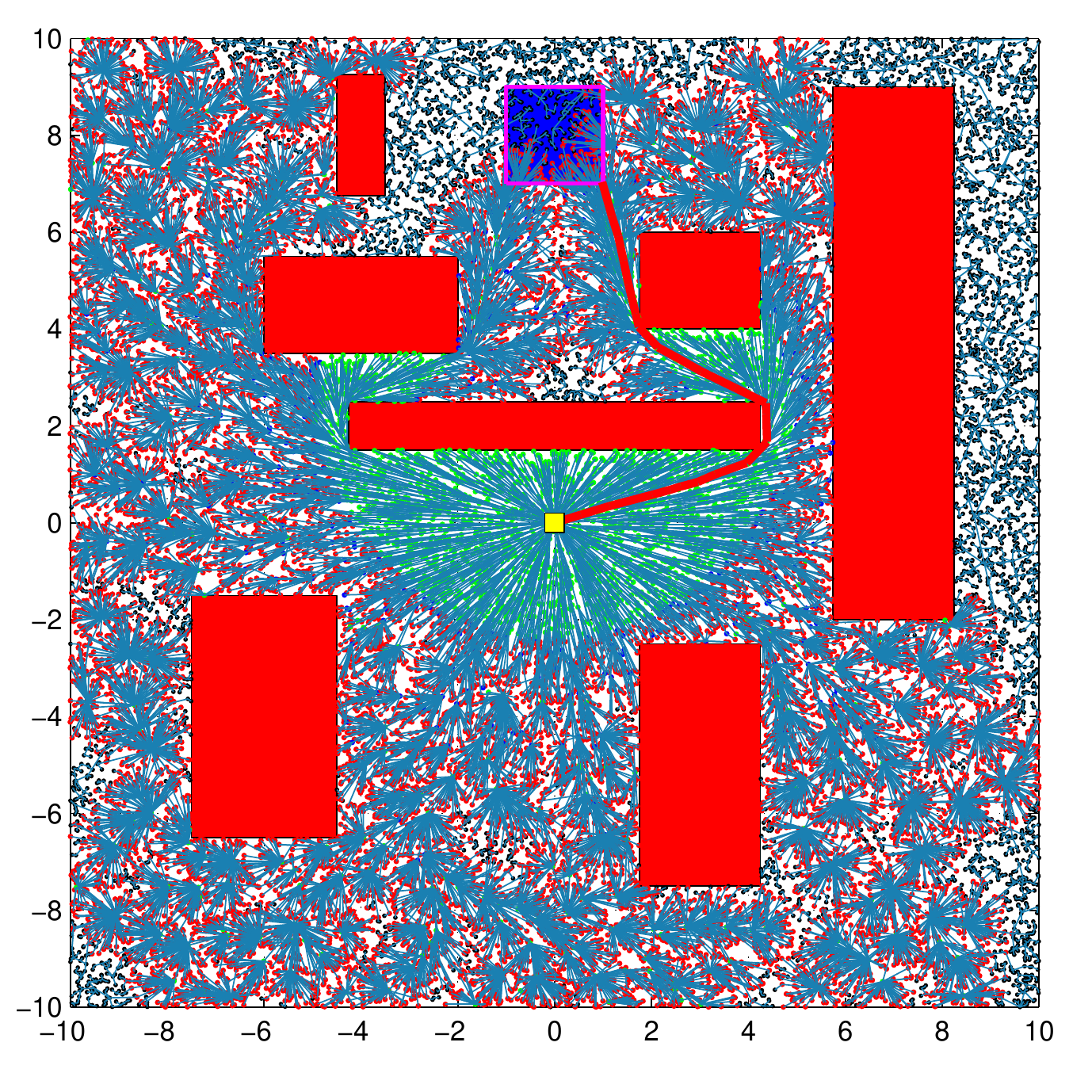}} \label{figure:pt2_rrtsharp_v0_it24999}}
   	}


    \caption{The evolution of the tree computed by \AlgRRTstar{} and \AlgRRTsharp{} algorithms is shown in  \subref{figure:pt2_rrtstar_it250}-\subref{figure:pt2_rrtstar_it24999} and \subref{figure:pt2_rrtsharp_v0_it250}-\subref{figure:pt2_rrtsharp_v0_it24999}, respectively. The configuration of the trees \subref{figure:pt2_rrtstar_it250}, \subref{figure:pt2_rrtsharp_v0_it250} is at 250 iterations, \subref{figure:pt2_rrtstar_it500}, \subref{figure:pt2_rrtsharp_v0_it500} is at 500 iterations, \subref{figure:pt2_rrtstar_it2500}, \subref{figure:pt2_rrtsharp_v0_it2500} is at 2500 iterations, 
    and \subref{figure:pt2_rrtstar_it24999}, \subref{figure:pt2_rrtsharp_v0_it24999} is at 25000 iterations.
    }
    \label{figure:sim_d2_pt2_rrtstar_rrtsharp_v0_iterations}
  \end{center}
\end{figure*}

In the second problem type, the same experiment was carried out and both algorithms were run in an environment with several obstacles. The configuration of the trees for both the \AlgRRTstar{} and \AlgRRTsharp{} algorithms at different stages are shown in Figure~\ref{figure:sim_d2_pt2_rrtstar_rrtsharp_v0_iterations}.

\begin{figure*}[htp]
  \begin{center}
	\mbox{
        \subfigure[]{\scalebox{0.26}{\includegraphics[trim = 4.0cm 3.0cm 4.0cm 3.0cm, clip =
          true]{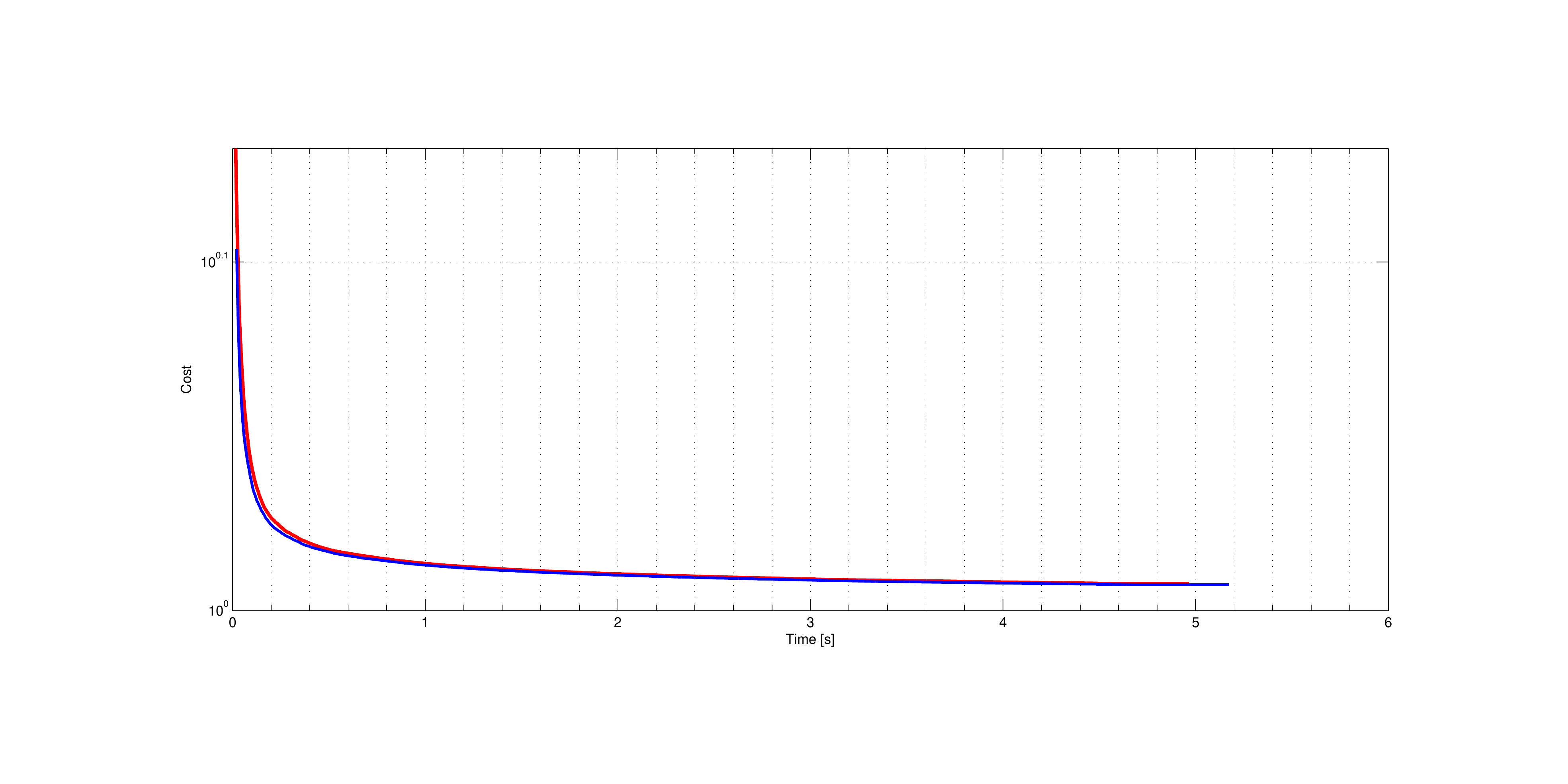}}\label{figure:time_cost_mean_d2_pt2_rrtstar_rrtsharp_v0}}

    	\subfigure[]{\scalebox{0.26}{\includegraphics[trim = 4.0cm 3.0cm 4.0cm 3.0cm, clip =
          true]{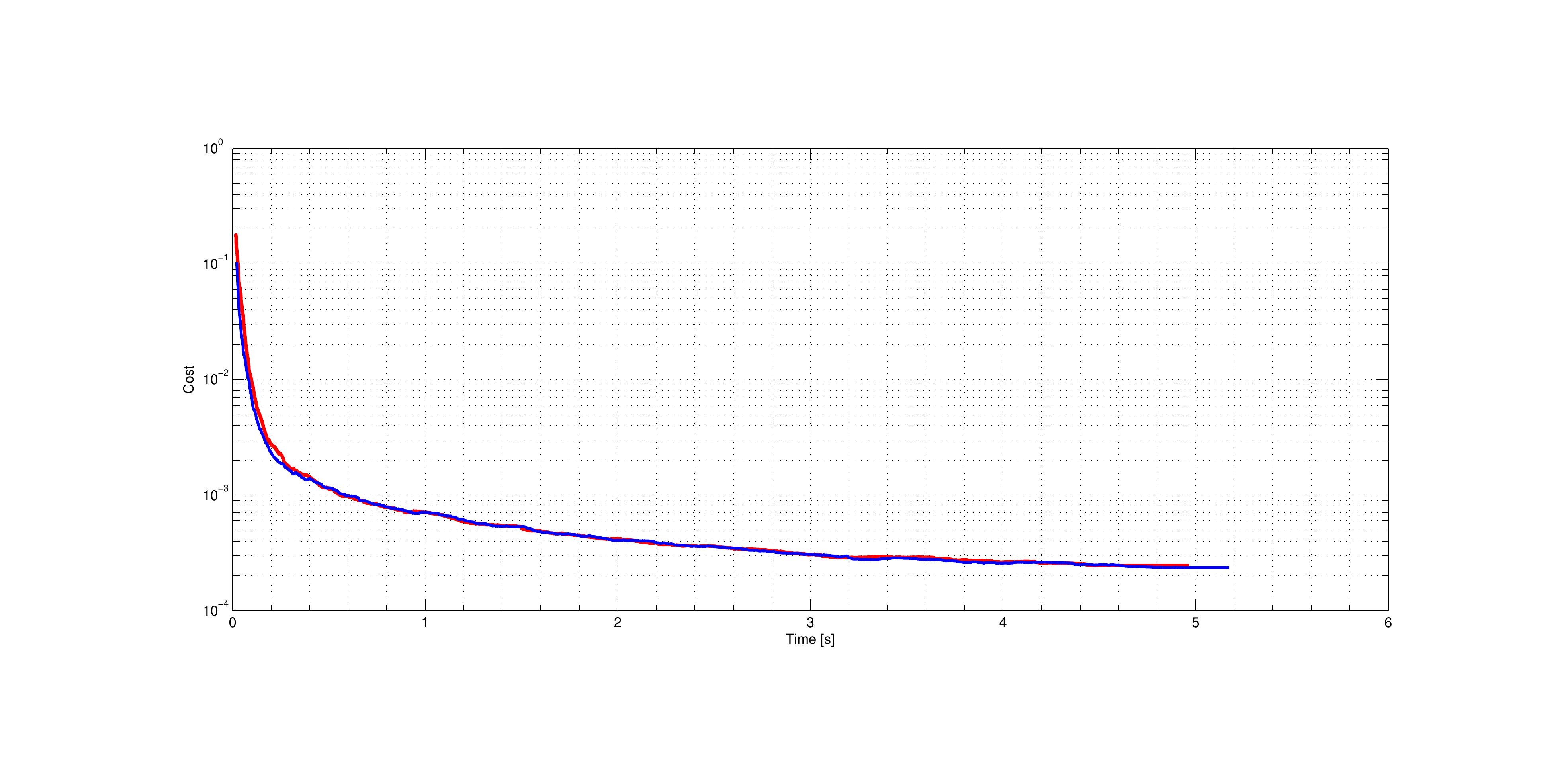}}\label{figure:time_cost_variance_d2_pt2_rrtstar_rrtsharp_v0}}
    }

%
    \caption{The change in the cost of the best paths computed by \AlgRRTstar{} and \AlgRRTsharp{} algorithms and the variance in the trials are shown in \subref{figure:time_cost_mean_d2_pt2_rrtstar_rrtsharp_v0} and \subref{figure:time_cost_variance_d2_pt2_rrtstar_rrtsharp_v0}, respectively.}
    \label{figure:sim_d2_pt2_rrtstar_rrtsharp_v0_histories}
  \end{center}
\end{figure*}

\FloatBarrier

\begin{figure*}[htp]
  \begin{center}

	\mbox{
    \subfigure[]{\scalebox{0.28}{\includegraphics[trim = 4.0cm 6.937cm 3.587cm 7.0cm, clip =
          true]{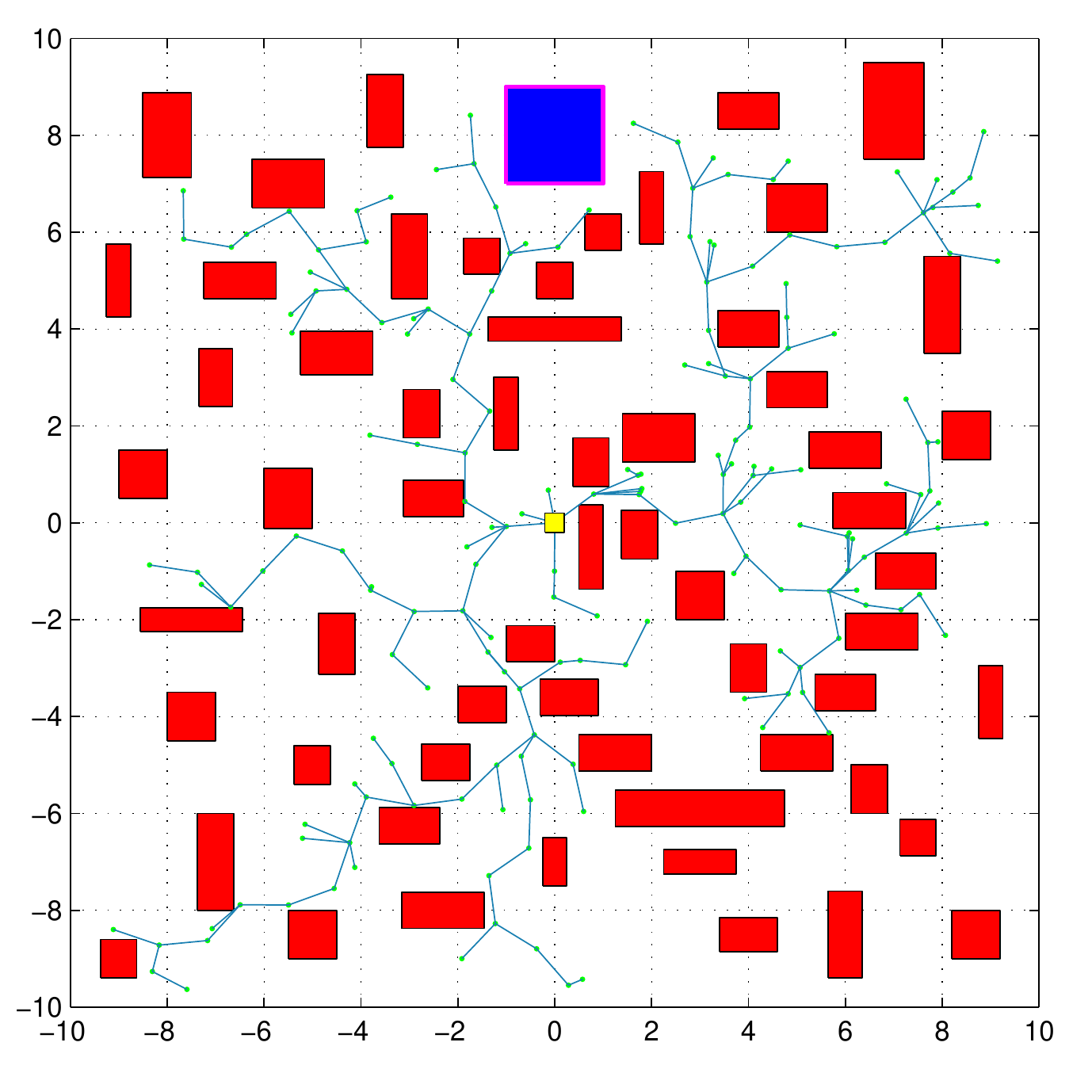}} \label{figure:pt3_rrtstar_it250}} 		
    \subfigure[]{\scalebox{0.28}{\includegraphics[trim = 4.0cm 6.937cm 3.587cm 7.0cm, clip =
          true]{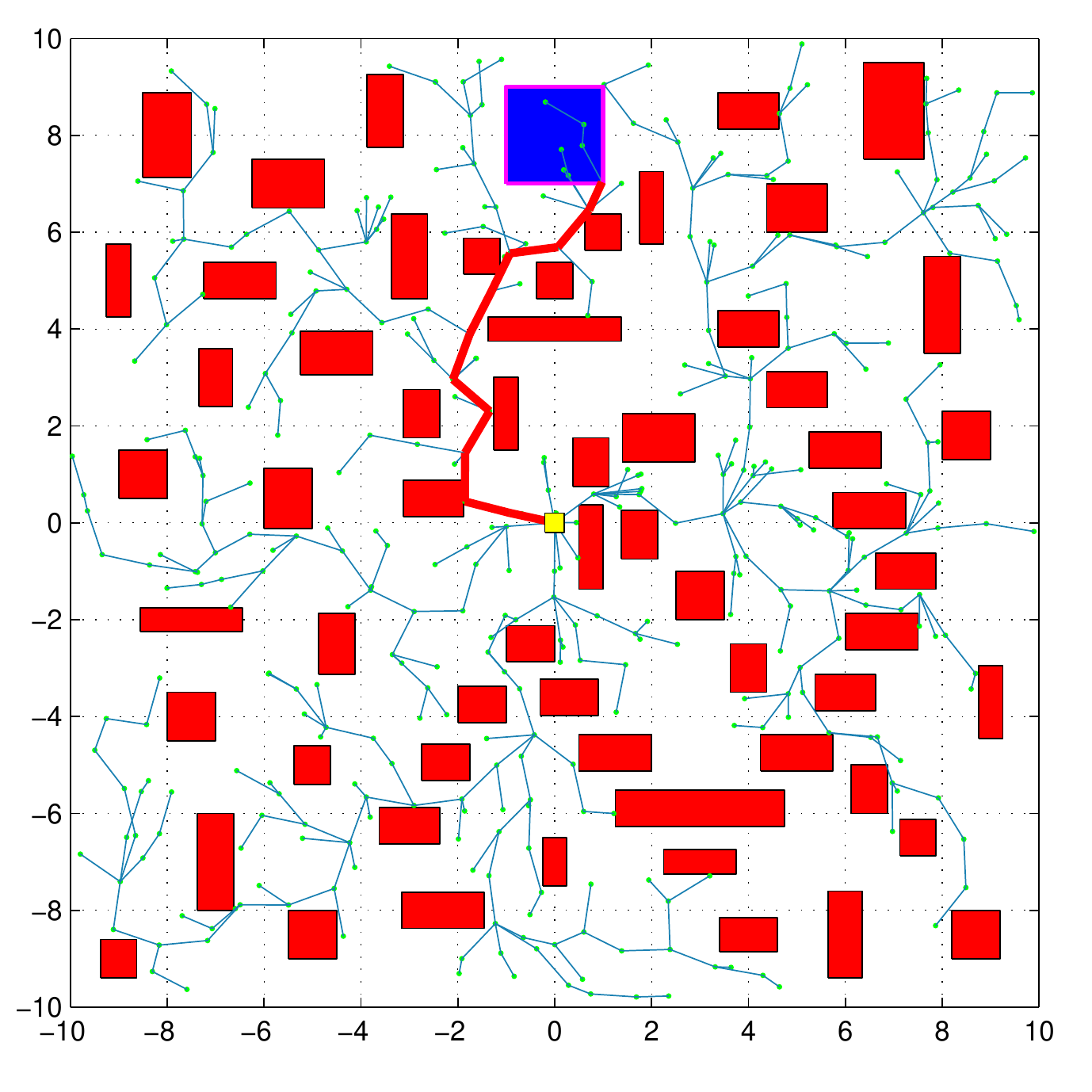}} \label{figure:pt3_rrtstar_it500}}
    \subfigure[]{\scalebox{0.28}{\includegraphics[trim = 4.0cm 6.937cm 3.587cm 7.0cm, clip =
          true]{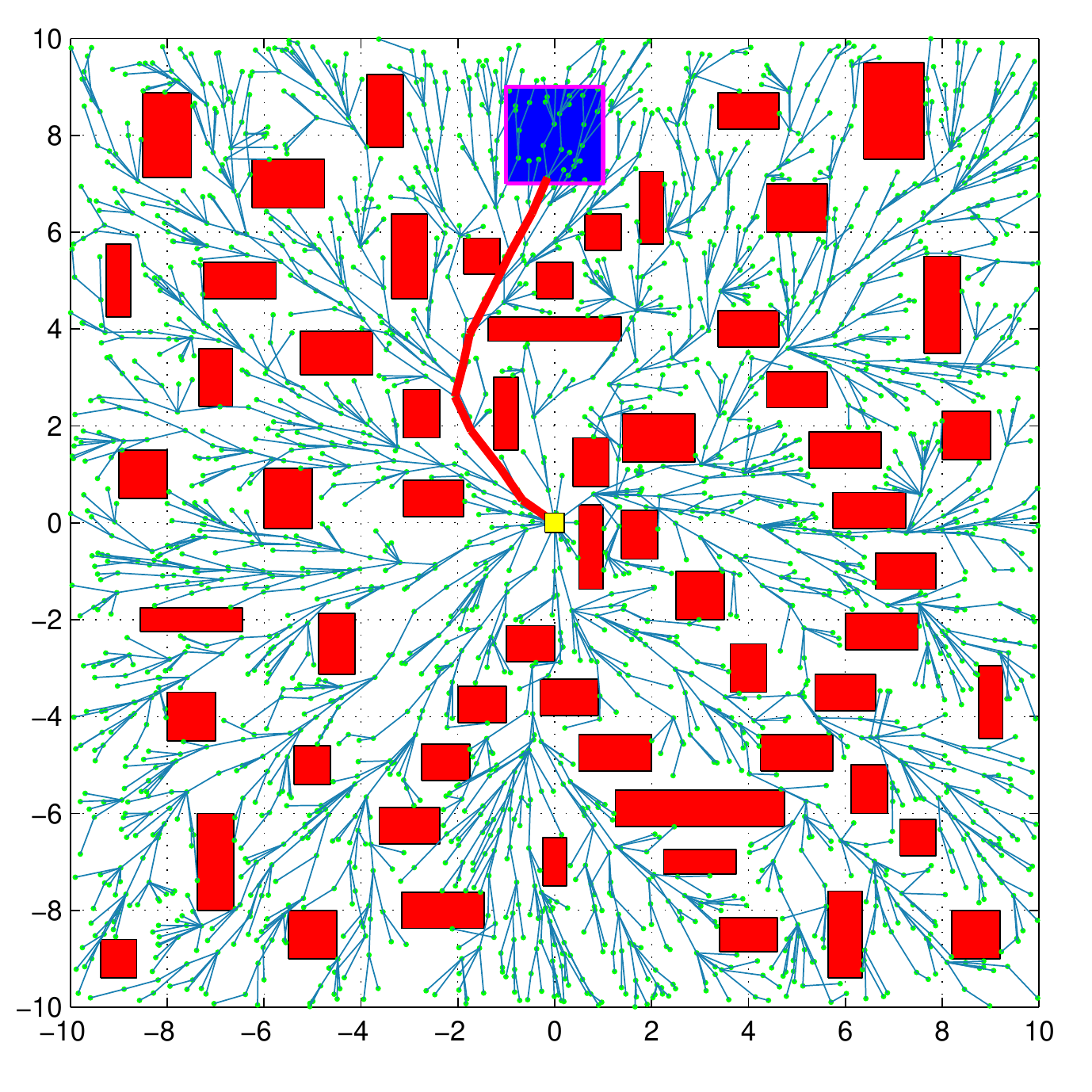}} \label{figure:pt3_rrtstar_it2500}}
    \subfigure[]{\scalebox{0.28}{\includegraphics[trim = 4.0cm 6.937cm 3.587cm 7.0cm, clip =
          true]{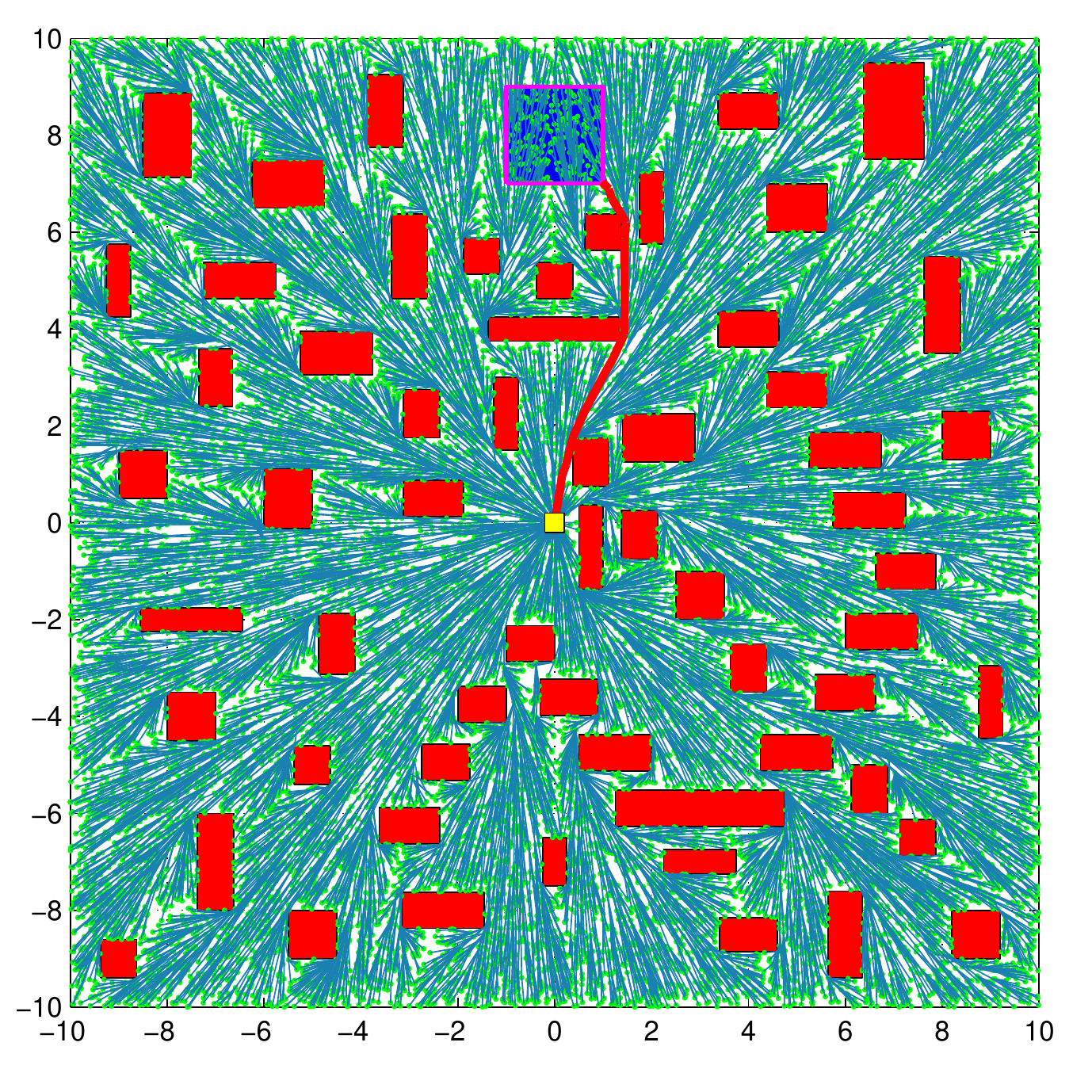}} \label{figure:pt3_rrtstar_it24999}}
    }

	\mbox{
    \subfigure[]{\scalebox{0.28}{\includegraphics[trim = 4.0cm 6.937cm 3.587cm 7.0cm, clip =
          true]{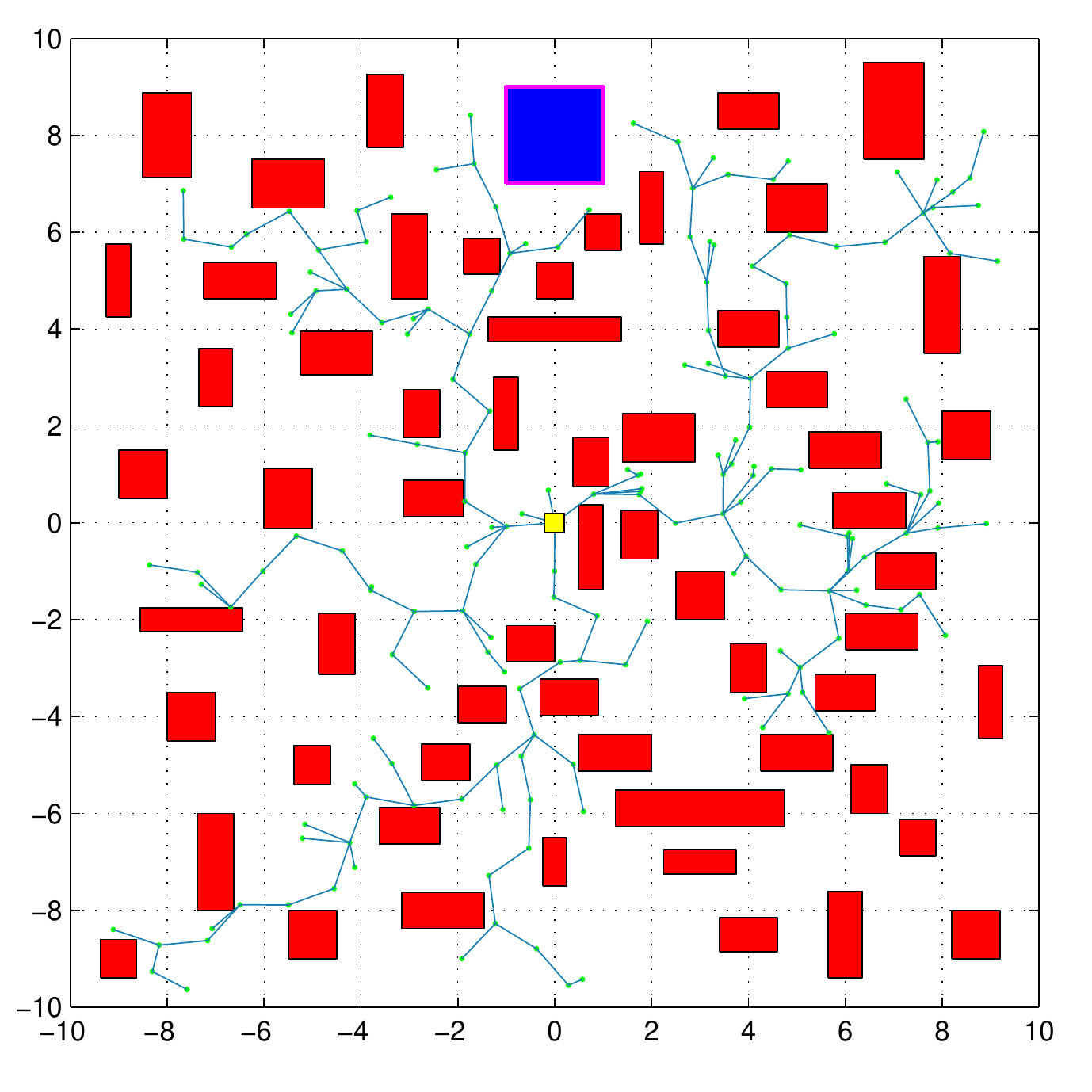}} \label{figure:pt3_rrtsharp_v0_it250}}
    \subfigure[]{\scalebox{0.28}{\includegraphics[trim = 4.0cm 6.937cm 3.587cm 7.0cm, clip =
          true]{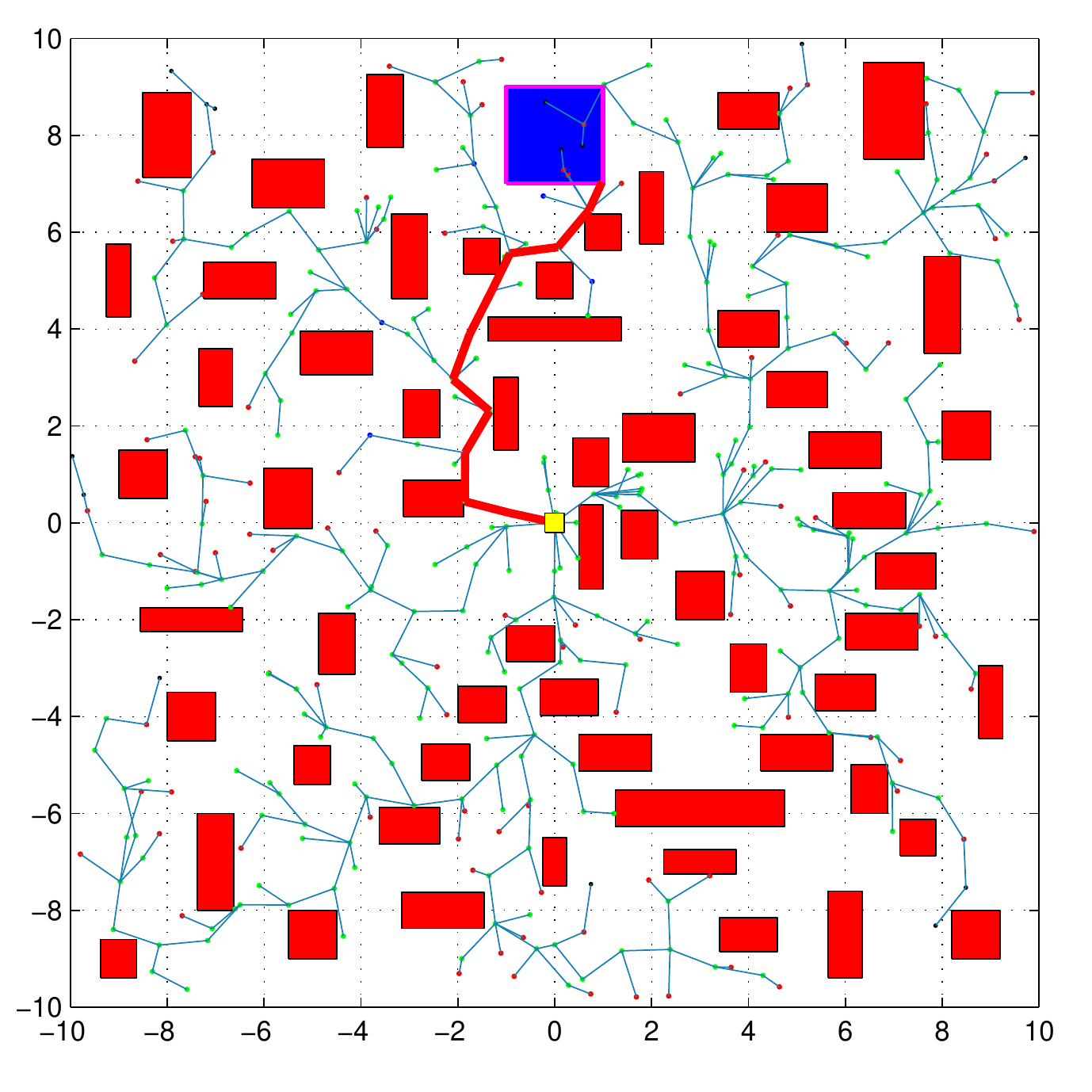}} \label{figure:pt3_rrtsharp_v0_it500}}
    \subfigure[]{\scalebox{0.28}{\includegraphics[trim = 4.0cm 6.937cm 3.587cm 7.0cm, clip =
          true]{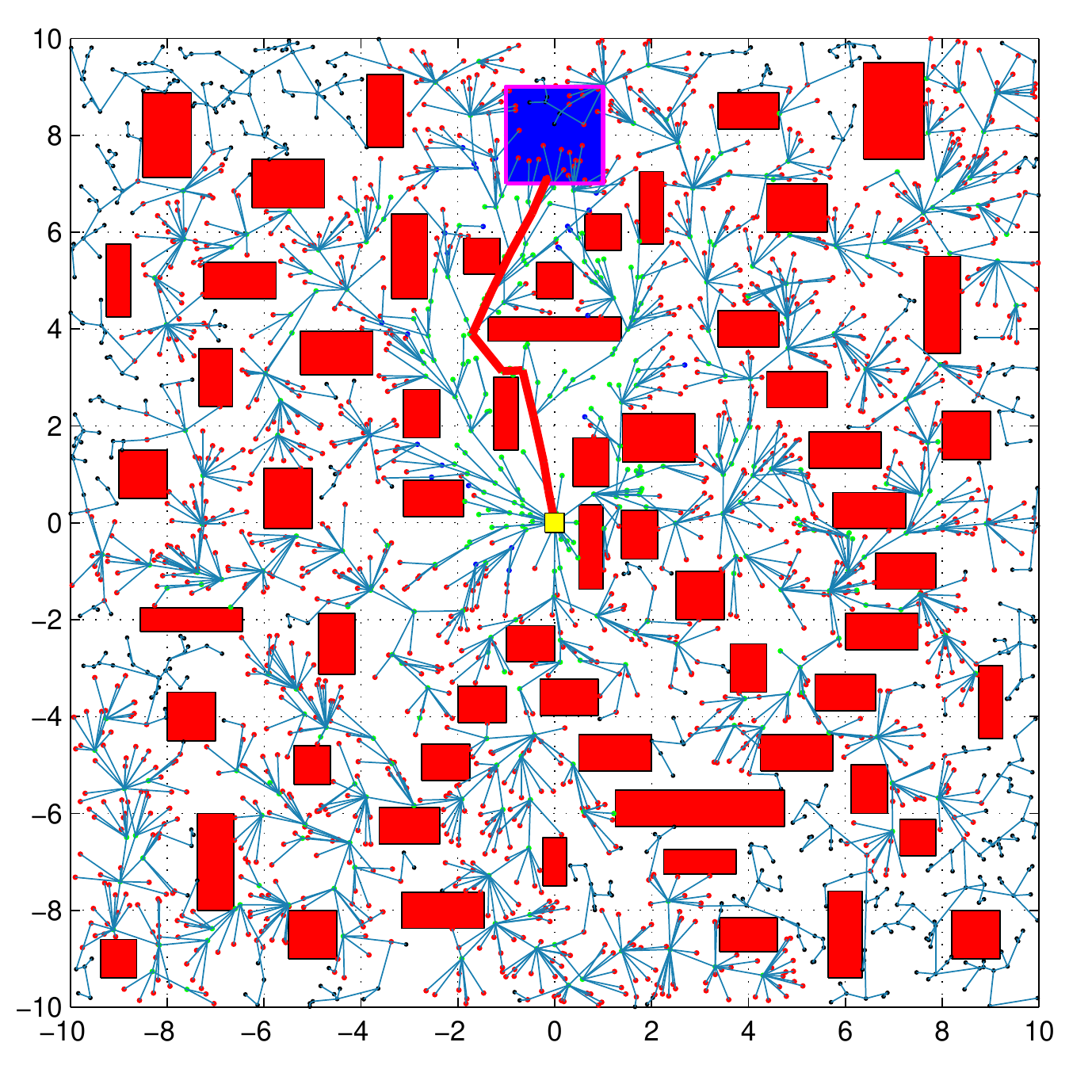}} \label{figure:pt3_rrtsharp_v0_it2500}}
    \subfigure[]{\scalebox{0.28}{\includegraphics[trim = 4.0cm 6.937cm 3.587cm 7.0cm, clip =
          true]{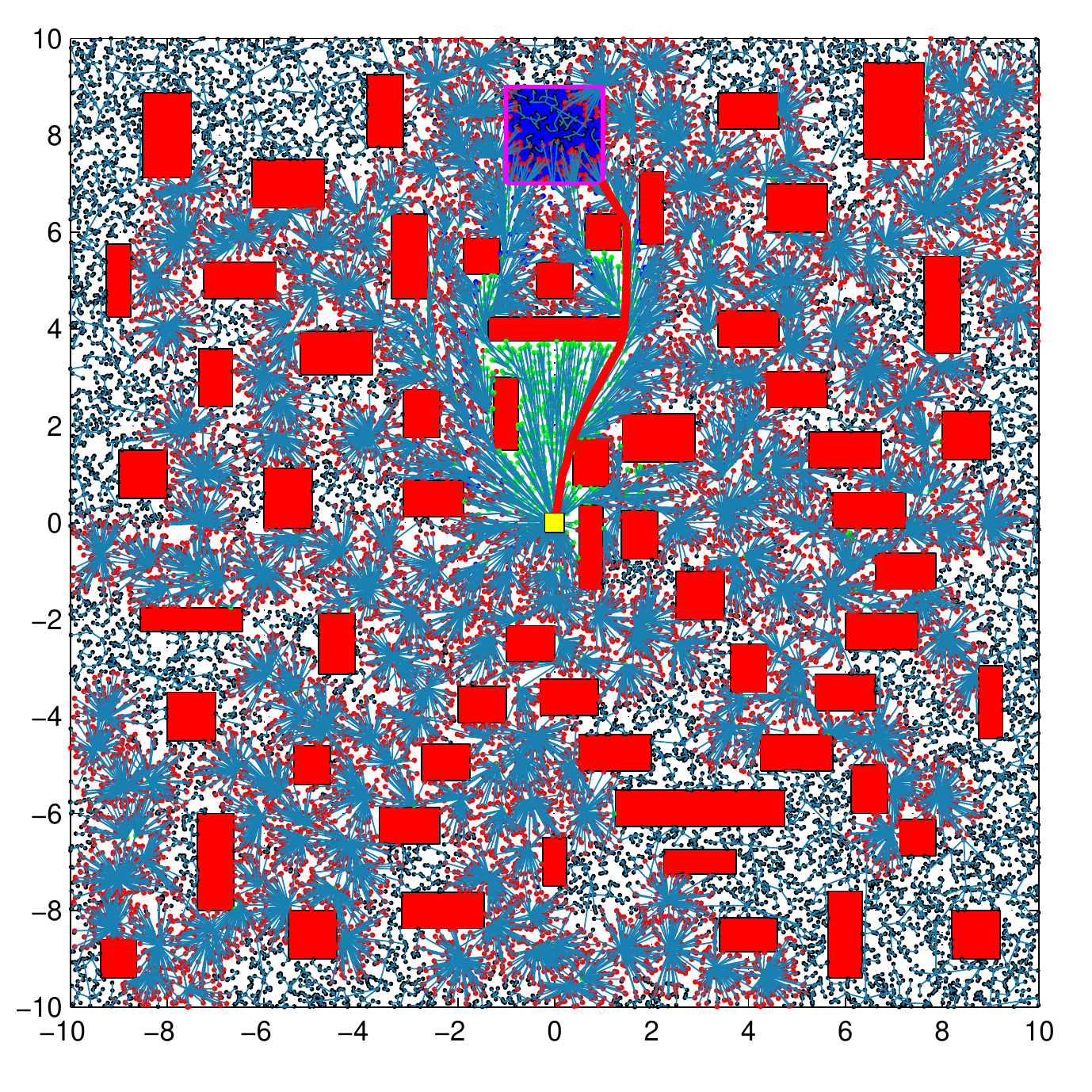}} \label{figure:pt3_rrtsharp_v0_it24999}}
   	}


    \caption{The evolution of the tree computed by \AlgRRTstar{} and \AlgRRTsharp{} algorithms is shown in  \subref{figure:pt3_rrtstar_it250}-\subref{figure:pt3_rrtstar_it24999} and \subref{figure:pt3_rrtsharp_v0_it250}-\subref{figure:pt3_rrtsharp_v0_it24999}, respectively. The configuration of the trees \subref{figure:pt3_rrtstar_it250}, \subref{figure:pt3_rrtsharp_v0_it250} is at 250 iterations, \subref{figure:pt3_rrtstar_it500}, \subref{figure:pt3_rrtsharp_v0_it500} is at 500 iterations, \subref{figure:pt3_rrtstar_it2500}, \subref{figure:pt3_rrtsharp_v0_it2500} is at 2500 iterations, 
    and \subref{figure:pt3_rrtstar_it24999}, \subref{figure:pt3_rrtsharp_v0_it24999} is at 25000 iterations.}
    \label{figure:sim_d2_pt3_rrtstar_rrtsharp_v0_iterations}
  \end{center}
\end{figure*}

In the third problem type, both algorithms were run in a more cluttered environment, where there are many different homotopy classes containing the local minimum solution for the problem. As shown in Figure~\ref{figure:sim_d2_pt3_rrtstar_rrtsharp_v0_iterations}, both algorithms switch between paths which have locally best cost, eventually converging to the optimal solution.

\begin{figure*}[htp]
  \begin{center}
	\mbox{
        \subfigure[]{\scalebox{0.26}{\includegraphics[trim = 4.0cm 3.0cm 4.0cm 3.0cm, clip =
          true]{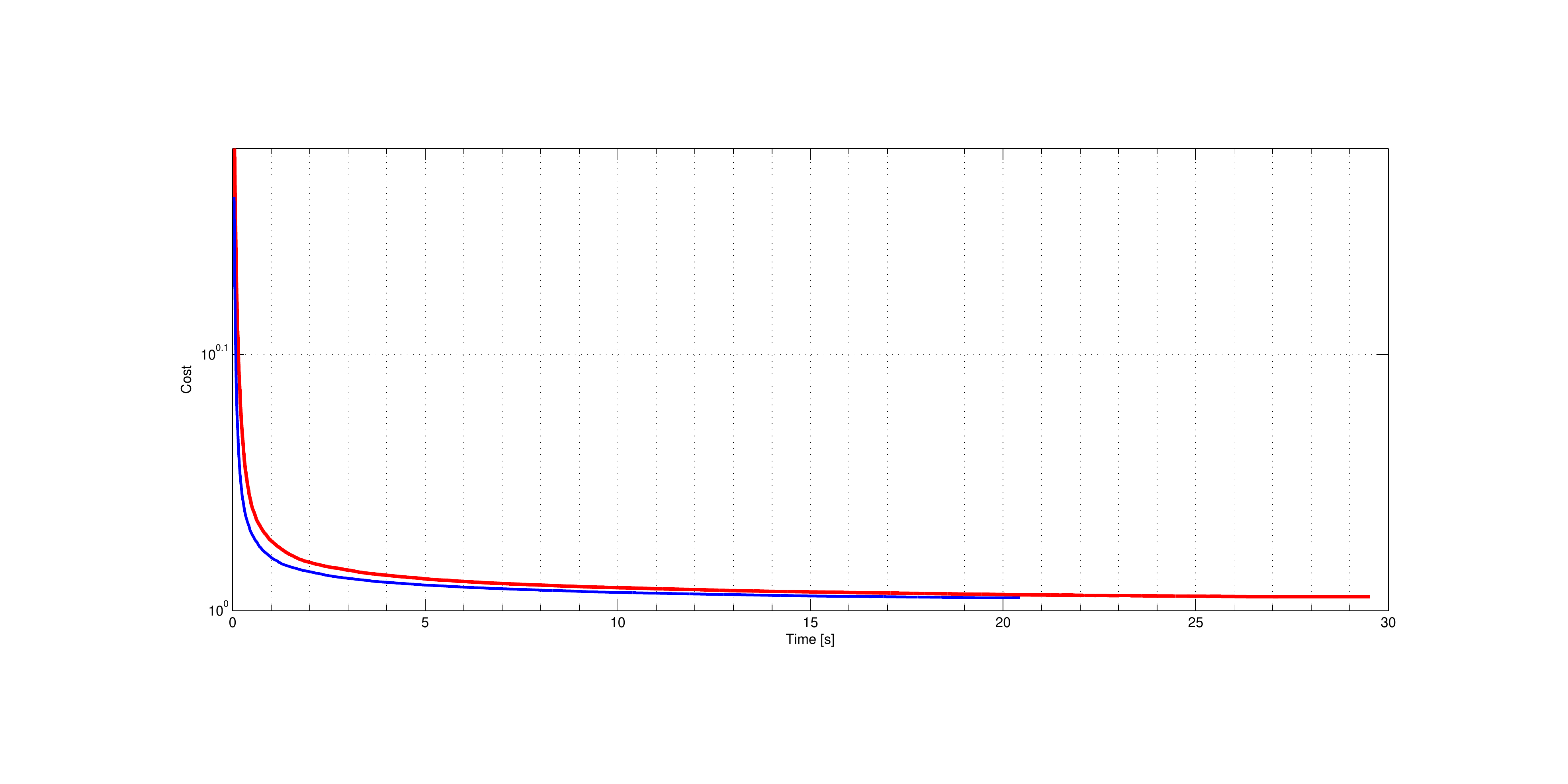}}\label{figure:time_cost_mean_d2_pt3_rrtstar_rrtsharp_v0}}

       	\subfigure[]{\scalebox{0.26}{\includegraphics[trim = 4.0cm 3.0cm 4.0cm 3.0cm, clip =
          true]{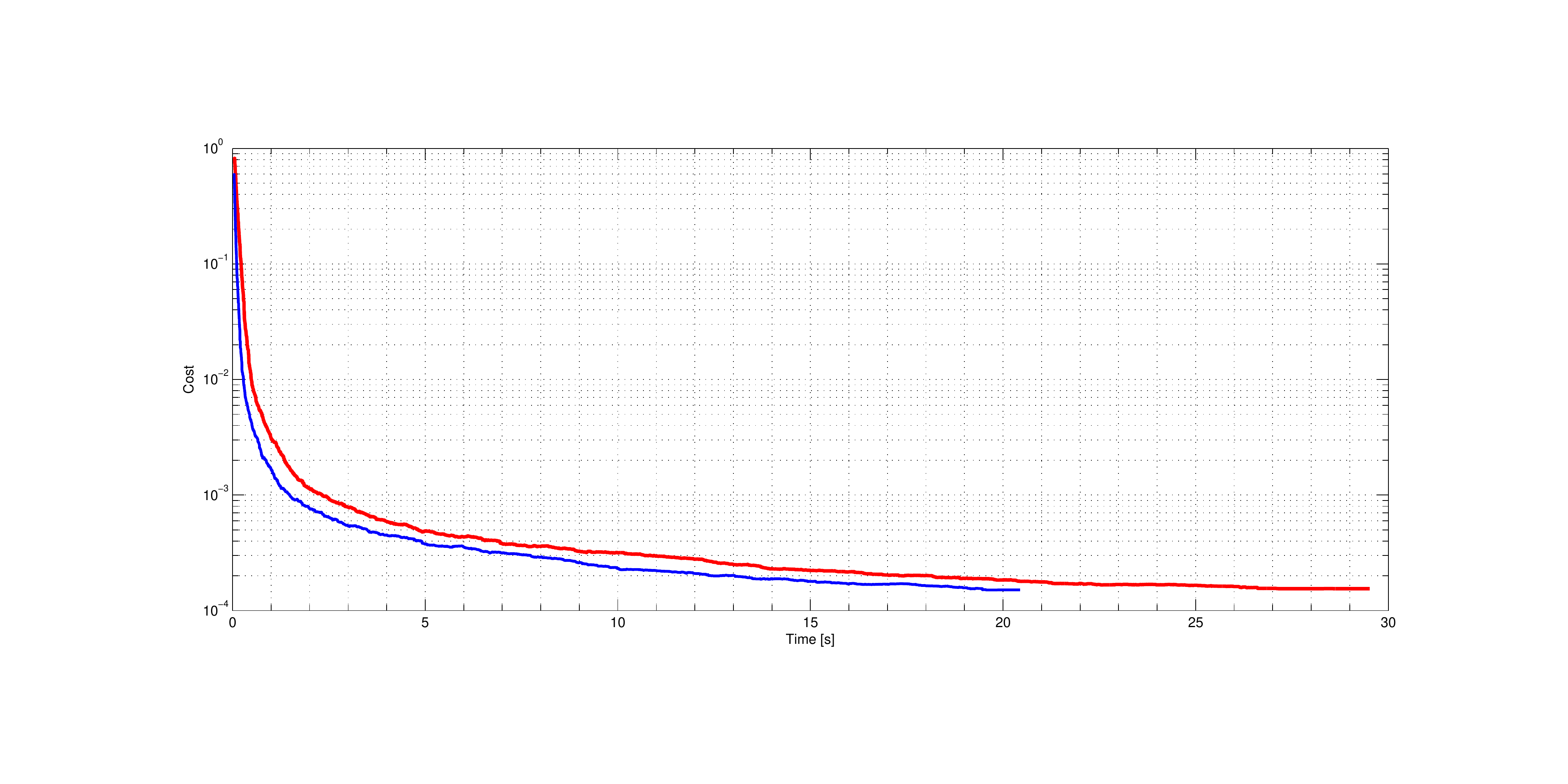}}\label{figure:time_cost_variance_d2_pt3_rrtstar_rrtsharp_v0}}

    }


    \caption{The change in the cost of the best paths computed by \AlgRRTstar{} and \AlgRRTsharp{} algorithms and the variance in the trials are shown in \subref{figure:time_cost_mean_d2_pt3_rrtstar_rrtsharp_v0} and \subref{figure:time_cost_variance_d2_pt3_rrtstar_rrtsharp_v0}, respectively.}
    \label{figure:sim_d2_pt3_rrtstar_rrtsharp_v0_histories}
  \end{center}
\end{figure*}

\FloatBarrier

\begin{figure*}[htp]
  \begin{center}

	\mbox{
    \subfigure[]{\scalebox{0.28}{\includegraphics[trim = 4.0cm 6.937cm 3.587cm 7.0cm, clip =
          true]{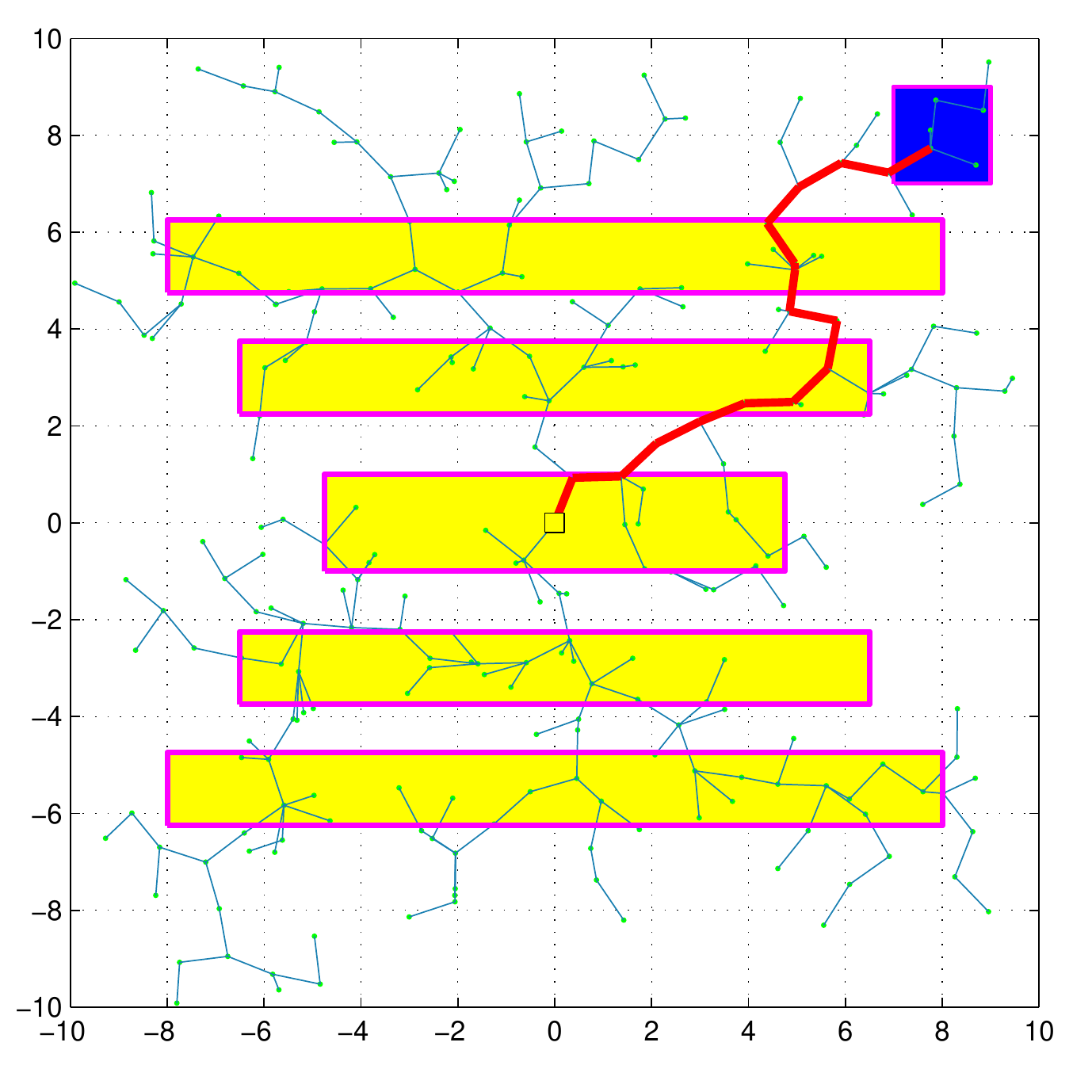}} \label{figure:pt4_rrtstar_it250}} 		
    \subfigure[]{\scalebox{0.28}{\includegraphics[trim = 4.0cm 6.937cm 3.587cm 7.0cm, clip =
          true]{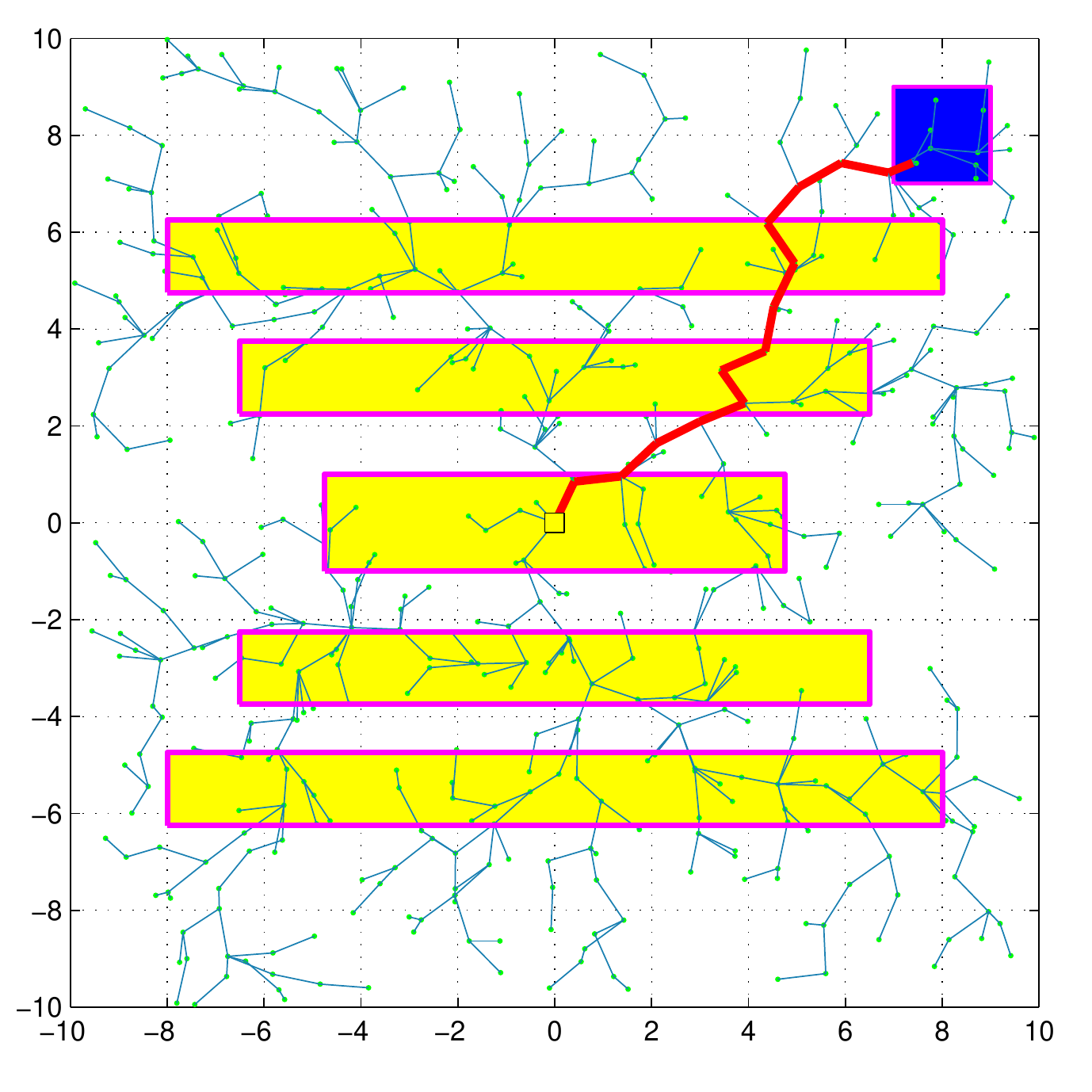}} \label{figure:pt4_rrtstar_it500}}
    \subfigure[]{\scalebox{0.28}{\includegraphics[trim = 4.0cm 6.937cm 3.587cm 7.0cm, clip =
          true]{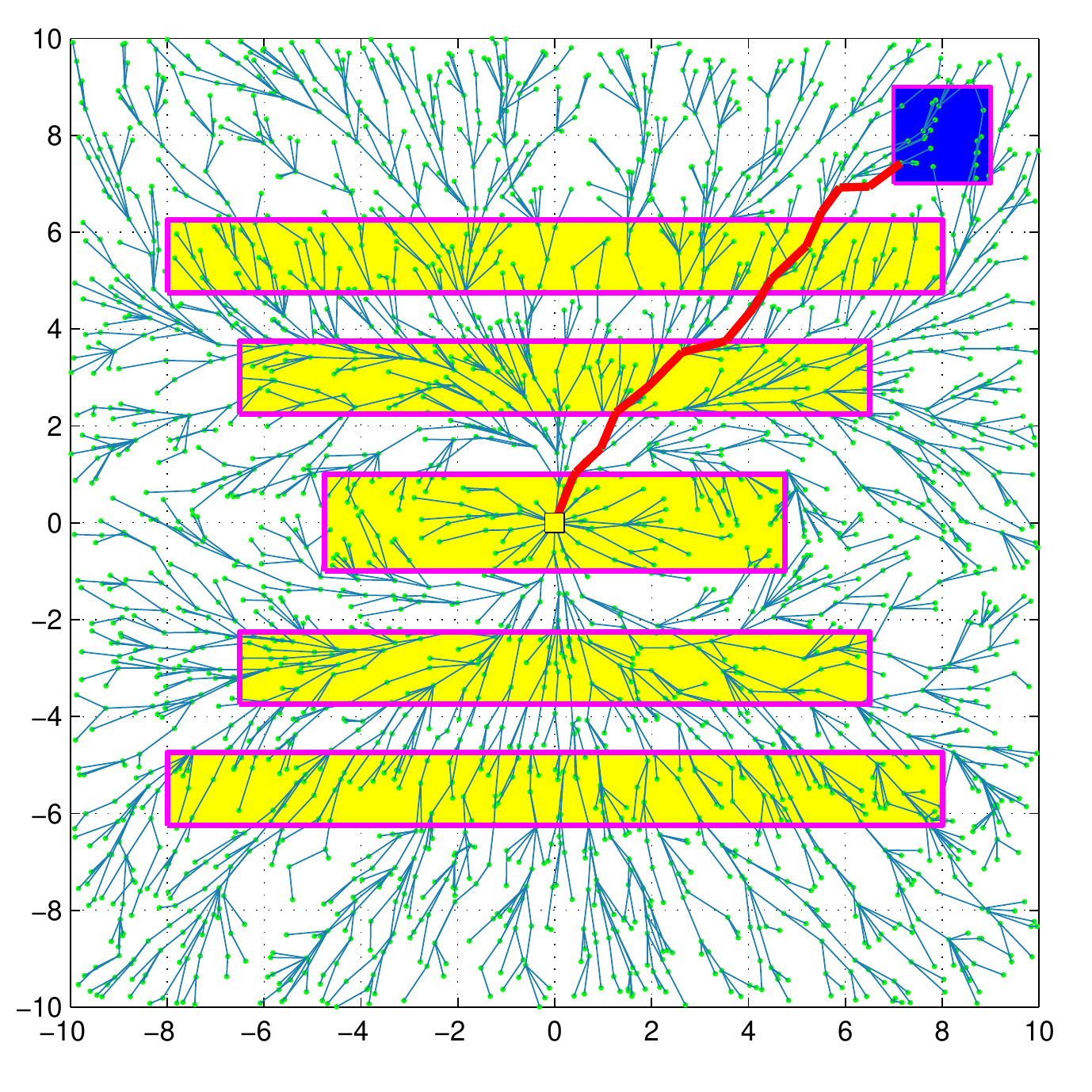}} \label{figure:pt4_rrtstar_it2500}}
    \subfigure[]{\scalebox{0.28}{\includegraphics[trim = 4.0cm 6.937cm 3.587cm 7.0cm, clip =
          true]{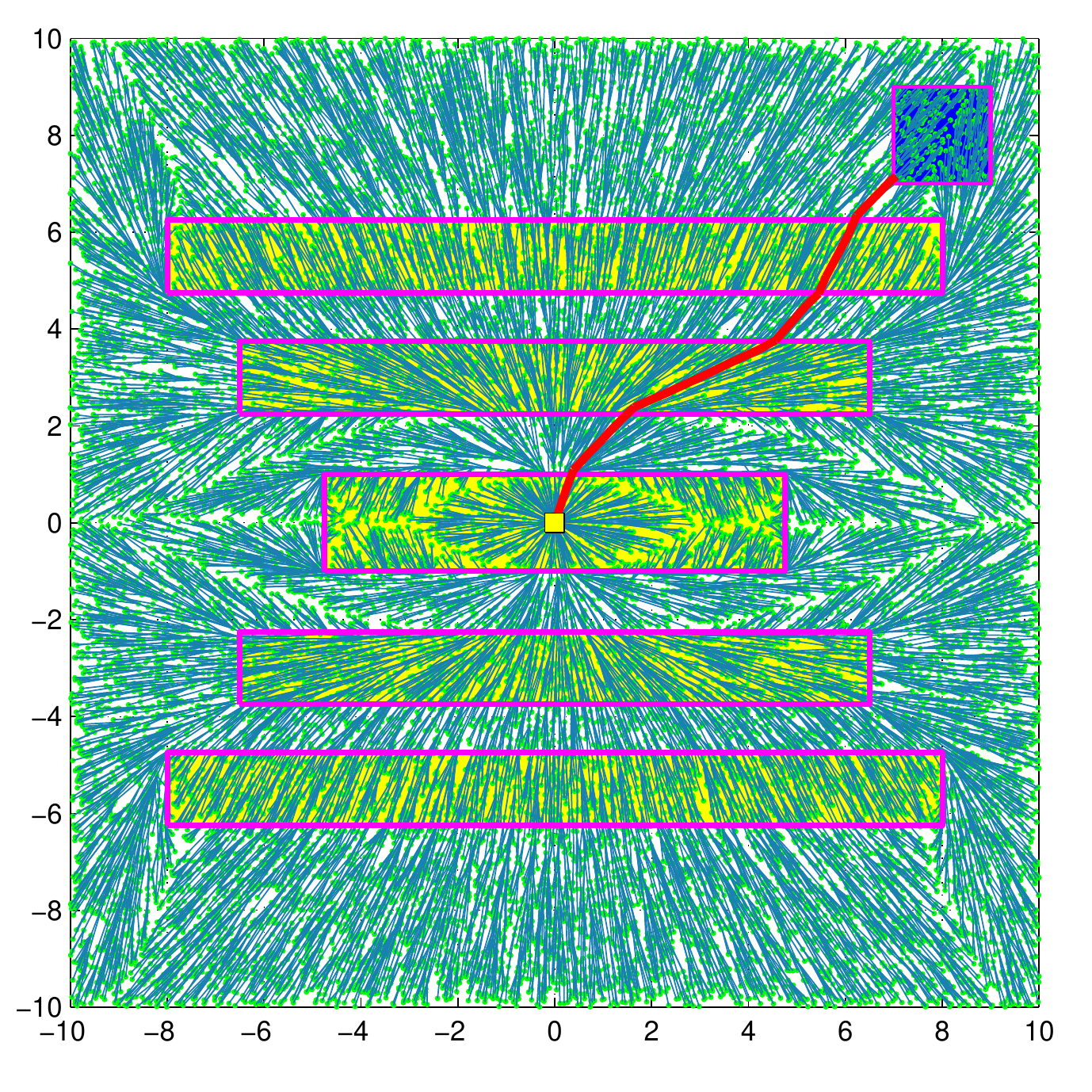}} \label{figure:pt4_rrtstar_it24999}}
    }

	\mbox{
    \subfigure[]{\scalebox{0.28}{\includegraphics[trim = 4.0cm 6.937cm 3.587cm 7.0cm, clip =
          true]{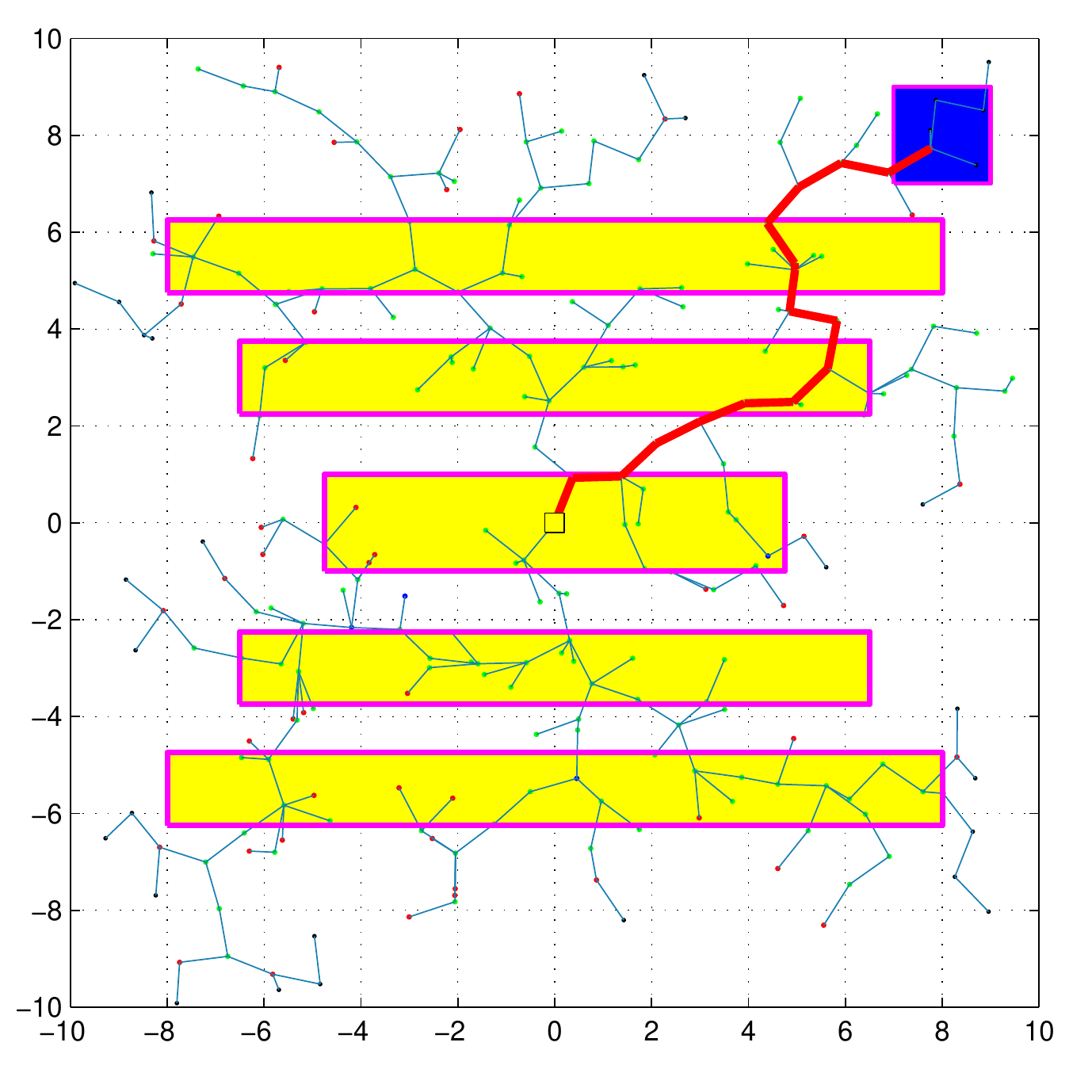}} \label{figure:pt4_rrtsharp_v0_it250}}
    \subfigure[]{\scalebox{0.28}{\includegraphics[trim = 4.0cm 6.937cm 3.587cm 7.0cm, clip =
          true]{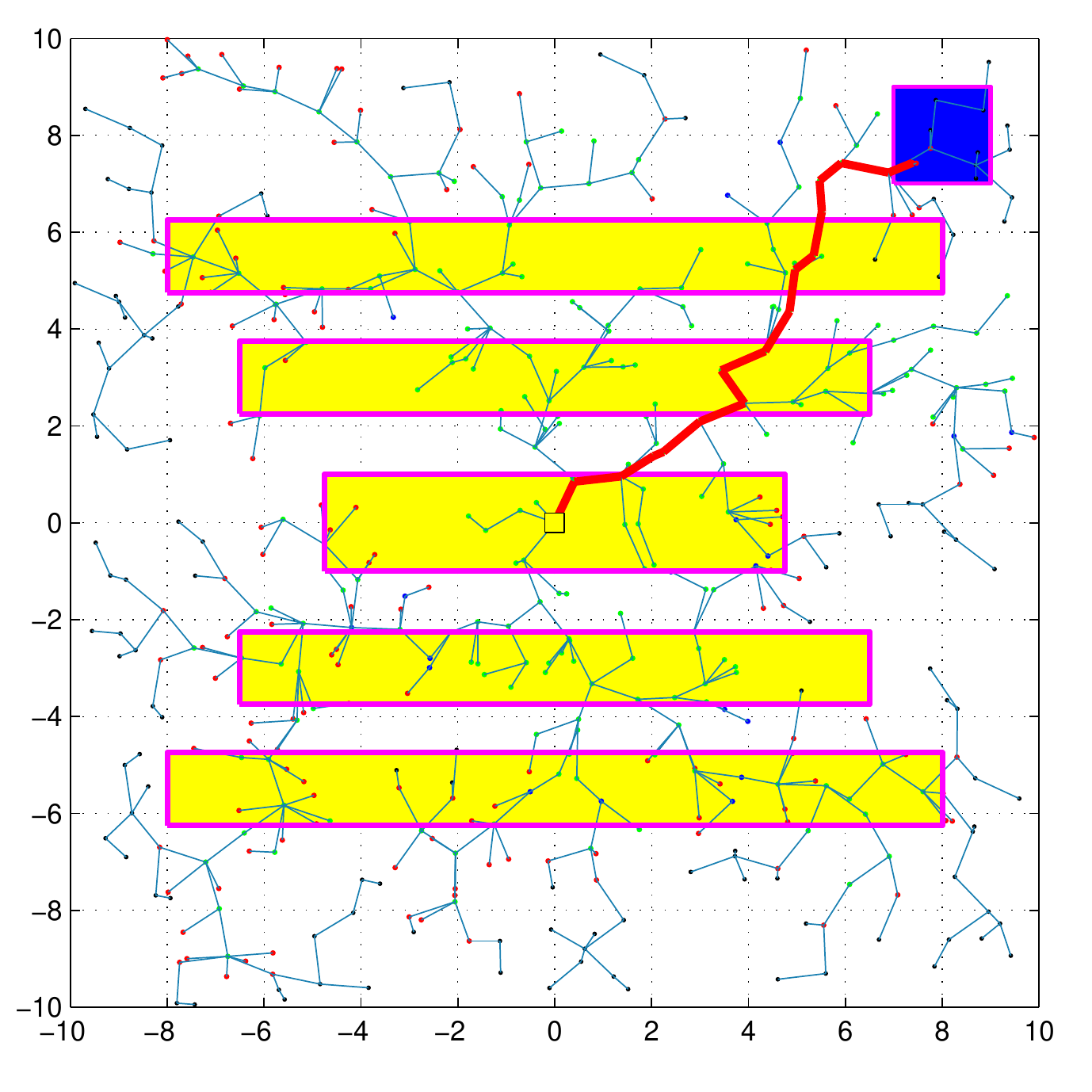}} \label{figure:pt4_rrtsharp_v0_it500}}
    \subfigure[]{\scalebox{0.28}{\includegraphics[trim = 4.0cm 6.937cm 3.587cm 7.0cm, clip =
          true]{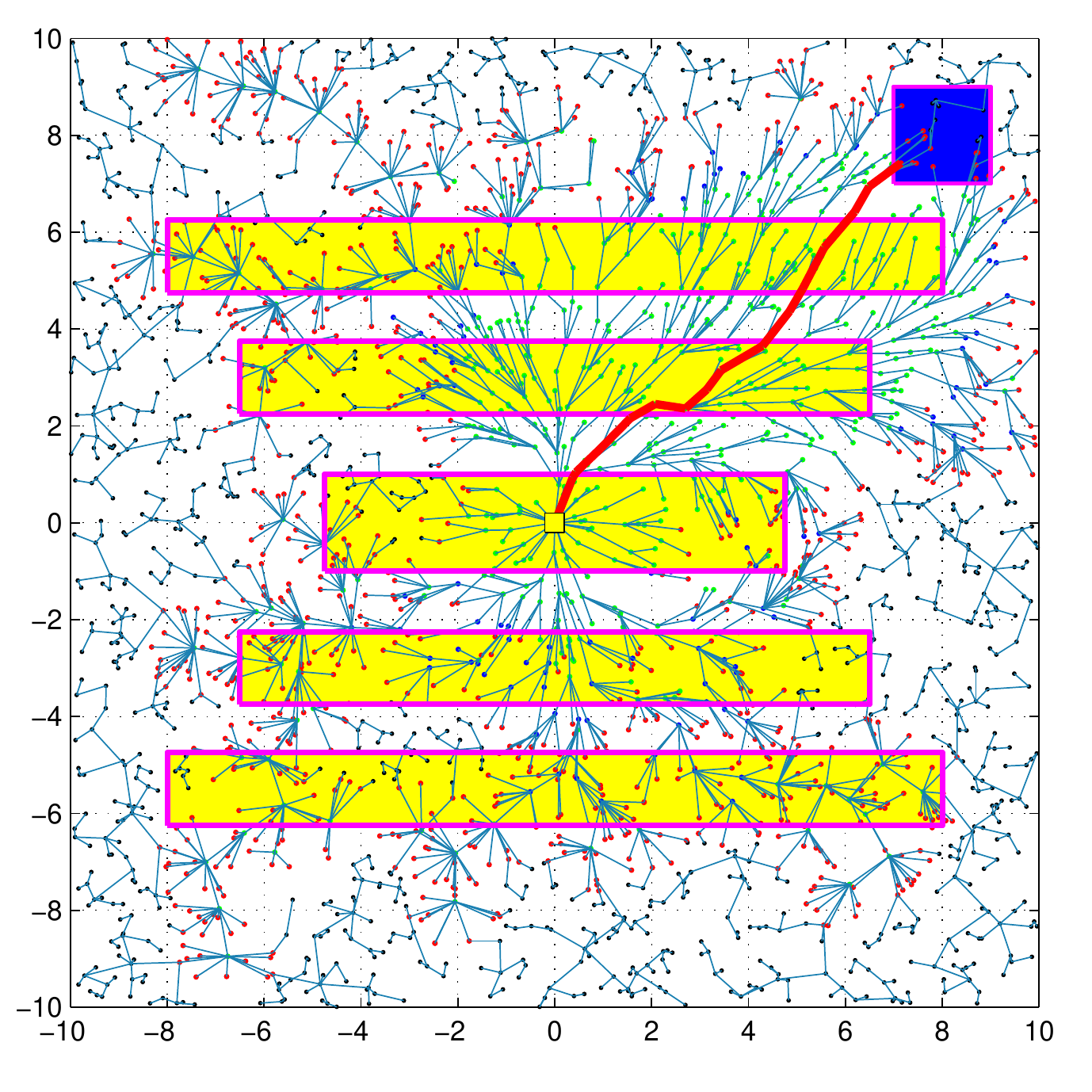}} \label{figure:pt4_rrtsharp_v0_it2500}}
    \subfigure[]{\scalebox{0.28}{\includegraphics[trim = 4.0cm 6.937cm 3.587cm 7.0cm, clip =
          true]{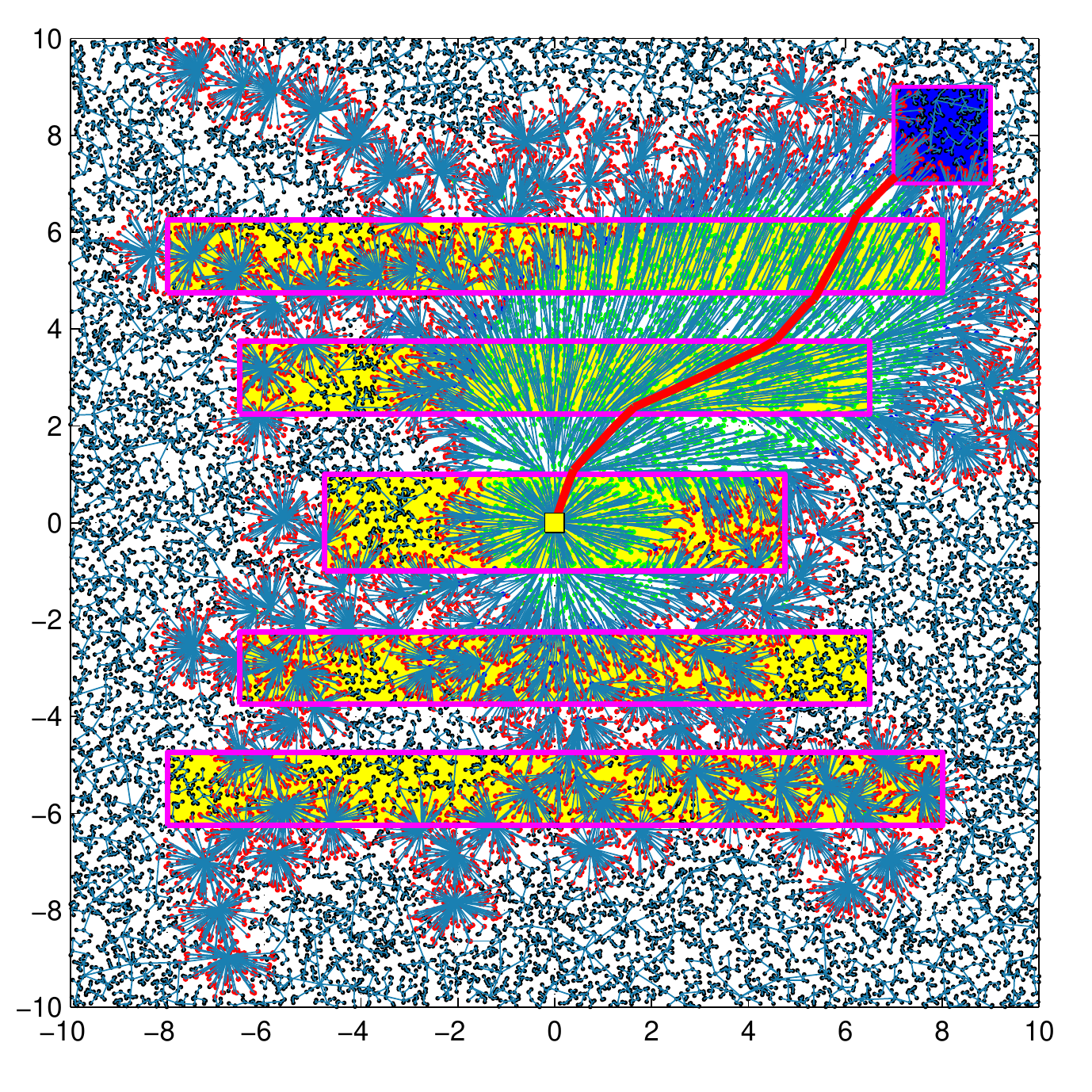}} \label{figure:pt4_rrtsharp_v0_it24999}}
   	}


\caption{The evolution of the tree computed by \AlgRRTstar{} and \AlgRRTsharp{} algorithms is shown in \subref{figure:pt4_rrtstar_it250}-\subref{figure:pt4_rrtstar_it24999} and \subref{figure:pt4_rrtsharp_v0_it250}-\subref{figure:pt4_rrtsharp_v0_it24999}, respectively. The configuration of the trees \subref{figure:pt4_rrtstar_it250}, \subref{figure:pt4_rrtsharp_v0_it250} is at 250 iterations, \subref{figure:pt4_rrtstar_it500}, \subref{figure:pt4_rrtsharp_v0_it500} is at 500 iterations, \subref{figure:pt4_rrtstar_it2500}, \subref{figure:pt4_rrtsharp_v0_it2500} is at 2500 iterations, 
and \subref{figure:pt4_rrtstar_it24999}, \subref{figure:pt4_rrtsharp_v0_it24999} is at 25000 iterations.}
    \label{figure:sim_d2_pt4_rrtstar_rrtsharp_v0_iterations}
  \end{center}
\end{figure*}

Finally, in the fourth problem type, both algorithms were run in a obstacle-free environment where there are different cost zones. The cost coefficient of each zone from top to bottom is 1.5, 0.75, 2.5, 0.75, and 1.5, respectively and 1 elsewhere. As seen in Figure~\ref{figure:sim_d2_pt4_rrtstar_rrtsharp_v0_iterations}, both algorithms compute the optimal path which has longer segments in low-cost zones.

\begin{figure*}[htp]
  \begin{center}
	\mbox{
        \subfigure[]{\scalebox{0.26}{\includegraphics[trim = 4.0cm 3.0cm 4.0cm 3.0cm, clip =
          true]{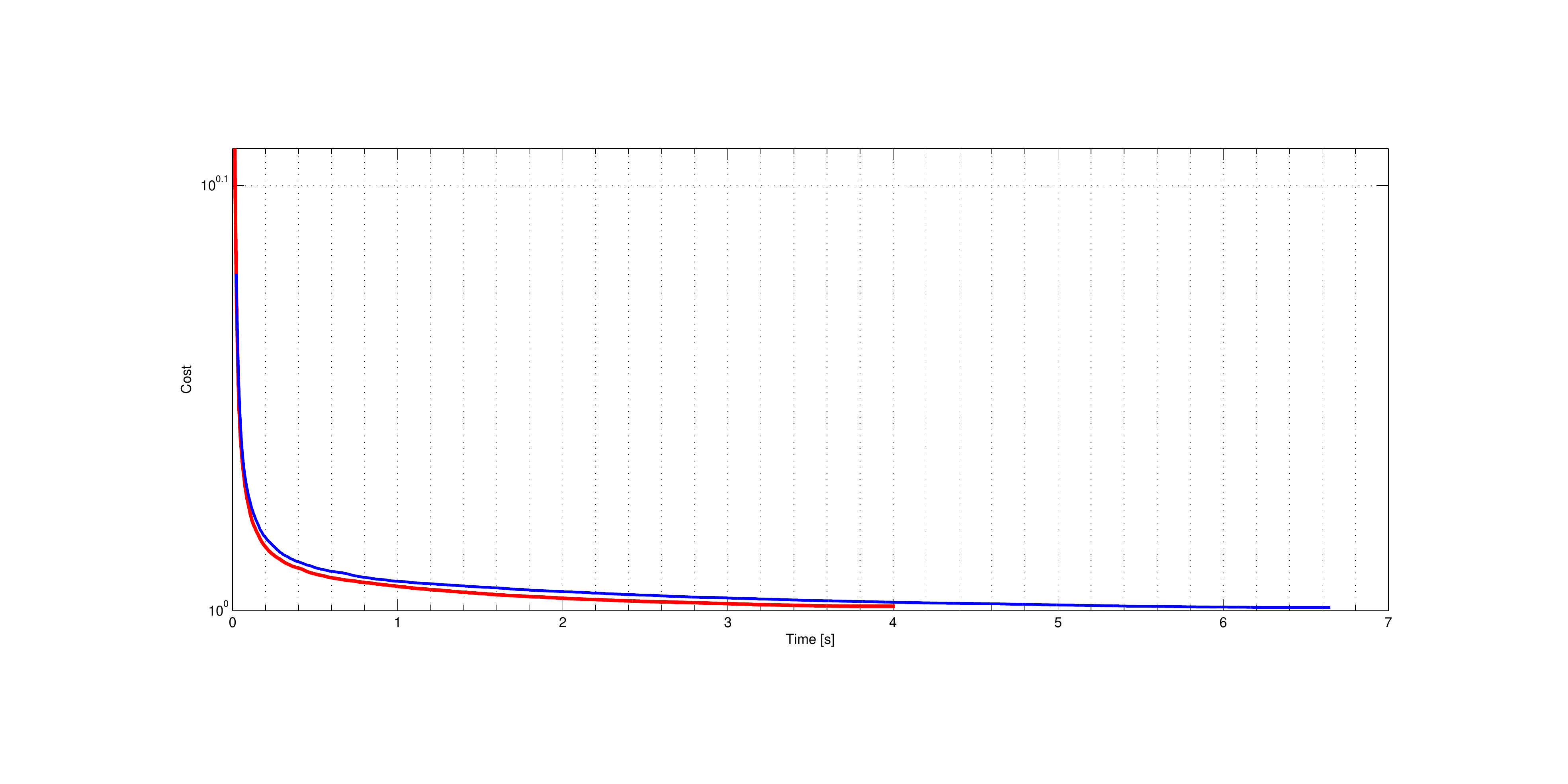}}\label{figure:time_cost_mean_d2_pt4_rrtstar_rrtsharp_v0}}

        \subfigure[]{\scalebox{0.26}{\includegraphics[trim = 4.0cm 3.0cm 4.0cm 3.0cm, clip =
          true]{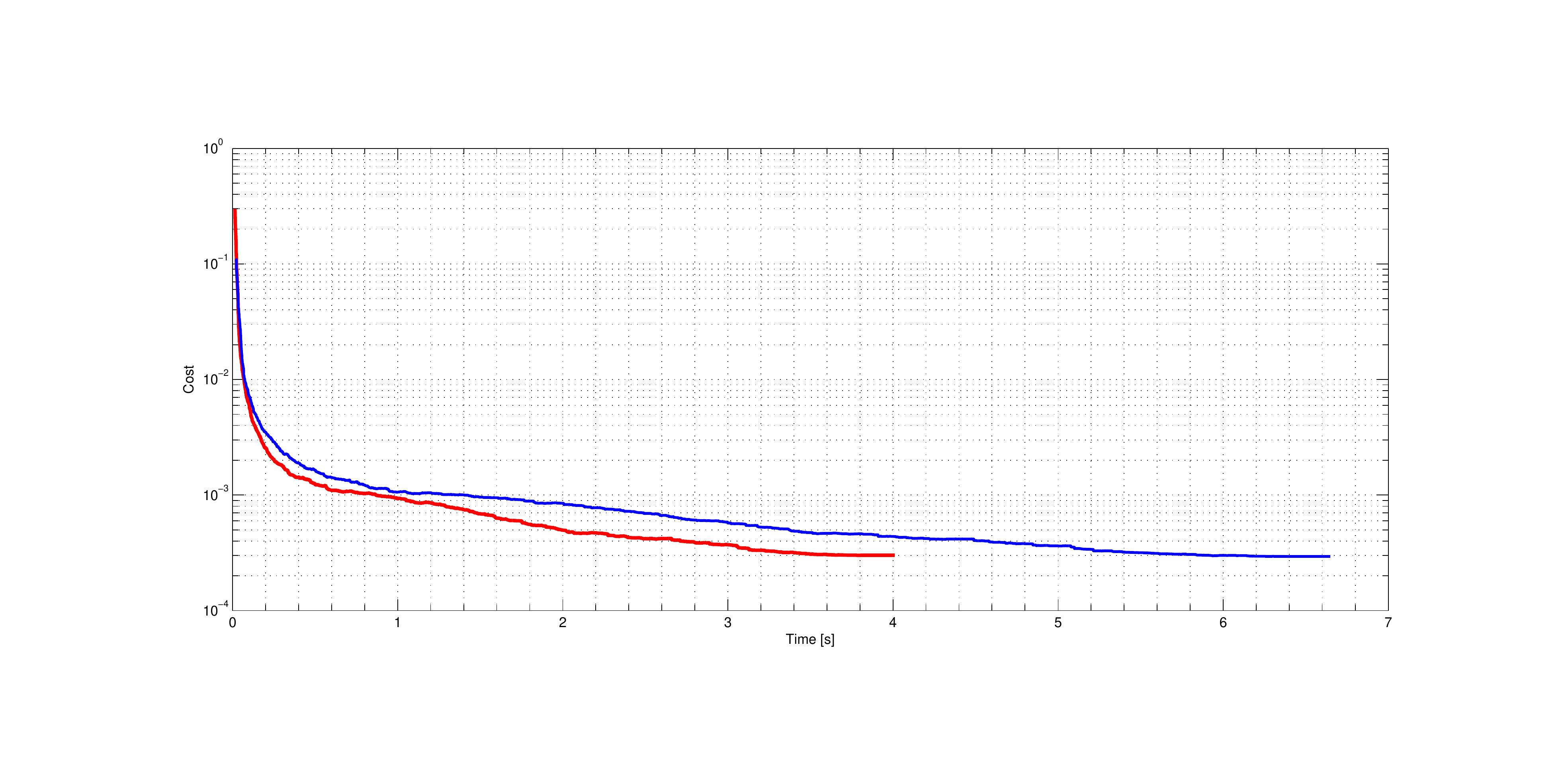}}\label{figure:time_cost_variance_d2_pt4_rrtstar_rrtsharp_v0}}

    }


    \caption{The change in the cost of the best paths computed by \AlgRRTstar{} and \AlgRRTsharp{} algorithms, and the variance in the trials are shown in \subref{figure:time_cost_mean_d2_pt4_rrtstar_rrtsharp_v0} and \subref{figure:time_cost_variance_d2_pt4_rrtstar_rrtsharp_v0}, respectively.}
    \label{figure:sim_d2_pt4_rrtstar_rrtsharp_v0_histories}
  \end{center}
\end{figure*}

\FloatBarrier

\section{Variants of the \AlgRRTsharp{} Algorithm}

Too many non-promising vertices are included in the tree computed by the \AlgRRTsharp{} algorithm as observed in the previous simulations. This is owing to the fact that the \AlgRRTsharp{} algorithm includes all new vertices in the graph regardless of their type. A simple vertex selection criterion can be used in the $\PrcExtend$ procedure in order to prevent the algorithm from growing the tree towards the region outside $\Xrel$. However, being over-selective on vertex inclusion may degrade the performance of the algorithm~--~and thus lead to a suboptimal solution~--~since the cost-to-come value of all vertices, which is used to decide if a new vertex is promising or not, is an estimate of the optimal one. In this section, we propose three variants of the baseline \AlgRRTsharp{} algorithm.

\begin{description}

\item [\AlgRRTsharpNoBlackVertex]: In the first variant, which is given in Algorithm~\ref{alg:extend_rrtsharp_v1}, if a new vertex happens to be consistent with infinite key value (black vertex), it is not included in the graph. This situation can happen if all of the neighbor vertices of the new vertex happen to be inconsistent with infinite g-value and finite lmc-value (red vertices). First, the estimates of the cost-to-come-value of the new vertex $\xnew$ are initialized with infinite cost, and its parent vertex is set to `null' in Line 6. Then, a better value for the lmc-value of the new vertex is searched among its neighbor vertices. During this search, the parent of the new vertex remains unassigned only if there are no any neighboring vertices with finite g-value.

\item [\AlgRRTsharpPromisingParent]: In the second variant, the algorithm becomes more selective on vertices to be added to the graph and the ``$\PrcParent(\xnew) \not= \emptyset ~\wedge~ \PrcKey(\PrcParent(\xnew)) \prec \PrcKey(\xmingoal)$'' condition is checked in Line~~\ref{line:node_inclusion_condition}. Simply, a new vertex is included to the graph only if its parent is a promising vertex.

\item [\AlgRRTsharpPromisingNewVertex]: Lastly, the third variant is most selective on vertex for inclusion and $\PrcKey(\xnew) \prec \PrcKey(\xmingoal)$ condition is checked, that is, only promising new vertices are included in the graph.

\end{description}

\input{extend_rrtsharp_v1.tex}

\FloatBarrier

\section{Numerical Simulations 2}

\begin{figure*}[htp]

  \begin{center}
  	\mbox{
	\setcounter{subfigure}{0}
	\renewcommand{\thesubfigure}{(\alph{subfigure})}
	\subfigure[]{\scalebox{0.28}{\includegraphics[trim = 4.0cm 6.937cm 3.587cm 7.0cm, clip =
          true]{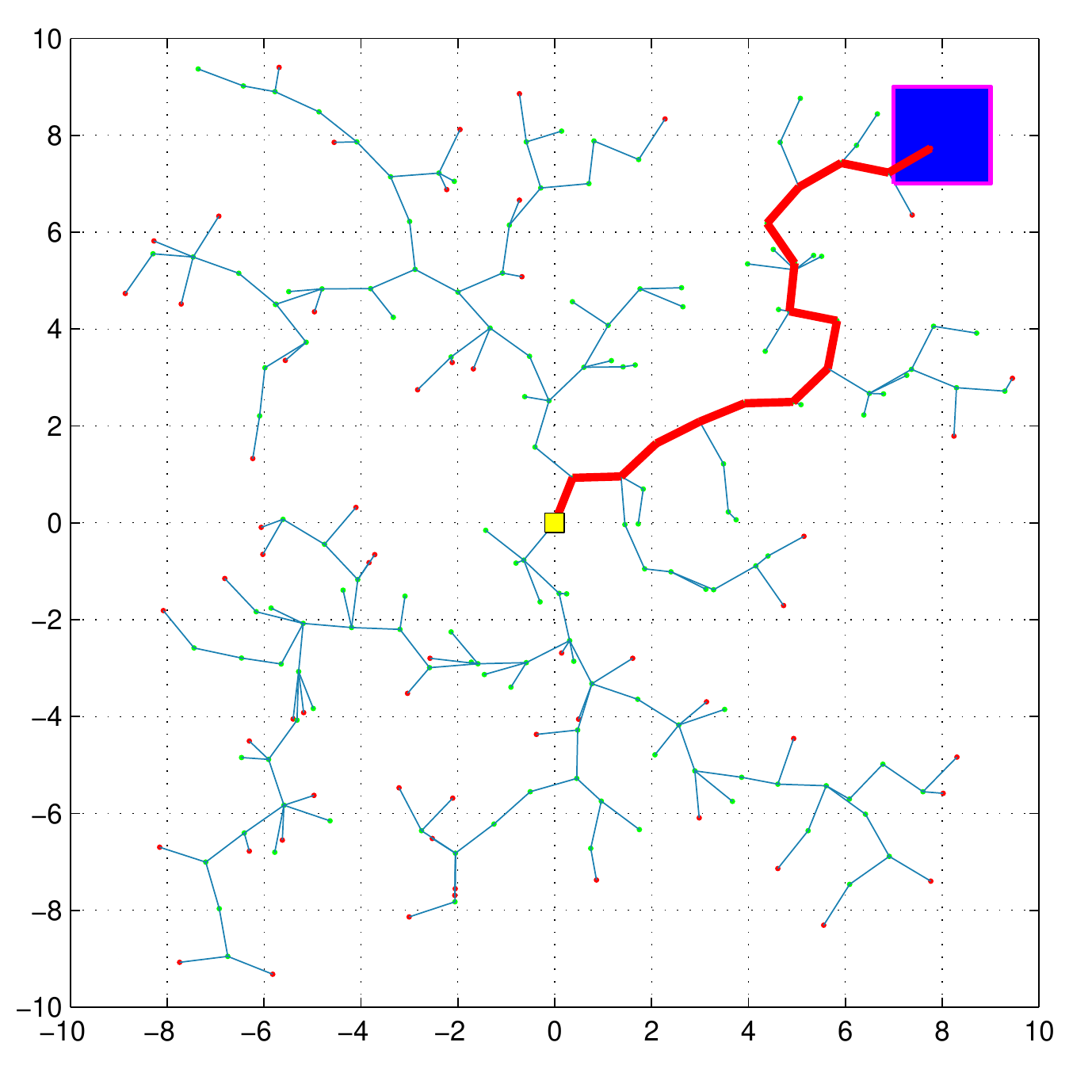}} \label{figure:pt1_rrtsharp_v1_it250}}
    \subfigure[]{\scalebox{0.28}{\includegraphics[trim = 4.0cm 6.937cm 3.587cm 7.0cm, clip =
          true]{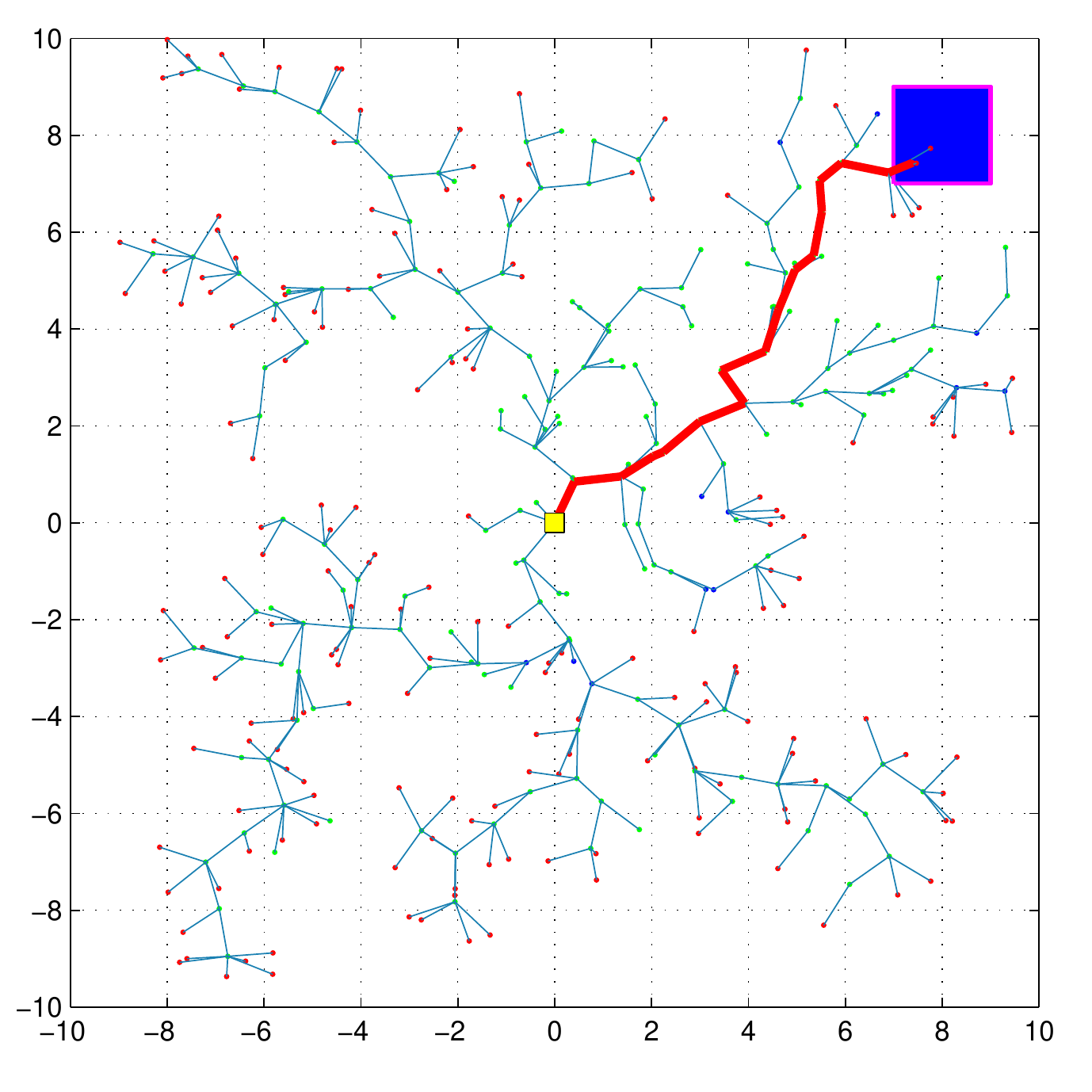}} \label{figure:pt1_rrtsharp_v1_it500}}
    \subfigure[]{\scalebox{0.28}{\includegraphics[trim = 4.0cm 6.937cm 3.587cm 7.0cm, clip =
          true]{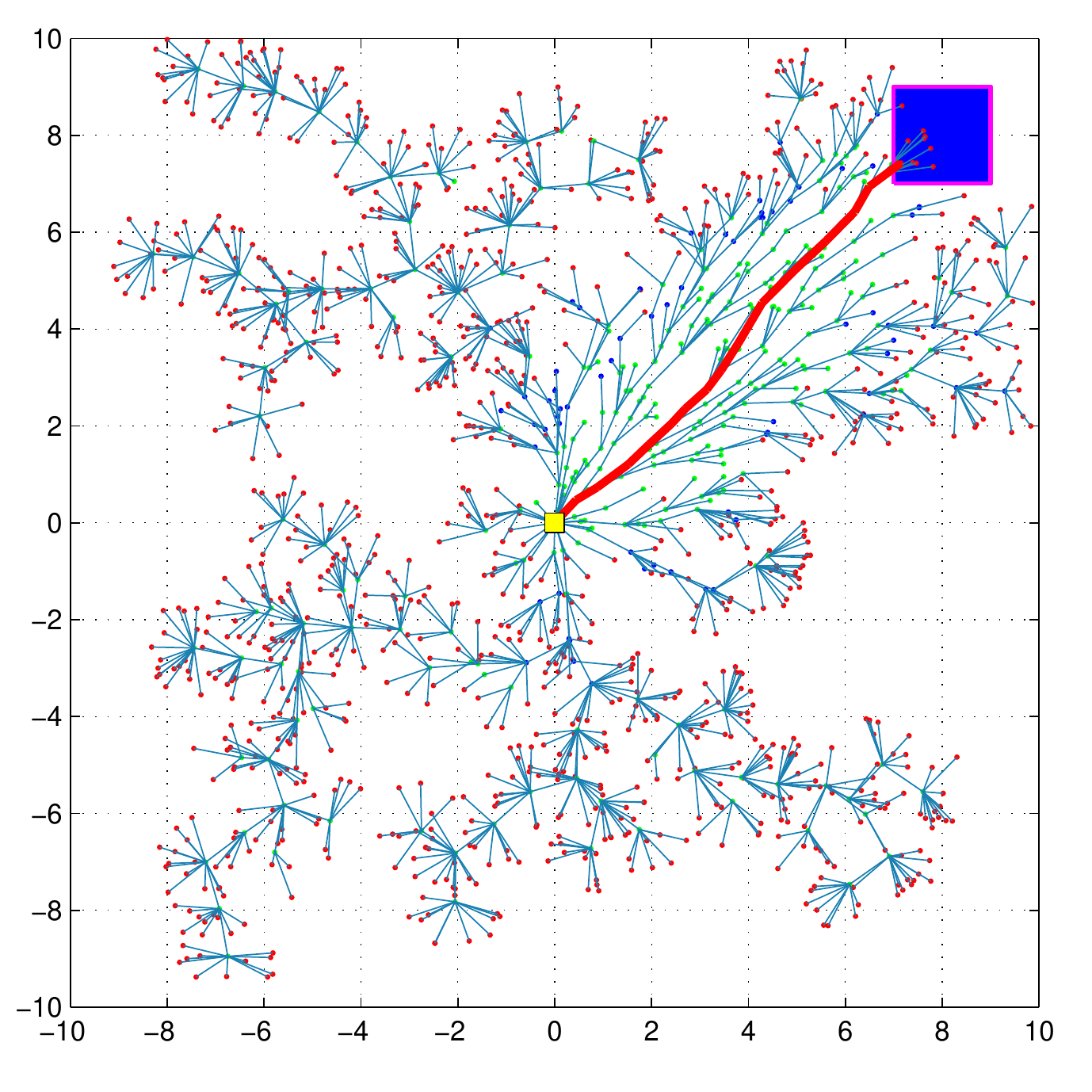}} \label{figure:pt1_rrtsharp_v1_it2500}}
    \subfigure[]{\scalebox{0.28}{\includegraphics[trim = 4.0cm 6.937cm 3.587cm 7.0cm, clip =
          true]{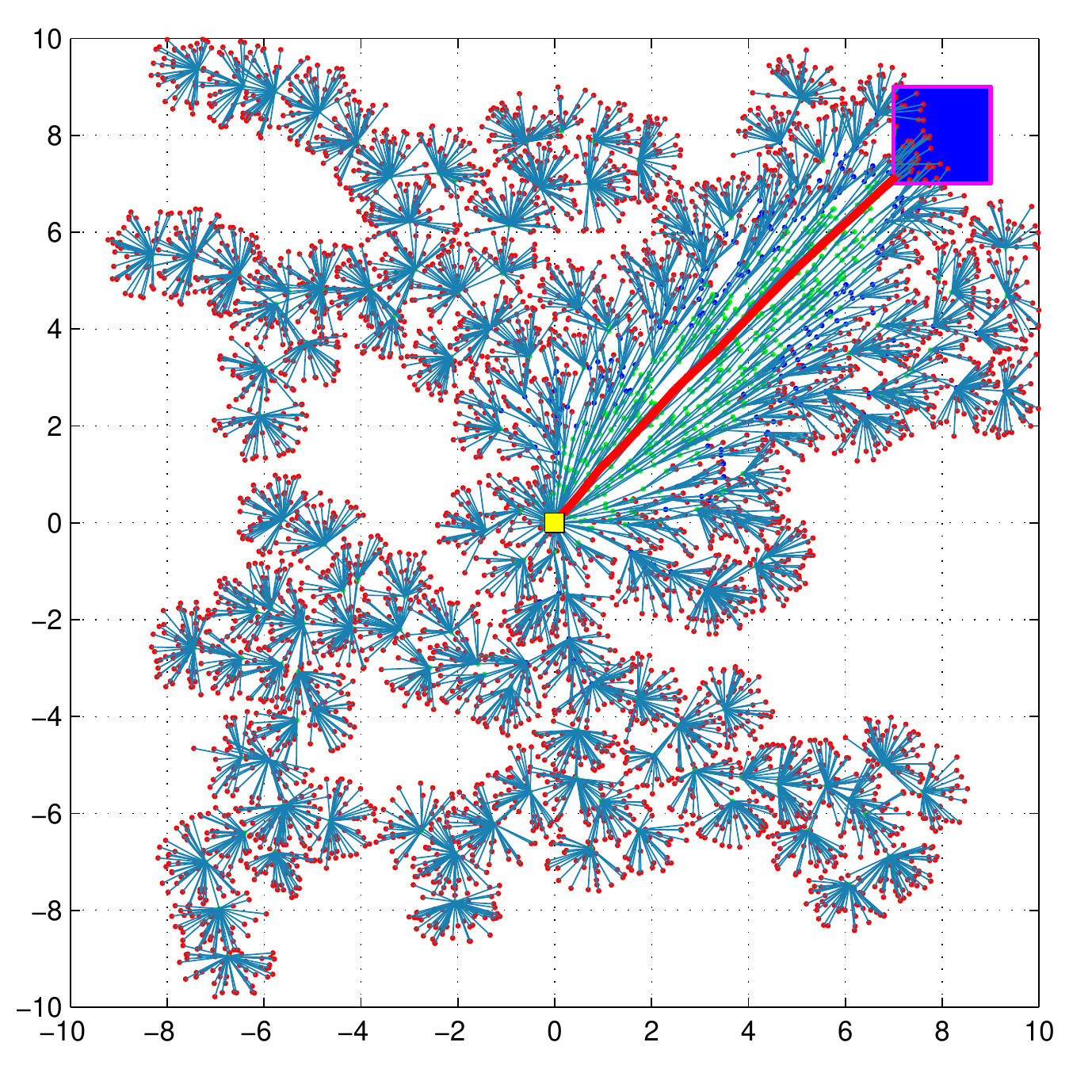}} \label{figure:pt1_rrtsharp_v1_it10000}}
    }
	\mbox{
	\setcounter{subfigure}{0}
	\renewcommand{\thesubfigure}{(\roman{subfigure})}
	\subfigure[]{\scalebox{0.28}{\includegraphics[trim = 4.0cm 6.937cm 3.587cm 7.0cm, clip =
          true]{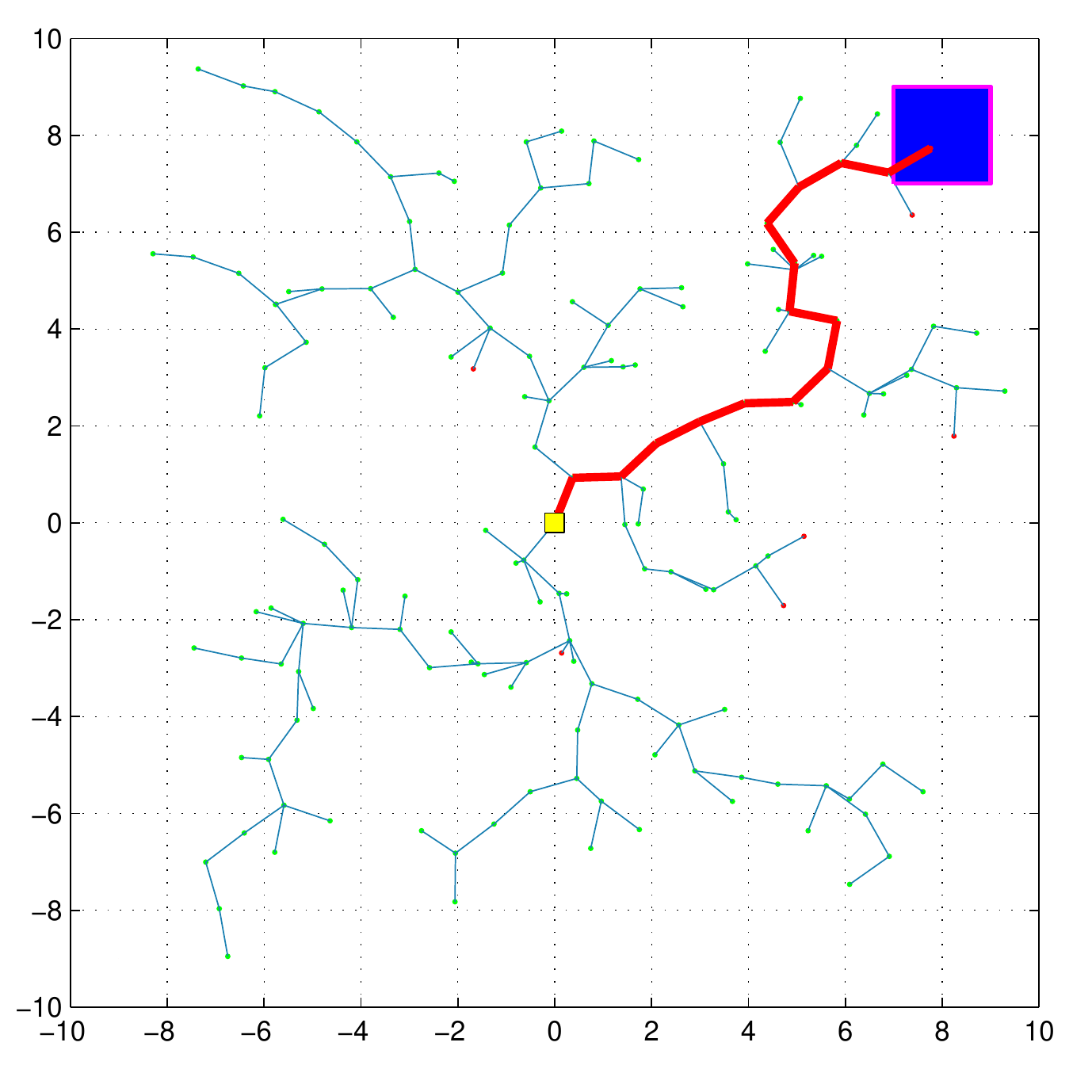}} \label{figure:pt1_rrtsharp_v2_it250}}
    \subfigure[]{\scalebox{0.28}{\includegraphics[trim = 4.0cm 6.937cm 3.587cm 7.0cm, clip =
          true]{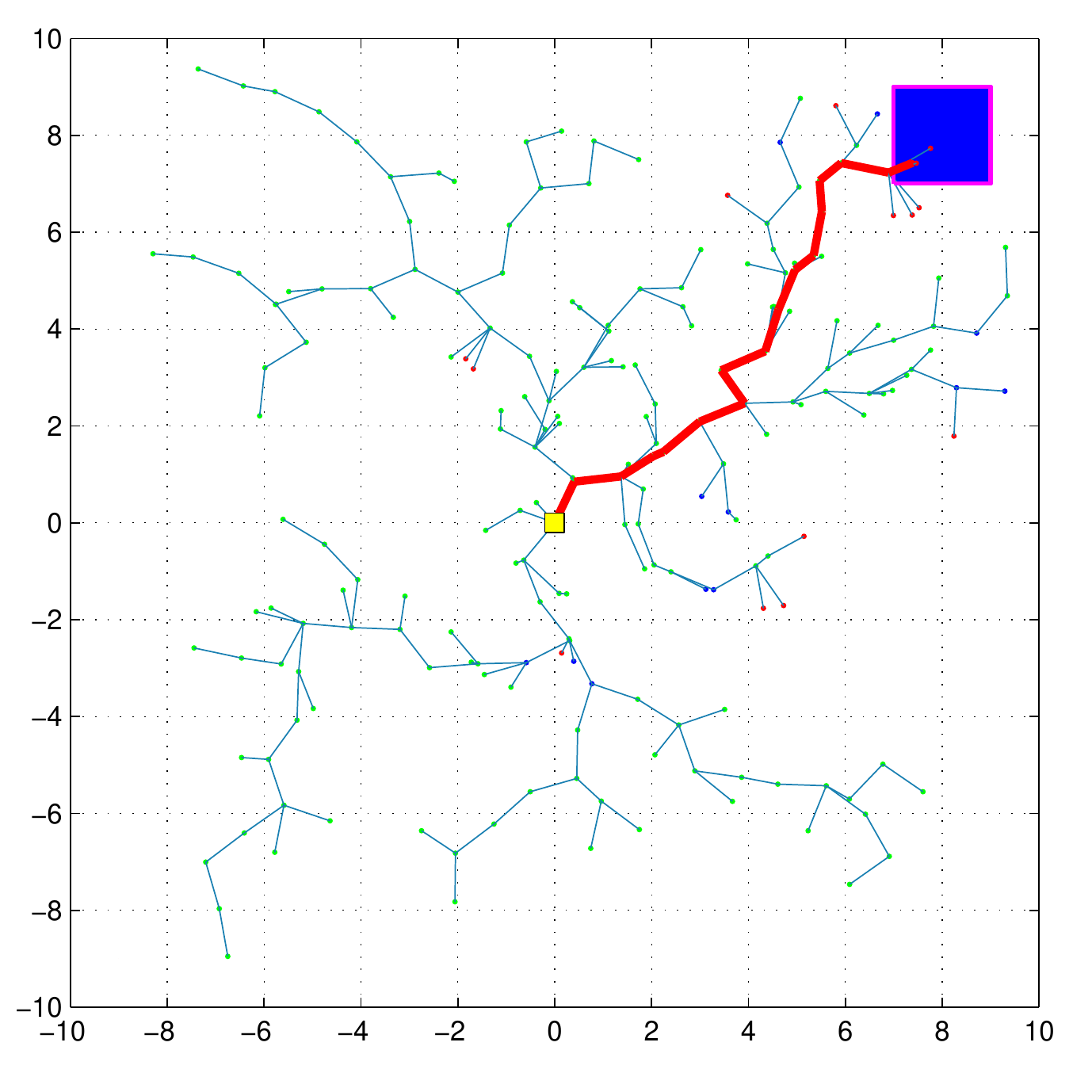}} \label{figure:pt1_rrtsharp_v2_it500}}
    \subfigure[]{\scalebox{0.28}{\includegraphics[trim = 4.0cm 6.937cm 3.587cm 7.0cm, clip =
          true]{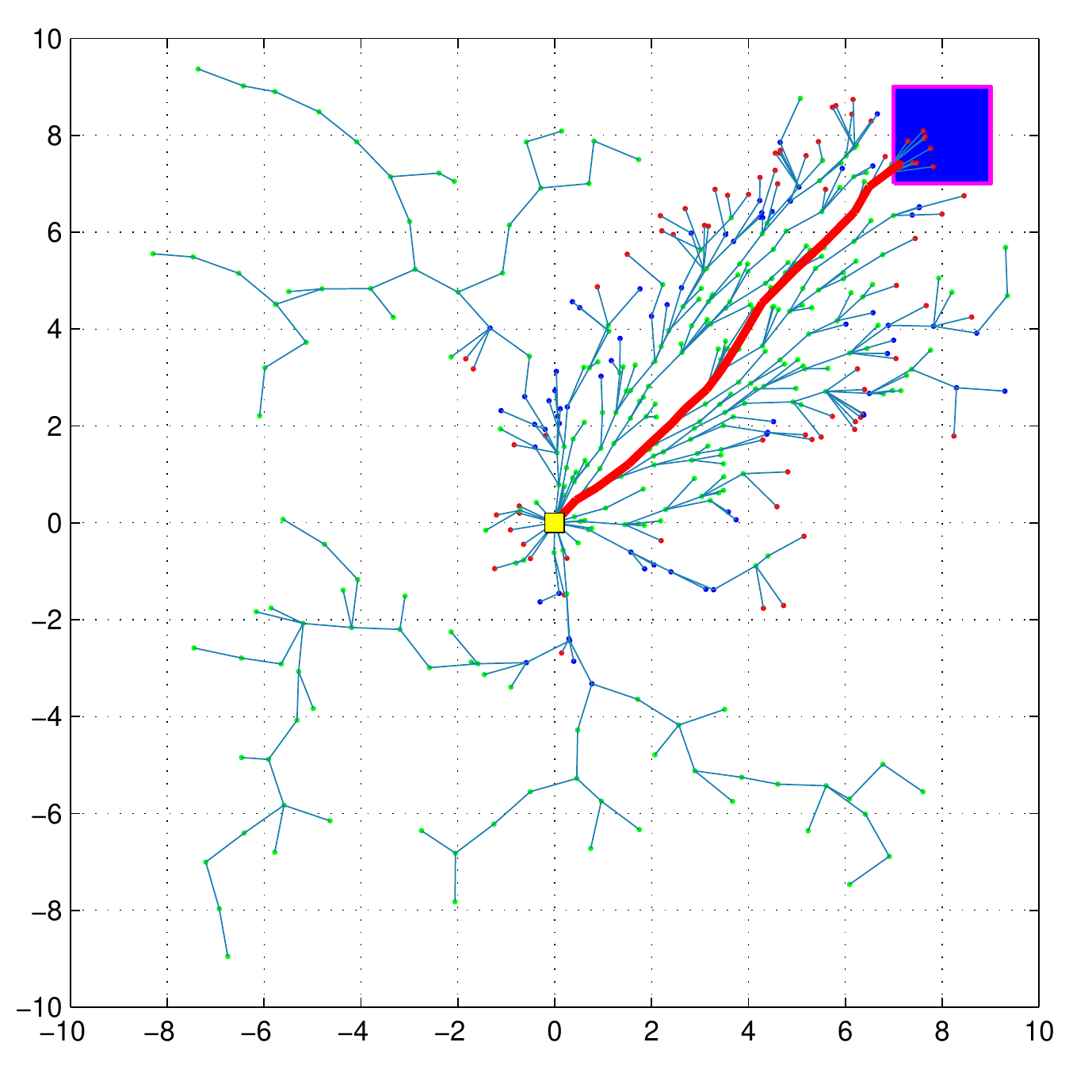}} \label{figure:pt1_rrtsharp_v2_it2500}}
    \subfigure[]{\scalebox{0.28}{\includegraphics[trim = 4.0cm 6.937cm 3.587cm 7.0cm, clip =
          true]{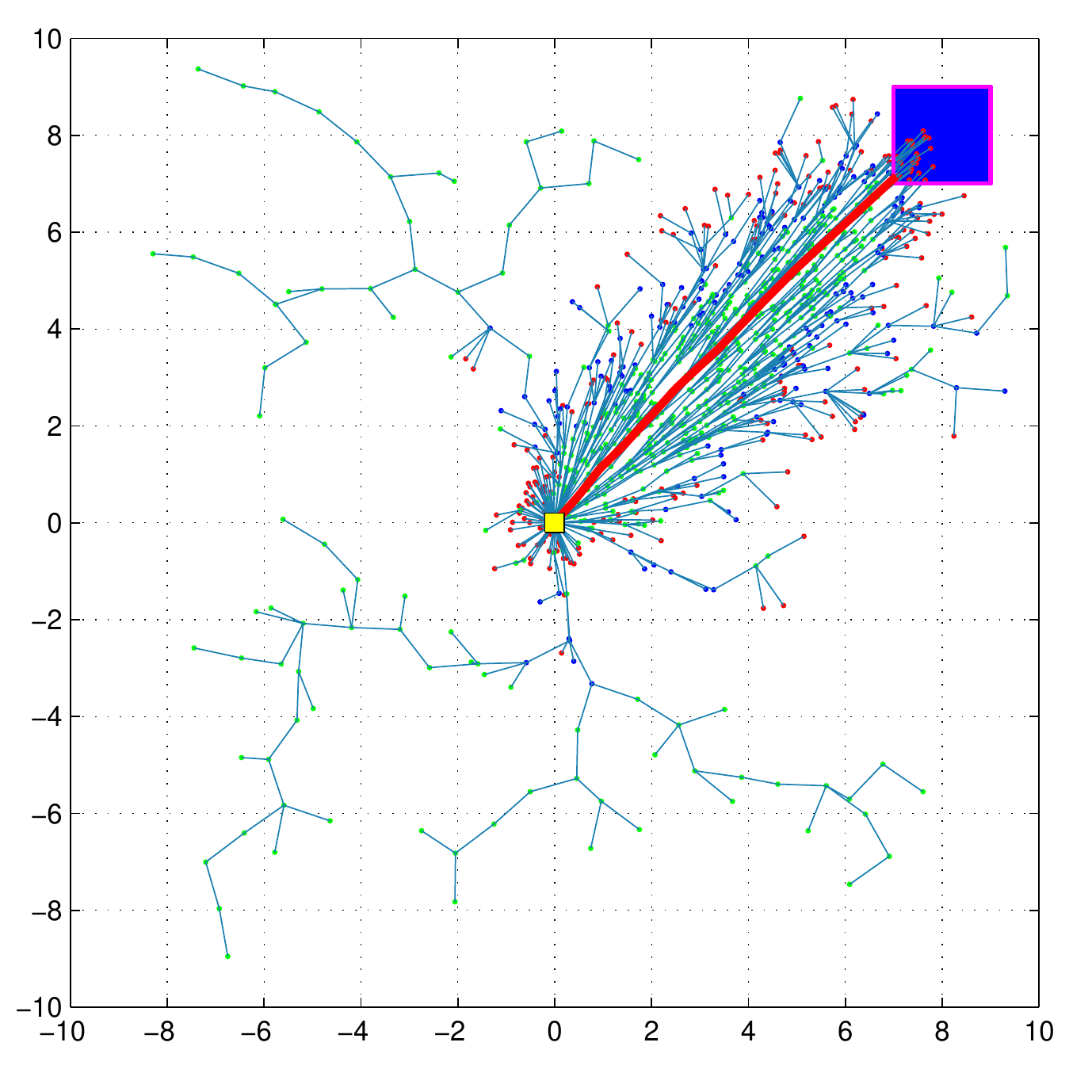}} \label{figure:pt1_rrtsharp_v2_it10000}}
   	}	
	\mbox{
	\setcounter{subfigure}{4}
	\renewcommand{\thesubfigure}{(\alph{subfigure})}
    \subfigure[]{\scalebox{0.57}{\includegraphics[trim = 4.0cm 6.937cm 3.587cm 7.0cm, clip =
          true]{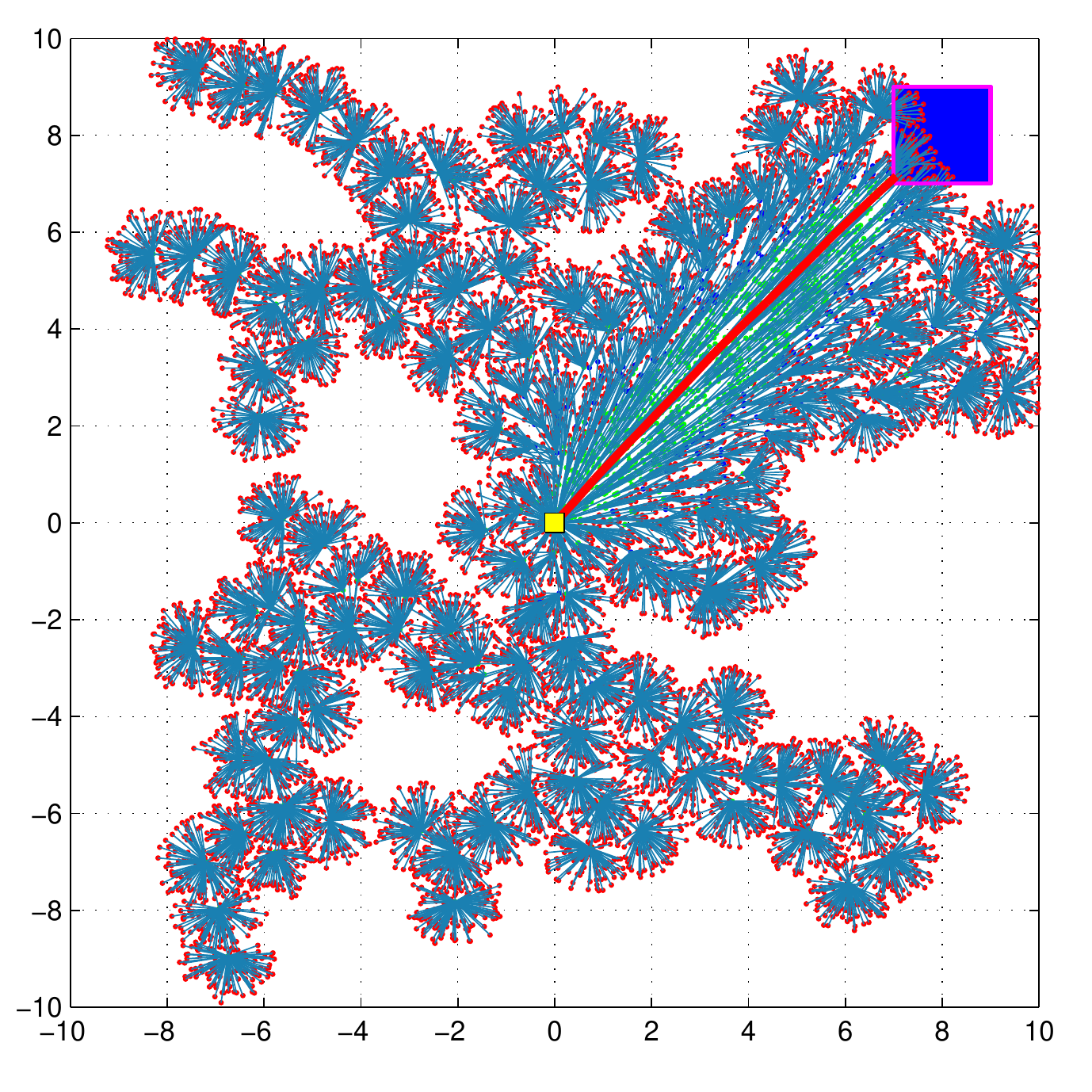}} \label{figure:pt1_rrtsharp_v1_it24999}}
    \setcounter{subfigure}{4}
    \renewcommand{\thesubfigure}{(\roman{subfigure})}
    \subfigure[]{\scalebox{0.57}{\includegraphics[trim = 4.0cm 6.937cm 3.587cm 7.0cm, clip =
          true]{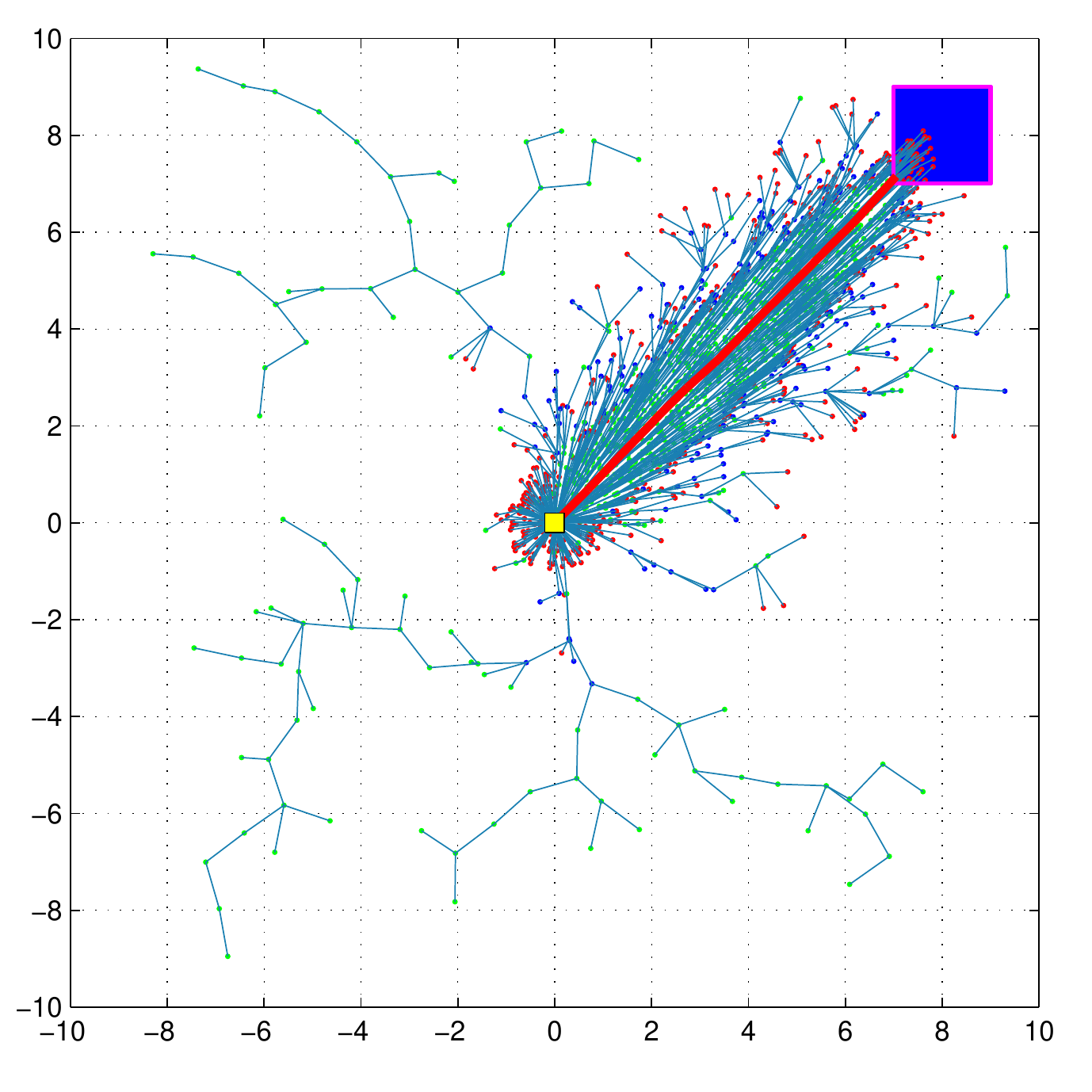}} \label{figure:pt1_rrtsharp_v2_it24999}}
    }

\caption{The evolution of the tree computed by \AlgRRTsharpNoBlackVertex{} and \AlgRRTsharpPromisingParent{} algorithms is shown in \subref{figure:pt1_rrtsharp_v1_it250}-\subref{figure:pt1_rrtsharp_v1_it24999} and \subref{figure:pt1_rrtsharp_v2_it250}-\subref{figure:pt1_rrtsharp_v2_it24999}, respectively. The configuration of the trees \subref{figure:pt1_rrtsharp_v1_it250}, \subref{figure:pt1_rrtsharp_v2_it250} is at 250 iterations, \subref{figure:pt1_rrtsharp_v1_it500}, \subref{figure:pt1_rrtsharp_v2_it500} is at 500 iterations, \subref{figure:pt1_rrtsharp_v1_it2500}, \subref{figure:pt1_rrtsharp_v2_it2500} is at 2500 iterations, \subref{figure:pt1_rrtsharp_v1_it10000}, \subref{figure:pt1_rrtsharp_v2_it10000} is at 10000 iterations,
and \subref{figure:pt1_rrtsharp_v1_it24999}, \subref{figure:pt1_rrtsharp_v2_it24999} is at 25000 iterations.}

    \label{figure:sim_d2_pt1_rrtsharp_v1_v2_iterations}
  \end{center}
\end{figure*}

\begin{figure*}[htp]
  \begin{center}

	\mbox{
	\setcounter{subfigure}{0}
	\renewcommand{\thesubfigure}{(\alph{subfigure})}
    \subfigure[]{\scalebox{0.28}{\includegraphics[trim = 4.0cm 6.937cm 3.587cm 7.0cm, clip =
          true]{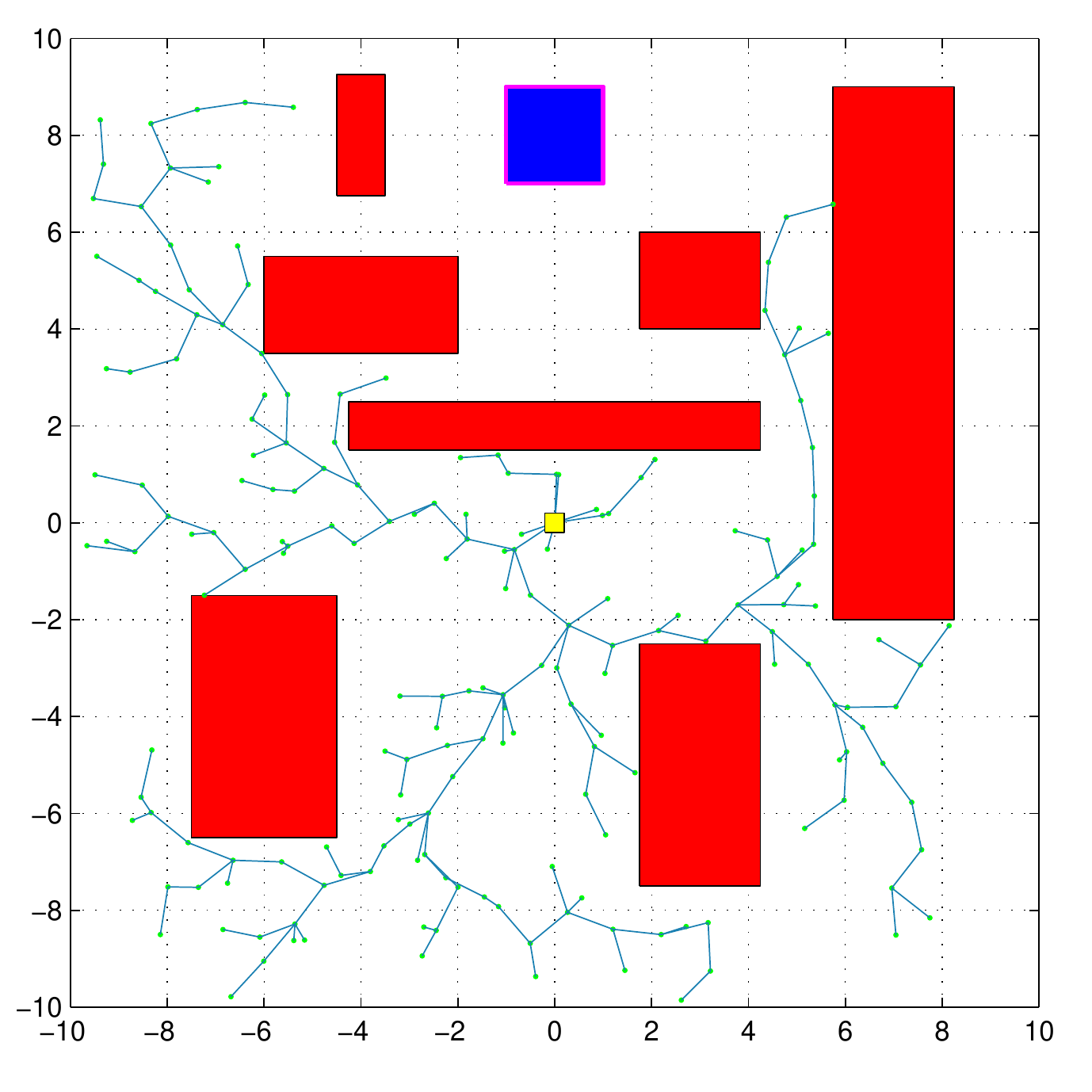}} \label{figure:pt2_rrtsharp_v1_it250}} 		
    \subfigure[]{\scalebox{0.28}{\includegraphics[trim = 4.0cm 6.937cm 3.587cm 7.0cm, clip =
          true]{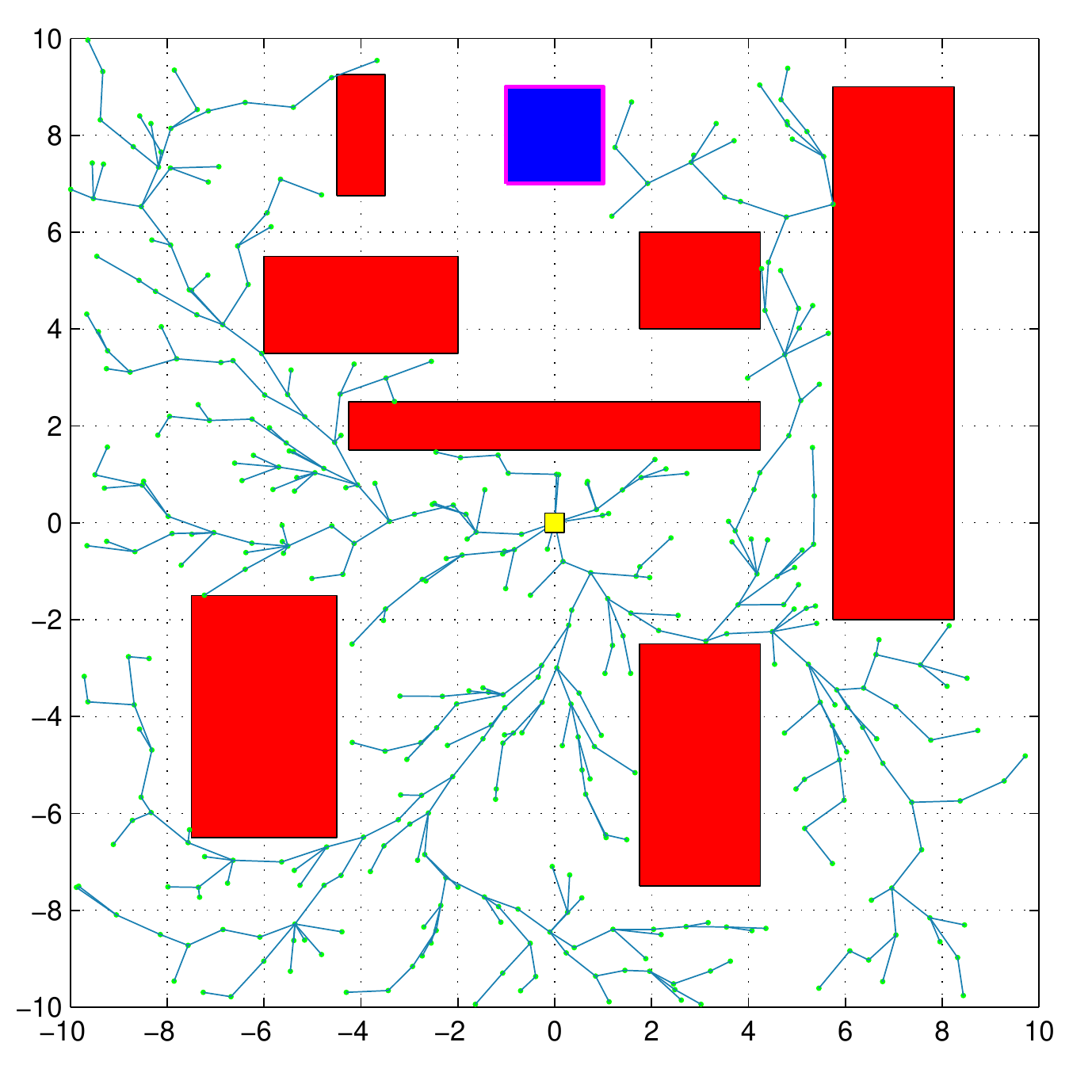}} \label{figure:pt2_rrtsharp_v1_it500}}
    \subfigure[]{\scalebox{0.28}{\includegraphics[trim = 4.0cm 6.937cm 3.587cm 7.0cm, clip =
          true]{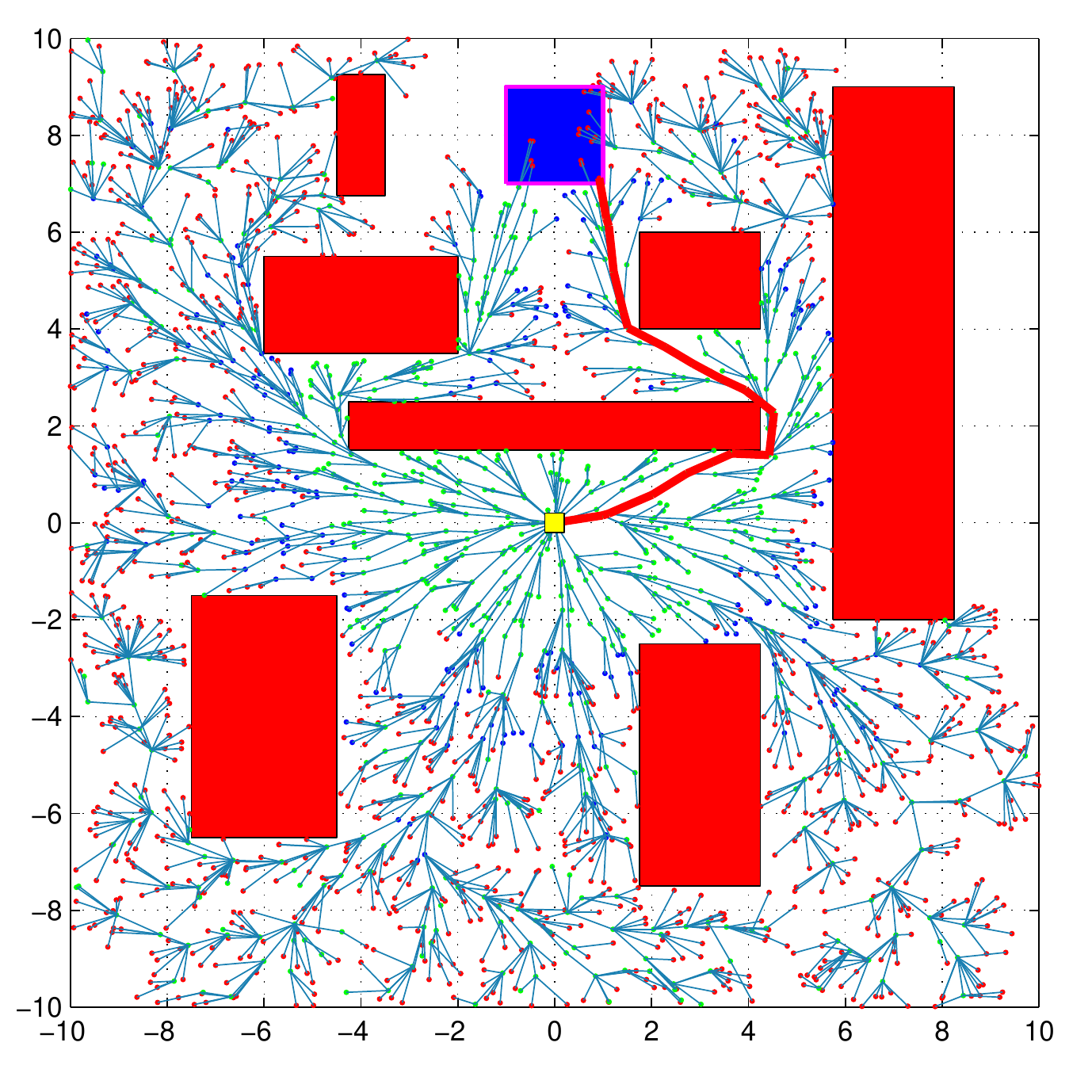}} \label{figure:pt2_rrtsharp_v1_it2500}}
    \subfigure[]{\scalebox{0.28}{\includegraphics[trim = 4.0cm 6.937cm 3.587cm 7.0cm, clip =
          true]{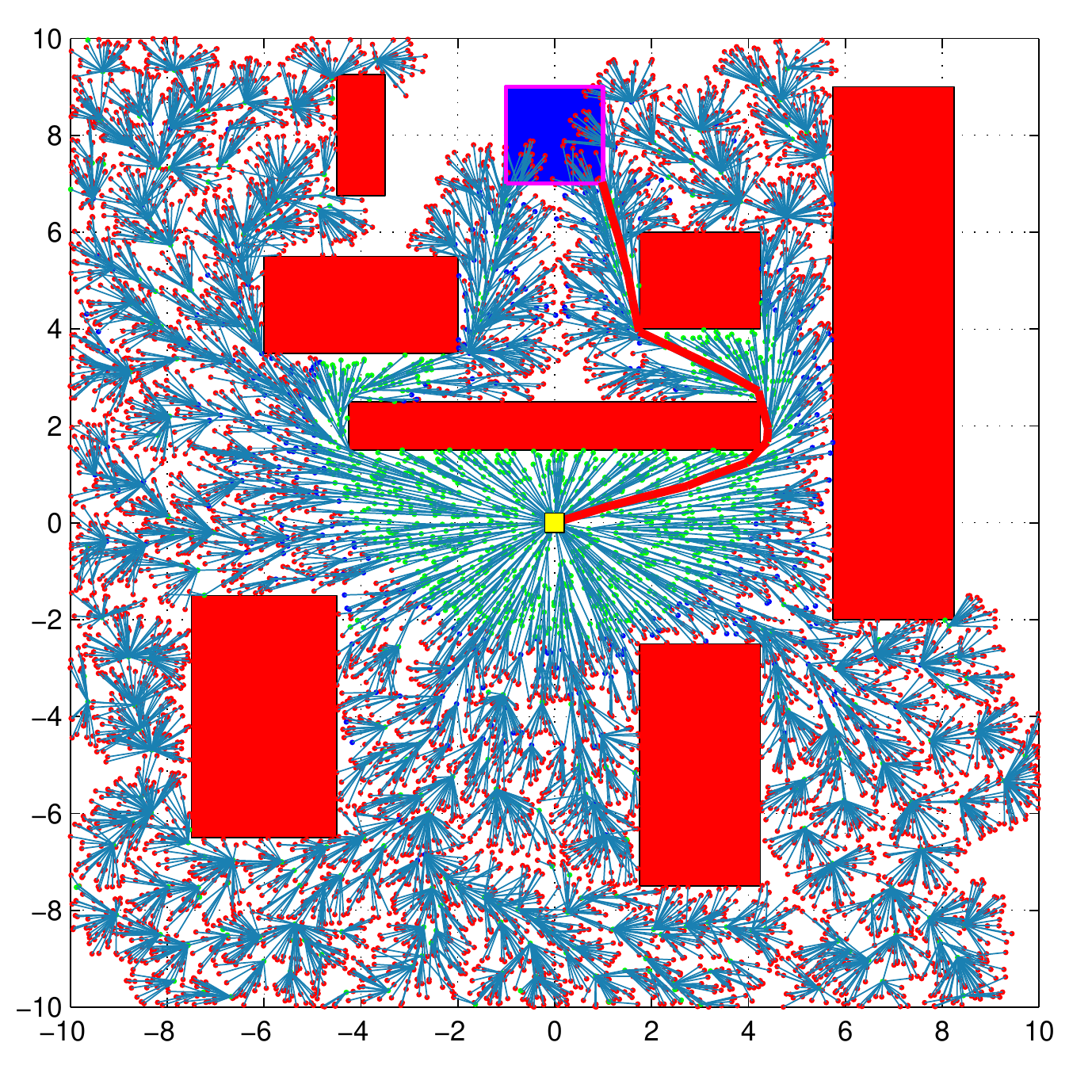}} \label{figure:pt2_rrtsharp_v1_it10000}}
    }
    \mbox{
	\setcounter{subfigure}{0}
	\renewcommand{\thesubfigure}{(\roman{subfigure})}
    \subfigure[]{\scalebox{0.28}{\includegraphics[trim = 4.0cm 6.937cm 3.587cm 7.0cm, clip =
          true]{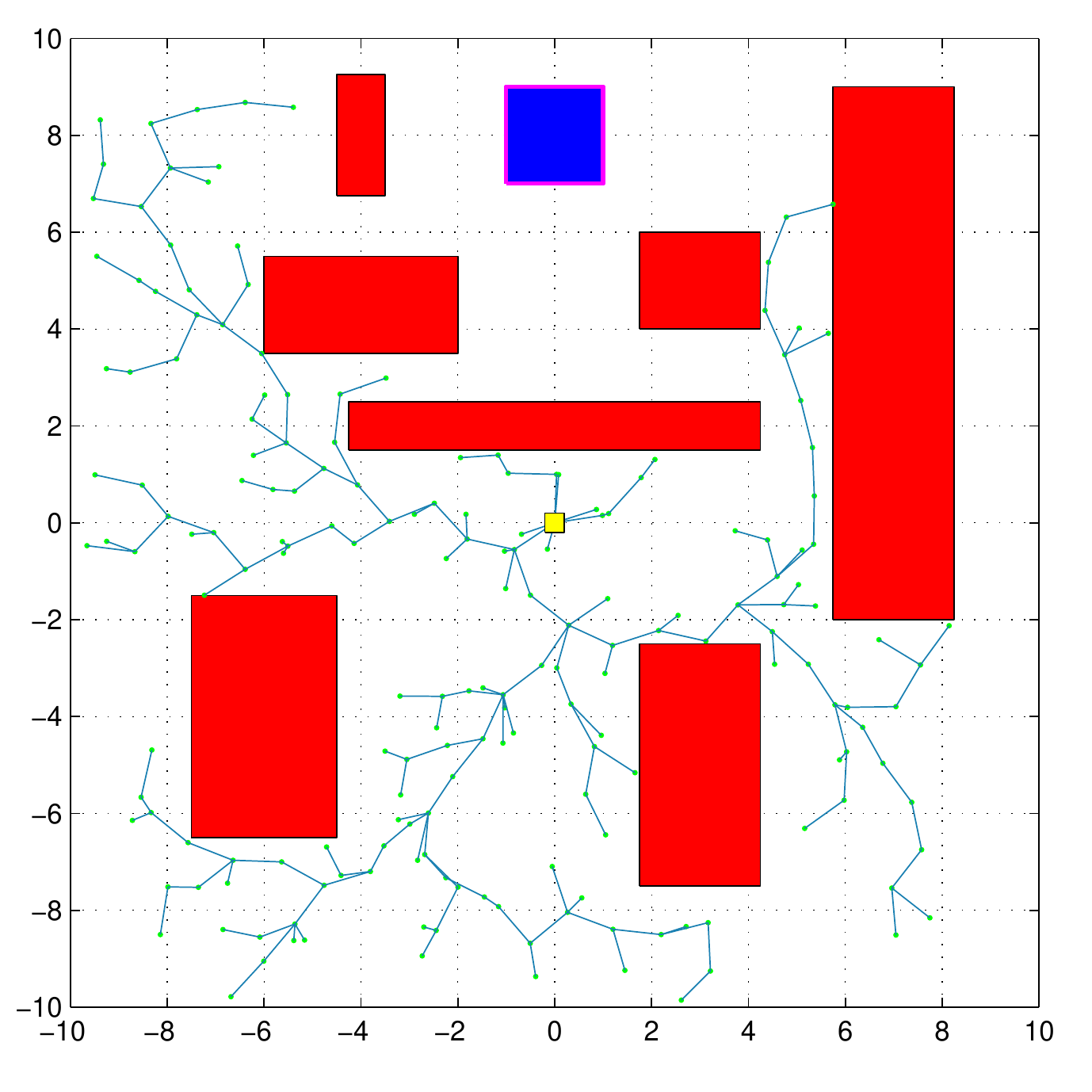}} \label{figure:pt2_rrtsharp_v2_it250}}
    \subfigure[]{\scalebox{0.28}{\includegraphics[trim = 4.0cm 6.937cm 3.587cm 7.0cm, clip =
          true]{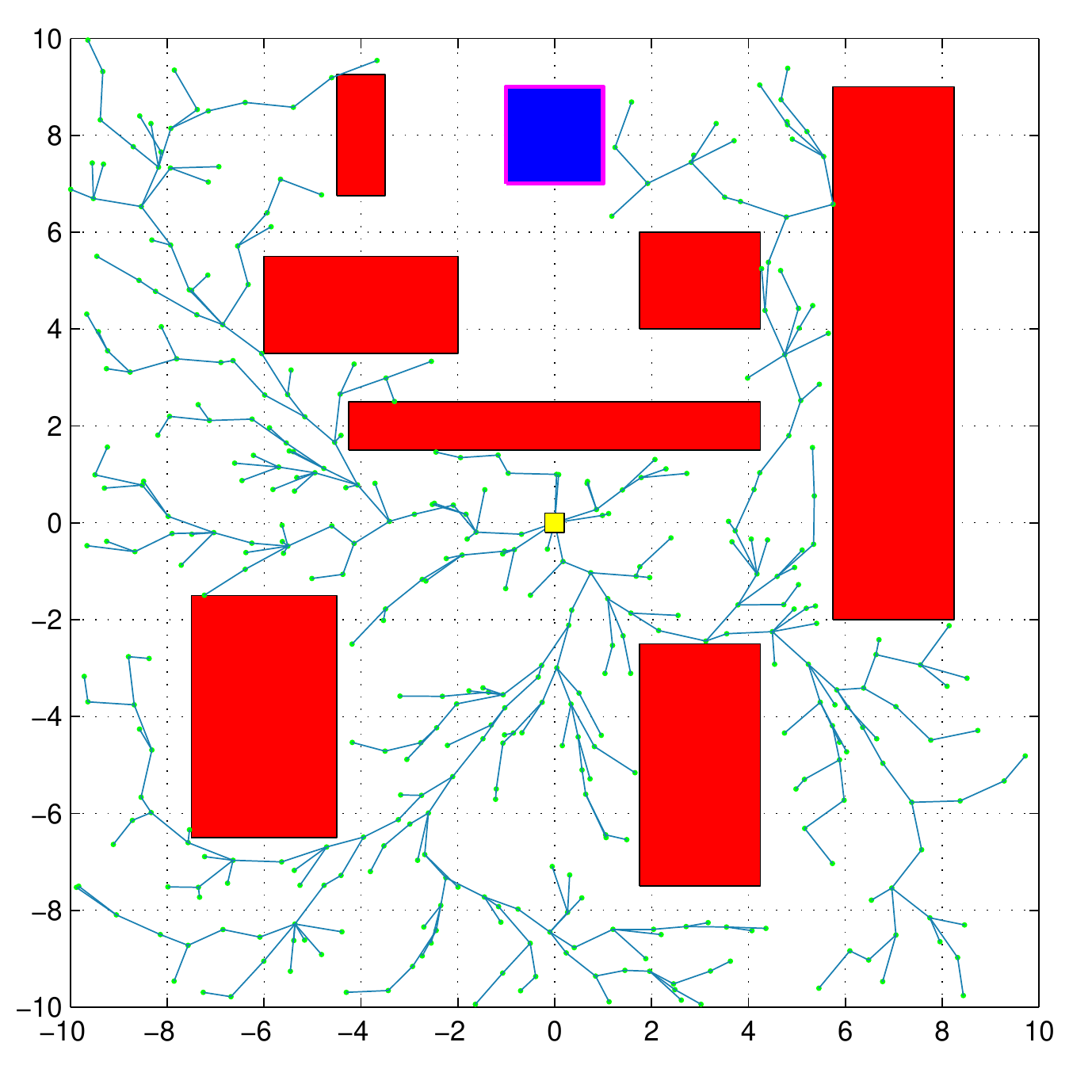}} \label{figure:pt2_rrtsharp_v2_it500}}
    \subfigure[]{\scalebox{0.28}{\includegraphics[trim = 4.0cm 6.937cm 3.587cm 7.0cm, clip =
          true]{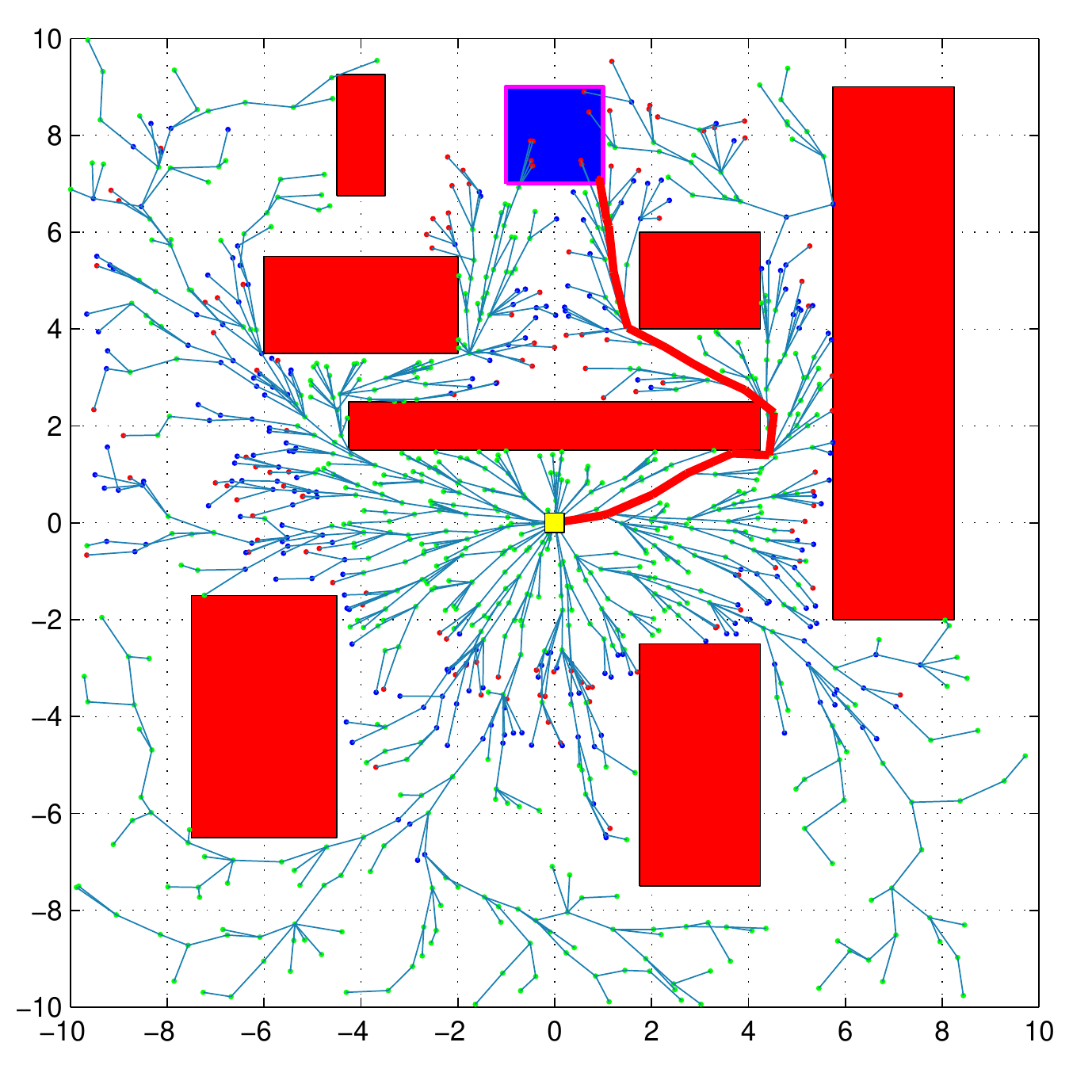}} \label{figure:pt2_rrtsharp_v2_it2500}}
    \subfigure[]{\scalebox{0.28}{\includegraphics[trim = 4.0cm 6.937cm 3.587cm 7.0cm, clip =
          true]{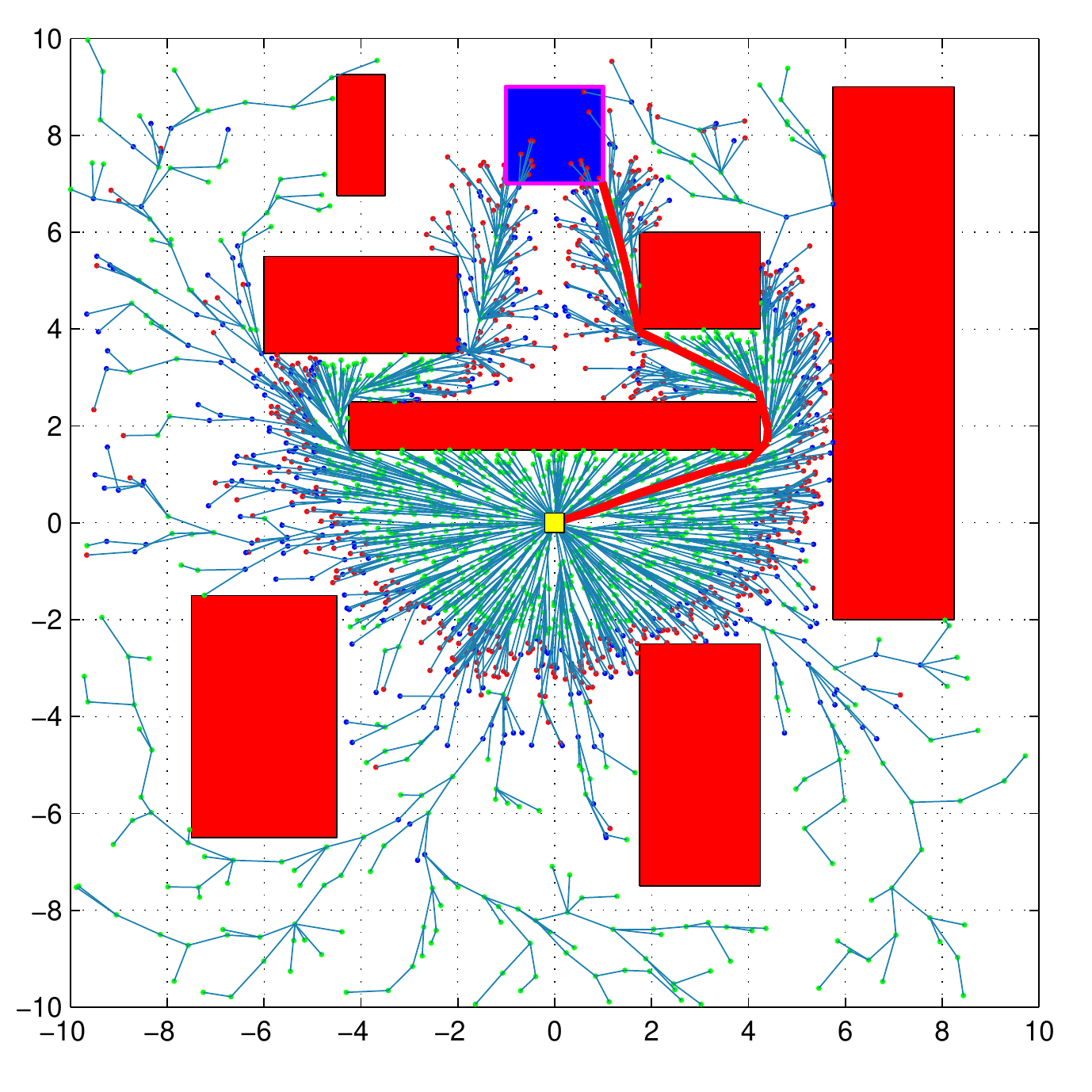}} \label{figure:pt2_rrtsharp_v2_it10000}}
   	}
	\mbox{
	\setcounter{subfigure}{4}
	\renewcommand{\thesubfigure}{(\alph{subfigure})} 	
    \subfigure[]{\scalebox{0.57}{\includegraphics[trim = 4.0cm 6.937cm 3.587cm 7.0cm, clip =
          true]{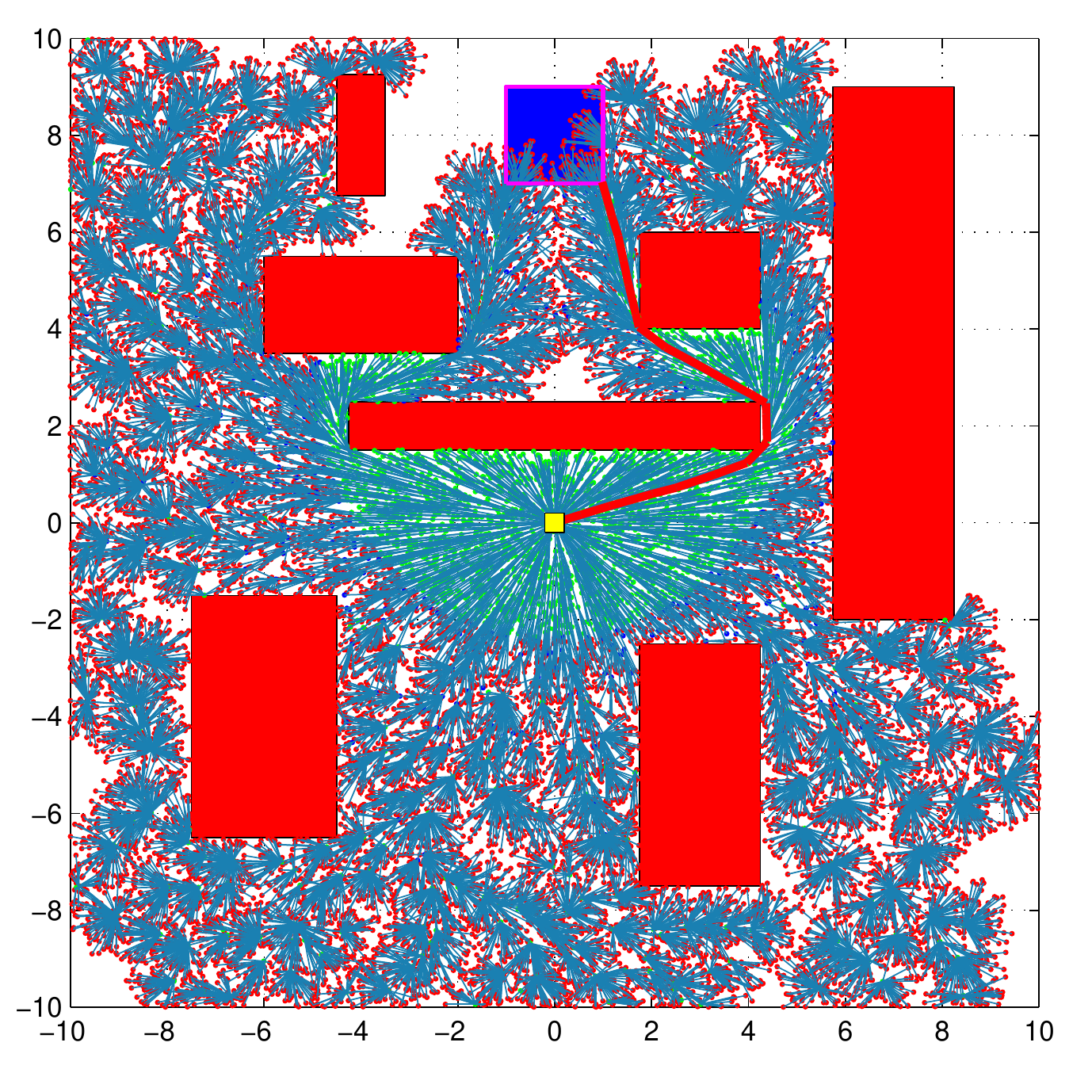}} \label{figure:pt2_rrtsharp_v1_it24999}}
	\setcounter{subfigure}{4}
	\renewcommand{\thesubfigure}{(\roman{subfigure})}
    \subfigure[]{\scalebox{0.57}{\includegraphics[trim = 4.0cm 6.937cm 3.587cm 7.0cm, clip =
          true]{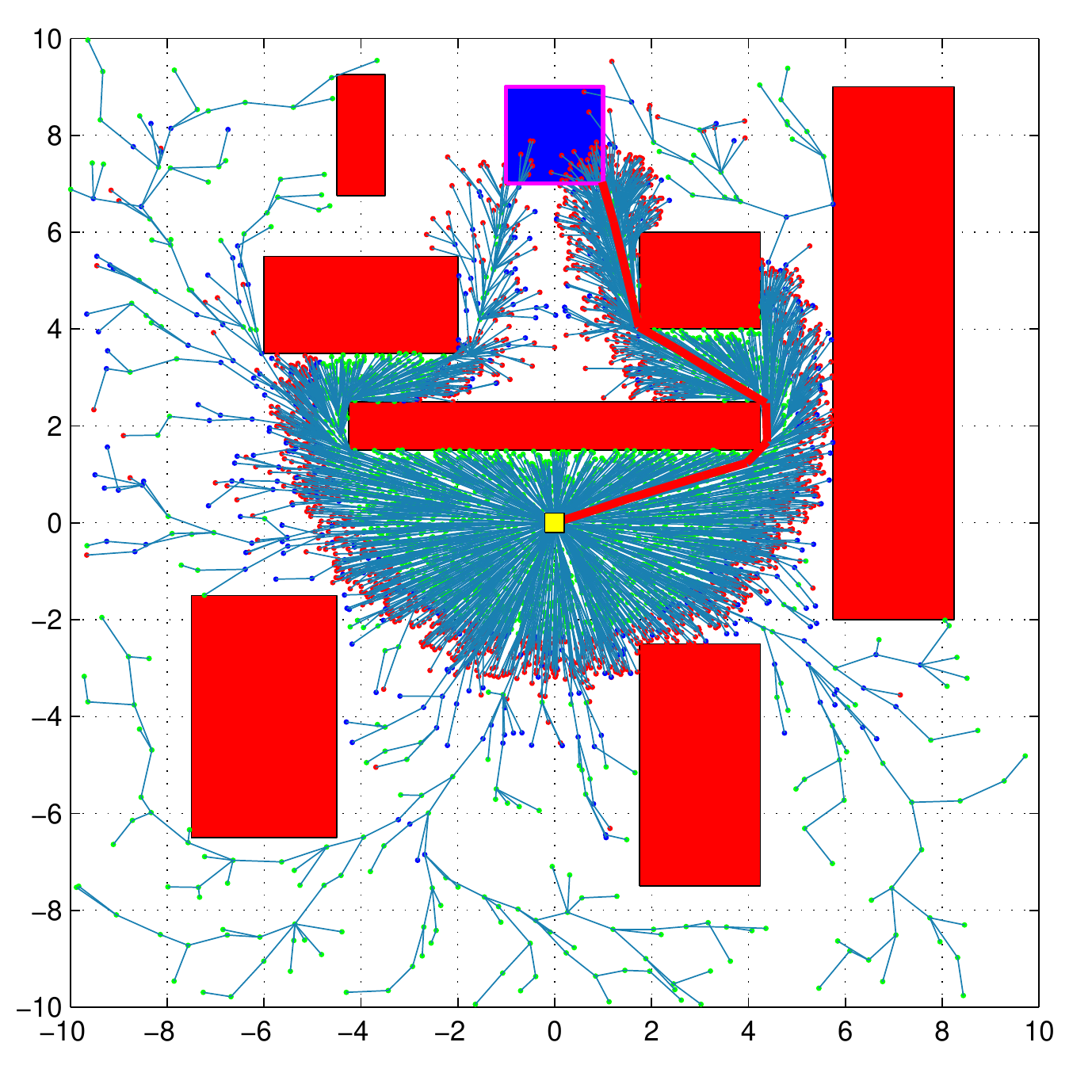}} \label{figure:pt2_rrtsharp_v2_it24999}}
    }

\caption{The evolution of the tree computed by \AlgRRTsharpNoBlackVertex{} and \AlgRRTsharpPromisingParent{} algorithms is shown in \subref{figure:pt2_rrtsharp_v1_it250}-\subref{figure:pt2_rrtsharp_v1_it24999} and \subref{figure:pt2_rrtsharp_v2_it250}-\subref{figure:pt2_rrtsharp_v2_it24999}, respectively. The configuration of the trees \subref{figure:pt2_rrtsharp_v1_it250}, \subref{figure:pt2_rrtsharp_v2_it250} is at 250 iterations, \subref{figure:pt2_rrtsharp_v1_it500}, \subref{figure:pt2_rrtsharp_v2_it500} is at 500 iterations, \subref{figure:pt2_rrtsharp_v1_it2500}, \subref{figure:pt2_rrtsharp_v2_it2500} is at 2500 iterations, \subref{figure:pt2_rrtsharp_v1_it10000}, \subref{figure:pt2_rrtsharp_v2_it10000} is at 10000 iterations,
 and \subref{figure:pt2_rrtsharp_v1_it24999}, \subref{figure:pt2_rrtsharp_v2_it24999} is at 25000 iterations.}
    \label{figure:sim_d2_pt2_rrtsharp_v1_v2_iterations}
  \end{center}
\end{figure*}

\begin{figure*}[htp]
  \begin{center}

	\mbox{
	\setcounter{subfigure}{0}
	\renewcommand{\thesubfigure}{(\alph{subfigure})}
    \subfigure[]{\scalebox{0.28}{\includegraphics[trim = 4.0cm 6.937cm 3.587cm 7.0cm, clip =
          true]{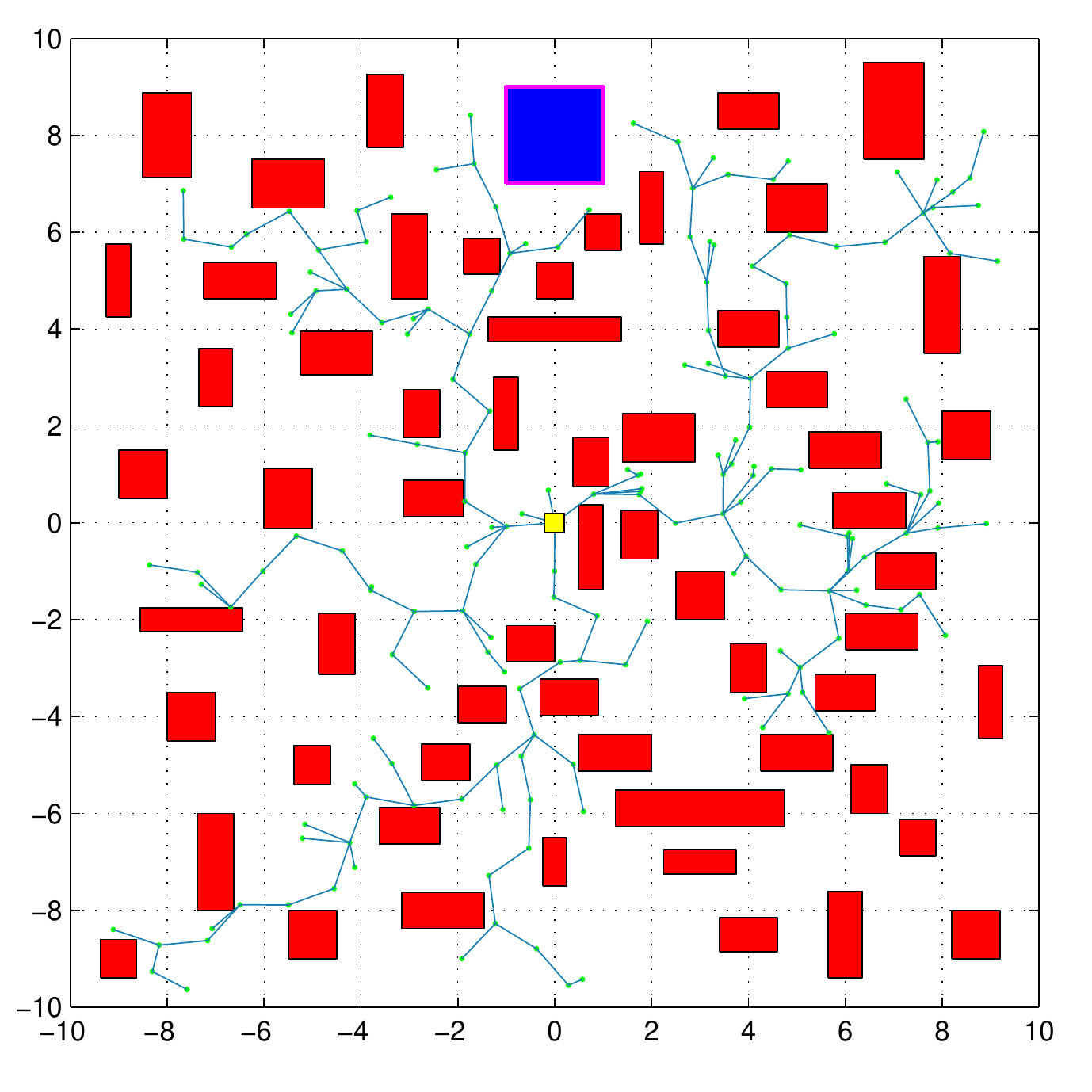}} \label{figure:pt3_rrtsharp_v1_it250}} 		
    \subfigure[]{\scalebox{0.28}{\includegraphics[trim = 4.0cm 6.937cm 3.587cm 7.0cm, clip =
          true]{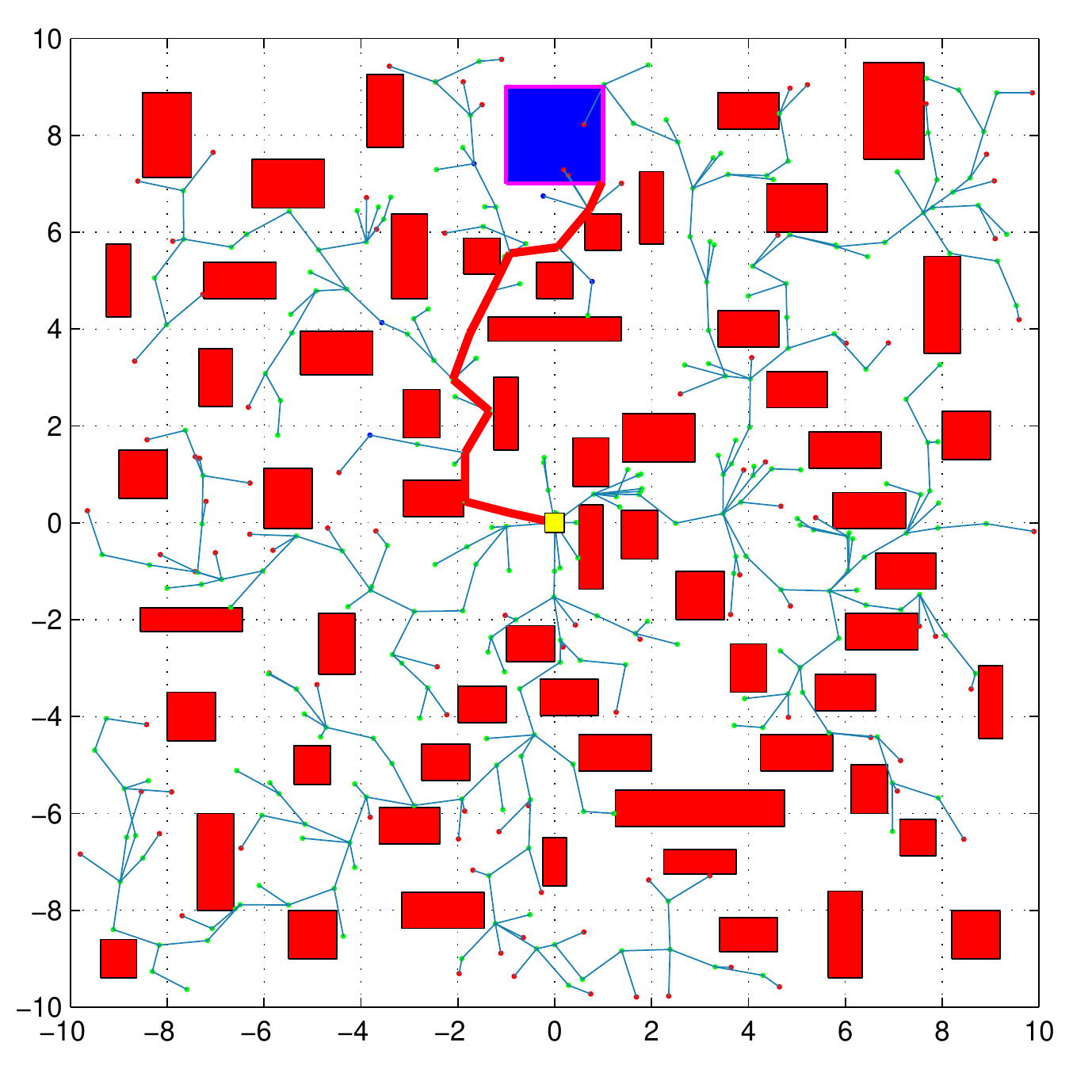}} \label{figure:pt3_rrtsharp_v1_it500}}
    \subfigure[]{\scalebox{0.28}{\includegraphics[trim = 4.0cm 6.937cm 3.587cm 7.0cm, clip =
          true]{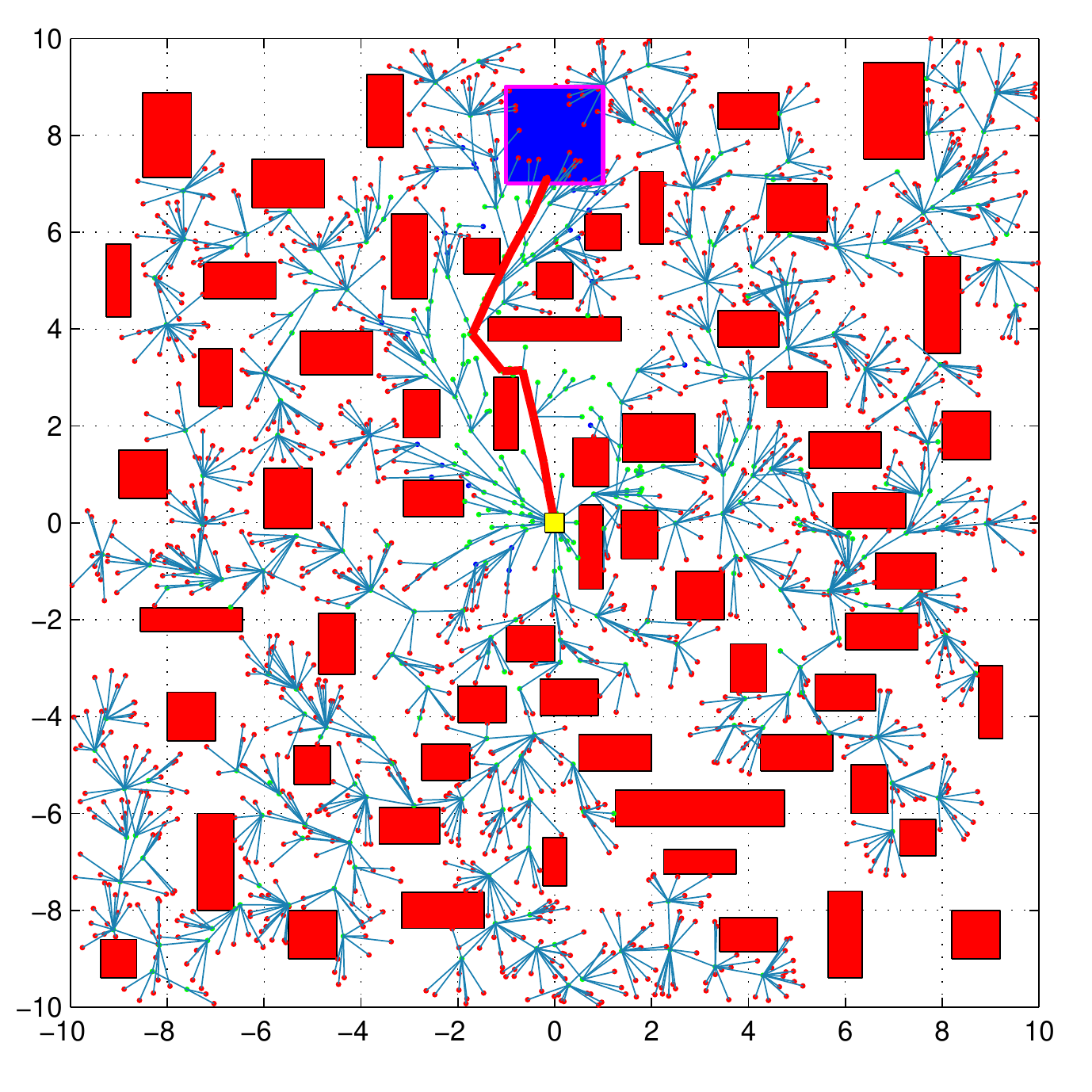}} \label{figure:pt3_rrtsharp_v1_it2500}}
    \subfigure[]{\scalebox{0.28}{\includegraphics[trim = 4.0cm 6.937cm 3.587cm 7.0cm, clip =
          true]{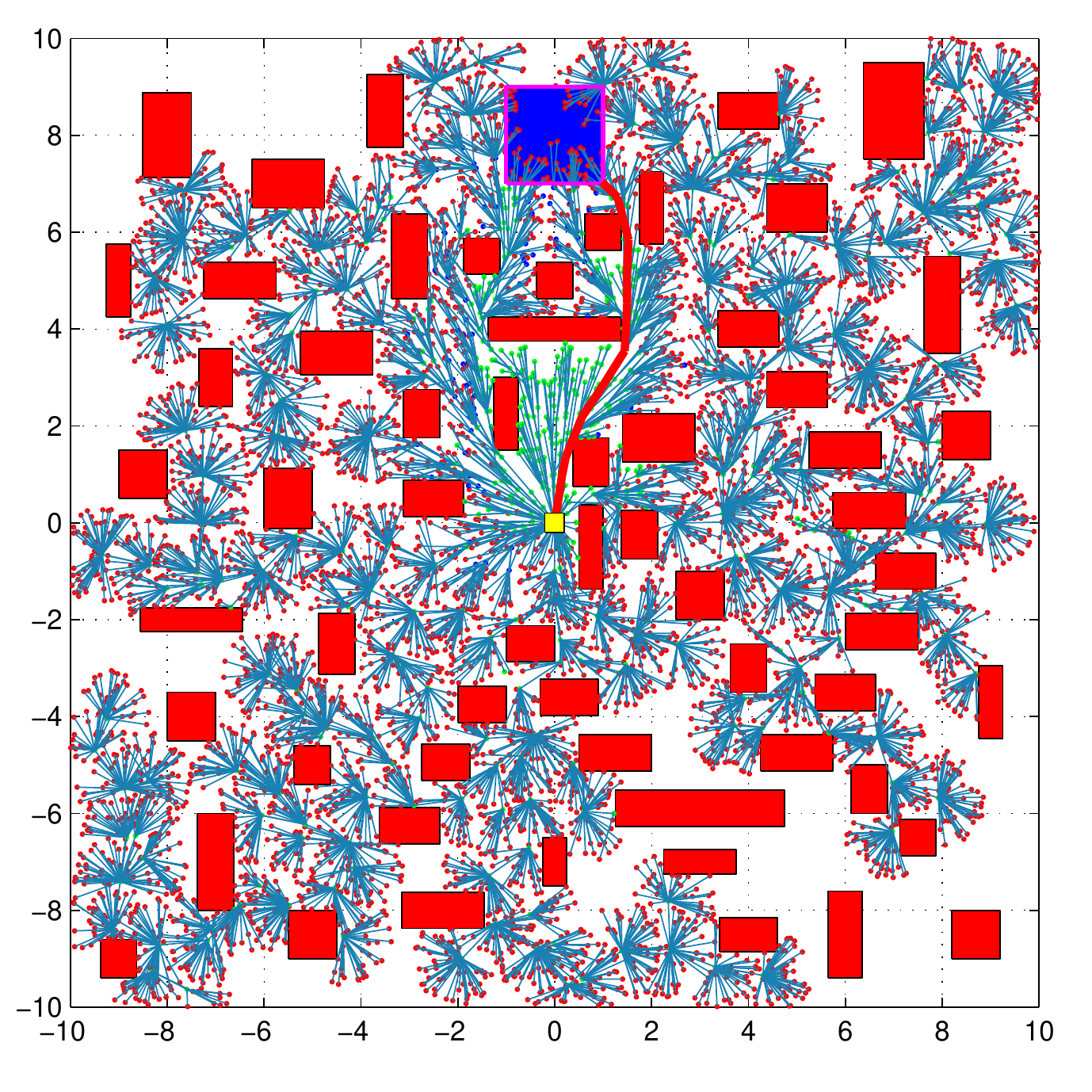}} \label{figure:pt3_rrtsharp_v1_it10000}}
    }
	\mbox{
	\setcounter{subfigure}{0}
	\renewcommand{\thesubfigure}{(\roman{subfigure})}
    \subfigure[]{\scalebox{0.28}{\includegraphics[trim = 4.0cm 6.937cm 3.587cm 7.0cm, clip =
          true]{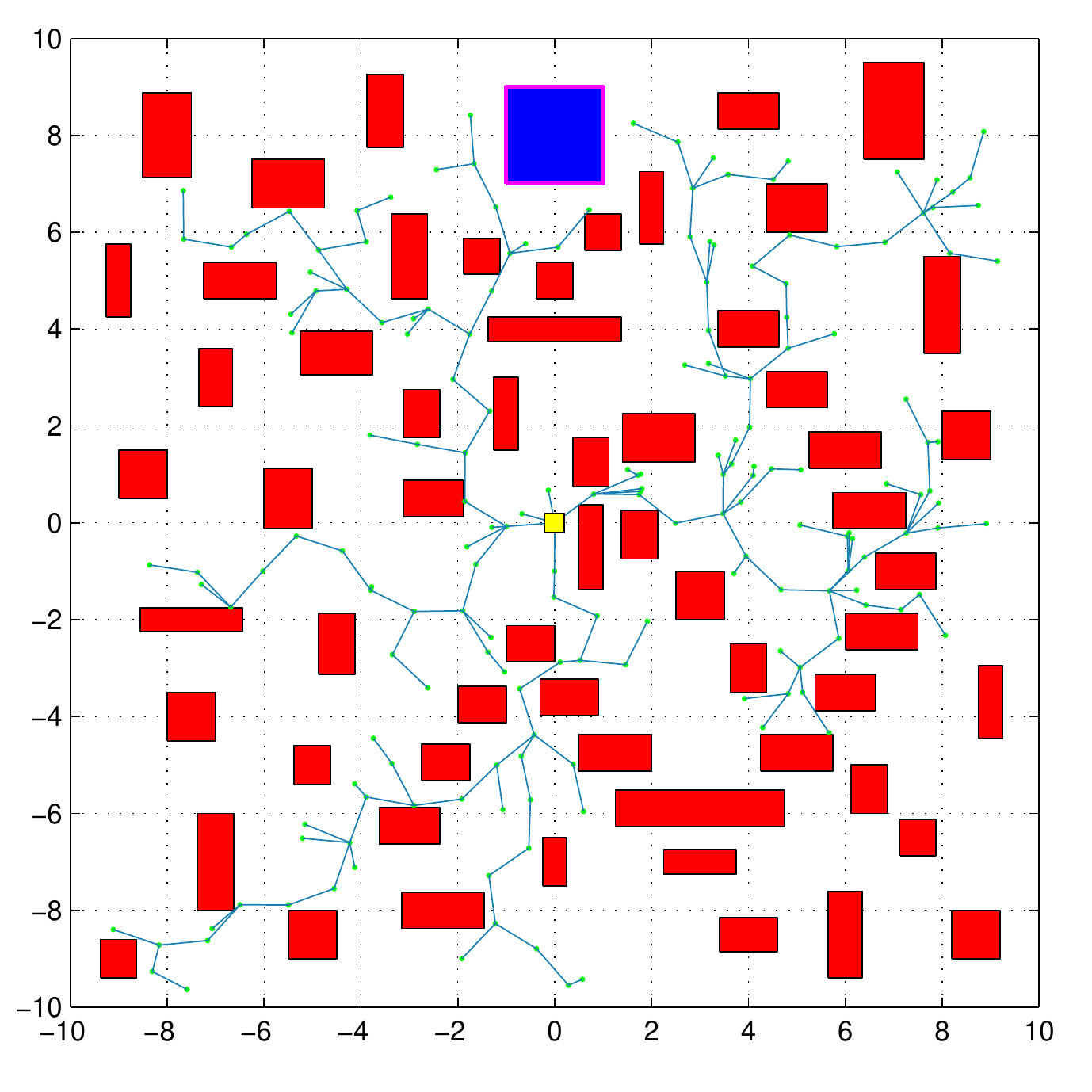}} \label{figure:pt3_rrtsharp_v2_it250}}
    \subfigure[]{\scalebox{0.28}{\includegraphics[trim = 4.0cm 6.937cm 3.587cm 7.0cm, clip =
          true]{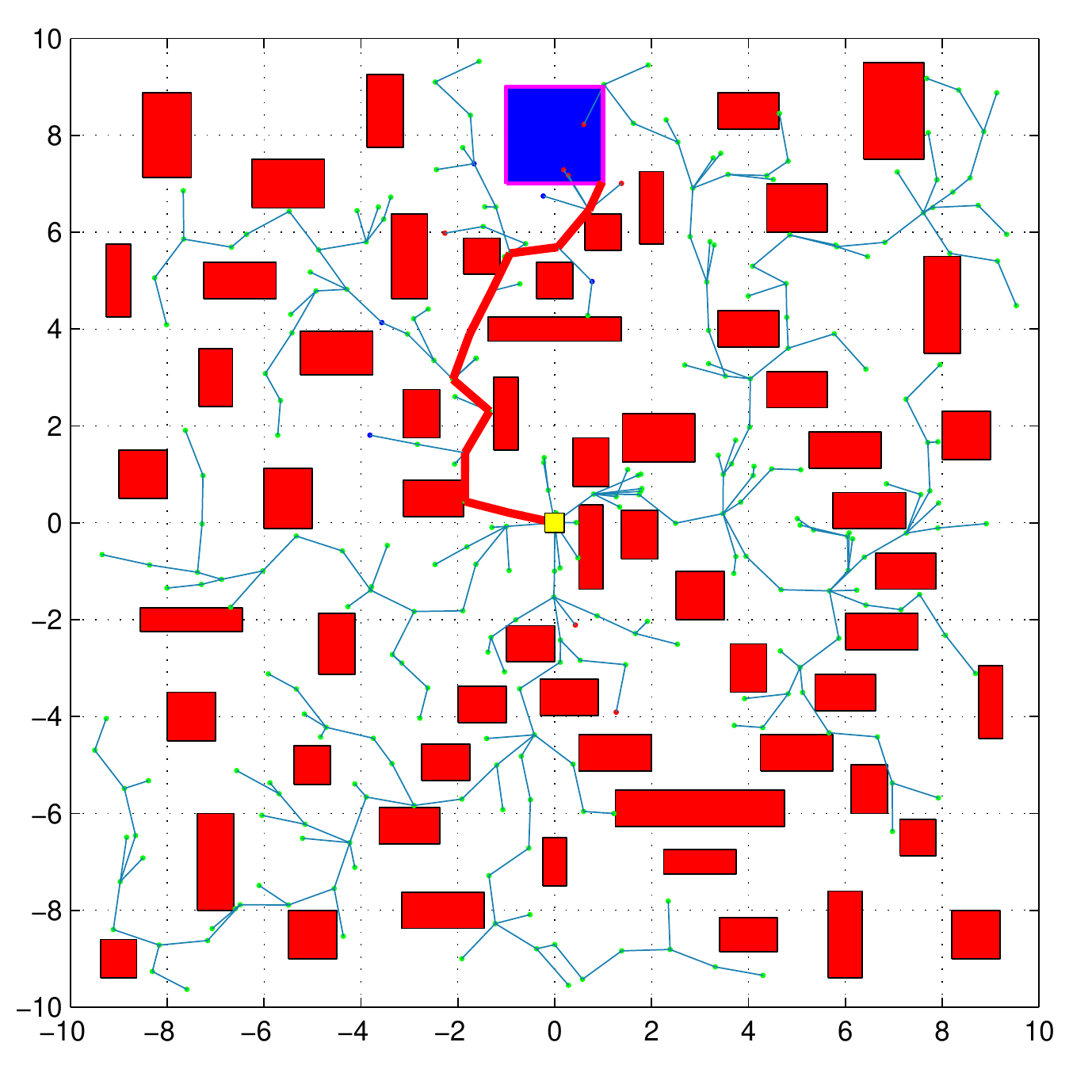}} \label{figure:pt3_rrtsharp_v2_it500}}
    \subfigure[]{\scalebox{0.28}{\includegraphics[trim = 4.0cm 6.937cm 3.587cm 7.0cm, clip =
          true]{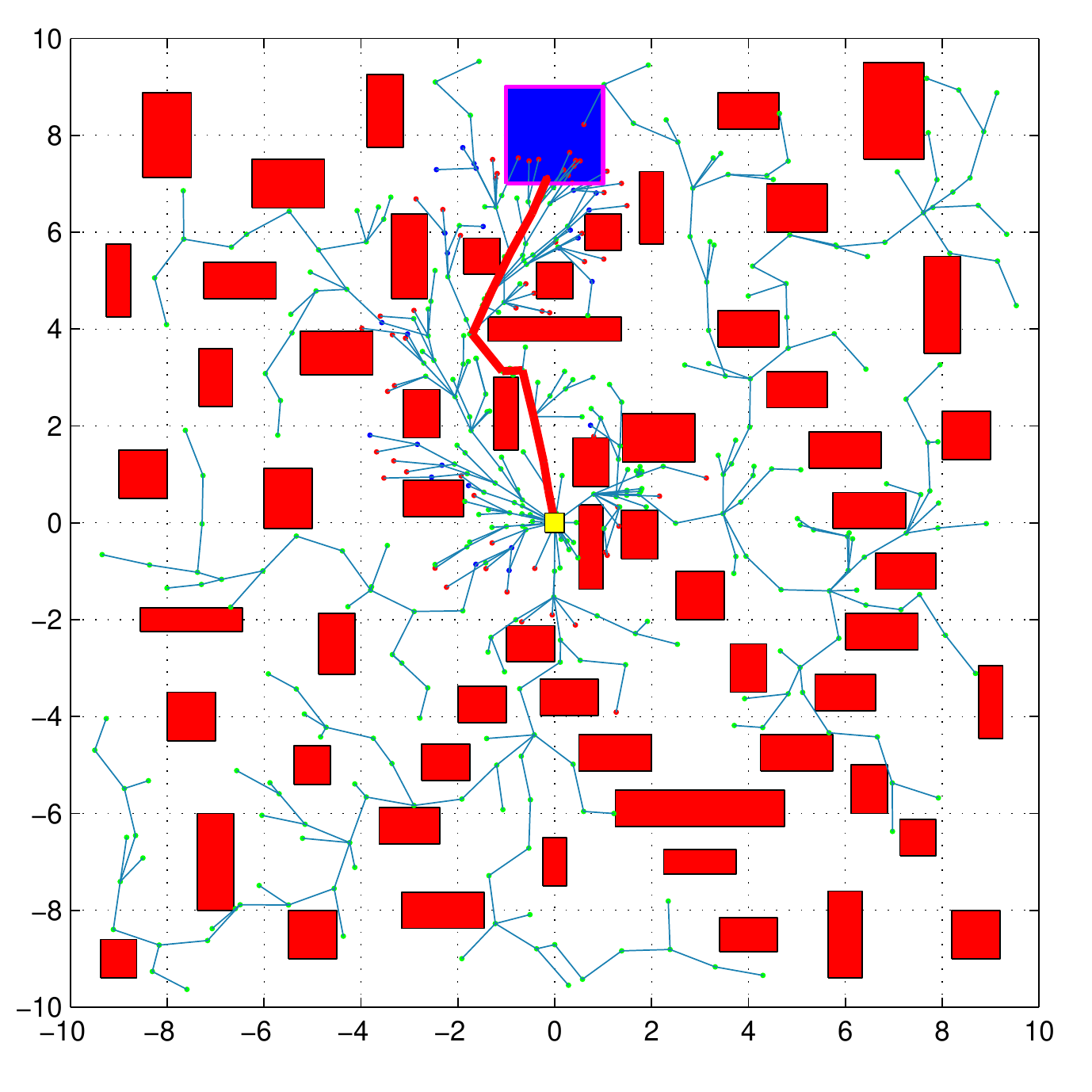}} \label{figure:pt3_rrtsharp_v2_it2500}}
    \subfigure[]{\scalebox{0.28}{\includegraphics[trim = 4.0cm 6.937cm 3.587cm 7.0cm, clip =
          true]{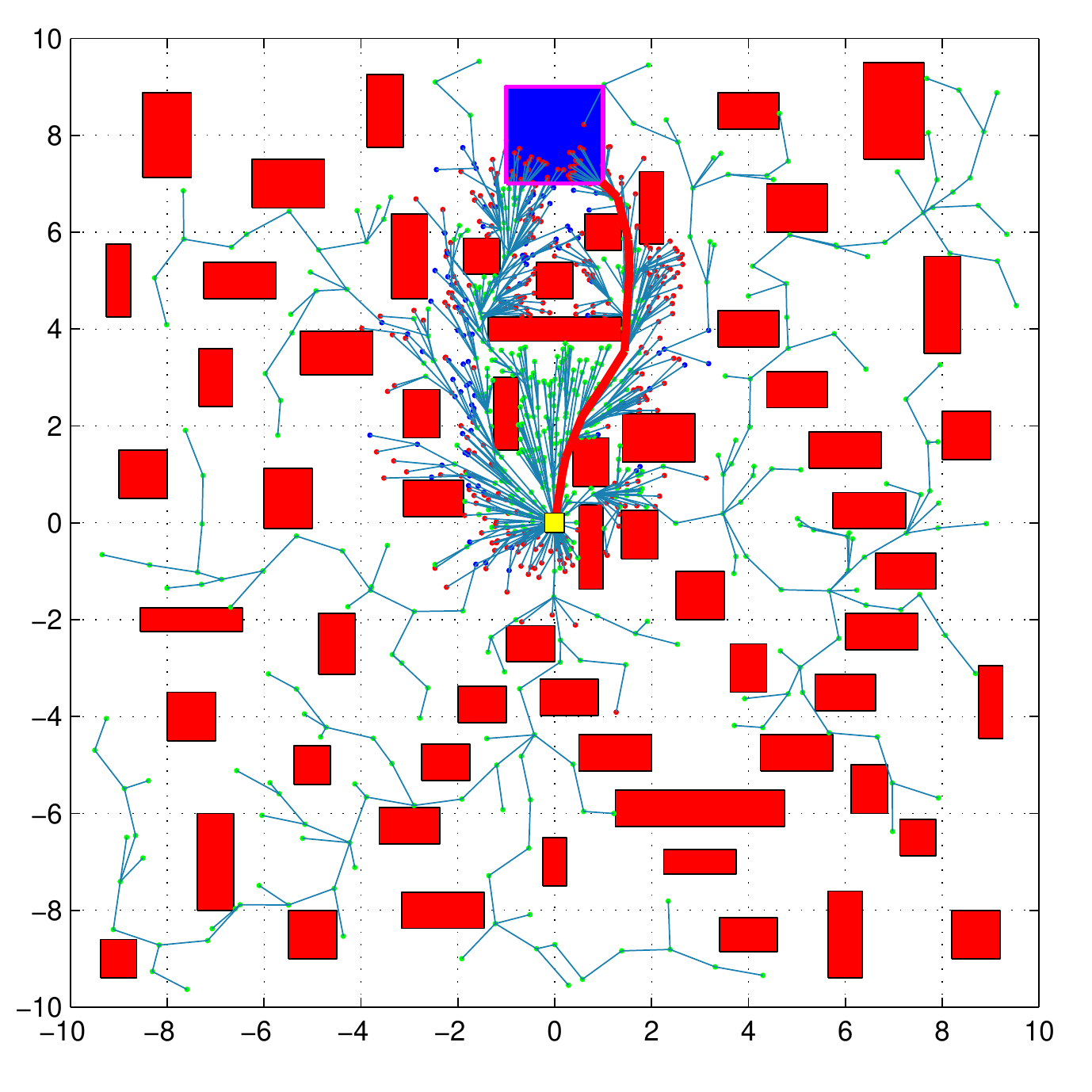}} \label{figure:pt3_rrtsharp_v2_it10000}}
   	}
	\mbox{
	\setcounter{subfigure}{4}
	\renewcommand{\thesubfigure}{(\alph{subfigure})}
    \subfigure[]{\scalebox{0.57}{\includegraphics[trim = 4.0cm 6.937cm 3.587cm 7.0cm, clip =
          true]{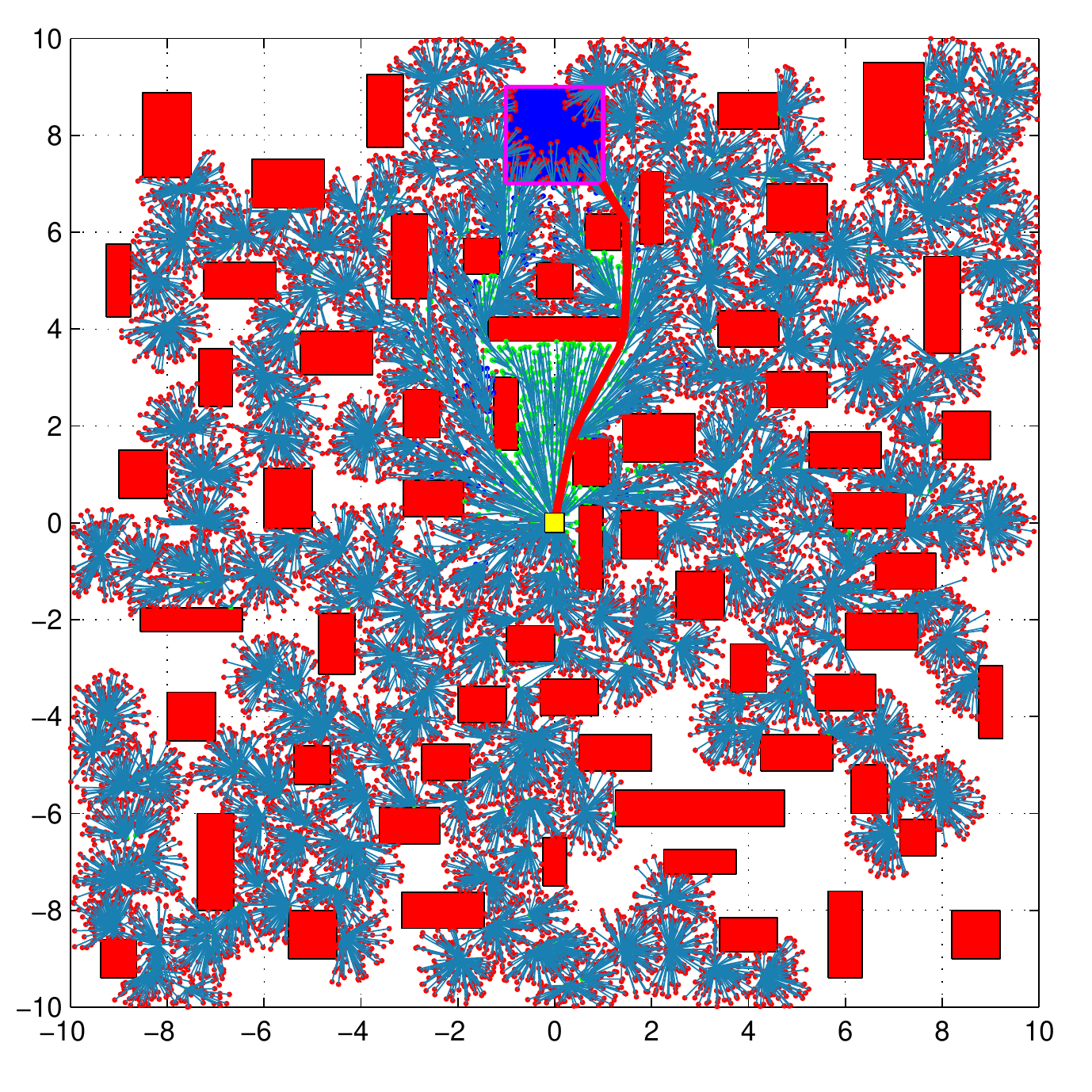}} \label{figure:pt3_rrtsharp_v1_it24999}}
	\setcounter{subfigure}{4}
	\renewcommand{\thesubfigure}{(\roman{subfigure})}
    \subfigure[]{\scalebox{0.57}{\includegraphics[trim = 4.0cm 6.937cm 3.587cm 7.0cm, clip =
          true]{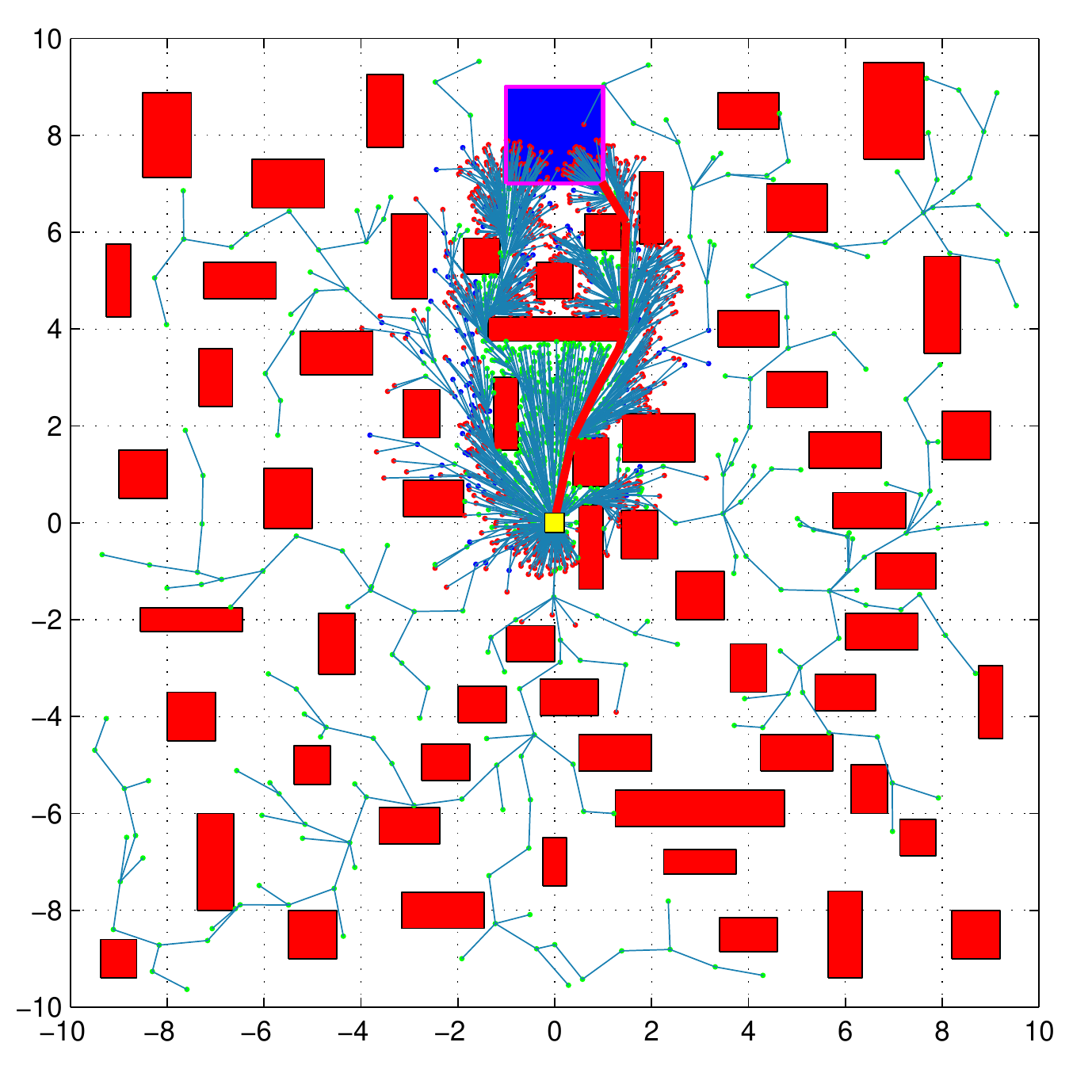}} \label{figure:pt3_rrtsharp_v2_it24999}}
    }

\caption{The evolution of the tree computed by \AlgRRTsharpNoBlackVertex{} and \AlgRRTsharpPromisingParent{} algorithms is shown in \subref{figure:pt3_rrtsharp_v1_it250}-\subref{figure:pt3_rrtsharp_v1_it24999} and \subref{figure:pt3_rrtsharp_v2_it250}-\subref{figure:pt3_rrtsharp_v2_it24999}, respectively. The configuration of the trees \subref{figure:pt3_rrtsharp_v1_it250}, \subref{figure:pt3_rrtsharp_v2_it250} is at 250 iterations, \subref{figure:pt3_rrtsharp_v1_it500}, \subref{figure:pt3_rrtsharp_v2_it500} is at 500 iterations, \subref{figure:pt3_rrtsharp_v1_it2500}, \subref{figure:pt3_rrtsharp_v2_it2500} is at 2500 iterations, \subref{figure:pt3_rrtsharp_v1_it10000}, \subref{figure:pt3_rrtsharp_v2_it10000} is at 10000 iterations,
and \subref{figure:pt3_rrtsharp_v1_it24999}, \subref{figure:pt3_rrtsharp_v2_it24999} is at 25000 iterations.}
    \label{figure:sim_d2_pt3_rrtsharp_v1_v2_iterations}
  \end{center}
\end{figure*}

\begin{figure*}[htp]
  \begin{center}

	\mbox{
	\setcounter{subfigure}{0}
	\renewcommand{\thesubfigure}{(\alph{subfigure})}
    \subfigure[]{\scalebox{0.28}{\includegraphics[trim = 4.0cm 6.937cm 3.587cm 7.0cm, clip =
          true]{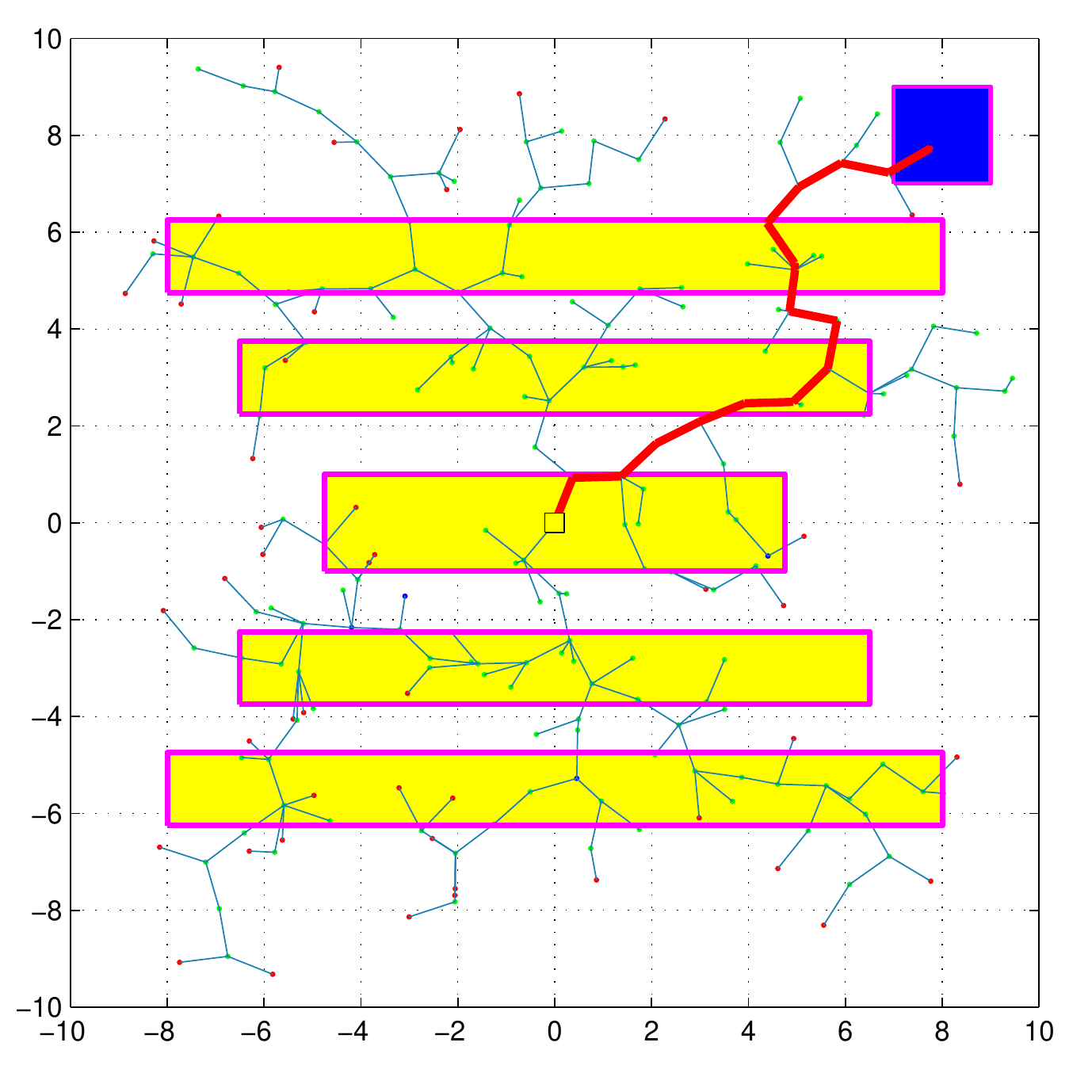}} \label{figure:pt4_rrtsharp_v1_it250}} 		
    \subfigure[]{\scalebox{0.28}{\includegraphics[trim = 4.0cm 6.937cm 3.587cm 7.0cm, clip =
          true]{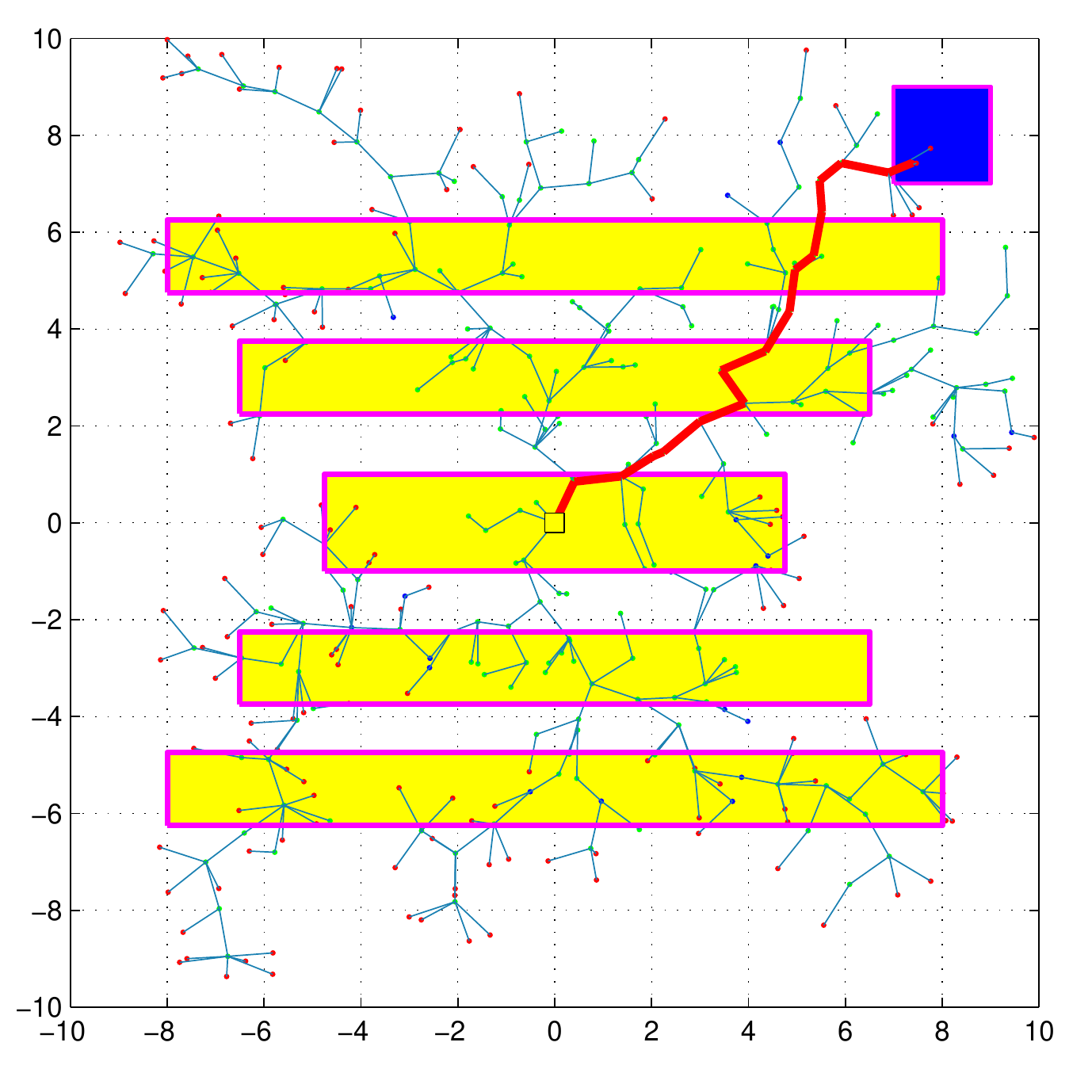}} \label{figure:pt4_rrtsharp_v1_it500}}
    \subfigure[]{\scalebox{0.28}{\includegraphics[trim = 4.0cm 6.937cm 3.587cm 7.0cm, clip =
          true]{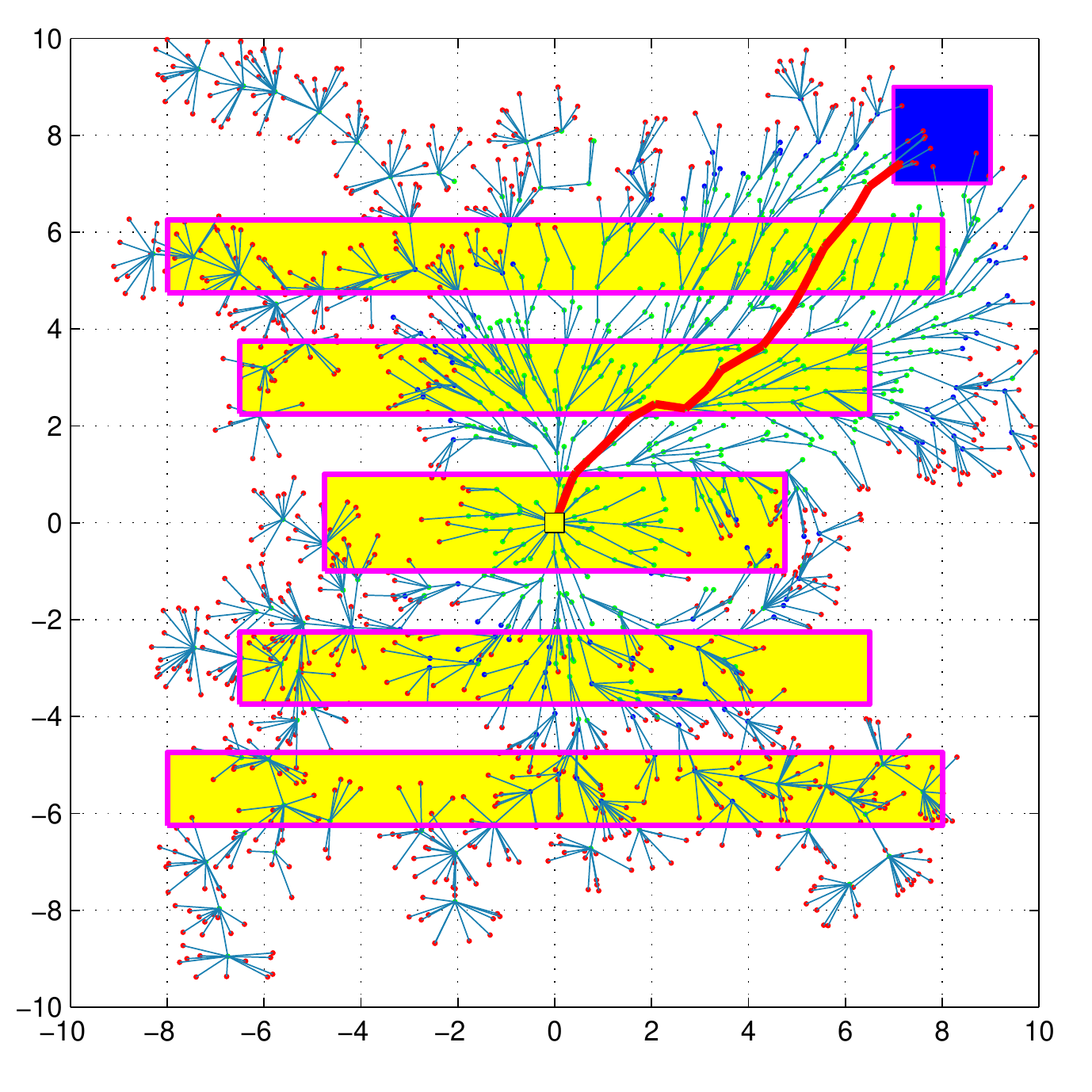}} \label{figure:pt4_rrtsharp_v1_it2500}}
    \subfigure[]{\scalebox{0.28}{\includegraphics[trim = 4.0cm 6.937cm 3.587cm 7.0cm, clip =
          true]{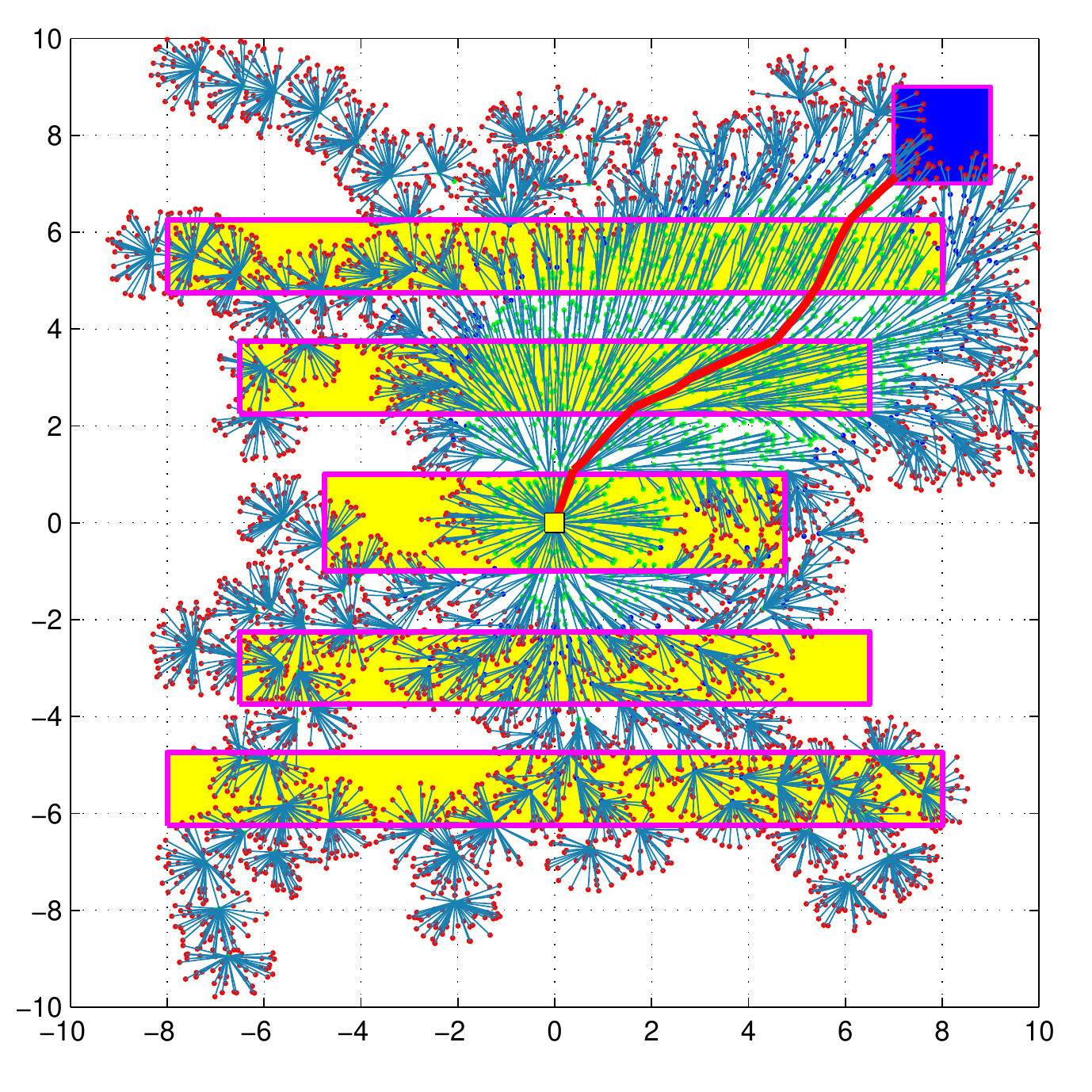}} \label{figure:pt4_rrtsharp_v1_it10000}}
    }
	\mbox{
	\setcounter{subfigure}{0}
	\renewcommand{\thesubfigure}{(\roman{subfigure})}
    \subfigure[]{\scalebox{0.28}{\includegraphics[trim = 4.0cm 6.937cm 3.587cm 7.0cm, clip =
          true]{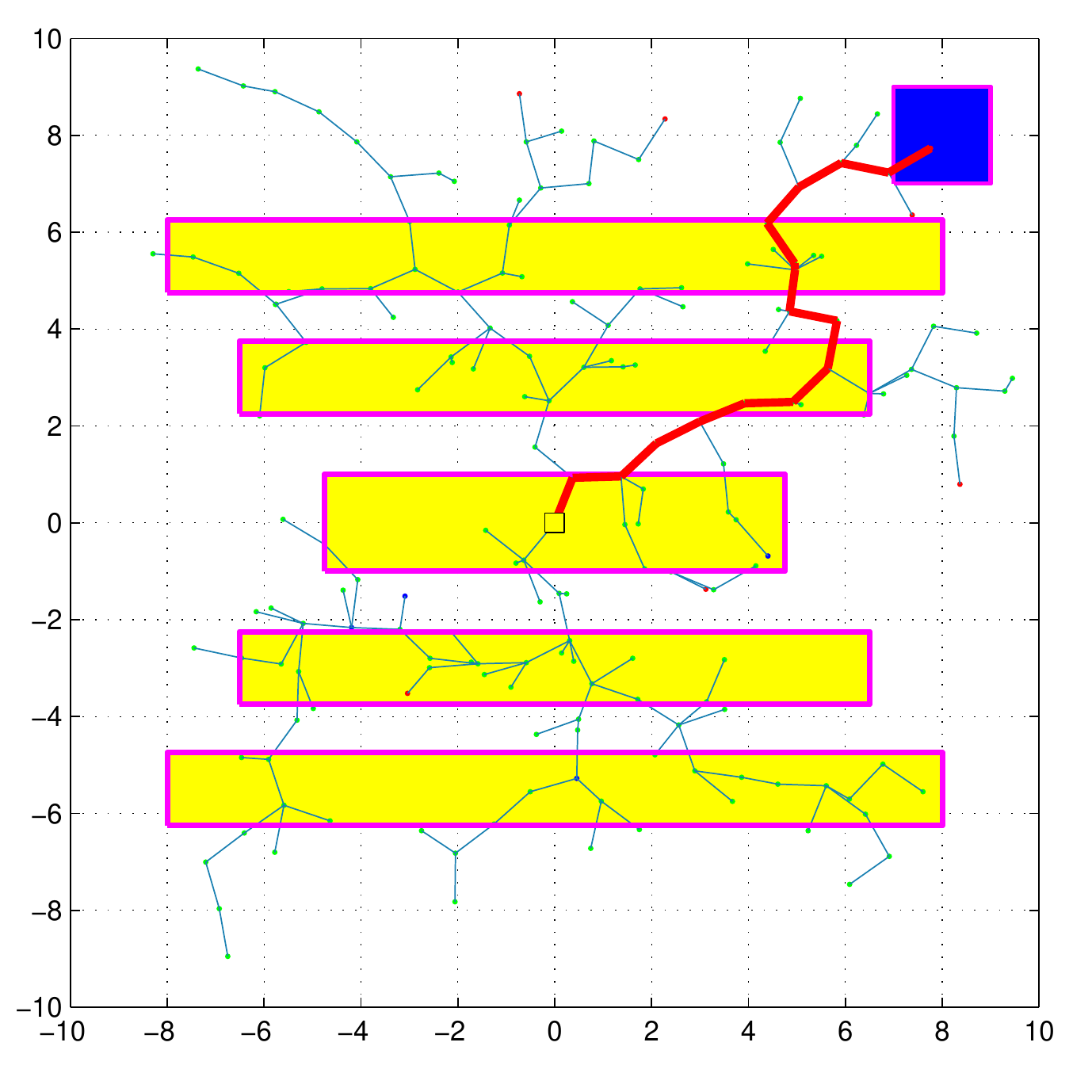}} \label{figure:pt4_rrtsharp_v2_it250}}
    \subfigure[]{\scalebox{0.28}{\includegraphics[trim = 4.0cm 6.937cm 3.587cm 7.0cm, clip =
          true]{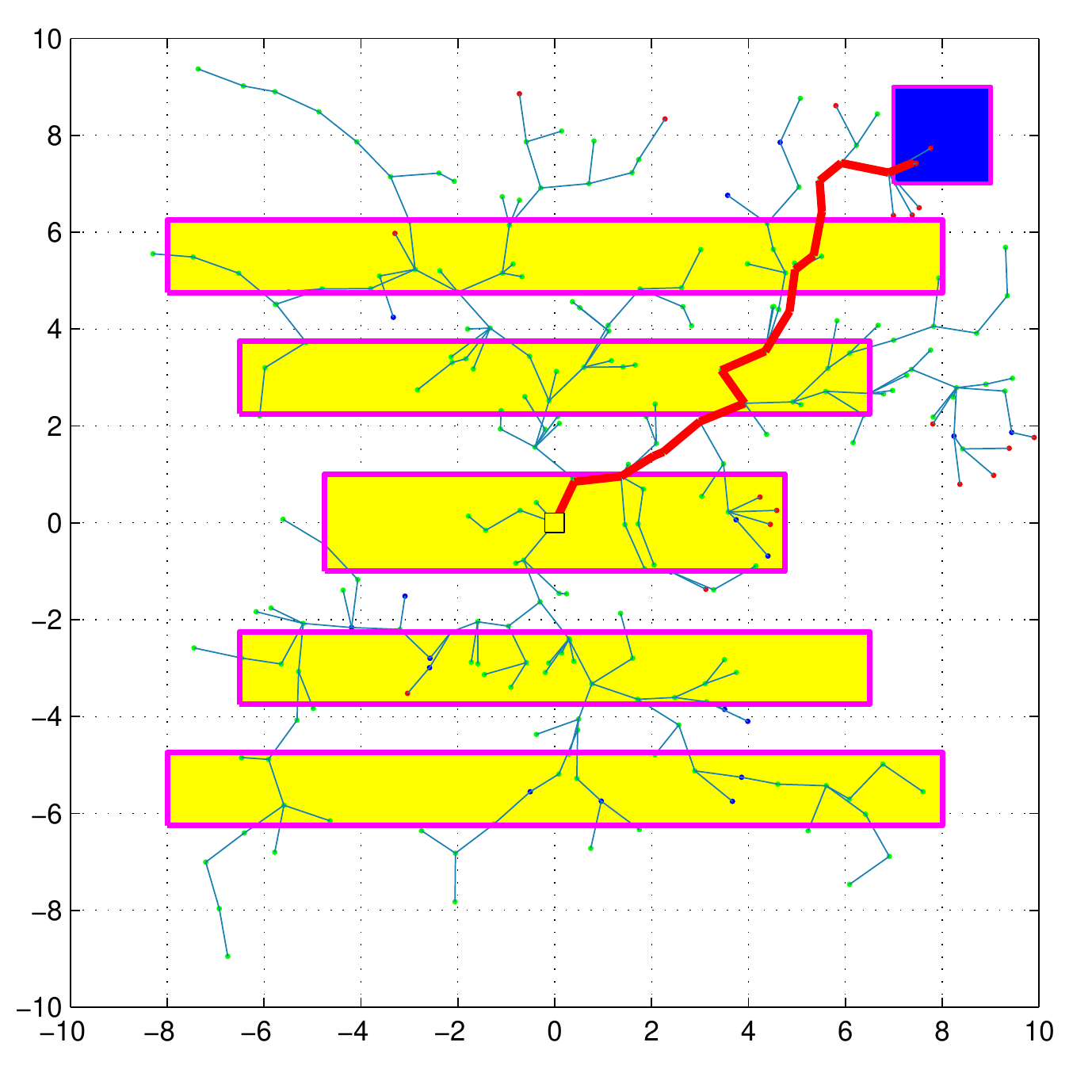}} \label{figure:pt4_rrtsharp_v2_it500}}
    \subfigure[]{\scalebox{0.28}{\includegraphics[trim = 4.0cm 6.937cm 3.587cm 7.0cm, clip =
          true]{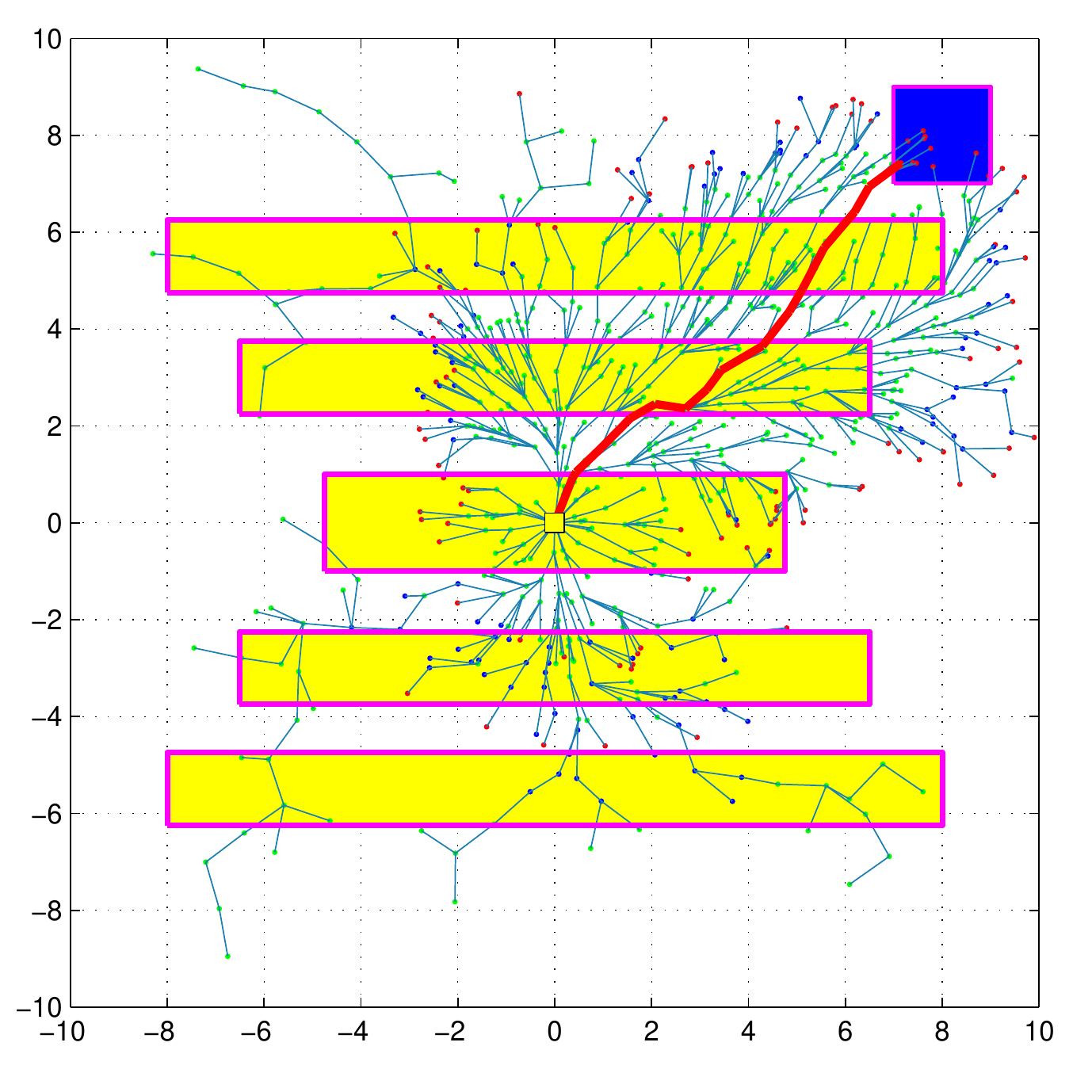}} \label{figure:pt4_rrtsharp_v2_it2500}}
    \subfigure[]{\scalebox{0.28}{\includegraphics[trim = 4.0cm 6.937cm 3.587cm 7.0cm, clip =
          true]{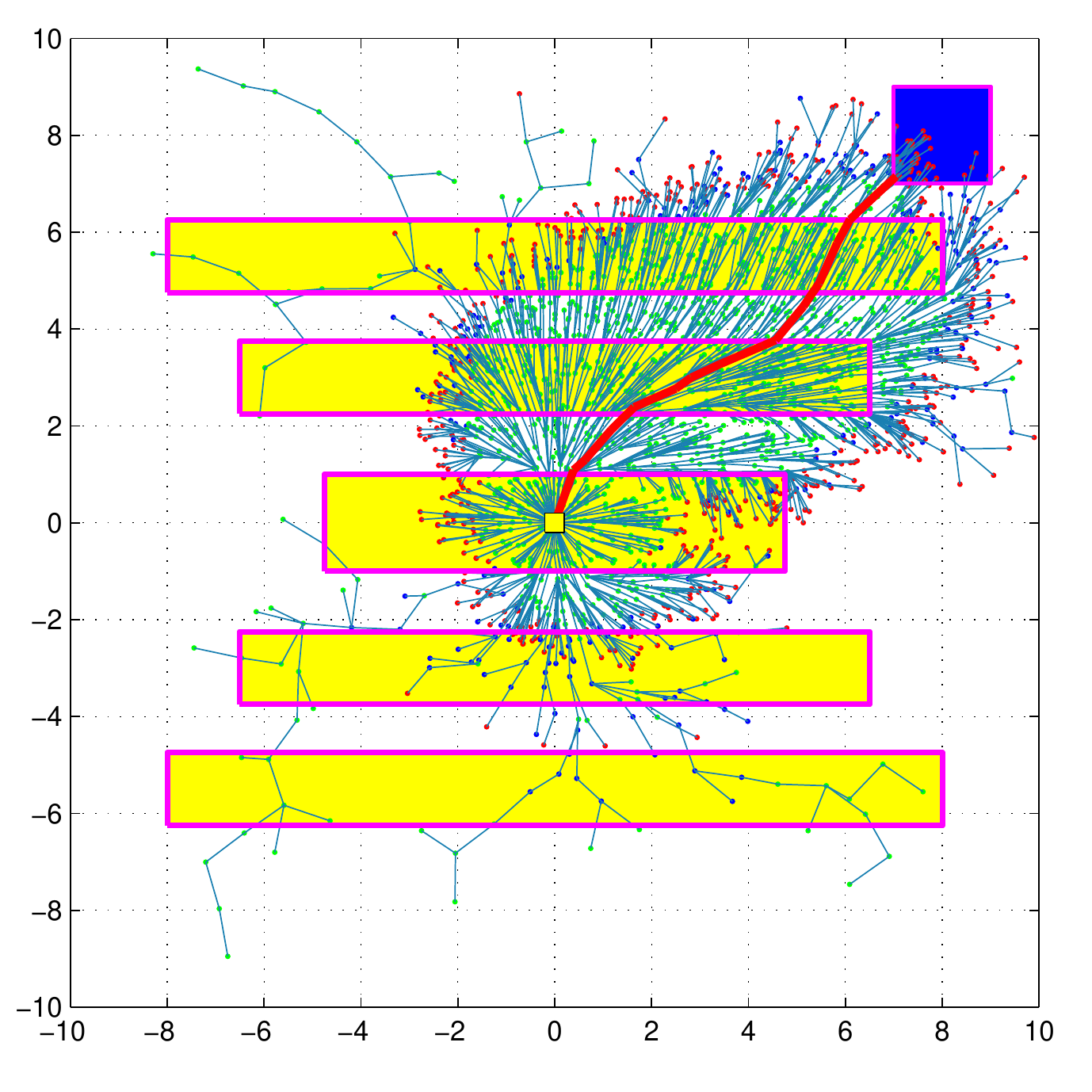}} \label{figure:pt4_rrtsharp_v2_it10000}}
   	}
	\mbox{
	\setcounter{subfigure}{4}
	\renewcommand{\thesubfigure}{(\alph{subfigure})}
    \subfigure[]{\scalebox{0.57}{\includegraphics[trim = 4.0cm 6.937cm 3.587cm 7.0cm, clip =
          true]{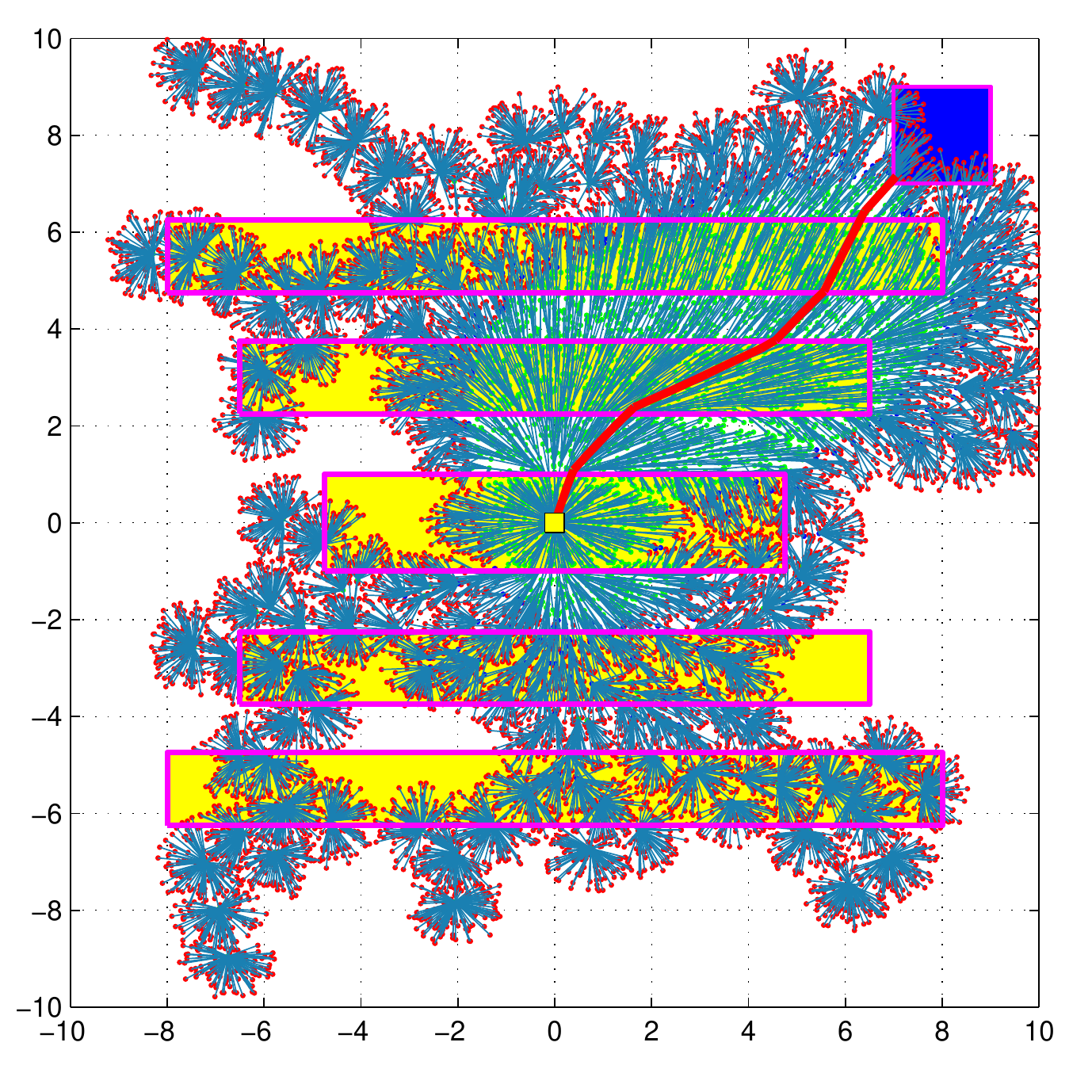}} \label{figure:pt4_rrtsharp_v1_it24999}}
\setcounter{subfigure}{4}
\renewcommand{\thesubfigure}{(\roman{subfigure})}
    \subfigure[]{\scalebox{0.57}{\includegraphics[trim = 4.0cm 6.937cm 3.587cm 7.0cm, clip =
          true]{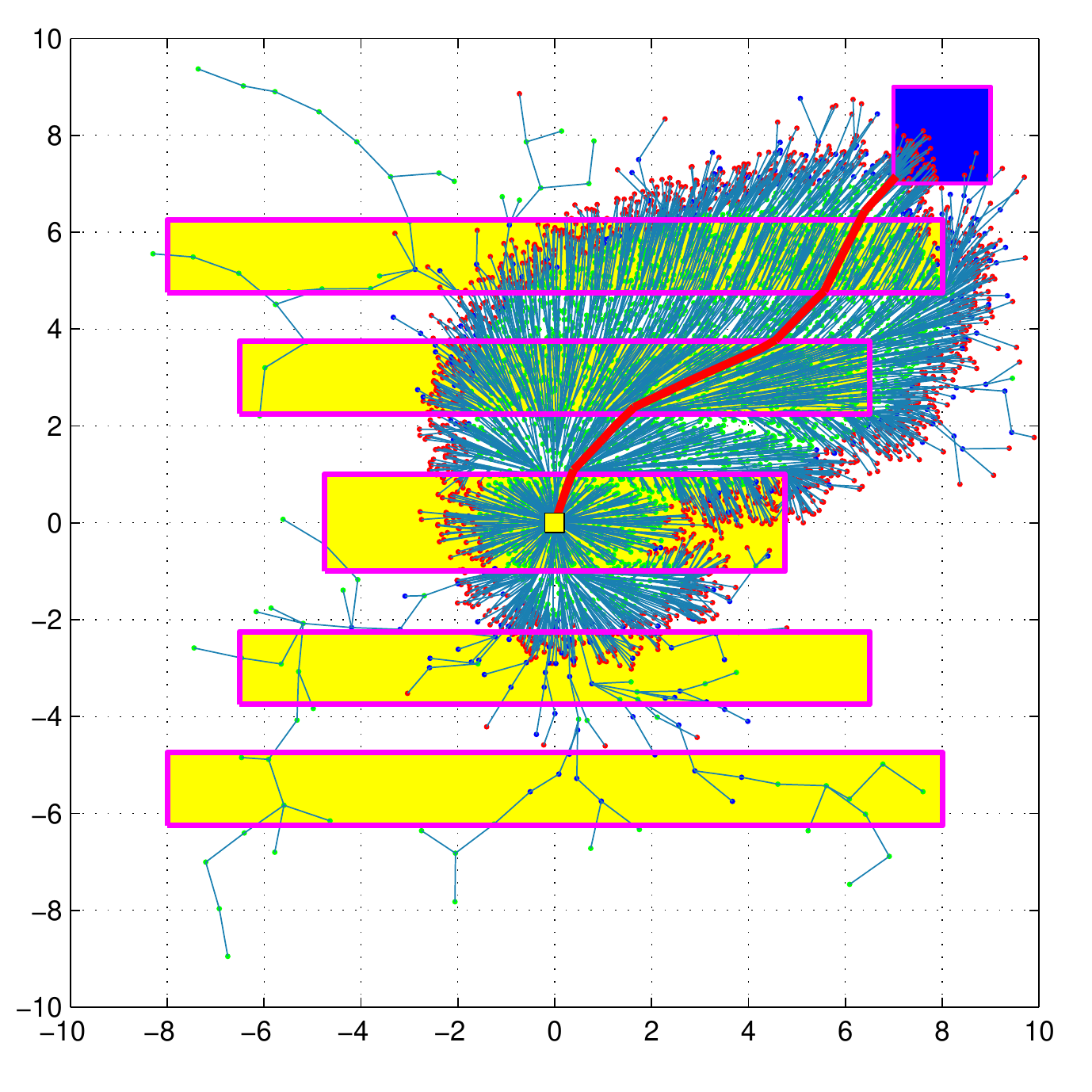}} \label{figure:pt4_rrtsharp_v2_it24999}}
    }

\caption{The evolution of the tree computed by \AlgRRTsharpNoBlackVertex{} and \AlgRRTsharpPromisingParent{} algorithms is shown in \subref{figure:pt4_rrtsharp_v1_it250}-\subref{figure:pt4_rrtsharp_v1_it24999} and \subref{figure:pt4_rrtsharp_v2_it250}-\subref{figure:pt4_rrtsharp_v2_it24999}, respectively. The configuration of the trees \subref{figure:pt4_rrtsharp_v1_it250}, \subref{figure:pt4_rrtsharp_v2_it250} is at 250 iterations, \subref{figure:pt4_rrtsharp_v1_it500}, \subref{figure:pt4_rrtsharp_v2_it500} is at 500 iterations, \subref{figure:pt4_rrtsharp_v1_it2500}, \subref{figure:pt4_rrtsharp_v2_it2500} is at 2500 iterations, \subref{figure:pt4_rrtsharp_v1_it10000}, \subref{figure:pt4_rrtsharp_v2_it10000} is at 10000 iterations,
and \subref{figure:pt4_rrtsharp_v1_it24999}, \subref{figure:pt4_rrtsharp_v2_it24999} is at 25000 iterations.}
    \label{figure:sim_d2_pt4_rrtsharp_v1_v2_iterations}
  \end{center}
\end{figure*}

The same experiments as before were carried out for the three variants of the \AlgRRTsharp{} algorithm. As seen in the figures below, all variants successfully prevent the inclusion of vertices which lie in the unfavorable regions of the search space. As seen in Figures~\ref{figure:pt1_rrtsharp_v1_it24999}, \ref{figure:pt2_rrtsharp_v1_it24999}, \ref{figure:pt3_rrtsharp_v1_it24999}, and \ref{figure:pt4_rrtsharp_v1_it24999}, the \AlgRRTsharpNoBlackVertex{} algorithm does not include any black vertices in the tree (these are the vertices that are consistent with infinite key value, hence non-promising), but still computes a solution to the problem, which is as good as the one computed by the \AlgRRTstar{} and \AlgRRTsharp{} algorithms. However, there are still many red (i.e., non-promising and inconsistent with infinite g-value and finite lmc-value) vertices included in the tree. This is owing to the fact that they are never made consistent until the last iteration, since they mostly lie outside of $\Xrel$. Therefore, they remain in the priority queue and need to be sorted during each iteration. This makes the $\PrcReduceInconsistency$ procedure slower. In the \AlgRRTsharpPromisingParent{} algorithm, the number of red vertices included into the tree is reduced by simply enforcing to have a promising parent vertex for the new vertex that is considered for extension. Red vertices are mostly included into the branches of the tree that are formed outside of the $\Xrel$ during exploration phase. As seen in Figures~\ref{figure:pt1_rrtsharp_v2_it24999}, \ref{figure:pt2_rrtsharp_v2_it24999}, \ref{figure:pt3_rrtsharp_v2_it24999}, and \ref{figure:pt4_rrtsharp_v2_it24999}, the \AlgRRTsharpPromisingParent{} algorithm tends not to include vertices into the branches of the tree which are very far away from the optimal solution. Lastly, the \AlgRRTsharpPromisingNewVertex{} algorithm includes a new vertex into the tree only if it is a promising one. Therefore, all vertices in the tree, other than the goal vertices, are either green or blue, which are located around the boundary of $\Xrel$.

\begin{figure*}[htp]
  \begin{center}
	\mbox{
        \tikzmark{lr1st}\subfigure[]{\scalebox{0.26}{\includegraphics[trim = 4.0cm 3.0cm 4.0cm 3.0cm, clip =
          true]{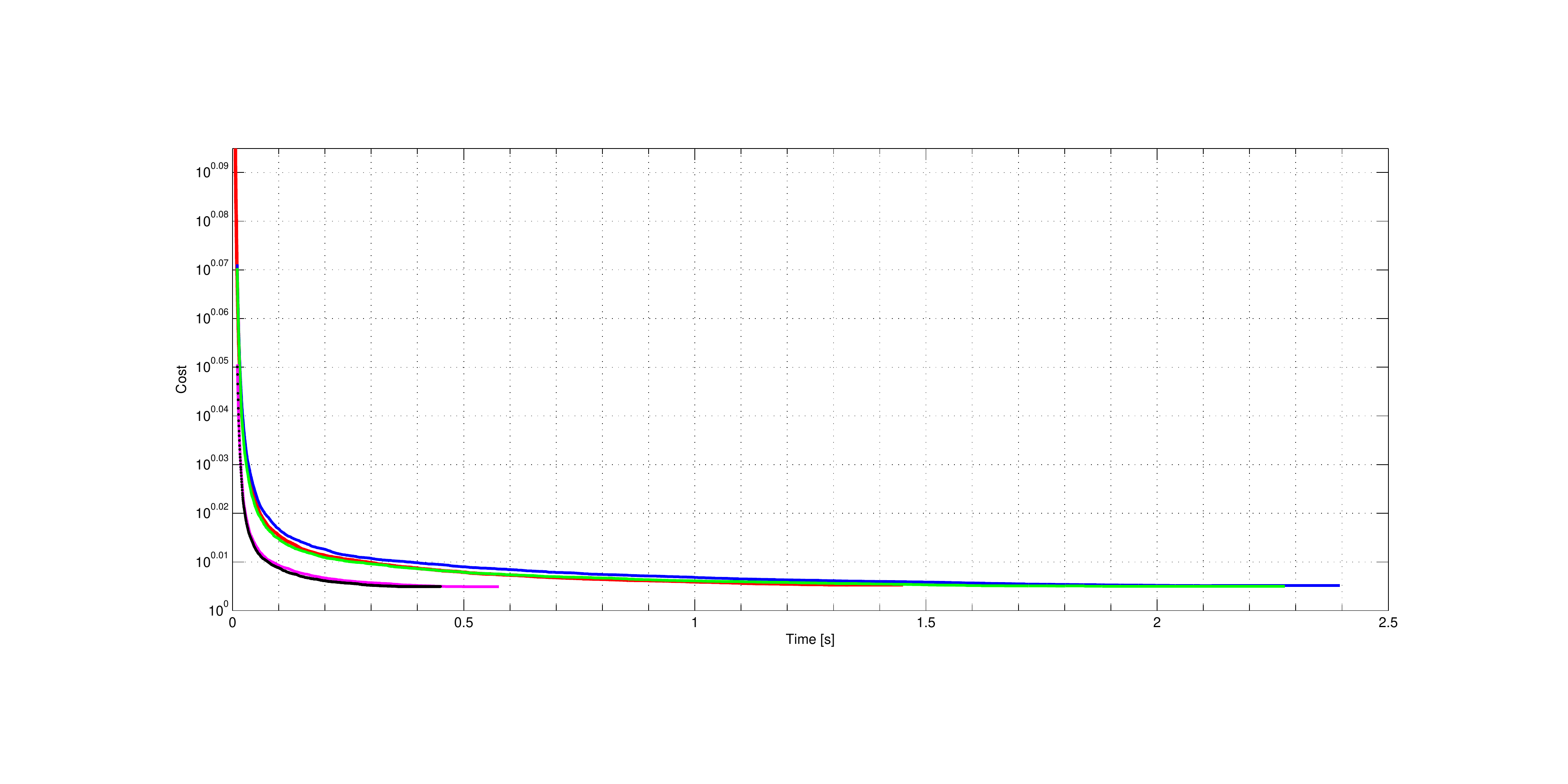}}\label{figure:time_cost_mean_d2_pt1_all}}\tikzmark{lr1en}

        \tikzmark{lr2st}\subfigure[]{\scalebox{0.26}{\includegraphics[trim = 4.0cm 3.0cm 4.0cm 3.0cm, clip =
          true]{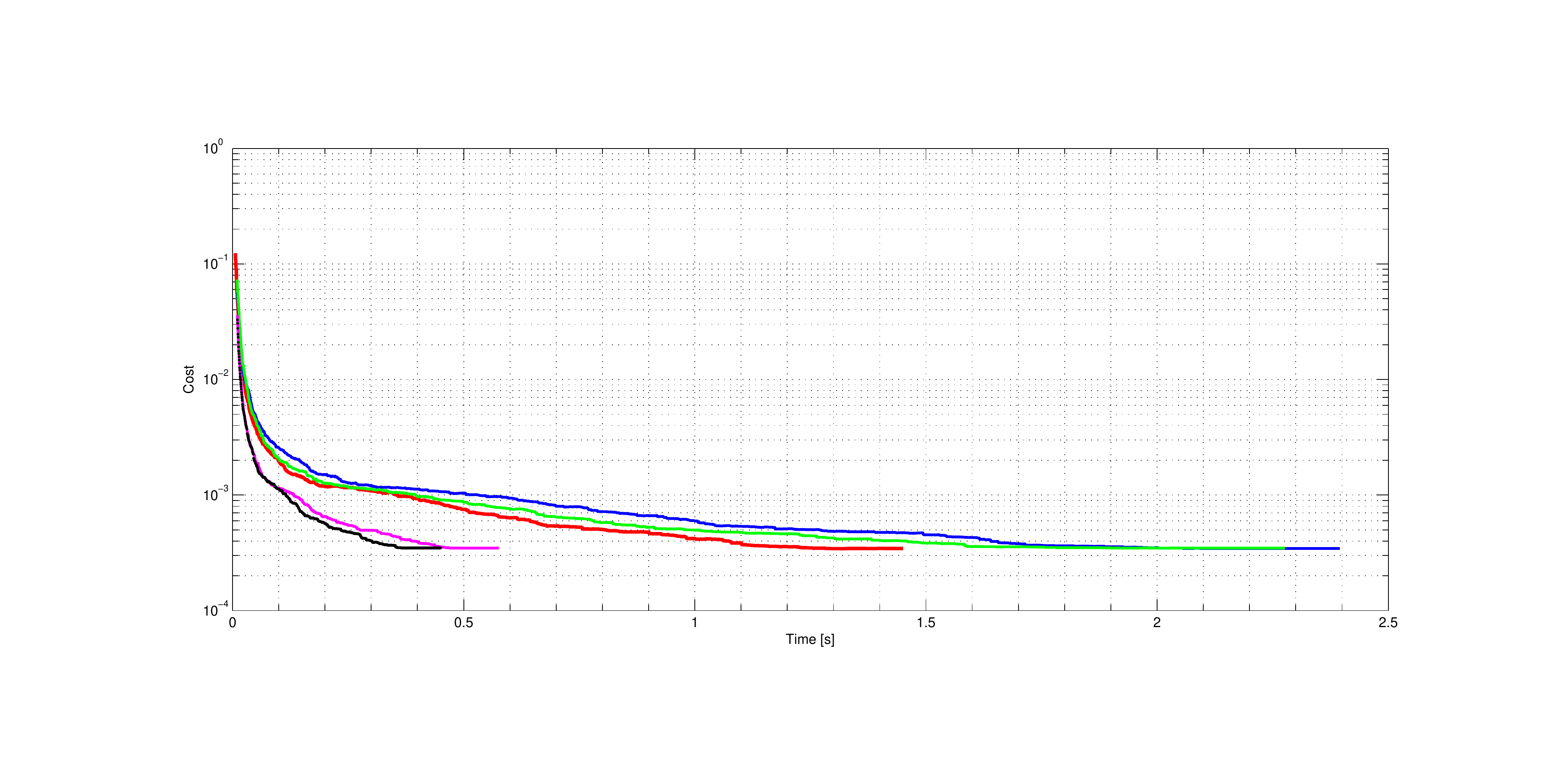}}\label{figure:time_cost_variance_d2_pt1_all}}\tikzmark{lr2en}

	\begin{tikzpicture}[
	remember picture,
	overlay,
	init/.style={inner sep=0pt}]
	\tiny
	\coordinate (lr1s) at ($(lr1st)+(6.7,1.95)$);
	\coordinate (lr1e) at ($(lr1en)+(-0.3,3.3)$);
	\coordinate (l1s) at ($(lr1st)+(6.7,3.25)$);
	\coordinate (l1e) at ($(lr1st)+(7.1,3.25)$);
	\coordinate (l2s) at ($(lr1st)+(6.7,2.95)$);
	\coordinate (l2e) at ($(lr1st)+(7.1,2.95)$);
	\coordinate (l3s) at ($(lr1st)+(6.7,2.65)$);
	\coordinate (l3e) at ($(lr1st)+(7.1,2.65)$);
	\coordinate (l4s) at ($(lr1st)+(6.7,2.35)$);
	\coordinate (l4e) at ($(lr1st)+(7.1,2.35)$);
	\coordinate (l5s) at ($(lr1st)+(6.7,2.05)$);
	\coordinate (l5e) at ($(lr1st)+(7.1,2.05)$);
	
	\node[draw=black,rectangle, fit=(lr1s) (lr1e)] {};
	\node[init] at (l1s) (n1s) {};
	\node[init] at (l1e) (n1e) [label=right:{\fontsize{0.5mm}{1mm}\selectfont\AlgRRTstar}]{};
	\node[init] at (l2s) (n2s) {};
	\node[init] at (l2e) (n2e) [label=right:{\fontsize{0.5mm}{1mm}\selectfont\AlgRRTsharp}]{};
	\node[init] at (l3s) (n3s) {};
	\node[init] at (l3e) (n3e) [label=right:{\fontsize{0.5mm}{1mm}\selectfont\AlgRRTsharpNoBlackVertex}] {};
	\node[init] at (l4s) (n4s) {};
	\node[init] at (l4e) (n4e) [label=right:{\fontsize{0.5mm}{1mm}\selectfont\AlgRRTsharpPromisingParent}]{};
	\node[init] at (l5s) (n5s) {};
	\node[init] at (l5e) (n5e) [label=right:{\fontsize{0.5mm}{1mm}\selectfont\AlgRRTsharpPromisingNewVertex}] {};
	\draw[red,thick] (n1s) -- (n1e);
	\draw[blue,thick] (n2s) -- (n2e);
	\draw[green,thick] (n3s) -- (n3e);
	\draw[magenta,thick] (n4s) -- (n4e);
	\draw[black,thick] (n5s) -- (n5e);
	
	\coordinate (lr2s) at ($(lr2st)+(6.7,1.95)$);
	\coordinate (lr2e) at ($(lr2en)+(-0.3,3.3)$);
	\coordinate (l6s) at ($(lr2st)+(6.7,3.25)$);
	\coordinate (l6e) at ($(lr2st)+(7.1,3.25)$);
	\coordinate (l7s) at ($(lr2st)+(6.7,2.95)$);
	\coordinate (l7e) at ($(lr2st)+(7.1,2.95)$);
	\coordinate (l8s) at ($(lr2st)+(6.7,2.65)$);
	\coordinate (l8e) at ($(lr2st)+(7.1,2.65)$);
	\coordinate (l9s) at ($(lr2st)+(6.7,2.35)$);
	\coordinate (l9e) at ($(lr2st)+(7.1,2.35)$);
	\coordinate (l10s) at ($(lr2st)+(6.7,2.05)$);
	\coordinate (l10e) at ($(lr2st)+(7.1,2.05)$);
	
	\node[draw=black,rectangle, fit=(lr2s) (lr2e)] {};
	\node[init] at (l6s) (n6s) {};
	\node[init] at (l6e) (n6e) [label=right:{\fontsize{0.5mm}{1mm}\selectfont\AlgRRTstar}]{};
	\node[init] at (l7s) (n7s) {};
	\node[init] at (l7e) (n7e) [label=right:{\fontsize{0.5mm}{1mm}\selectfont\AlgRRTsharp}]{};
	\node[init] at (l8s) (n8s) {};
	\node[init] at (l8e) (n8e) [label=right:{\fontsize{0.5mm}{1mm}\selectfont\AlgRRTsharpNoBlackVertex}] {};
	\node[init] at (l9s) (n9s) {};
	\node[init] at (l9e) (n9e) [label=right:{\fontsize{0.5mm}{1mm}\selectfont\AlgRRTsharpPromisingParent}]{};
	\node[init] at (l10s) (n10s) {};
	\node[init] at (l10e) (n10e) [label=right:{\fontsize{0.5mm}{1mm}\selectfont\AlgRRTsharpPromisingNewVertex}] {};
	\draw[red,thick] (n6s) -- (n6e);
	\draw[blue,thick] (n7s) -- (n7e);
	\draw[green,thick] (n8s) -- (n8e);
	\draw[magenta,thick] (n9s) -- (n9e);
	\draw[black,thick] (n10s) -- (n10e);	
\end{tikzpicture}

    }


    \caption{The change in the cost of the best paths computed by \AlgRRTstar{}, \AlgRRTsharp{}, and its variant algorithms and the variance in the trials are shown in \subref{figure:time_cost_mean_d2_pt1_all} and \subref{figure:time_cost_variance_d2_pt1_all}, respectively.}
    \label{figure:sim_d2_pt1_all_histories}
  \end{center}
\end{figure*}

\begin{figure*}[htp]
  \begin{center}
	\mbox{
        \tikzmark{lr1st}\subfigure[]{\scalebox{0.26}{\includegraphics[trim = 4.0cm 3.0cm 4.0cm 3.0cm, clip =
          true]{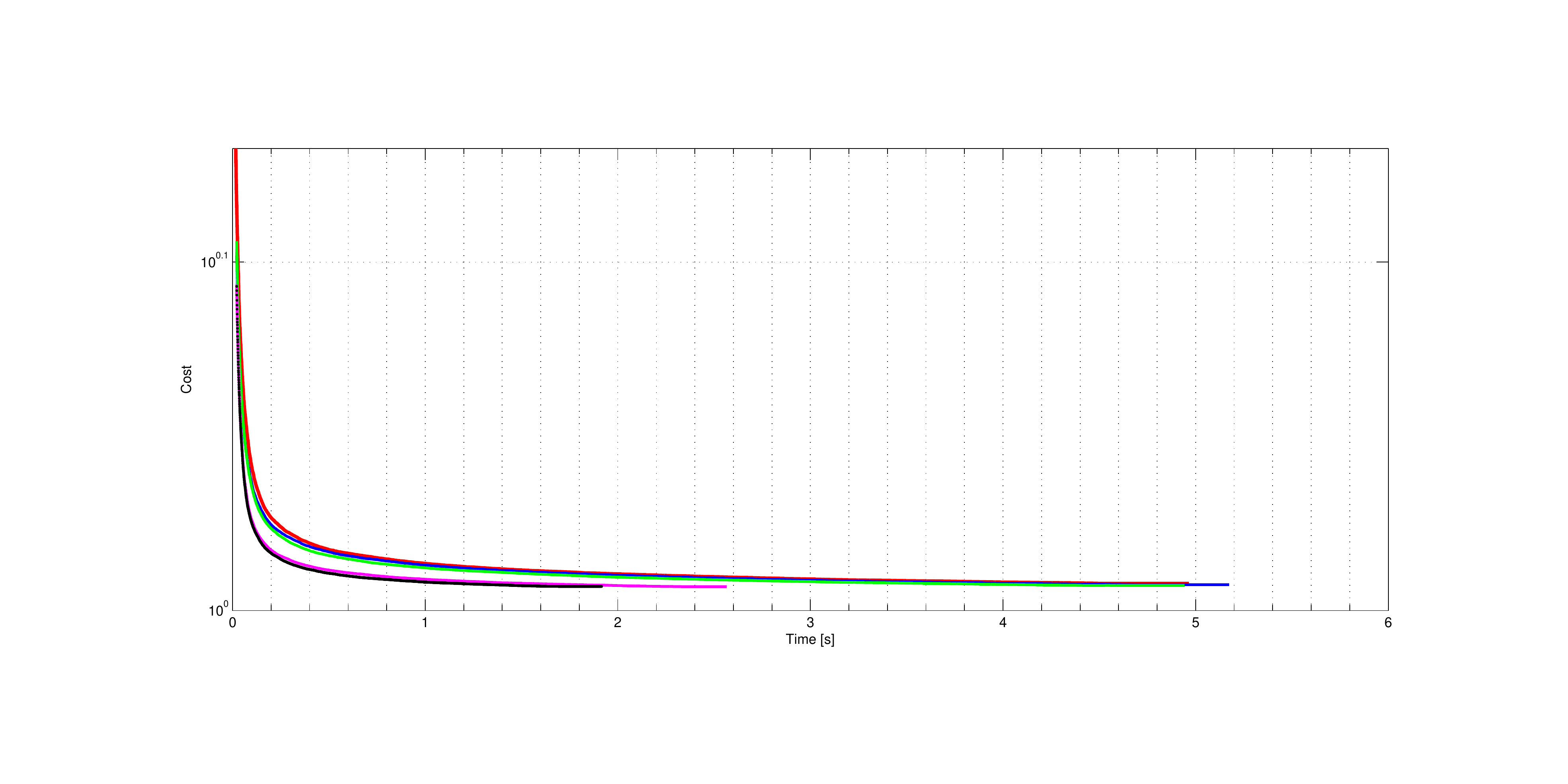}}\label{figure:time_cost_mean_d2_pt2_all}}\tikzmark{lr1en}
  		\tikzmark{lr2st}\subfigure[]{\scalebox{0.26}{\includegraphics[trim = 4.0cm 3.0cm 4.0cm 3.0cm, clip =
          true]{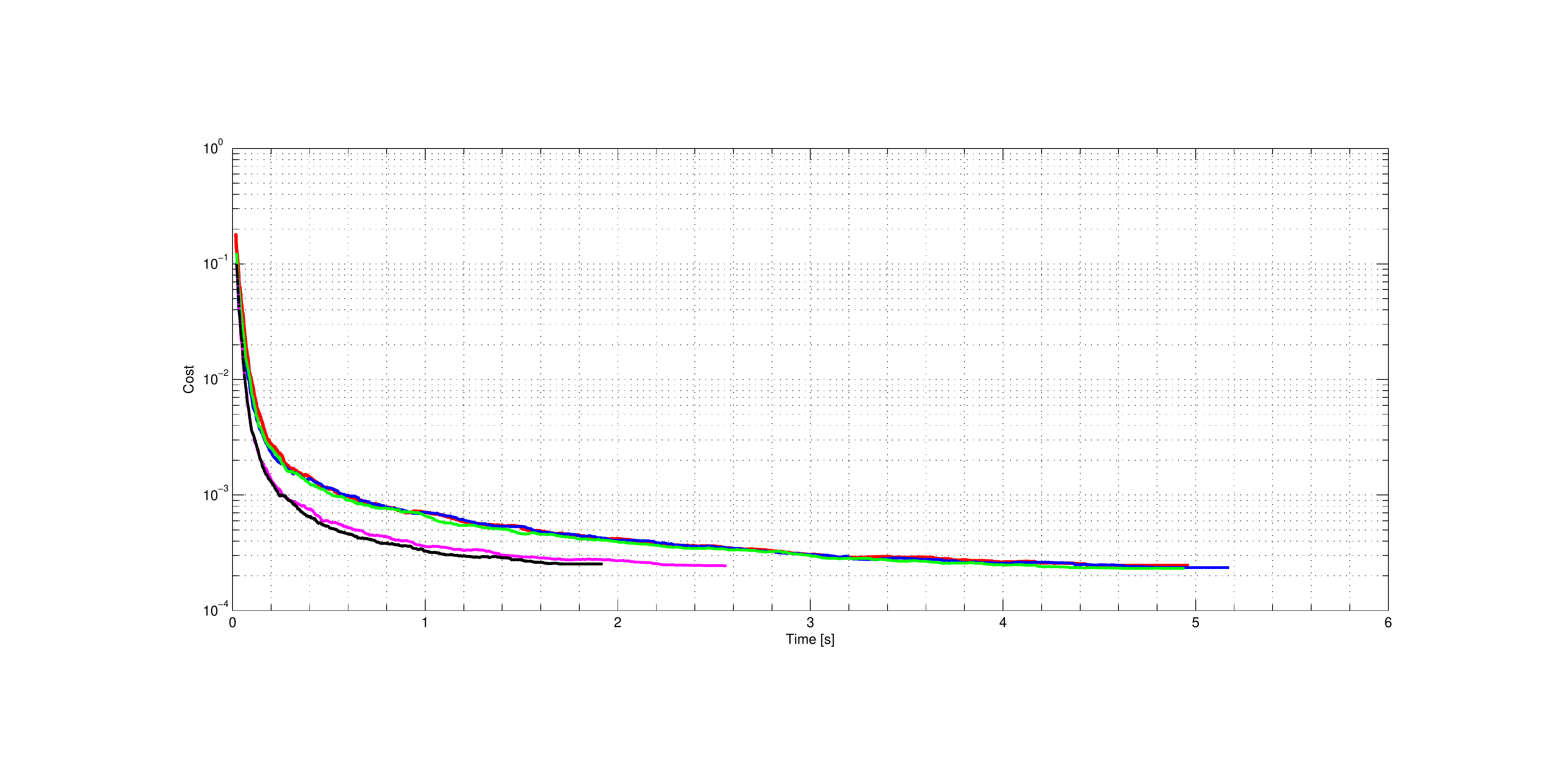}}\label{figure:time_cost_variance_d2_pt2_all}}\tikzmark{lr2en}
	}


    \caption{The change in the cost of the best paths computed by \AlgRRTstar{}, \AlgRRTsharp{}, and its variant algorithms and the variance in the trials are shown in \subref{figure:time_cost_mean_d2_pt2_all} and \subref{figure:time_cost_variance_d2_pt2_all}, respectively.}
    \label{figure:sim_d2_pt2_all_histories}

    \begin{tikzpicture}[
	remember picture,
	overlay,
	init/.style={inner sep=0pt}]
	\tiny
	\coordinate (lr1s) at ($(lr1st)+(6.7,1.95)$);
	\coordinate (lr1e) at ($(lr1en)+(-0.3,3.3)$);
	\coordinate (l1s) at ($(lr1st)+(6.7,3.25)$);
	\coordinate (l1e) at ($(lr1st)+(7.1,3.25)$);
	\coordinate (l2s) at ($(lr1st)+(6.7,2.95)$);
	\coordinate (l2e) at ($(lr1st)+(7.1,2.95)$);
	\coordinate (l3s) at ($(lr1st)+(6.7,2.65)$);
	\coordinate (l3e) at ($(lr1st)+(7.1,2.65)$);
	\coordinate (l4s) at ($(lr1st)+(6.7,2.35)$);
	\coordinate (l4e) at ($(lr1st)+(7.1,2.35)$);
	\coordinate (l5s) at ($(lr1st)+(6.7,2.05)$);
	\coordinate (l5e) at ($(lr1st)+(7.1,2.05)$);
	
	\node[draw=black,rectangle, fit=(lr1s) (lr1e)] {};
	\node[init] at (l1s) (n1s) {};
	\node[init] at (l1e) (n1e) [label=right:{\fontsize{0.5mm}{1mm}\selectfont\AlgRRTstar}]{};
	\node[init] at (l2s) (n2s) {};
	\node[init] at (l2e) (n2e) [label=right:{\fontsize{0.5mm}{1mm}\selectfont\AlgRRTsharp}]{};
	\node[init] at (l3s) (n3s) {};
	\node[init] at (l3e) (n3e) [label=right:{\fontsize{0.5mm}{1mm}\selectfont\AlgRRTsharpNoBlackVertex}] {};
	\node[init] at (l4s) (n4s) {};
	\node[init] at (l4e) (n4e) [label=right:{\fontsize{0.5mm}{1mm}\selectfont\AlgRRTsharpPromisingParent}]{};
	\node[init] at (l5s) (n5s) {};
	\node[init] at (l5e) (n5e) [label=right:{\fontsize{0.5mm}{1mm}\selectfont\AlgRRTsharpPromisingNewVertex}] {};
	\draw[red,thick] (n1s) -- (n1e);
	\draw[blue,thick] (n2s) -- (n2e);
	\draw[green,thick] (n3s) -- (n3e);
	\draw[magenta,thick] (n4s) -- (n4e);
	\draw[black,thick] (n5s) -- (n5e);
	
	\coordinate (lr2s) at ($(lr2st)+(6.7,1.95)$);
	\coordinate (lr2e) at ($(lr2en)+(-0.3,3.3)$);
	\coordinate (l6s) at ($(lr2st)+(6.7,3.25)$);
	\coordinate (l6e) at ($(lr2st)+(7.1,3.25)$);
	\coordinate (l7s) at ($(lr2st)+(6.7,2.95)$);
	\coordinate (l7e) at ($(lr2st)+(7.1,2.95)$);
	\coordinate (l8s) at ($(lr2st)+(6.7,2.65)$);
	\coordinate (l8e) at ($(lr2st)+(7.1,2.65)$);
	\coordinate (l9s) at ($(lr2st)+(6.7,2.35)$);
	\coordinate (l9e) at ($(lr2st)+(7.1,2.35)$);
	\coordinate (l10s) at ($(lr2st)+(6.7,2.05)$);
	\coordinate (l10e) at ($(lr2st)+(7.1,2.05)$);
	
	\node[draw=black,rectangle, fit=(lr2s) (lr2e)] {};
	\node[init] at (l6s) (n6s) {};
	\node[init] at (l6e) (n6e) [label=right:{\fontsize{0.5mm}{1mm}\selectfont\AlgRRTstar}]{};
	\node[init] at (l7s) (n7s) {};
	\node[init] at (l7e) (n7e) [label=right:{\fontsize{0.5mm}{1mm}\selectfont\AlgRRTsharp}]{};
	\node[init] at (l8s) (n8s) {};
	\node[init] at (l8e) (n8e) [label=right:{\fontsize{0.5mm}{1mm}\selectfont\AlgRRTsharpNoBlackVertex}] {};
	\node[init] at (l9s) (n9s) {};
	\node[init] at (l9e) (n9e) [label=right:{\fontsize{0.5mm}{1mm}\selectfont\AlgRRTsharpPromisingParent}]{};
	\node[init] at (l10s) (n10s) {};
	\node[init] at (l10e) (n10e) [label=right:{\fontsize{0.5mm}{1mm}\selectfont\AlgRRTsharpPromisingNewVertex}] {};
	\draw[red,thick] (n6s) -- (n6e);
	\draw[blue,thick] (n7s) -- (n7e);
	\draw[green,thick] (n8s) -- (n8e);
	\draw[magenta,thick] (n9s) -- (n9e);
	\draw[black,thick] (n10s) -- (n10e);	
\end{tikzpicture}
  \end{center}
\end{figure*}

\begin{figure*}[htp]
  \begin{center}
	\mbox{
        \tikzmark{lr1st}\subfigure[]{\scalebox{0.26}{\includegraphics[trim = 4.0cm 3.0cm 4.0cm 3.0cm, clip =
          true]{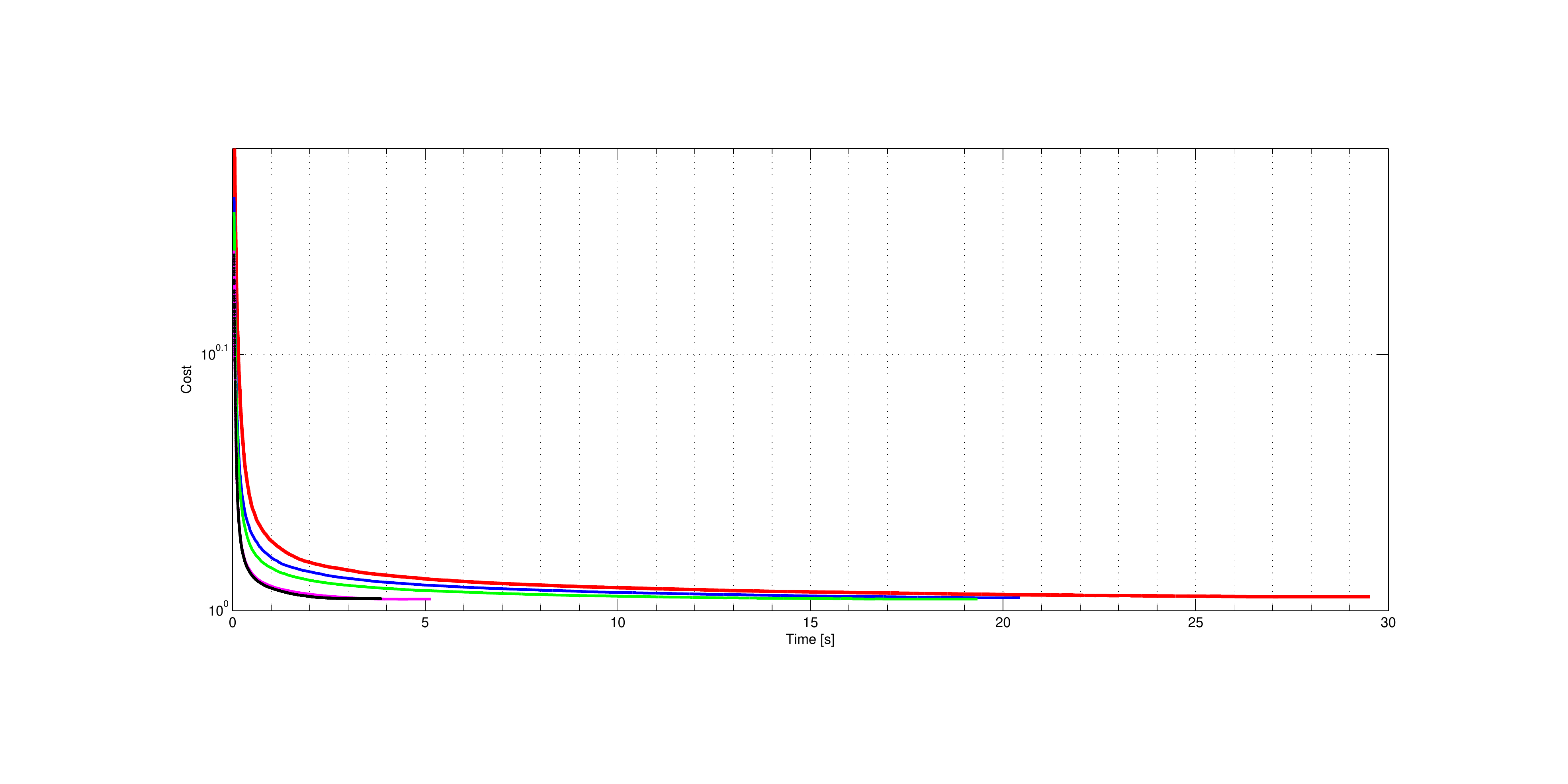}}\label{figure:time_cost_mean_d2_pt3_all}}\tikzmark{lr1en}
    	\tikzmark{lr2st}\subfigure[]{\scalebox{0.26}{\includegraphics[trim = 4.0cm 3.0cm 4.0cm 3.0cm, clip =
          true]{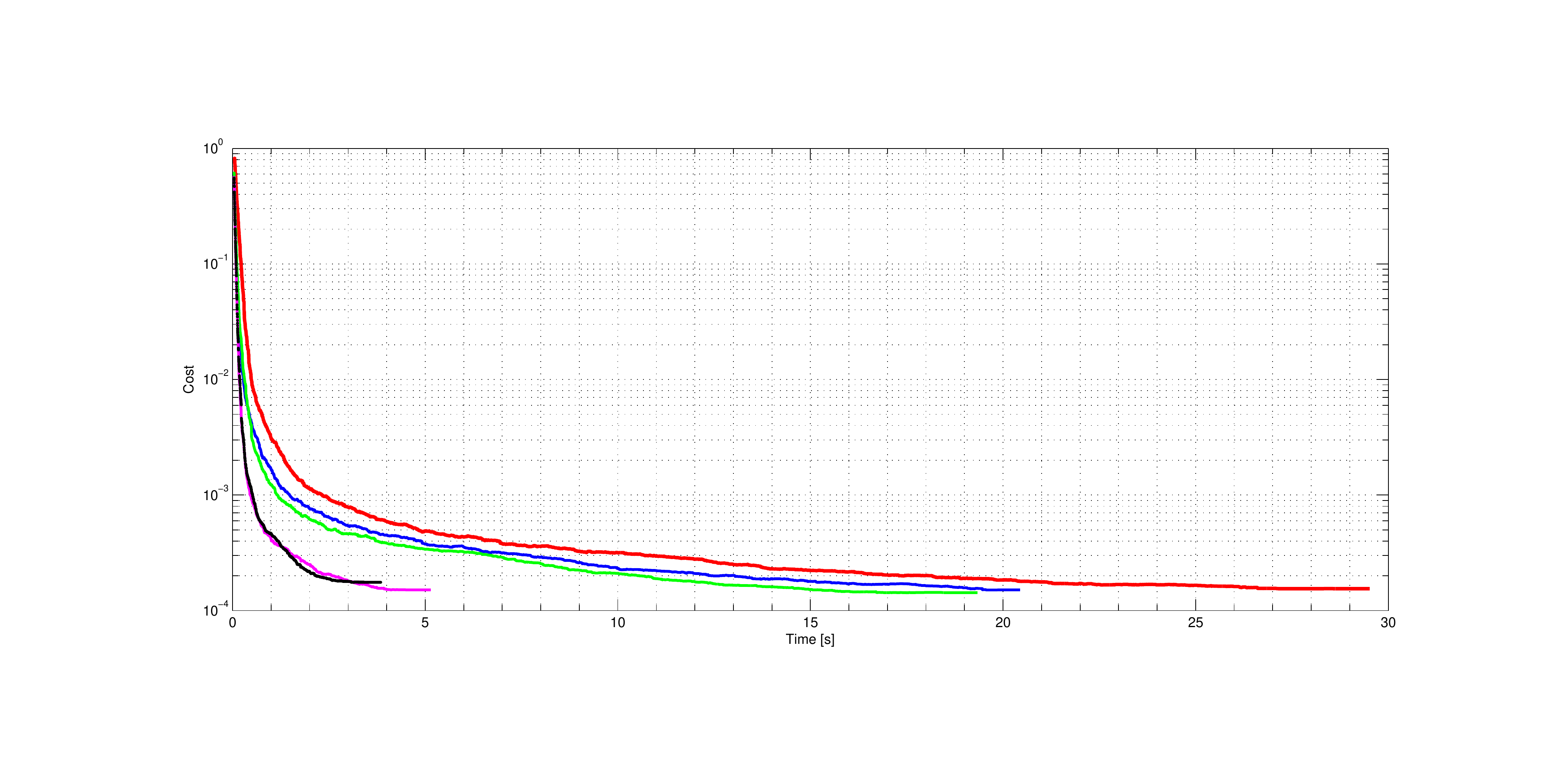}}\label{figure:time_cost_variance_d2_pt3_all}}\tikzmark{lr2en}

    }


    \caption{The change in the cost of the best paths computed by \AlgRRTstar{}, \AlgRRTsharp{}, and its variant algorithms and the variance in the trials are shown in \subref{figure:time_cost_mean_d2_pt3_all} and \subref{figure:time_cost_variance_d2_pt3_all}, respectively.}
    \label{figure:sim_d2_pt3_all_histories}

    \begin{tikzpicture}[
	remember picture,
	overlay,
	init/.style={inner sep=0pt}]
	\tiny
	\coordinate (lr1s) at ($(lr1st)+(6.7,1.95)$);
	\coordinate (lr1e) at ($(lr1en)+(-0.3,3.3)$);
	\coordinate (l1s) at ($(lr1st)+(6.7,3.25)$);
	\coordinate (l1e) at ($(lr1st)+(7.1,3.25)$);
	\coordinate (l2s) at ($(lr1st)+(6.7,2.95)$);
	\coordinate (l2e) at ($(lr1st)+(7.1,2.95)$);
	\coordinate (l3s) at ($(lr1st)+(6.7,2.65)$);
	\coordinate (l3e) at ($(lr1st)+(7.1,2.65)$);
	\coordinate (l4s) at ($(lr1st)+(6.7,2.35)$);
	\coordinate (l4e) at ($(lr1st)+(7.1,2.35)$);
	\coordinate (l5s) at ($(lr1st)+(6.7,2.05)$);
	\coordinate (l5e) at ($(lr1st)+(7.1,2.05)$);
	
	\node[draw=black,rectangle, fit=(lr1s) (lr1e)] {};
	\node[init] at (l1s) (n1s) {};
	\node[init] at (l1e) (n1e) [label=right:{\fontsize{0.5mm}{1mm}\selectfont\AlgRRTstar}]{};
	\node[init] at (l2s) (n2s) {};
	\node[init] at (l2e) (n2e) [label=right:{\fontsize{0.5mm}{1mm}\selectfont\AlgRRTsharp}]{};
	\node[init] at (l3s) (n3s) {};
	\node[init] at (l3e) (n3e) [label=right:{\fontsize{0.5mm}{1mm}\selectfont\AlgRRTsharpNoBlackVertex}] {};
	\node[init] at (l4s) (n4s) {};
	\node[init] at (l4e) (n4e) [label=right:{\fontsize{0.5mm}{1mm}\selectfont\AlgRRTsharpPromisingParent}]{};
	\node[init] at (l5s) (n5s) {};
	\node[init] at (l5e) (n5e) [label=right:{\fontsize{0.5mm}{1mm}\selectfont\AlgRRTsharpPromisingNewVertex}] {};
	\draw[red,thick] (n1s) -- (n1e);
	\draw[blue,thick] (n2s) -- (n2e);
	\draw[green,thick] (n3s) -- (n3e);
	\draw[magenta,thick] (n4s) -- (n4e);
	\draw[black,thick] (n5s) -- (n5e);
	
	\coordinate (lr2s) at ($(lr2st)+(6.7,1.95)$);
	\coordinate (lr2e) at ($(lr2en)+(-0.3,3.3)$);
	\coordinate (l6s) at ($(lr2st)+(6.7,3.25)$);
	\coordinate (l6e) at ($(lr2st)+(7.1,3.25)$);
	\coordinate (l7s) at ($(lr2st)+(6.7,2.95)$);
	\coordinate (l7e) at ($(lr2st)+(7.1,2.95)$);
	\coordinate (l8s) at ($(lr2st)+(6.7,2.65)$);
	\coordinate (l8e) at ($(lr2st)+(7.1,2.65)$);
	\coordinate (l9s) at ($(lr2st)+(6.7,2.35)$);
	\coordinate (l9e) at ($(lr2st)+(7.1,2.35)$);
	\coordinate (l10s) at ($(lr2st)+(6.7,2.05)$);
	\coordinate (l10e) at ($(lr2st)+(7.1,2.05)$);
	
	\node[draw=black,rectangle, fit=(lr2s) (lr2e)] {};
	\node[init] at (l6s) (n6s) {};
	\node[init] at (l6e) (n6e) [label=right:{\fontsize{0.5mm}{1mm}\selectfont\AlgRRTstar}]{};
	\node[init] at (l7s) (n7s) {};
	\node[init] at (l7e) (n7e) [label=right:{\fontsize{0.5mm}{1mm}\selectfont\AlgRRTsharp}]{};
	\node[init] at (l8s) (n8s) {};
	\node[init] at (l8e) (n8e) [label=right:{\fontsize{0.5mm}{1mm}\selectfont\AlgRRTsharpNoBlackVertex}] {};
	\node[init] at (l9s) (n9s) {};
	\node[init] at (l9e) (n9e) [label=right:{\fontsize{0.5mm}{1mm}\selectfont\AlgRRTsharpPromisingParent}]{};
	\node[init] at (l10s) (n10s) {};
	\node[init] at (l10e) (n10e) [label=right:{\fontsize{0.5mm}{1mm}\selectfont\AlgRRTsharpPromisingNewVertex}] {};
	\draw[red,thick] (n6s) -- (n6e);
	\draw[blue,thick] (n7s) -- (n7e);
	\draw[green,thick] (n8s) -- (n8e);
	\draw[magenta,thick] (n9s) -- (n9e);
	\draw[black,thick] (n10s) -- (n10e);	
\end{tikzpicture}
  \end{center}
\end{figure*}

\begin{figure*}[htp]
  \begin{center}

	\mbox{
        \tikzmark{lr1st}\subfigure[]{\scalebox{0.26}{\includegraphics[trim = 4.0cm 3.0cm 4.0cm 3.0cm, clip =
          true]{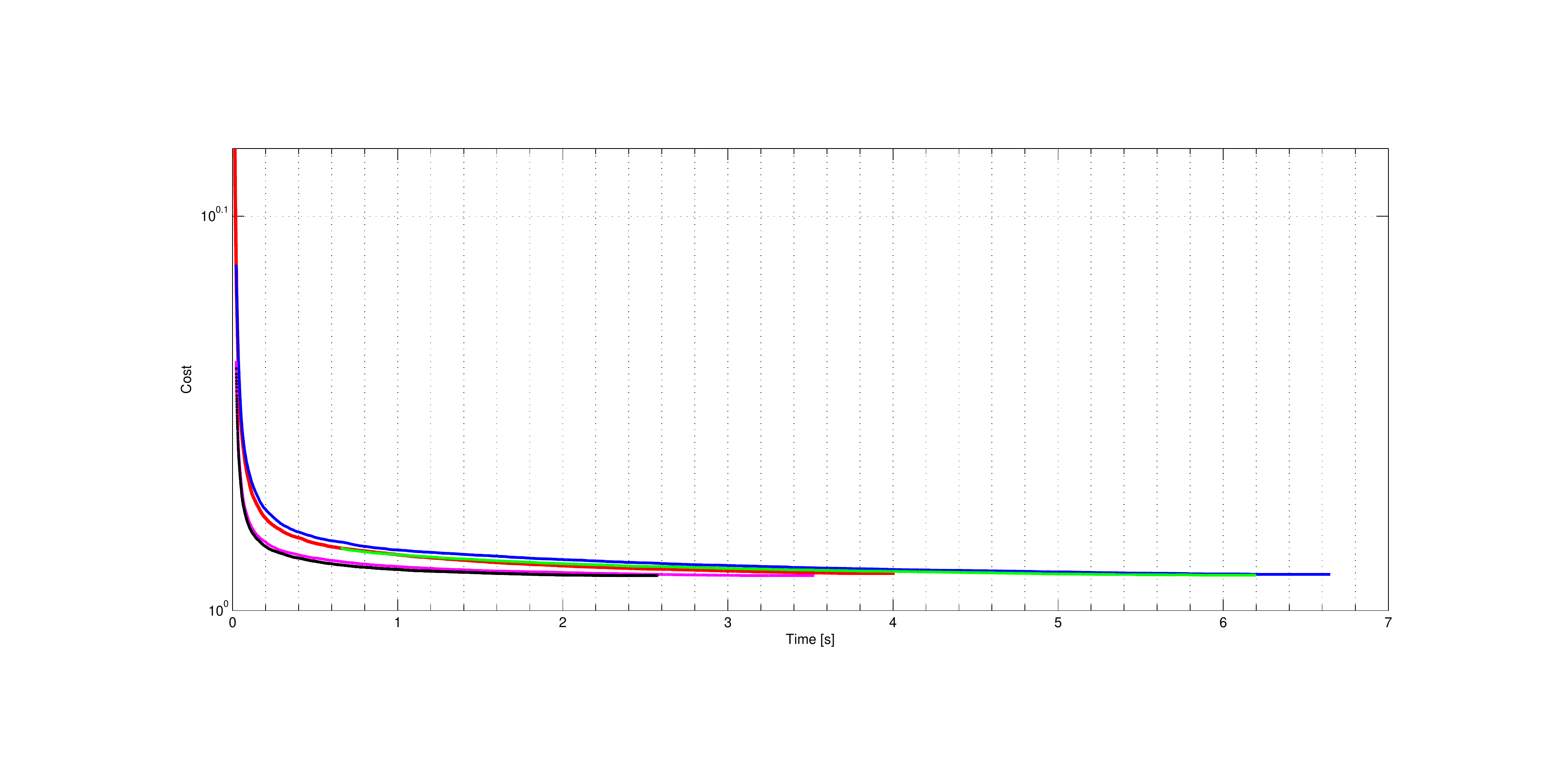}}\label{figure:time_cost_mean_d2_pt4_all}}\tikzmark{lr1en}
    	\tikzmark{lr2st}\subfigure[]{\scalebox{0.26}{\includegraphics[trim = 4.0cm 3.0cm 4.0cm 3.0cm, clip =
          true]{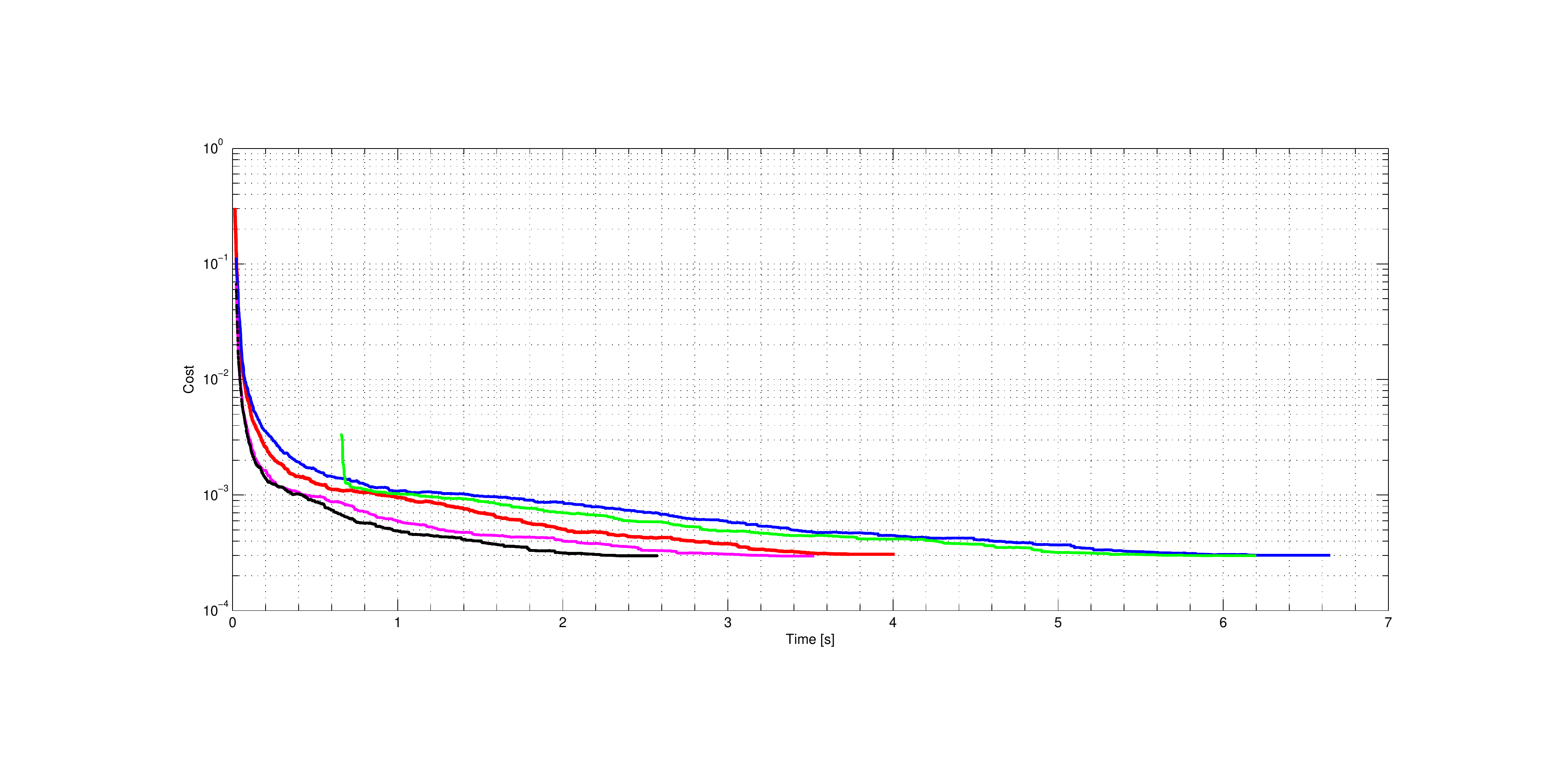}}\label{figure:time_cost_variance_d2_pt4_all}}\tikzmark{lr2en}
    }


    \caption{The change in the cost of the best paths computed by \AlgRRTstar{}, \AlgRRTsharp{}, and its variant algorithms and the variance in the trials are shown in \subref{figure:time_cost_mean_d2_pt4_all} and \subref{figure:time_cost_variance_d2_pt4_all}, respectively.}
    \label{figure:sim_d2_pt4_all_histories}

        \begin{tikzpicture}[
	remember picture,
	overlay,
	init/.style={inner sep=0pt}]
	\tiny
	\coordinate (lr1s) at ($(lr1st)+(6.7,1.95)$);
	\coordinate (lr1e) at ($(lr1en)+(-0.3,3.3)$);
	\coordinate (l1s) at ($(lr1st)+(6.7,3.25)$);
	\coordinate (l1e) at ($(lr1st)+(7.1,3.25)$);
	\coordinate (l2s) at ($(lr1st)+(6.7,2.95)$);
	\coordinate (l2e) at ($(lr1st)+(7.1,2.95)$);
	\coordinate (l3s) at ($(lr1st)+(6.7,2.65)$);
	\coordinate (l3e) at ($(lr1st)+(7.1,2.65)$);
	\coordinate (l4s) at ($(lr1st)+(6.7,2.35)$);
	\coordinate (l4e) at ($(lr1st)+(7.1,2.35)$);
	\coordinate (l5s) at ($(lr1st)+(6.7,2.05)$);
	\coordinate (l5e) at ($(lr1st)+(7.1,2.05)$);
	
	\node[draw=black,rectangle, fit=(lr1s) (lr1e)] {};
	\node[init] at (l1s) (n1s) {};
	\node[init] at (l1e) (n1e) [label=right:{\fontsize{0.5mm}{1mm}\selectfont\AlgRRTstar}]{};
	\node[init] at (l2s) (n2s) {};
	\node[init] at (l2e) (n2e) [label=right:{\fontsize{0.5mm}{1mm}\selectfont\AlgRRTsharp}]{};
	\node[init] at (l3s) (n3s) {};
	\node[init] at (l3e) (n3e) [label=right:{\fontsize{0.5mm}{1mm}\selectfont\AlgRRTsharpNoBlackVertex}] {};
	\node[init] at (l4s) (n4s) {};
	\node[init] at (l4e) (n4e) [label=right:{\fontsize{0.5mm}{1mm}\selectfont\AlgRRTsharpPromisingParent}]{};
	\node[init] at (l5s) (n5s) {};
	\node[init] at (l5e) (n5e) [label=right:{\fontsize{0.5mm}{1mm}\selectfont\AlgRRTsharpPromisingNewVertex}] {};
	\draw[red,thick] (n1s) -- (n1e);
	\draw[blue,thick] (n2s) -- (n2e);
	\draw[green,thick] (n3s) -- (n3e);
	\draw[magenta,thick] (n4s) -- (n4e);
	\draw[black,thick] (n5s) -- (n5e);
	
	\coordinate (lr2s) at ($(lr2st)+(6.7,1.95)$);
	\coordinate (lr2e) at ($(lr2en)+(-0.3,3.3)$);
	\coordinate (l6s) at ($(lr2st)+(6.7,3.25)$);
	\coordinate (l6e) at ($(lr2st)+(7.1,3.25)$);
	\coordinate (l7s) at ($(lr2st)+(6.7,2.95)$);
	\coordinate (l7e) at ($(lr2st)+(7.1,2.95)$);
	\coordinate (l8s) at ($(lr2st)+(6.7,2.65)$);
	\coordinate (l8e) at ($(lr2st)+(7.1,2.65)$);
	\coordinate (l9s) at ($(lr2st)+(6.7,2.35)$);
	\coordinate (l9e) at ($(lr2st)+(7.1,2.35)$);
	\coordinate (l10s) at ($(lr2st)+(6.7,2.05)$);
	\coordinate (l10e) at ($(lr2st)+(7.1,2.05)$);
	
	\node[draw=black,rectangle, fit=(lr2s) (lr2e)] {};
	\node[init] at (l6s) (n6s) {};
	\node[init] at (l6e) (n6e) [label=right:{\fontsize{0.5mm}{1mm}\selectfont\AlgRRTstar}]{};
	\node[init] at (l7s) (n7s) {};
	\node[init] at (l7e) (n7e) [label=right:{\fontsize{0.5mm}{1mm}\selectfont\AlgRRTsharp}]{};
	\node[init] at (l8s) (n8s) {};
	\node[init] at (l8e) (n8e) [label=right:{\fontsize{0.5mm}{1mm}\selectfont\AlgRRTsharpNoBlackVertex}] {};
	\node[init] at (l9s) (n9s) {};
	\node[init] at (l9e) (n9e) [label=right:{\fontsize{0.5mm}{1mm}\selectfont\AlgRRTsharpPromisingParent}]{};
	\node[init] at (l10s) (n10s) {};
	\node[init] at (l10e) (n10e) [label=right:{\fontsize{0.5mm}{1mm}\selectfont\AlgRRTsharpPromisingNewVertex}] {};
	\draw[red,thick] (n6s) -- (n6e);
	\draw[blue,thick] (n7s) -- (n7e);
	\draw[green,thick] (n8s) -- (n8e);
	\draw[magenta,thick] (n9s) -- (n9e);
	\draw[black,thick] (n10s) -- (n10e);	
\end{tikzpicture}
  \end{center}
\end{figure*}

The convergence rate and variance in the computation of the best path for all algorithms are shown in Figures~\ref{figure:sim_d2_pt1_all_histories}, \ref{figure:sim_d2_pt2_all_histories}, \ref{figure:sim_d2_pt3_all_histories}, and \ref{figure:sim_d2_pt4_all_histories}. Since this is a two-dimensional problem, the optimal path for each problem type can be computed visually and the cost of the paths for each algorithm is normalized with respect to the cost of the optimal solution. The ratio of the cost of the best path over the optimal cost for the \AlgRRTstar{}, \AlgRRTsharp{}, \AlgRRTsharpNoBlackVertex{}, \AlgRRTsharpPromisingParent{}, and \AlgRRTsharpPromisingNewVertex{} algorithms is shown in red, blue, green, magenta, and black colors, respectively.


\begin{figure*}[htp]
  \begin{center}
	\mbox{
    \subfigure[]{\scalebox{0.35}{\includegraphics[trim = 4.0cm 6.937cm 3.587cm 7.0cm, clip =
          true]{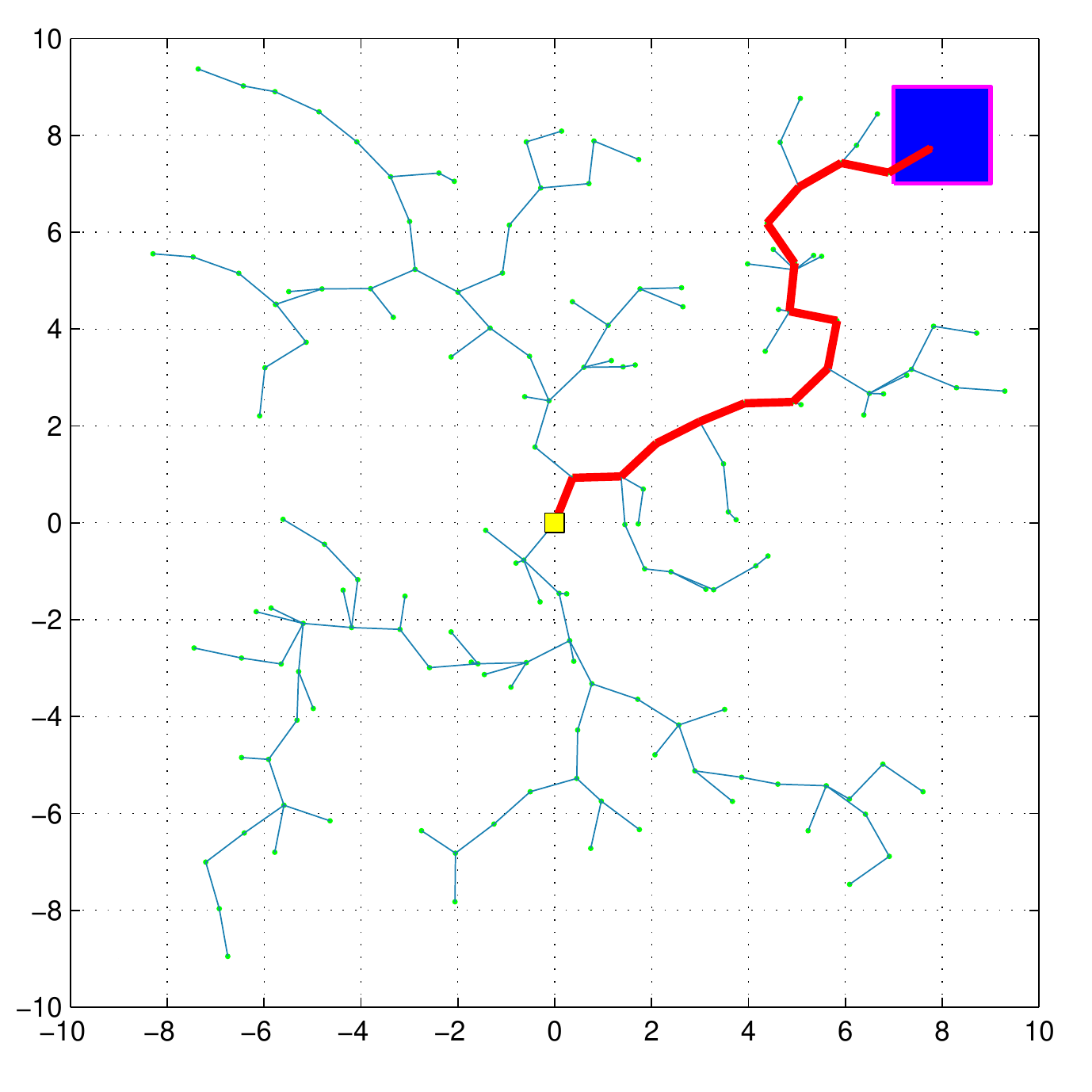}} \label{figure:pt1_rrtsharp_v3_it250}}
    \subfigure[]{\scalebox{0.35}{\includegraphics[trim = 4.0cm 6.937cm 3.587cm 7.0cm, clip =
          true]{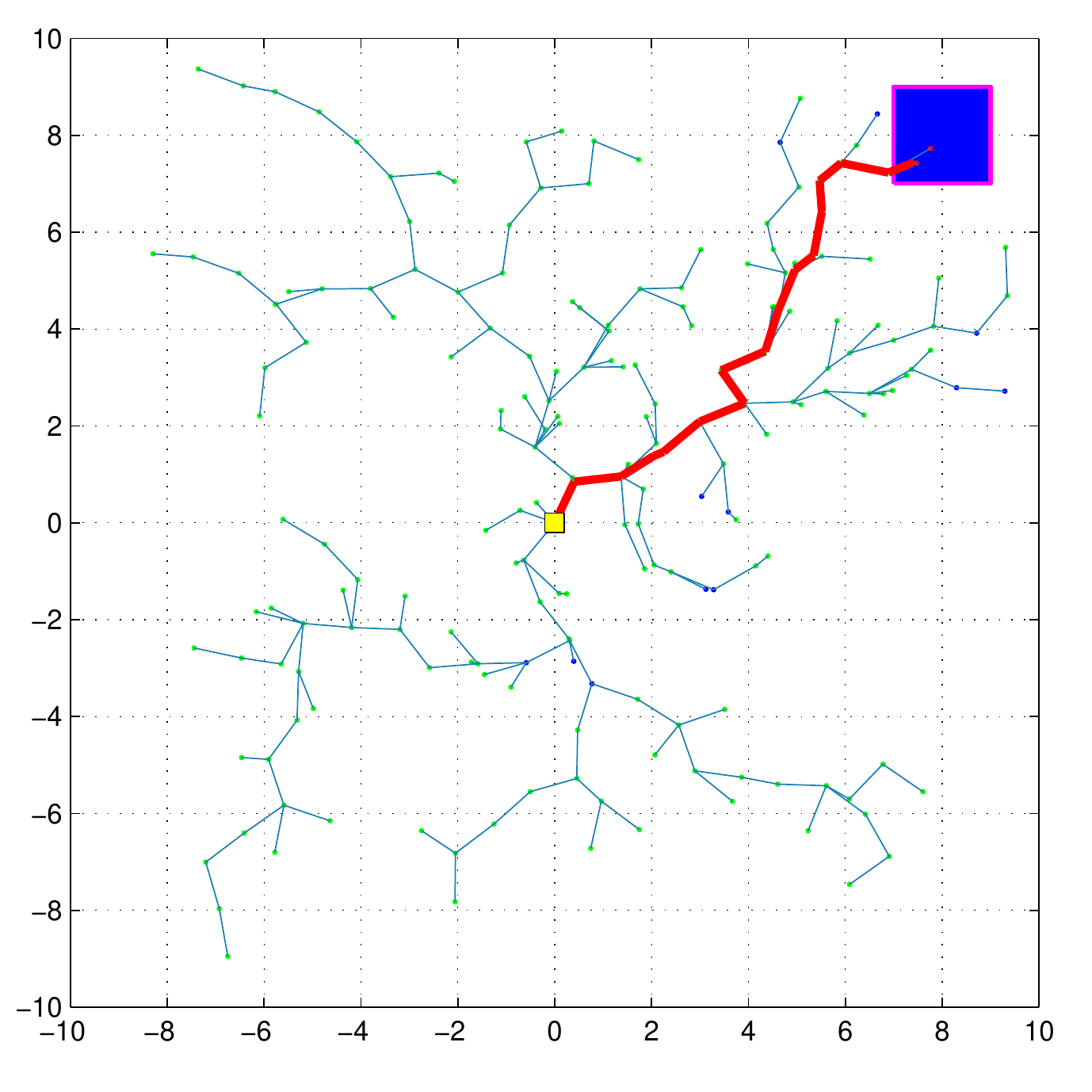}} \label{figure:pt1_rrtsharp_v3_it500}}
    \subfigure[]{\scalebox{0.35}{\includegraphics[trim = 4.0cm 6.937cm 3.587cm 7.0cm, clip =
          true]{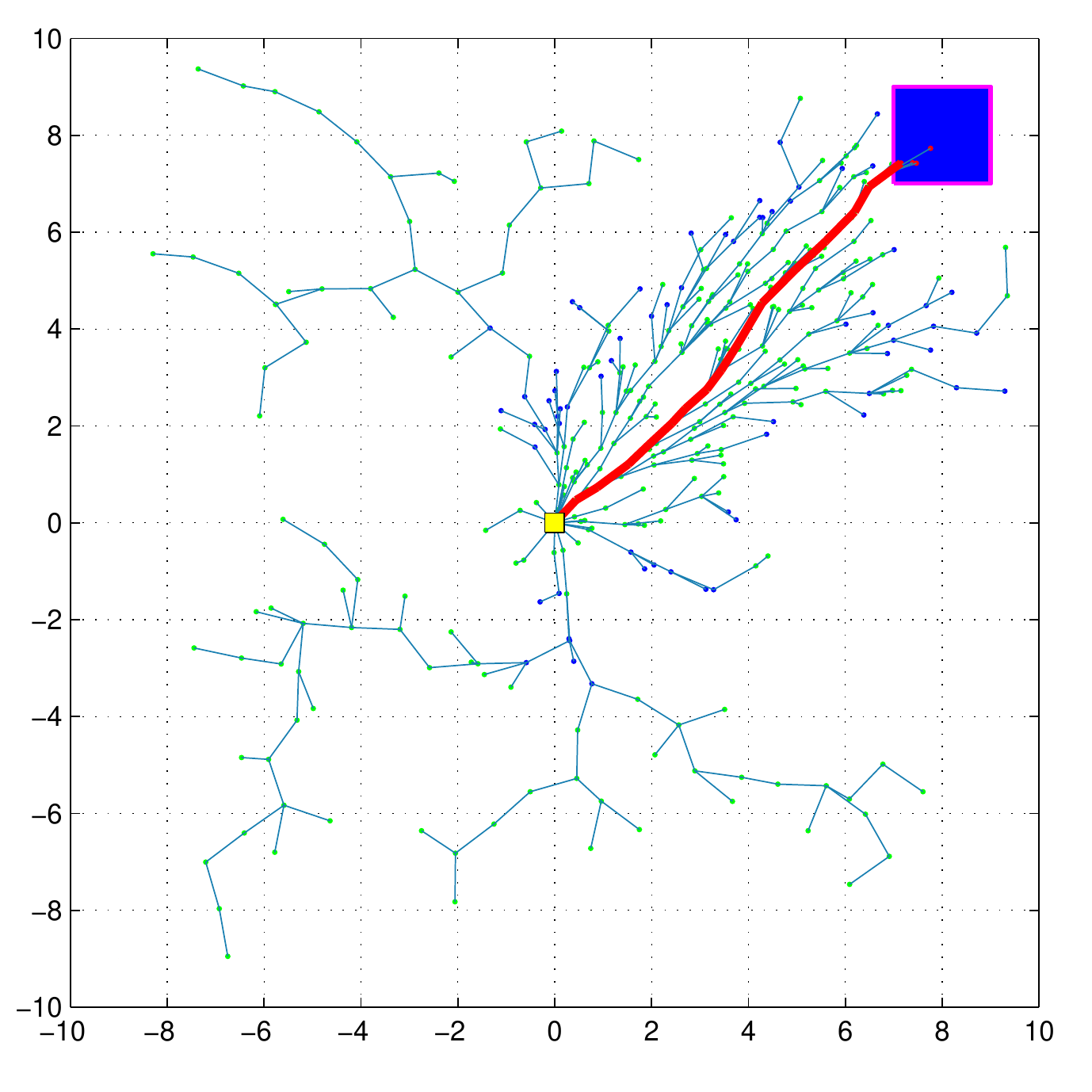}} \label{figure:pt1_rrtsharp_v3_it2500}}
   	}
   	
	\mbox{
	\subfigure[]{\scalebox{0.35}{\includegraphics[trim = 4.0cm 6.937cm 3.587cm 7.0cm, clip =
          true]{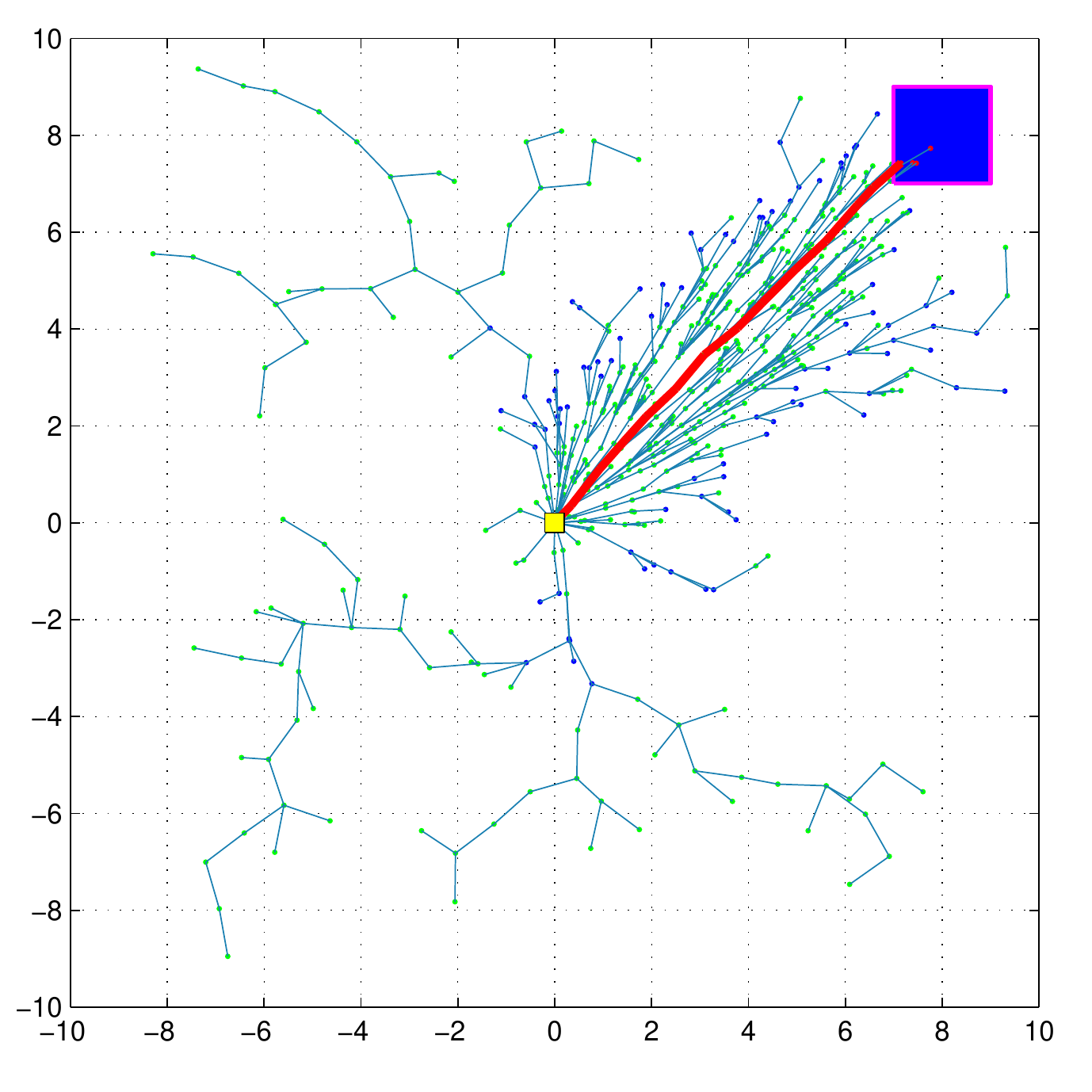}} \label{figure:pt1_rrtsharp_v3_it5000}}
    \subfigure[]{\scalebox{0.35}{\includegraphics[trim = 4.0cm 6.937cm 3.587cm 7.0cm, clip =
          true]{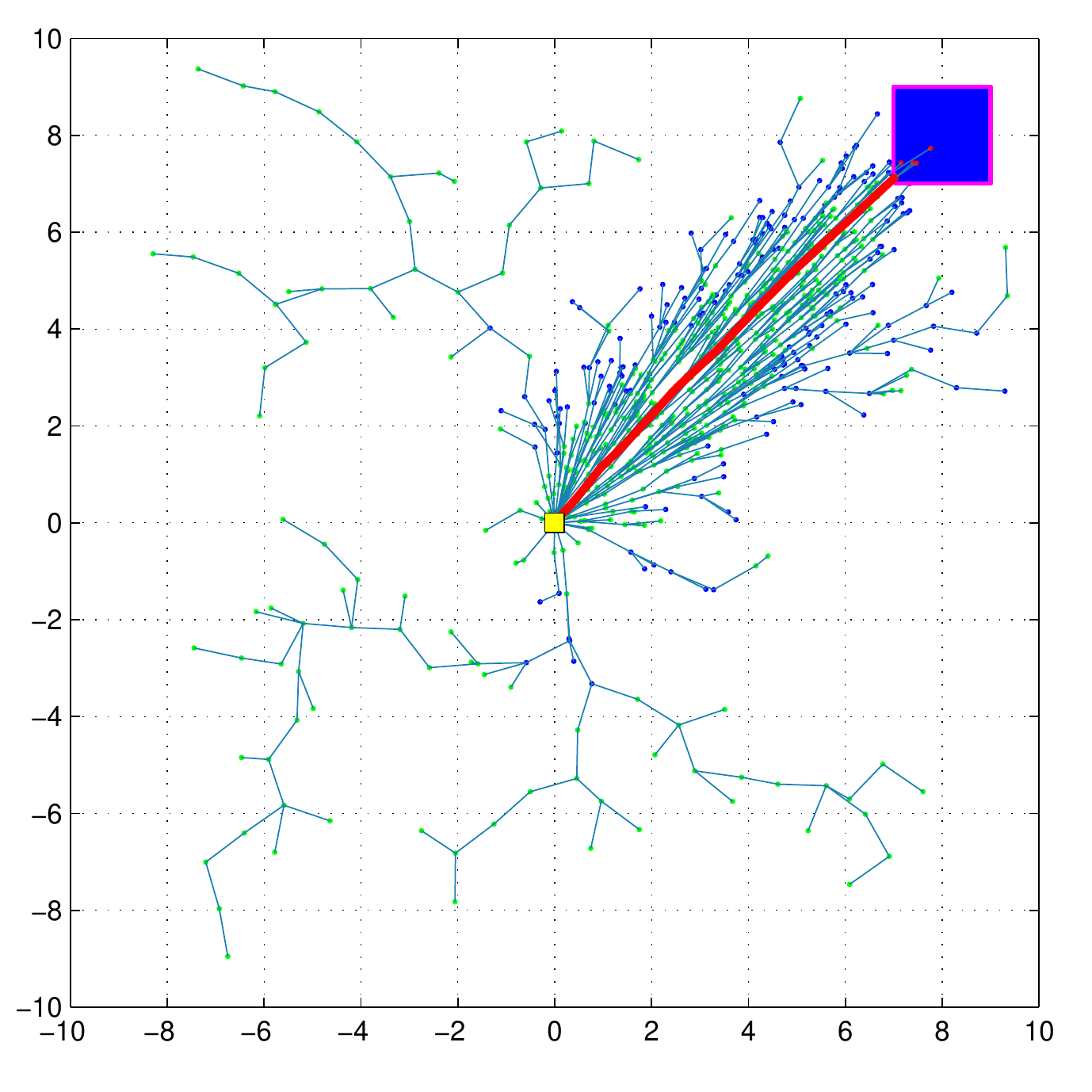}} \label{figure:pt1_rrtsharp_v3_it10000}}
    \subfigure[]{\scalebox{0.35}{\includegraphics[trim = 4.0cm 6.937cm 3.587cm 7.0cm, clip =
          true]{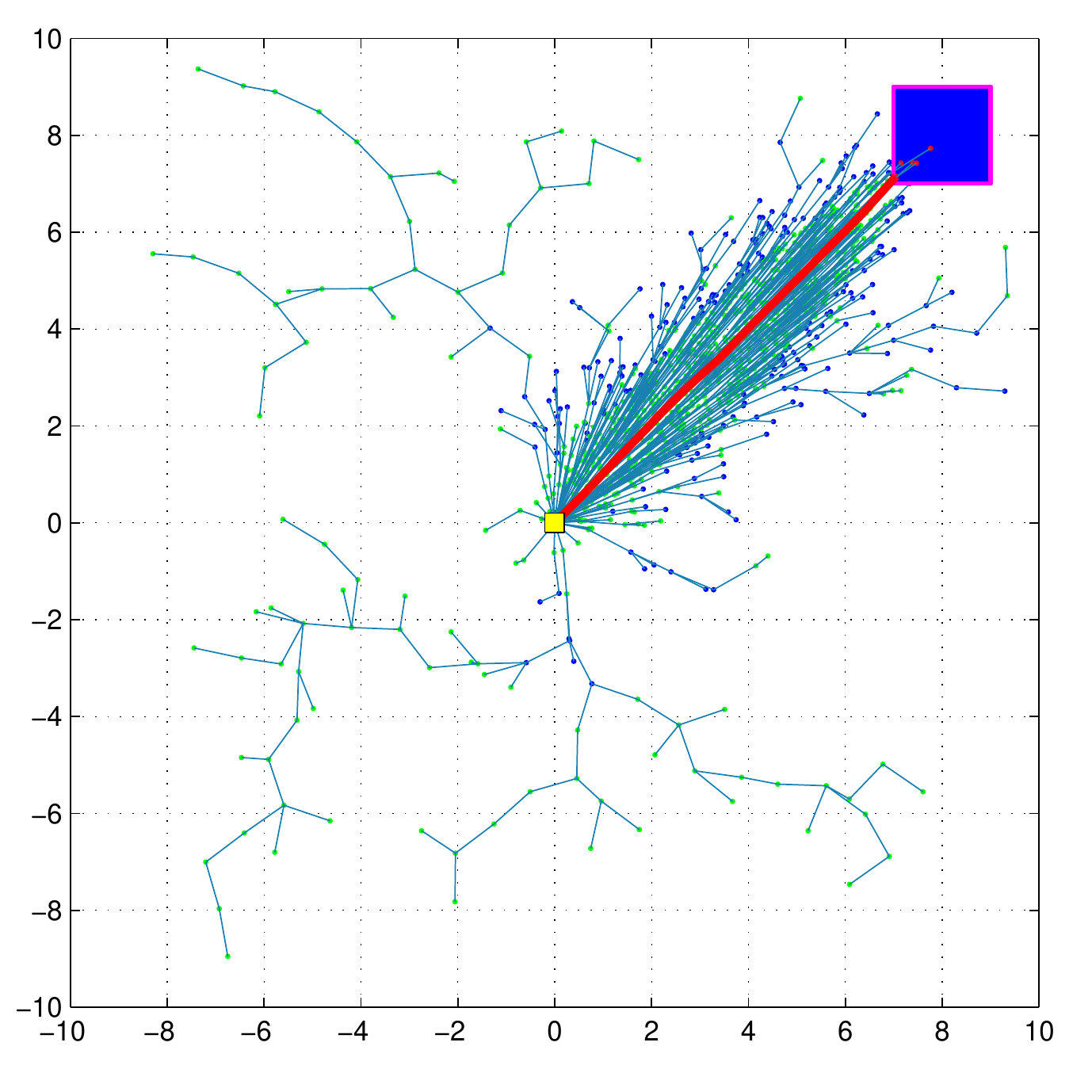}} \label{figure:pt1_rrtsharp_v3_it24999}}
    }

    \caption{The evolution of the tree computed by \AlgRRTsharpPromisingNewVertex{} algorithm is shown in   \subref{figure:pt1_rrtsharp_v3_it250}-\subref{figure:pt1_rrtsharp_v3_it24999}. The configuration of the trees in \subref{figure:pt1_rrtsharp_v3_it250} is at 250 iterations, in \subref{figure:pt1_rrtsharp_v3_it500} is at 500 iterations, in \subref{figure:pt1_rrtsharp_v3_it2500} is at 2500 iterations, in \subref{figure:pt1_rrtsharp_v3_it5000} is at 5000 iterations, in \subref{figure:pt1_rrtsharp_v3_it10000} is at 10000 iterations, and in \subref{figure:pt1_rrtsharp_v3_it24999} is at 25000 iterations.
    }
    \label{figure:sim_d2_pt1_rrtsharp_v3_iterations}
  \end{center}
\end{figure*}

\FloatBarrier

\begin{figure*}[htp]
  \begin{center}

	\mbox{
    \subfigure[]{\scalebox{0.35}{\includegraphics[trim = 4.0cm 6.937cm 3.587cm 7.0cm, clip =
          true]{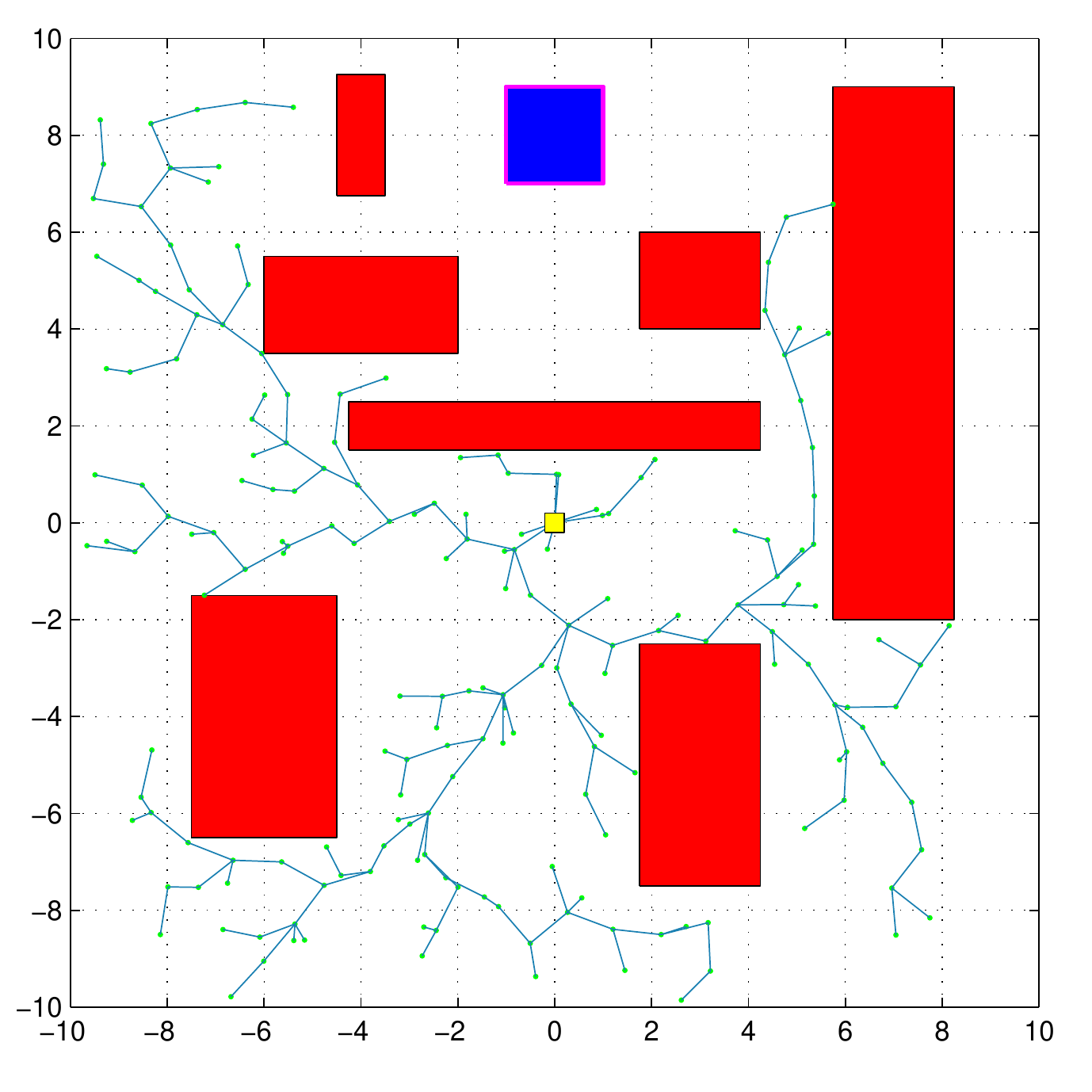}} \label{figure:pt2_rrtsharp_v3_it250}}
    \subfigure[]{\scalebox{0.35}{\includegraphics[trim = 4.0cm 6.937cm 3.587cm 7.0cm, clip =
          true]{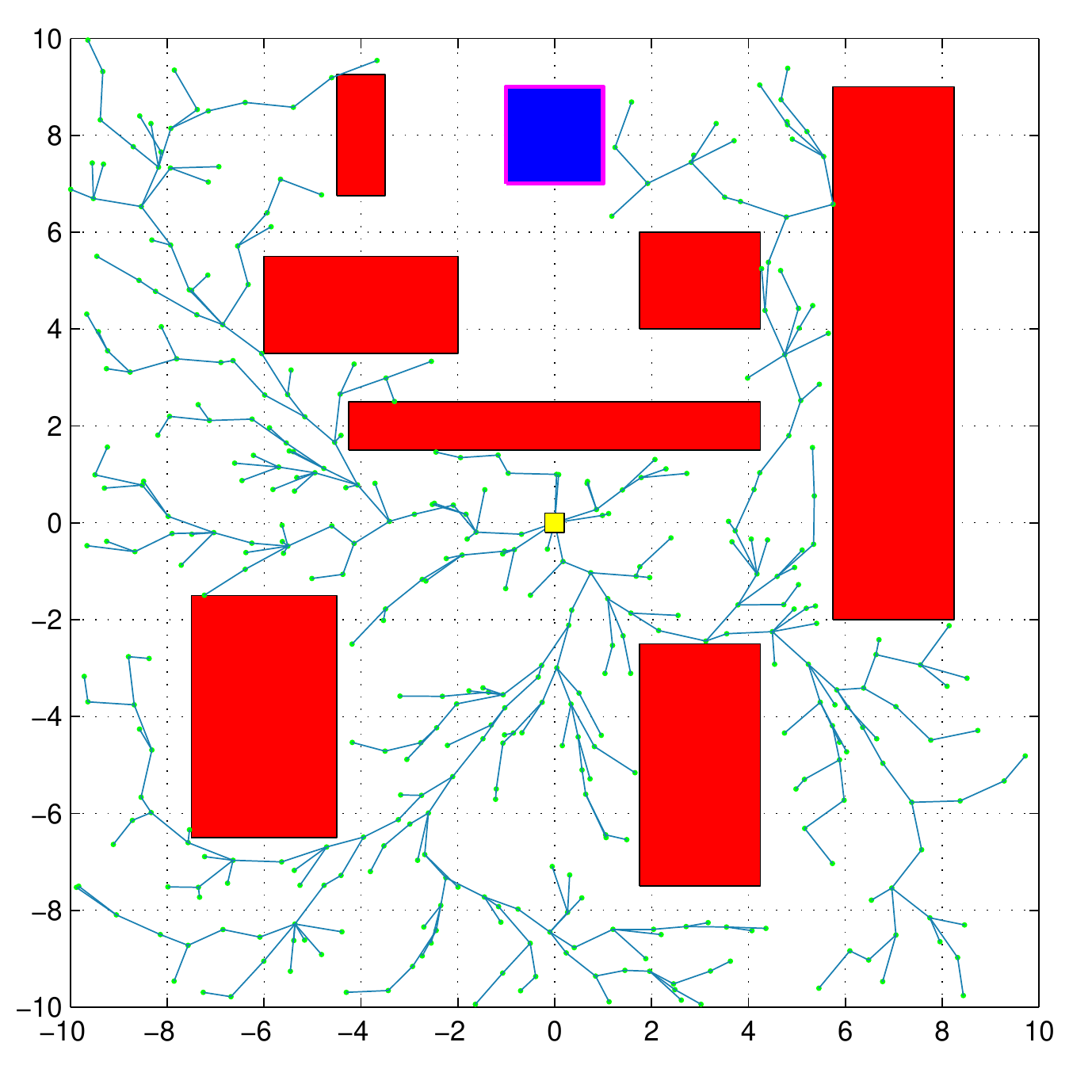}} \label{figure:pt2_rrtsharp_v3_it500}}
    \subfigure[]{\scalebox{0.35}{\includegraphics[trim = 4.0cm 6.937cm 3.587cm 7.0cm, clip =
          true]{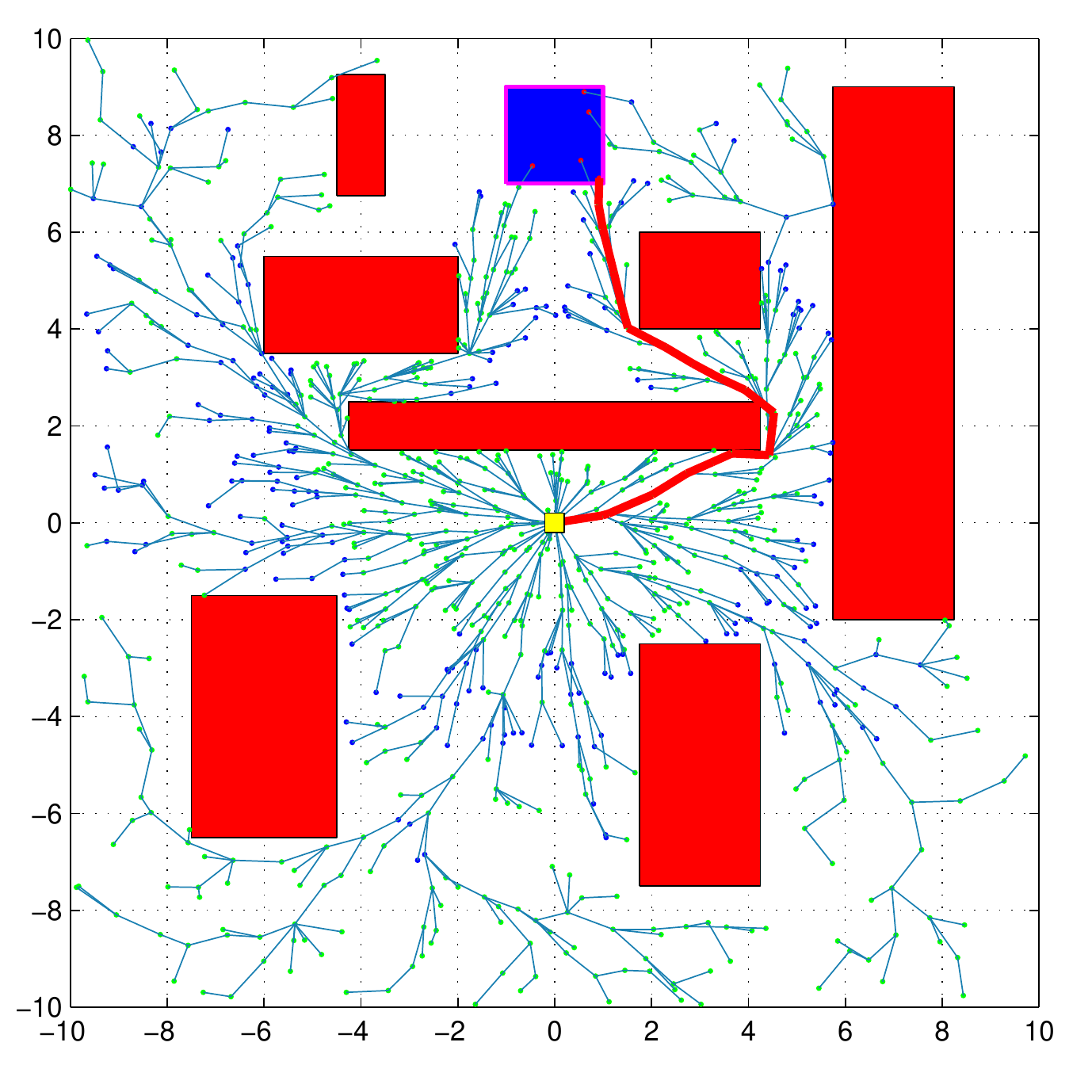}} \label{figure:pt2_rrtsharp_v3_it2500}}
   	}
   	
	\mbox{
	\subfigure[]{\scalebox{0.35}{\includegraphics[trim = 4.0cm 6.937cm 3.587cm 7.0cm, clip =
          true]{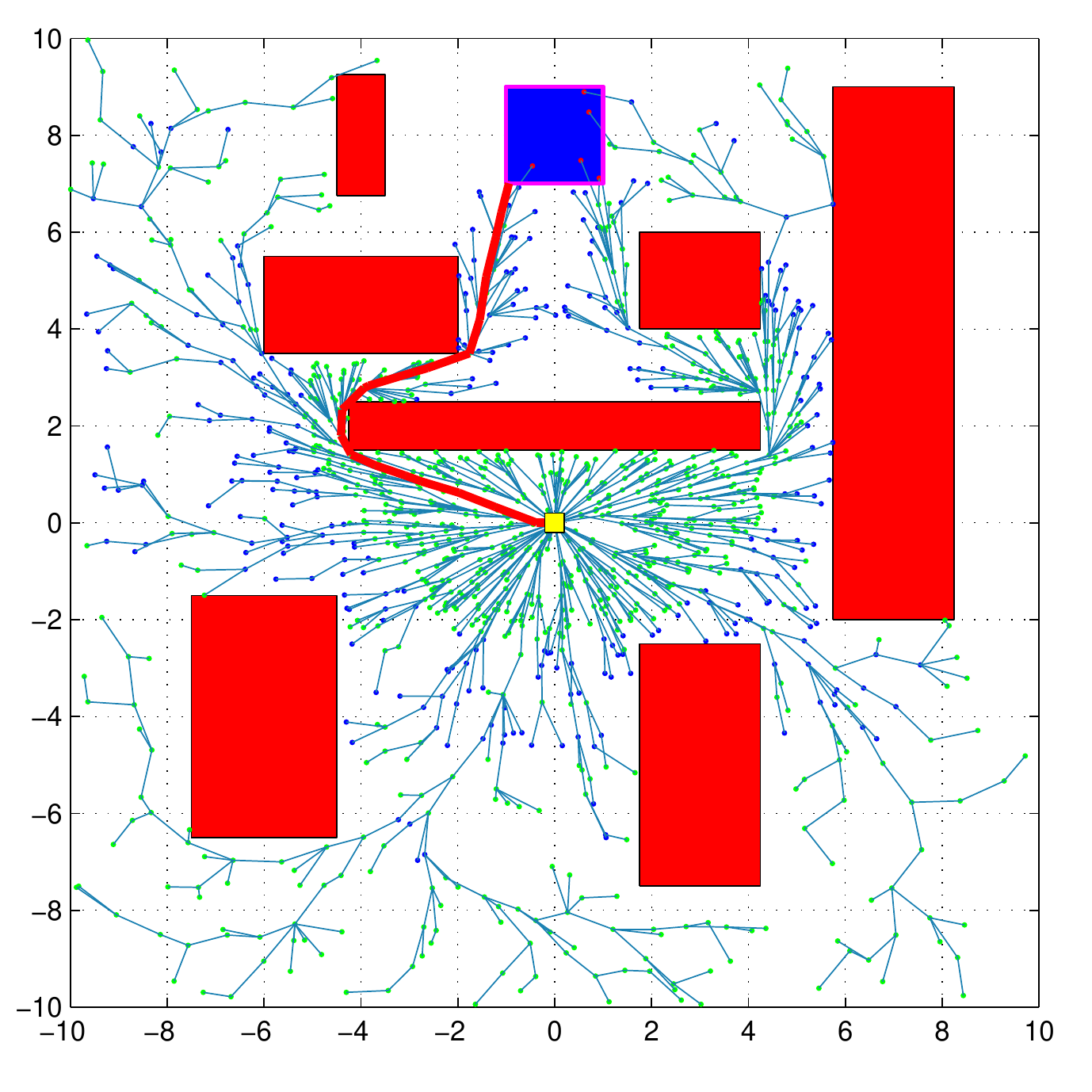}} \label{figure:pt2_rrtsharp_v3_it5000}}
    \subfigure[]{\scalebox{0.35}{\includegraphics[trim = 4.0cm 6.937cm 3.587cm 7.0cm, clip =
          true]{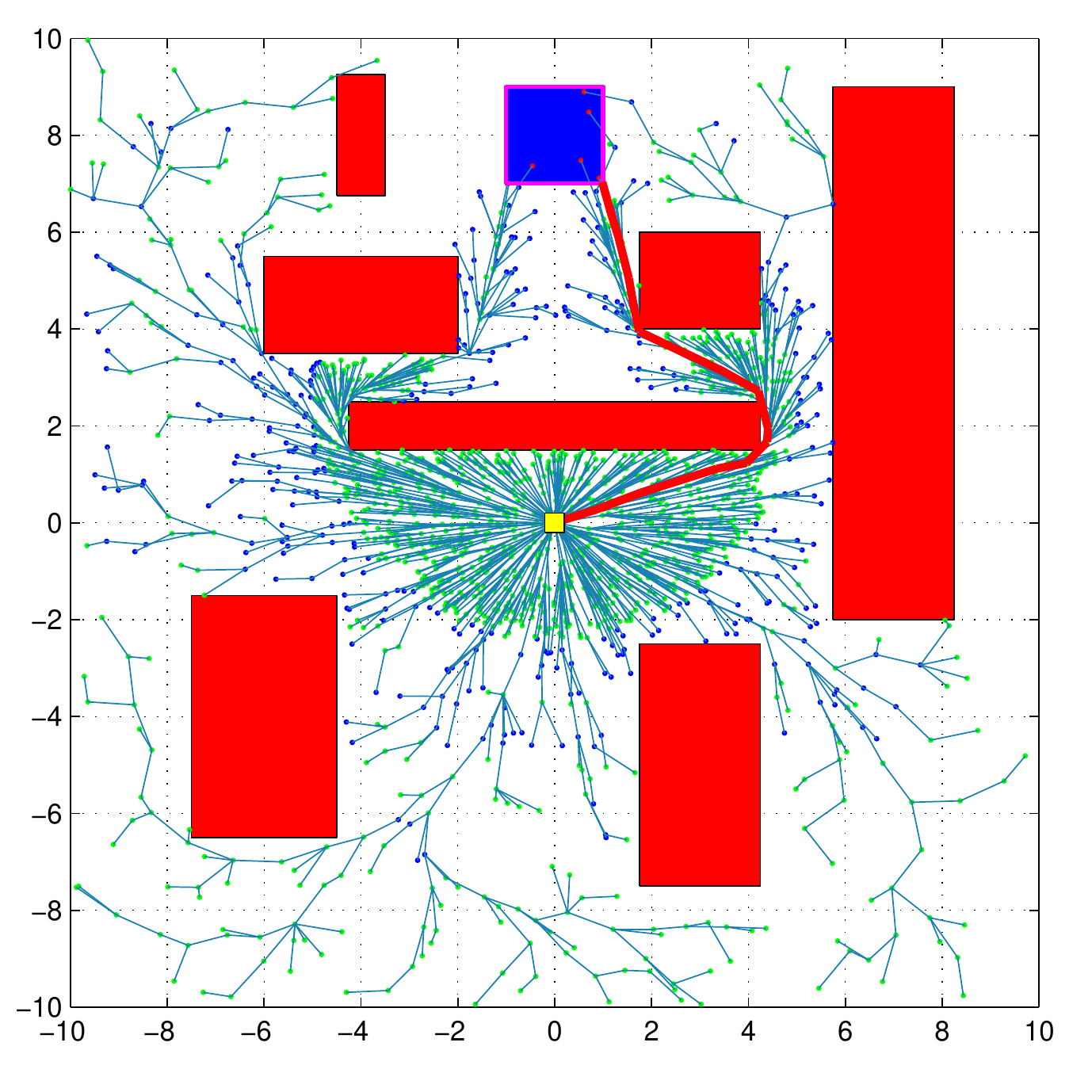}} \label{figure:pt2_rrtsharp_v3_it10000}}
    \subfigure[]{\scalebox{0.35}{\includegraphics[trim = 4.0cm 6.937cm 3.587cm 7.0cm, clip =
          true]{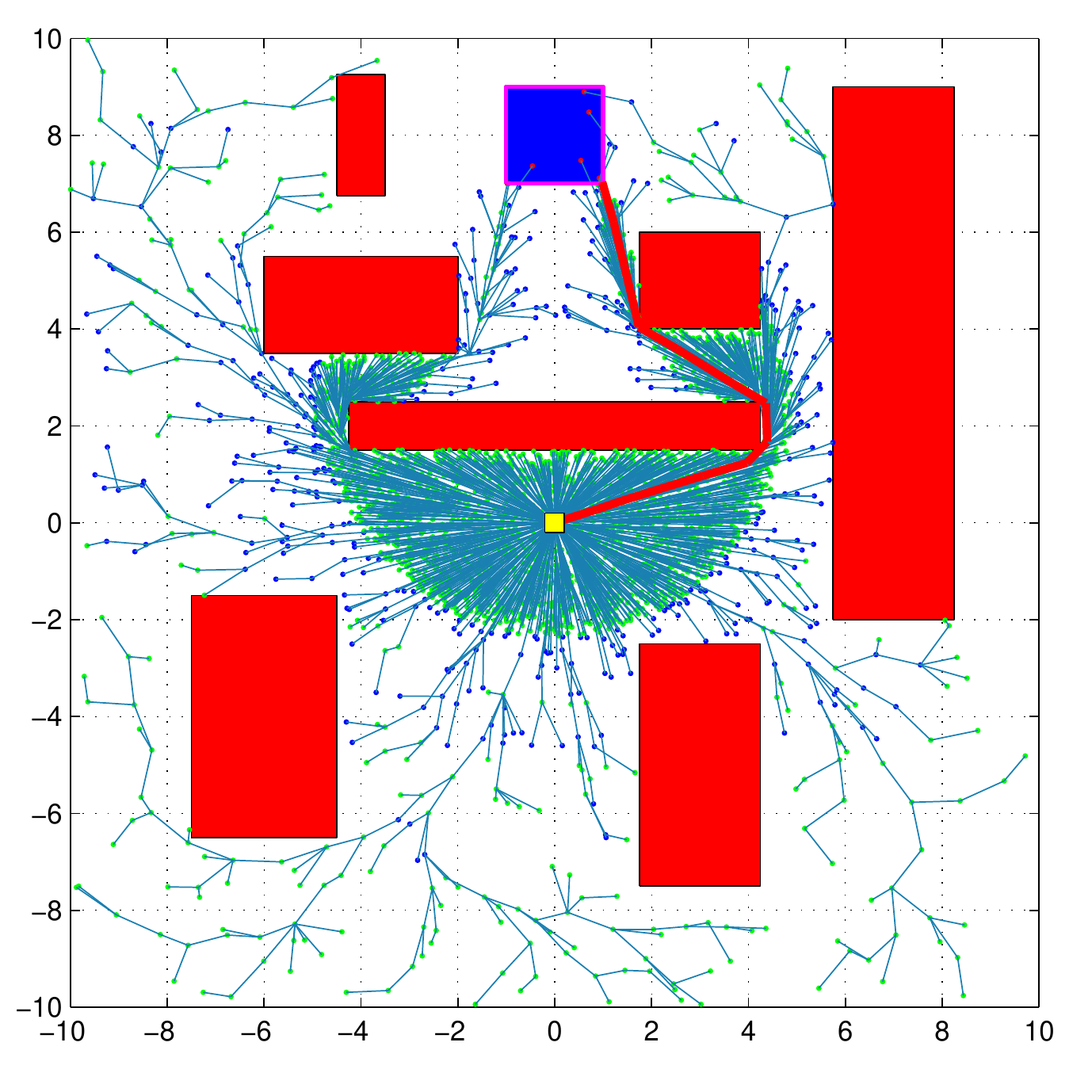}} \label{figure:pt2_rrtsharp_v3_it24999}}
    }

    \caption{The evolution of the tree computed by \AlgRRTsharpPromisingNewVertex{} algorithm is shown in   \subref{figure:pt2_rrtsharp_v3_it250}-\subref{figure:pt2_rrtsharp_v3_it24999}. The configuration of the trees in \subref{figure:pt2_rrtsharp_v3_it250} is at 250 iterations, in \subref{figure:pt2_rrtsharp_v3_it500} is at 500 iterations, in \subref{figure:pt2_rrtsharp_v3_it2500} is at 2500 iterations, in \subref{figure:pt2_rrtsharp_v3_it5000} is at 5000 iterations, in \subref{figure:pt2_rrtsharp_v3_it10000} is at 10000 iterations, and in \subref{figure:pt2_rrtsharp_v3_it24999} is at 25000 iterations.
    }
    \label{figure:sim_d2_pt2_rrtsharp_v3_iterations}
  \end{center}
\end{figure*}

\FloatBarrier

\begin{figure*}[htp]
  \begin{center}

	\mbox{
    \subfigure[]{\scalebox{0.35}{\includegraphics[trim = 4.0cm 6.937cm 3.587cm 7.0cm, clip =
          true]{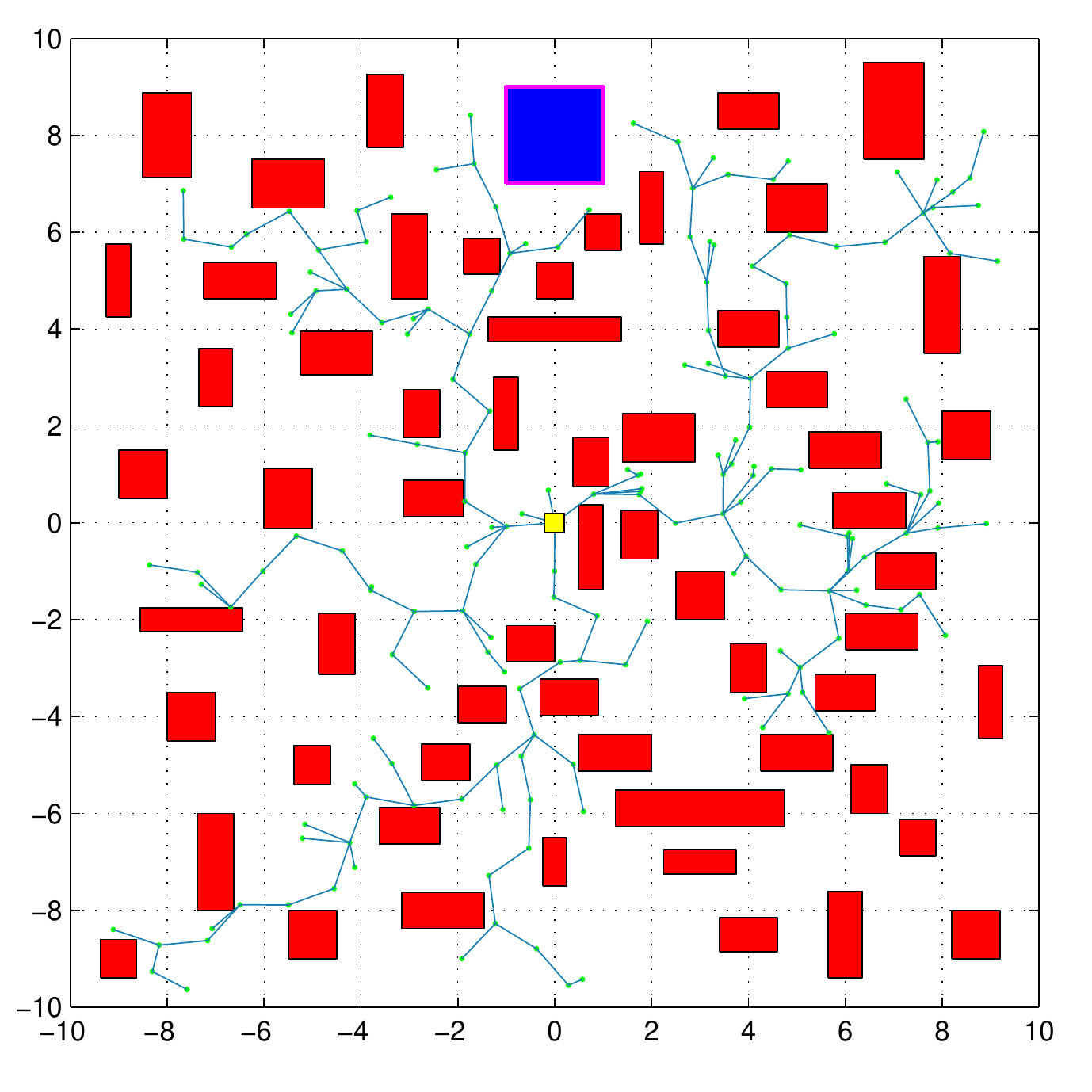}} \label{figure:pt3_rrtsharp_v3_it250}}
    \subfigure[]{\scalebox{0.35}{\includegraphics[trim = 4.0cm 6.937cm 3.587cm 7.0cm, clip =
          true]{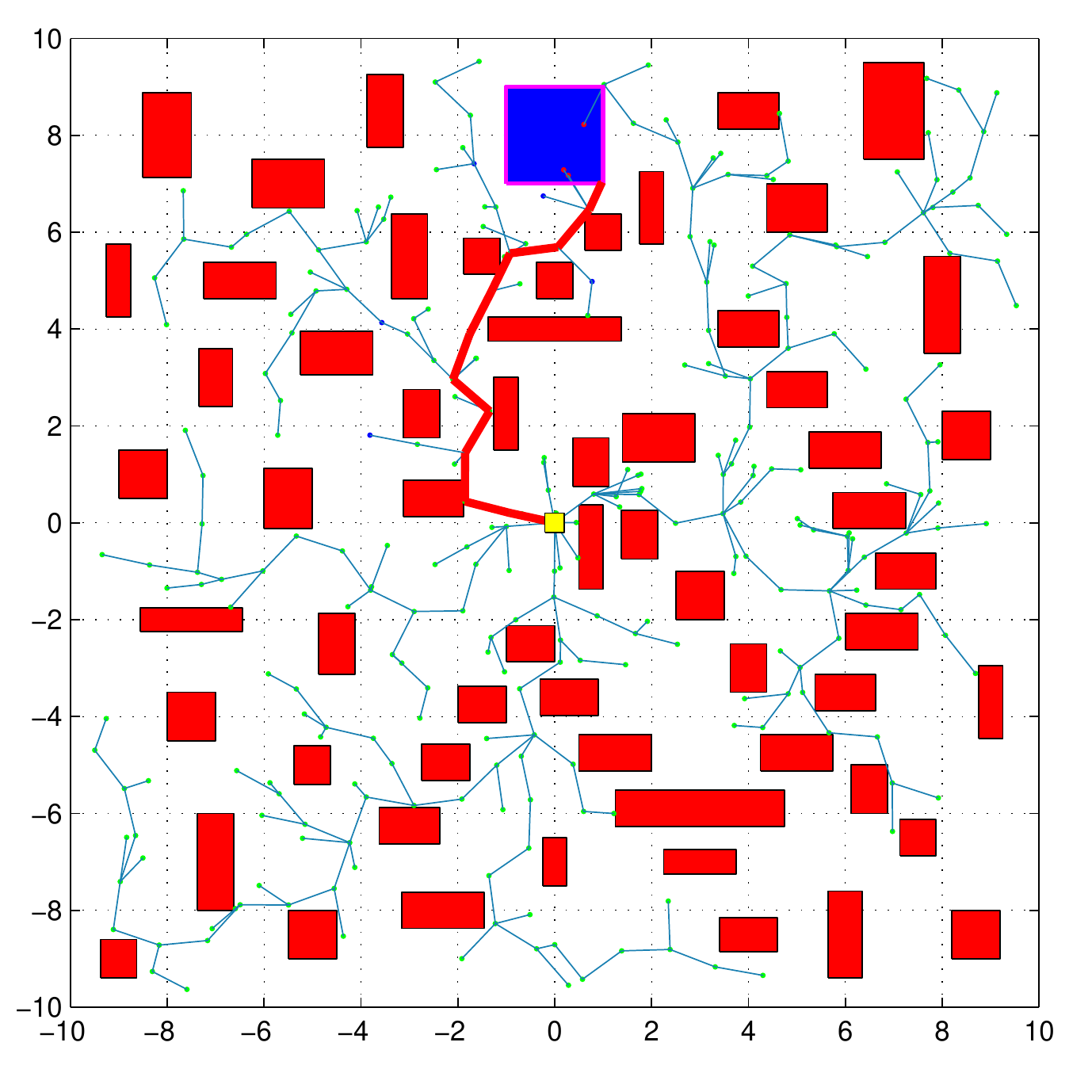}} \label{figure:pt3_rrtsharp_v3_it500}}
    \subfigure[]{\scalebox{0.35}{\includegraphics[trim = 4.0cm 6.937cm 3.587cm 7.0cm, clip =
          true]{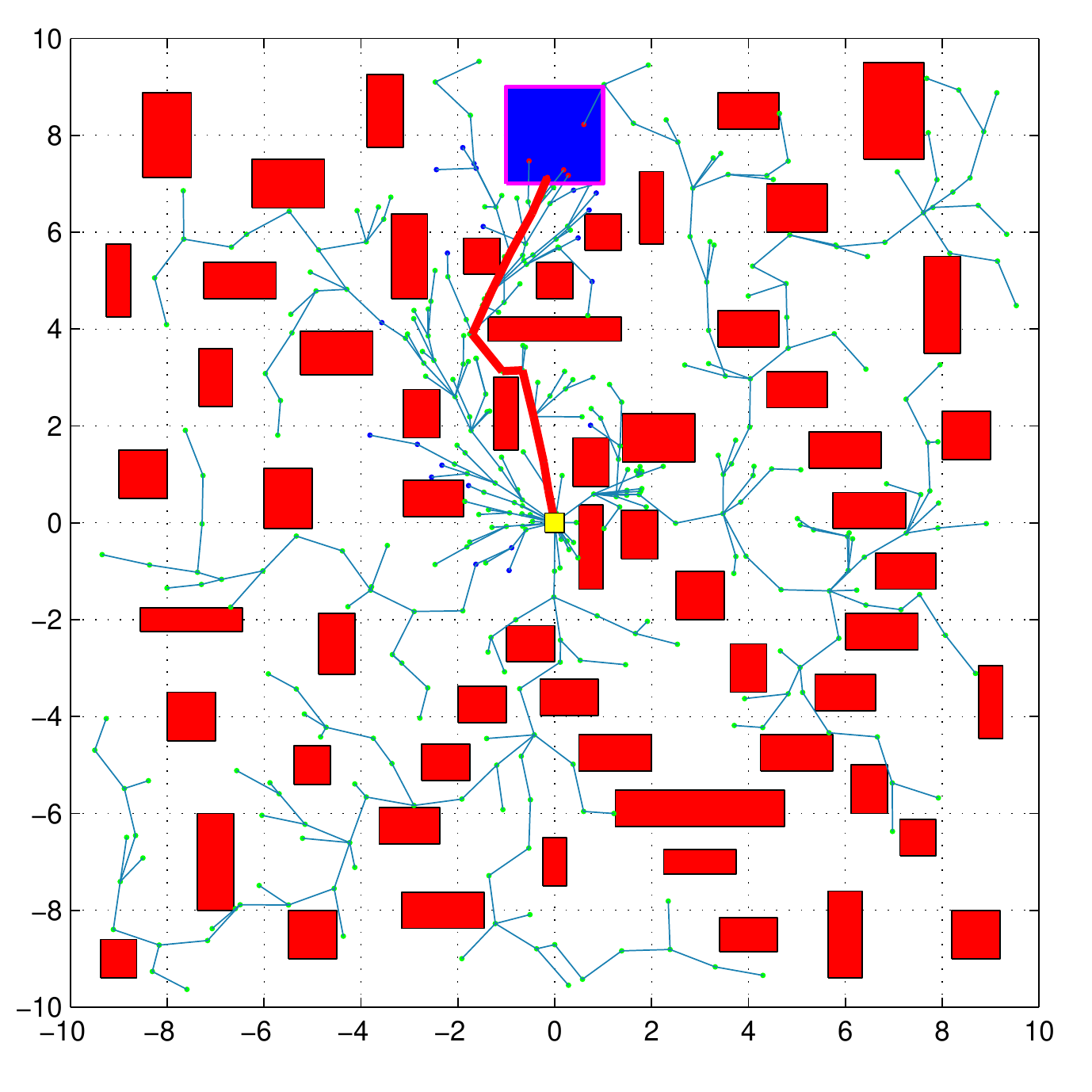}} \label{figure:pt3_rrtsharp_v3_it2500}}
   	}
   	
	\mbox{
	\subfigure[]{\scalebox{0.35}{\includegraphics[trim = 4.0cm 6.937cm 3.587cm 7.0cm, clip =
          true]{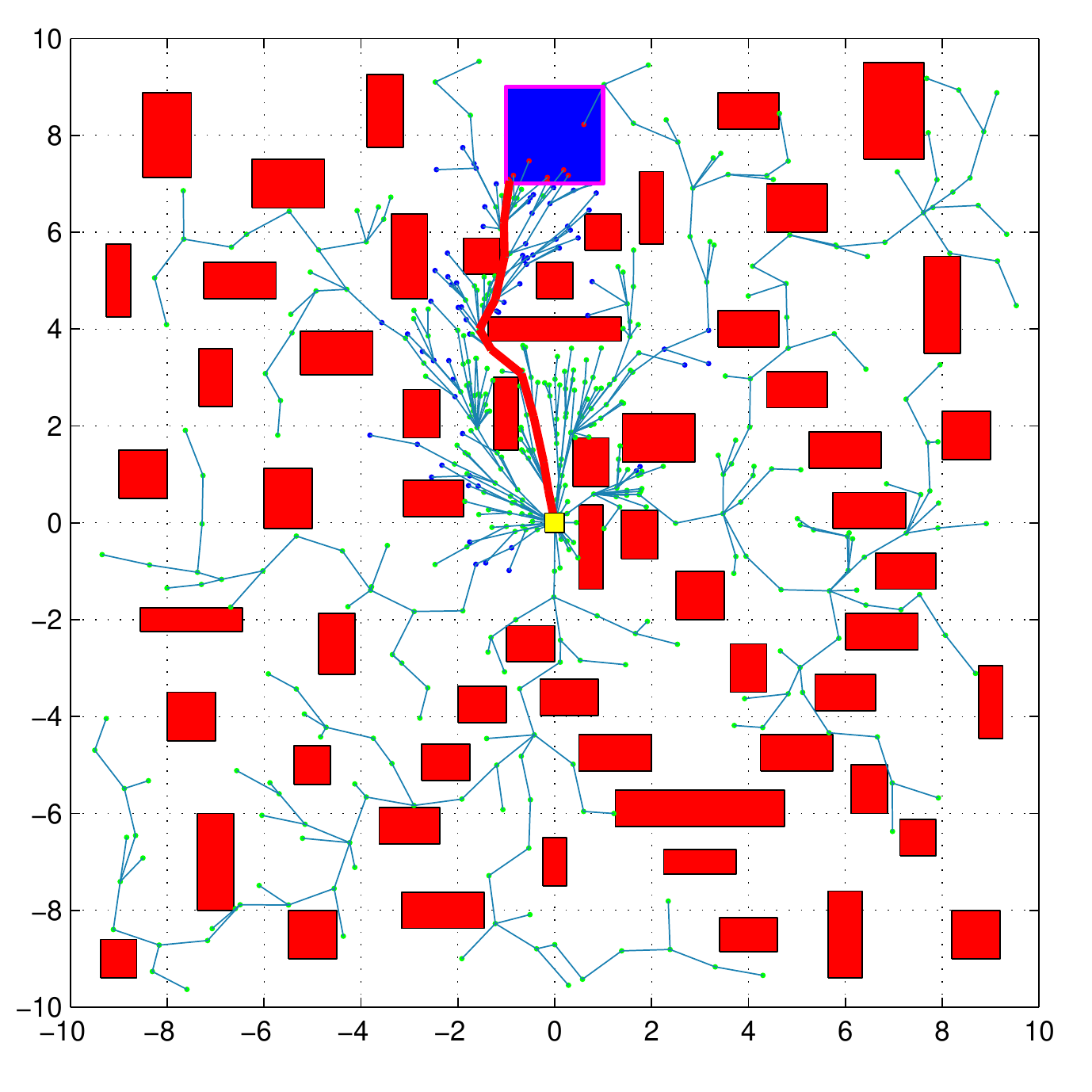}} \label{figure:pt3_rrtsharp_v3_it5000}}
    \subfigure[]{\scalebox{0.35}{\includegraphics[trim = 4.0cm 6.937cm 3.587cm 7.0cm, clip =
          true]{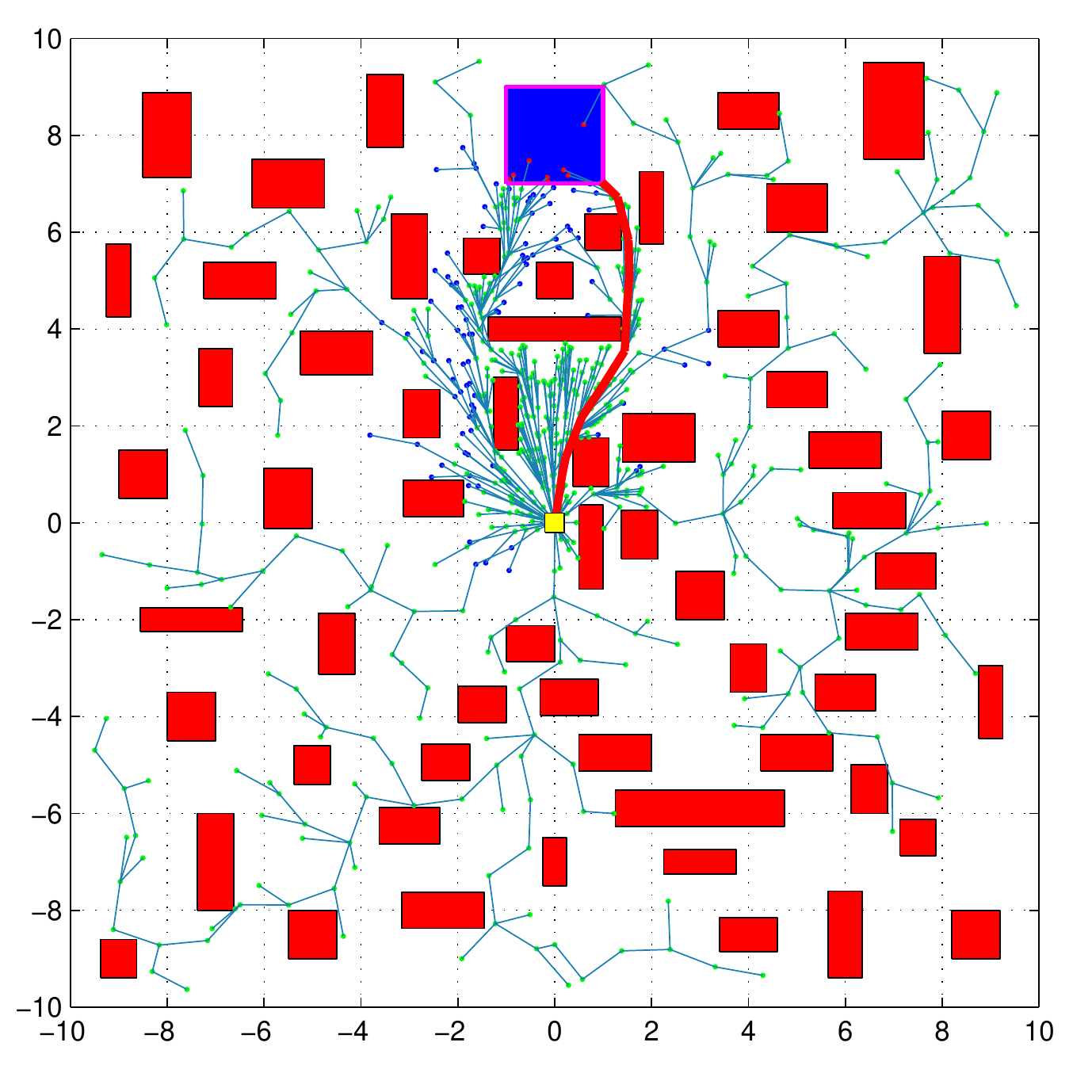}} \label{figure:pt3_rrtsharp_v3_it10000}}
    \subfigure[]{\scalebox{0.35}{\includegraphics[trim = 4.0cm 6.937cm 3.587cm 7.0cm, clip =
          true]{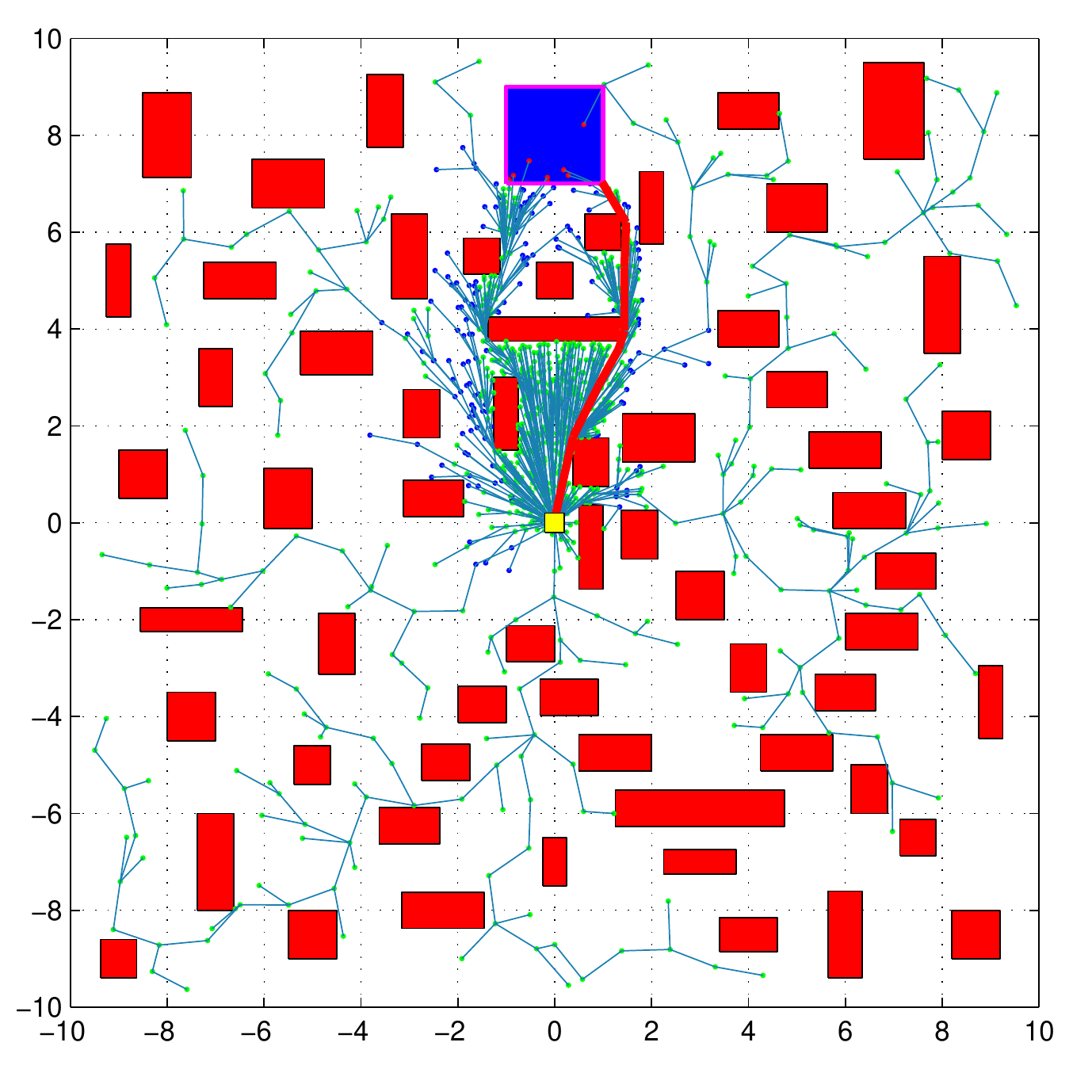}} \label{figure:pt3_rrtsharp_v3_it24999}}
    }

    \caption{The evolution of the tree computed by \AlgRRTsharpPromisingNewVertex{} algorithm is shown in   \subref{figure:pt3_rrtsharp_v3_it250}-\subref{figure:pt3_rrtsharp_v3_it24999}. The configuration of the trees in \subref{figure:pt3_rrtsharp_v3_it250} is at 250 iterations, in \subref{figure:pt3_rrtsharp_v3_it500} is at 500 iterations, in \subref{figure:pt3_rrtsharp_v3_it2500} is at 2500 iterations, in \subref{figure:pt3_rrtsharp_v3_it5000} is at 5000 iterations, in \subref{figure:pt3_rrtsharp_v3_it10000} is at 10000 iterations, and in \subref{figure:pt3_rrtsharp_v3_it24999} is at 25000 iterations.
    }
    \label{figure:sim_d2_pt3_rrtsharp_v3_iterations}
  \end{center}
\end{figure*}

\FloatBarrier

\begin{figure*}[htp]
  \begin{center}

	\mbox{
    \subfigure[]{\scalebox{0.35}{\includegraphics[trim = 4.0cm 6.937cm 3.587cm 7.0cm, clip =
          true]{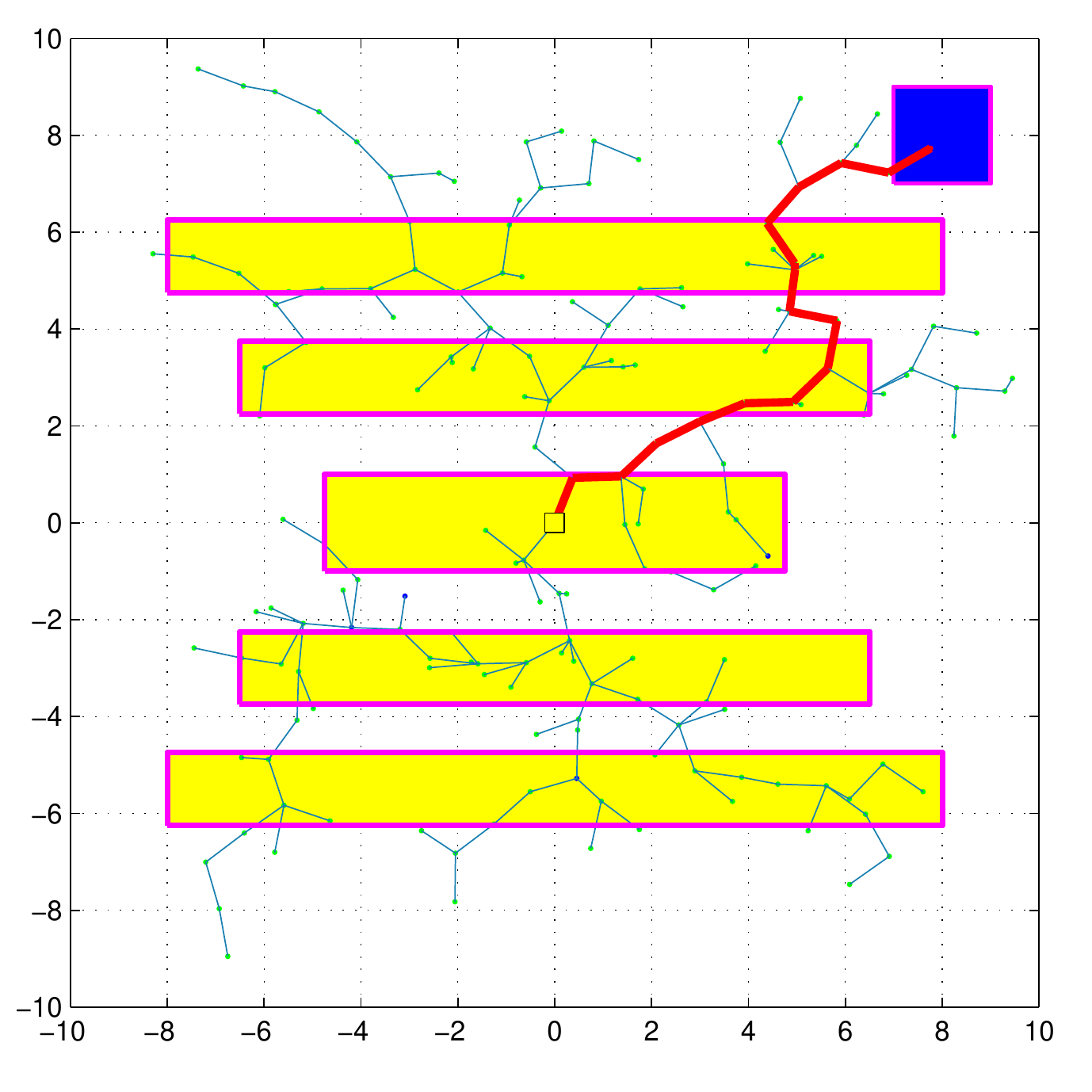}} \label{figure:pt4_rrtsharp_v3_it250}}
    \subfigure[]{\scalebox{0.35}{\includegraphics[trim = 4.0cm 6.937cm 3.587cm 7.0cm, clip =
          true]{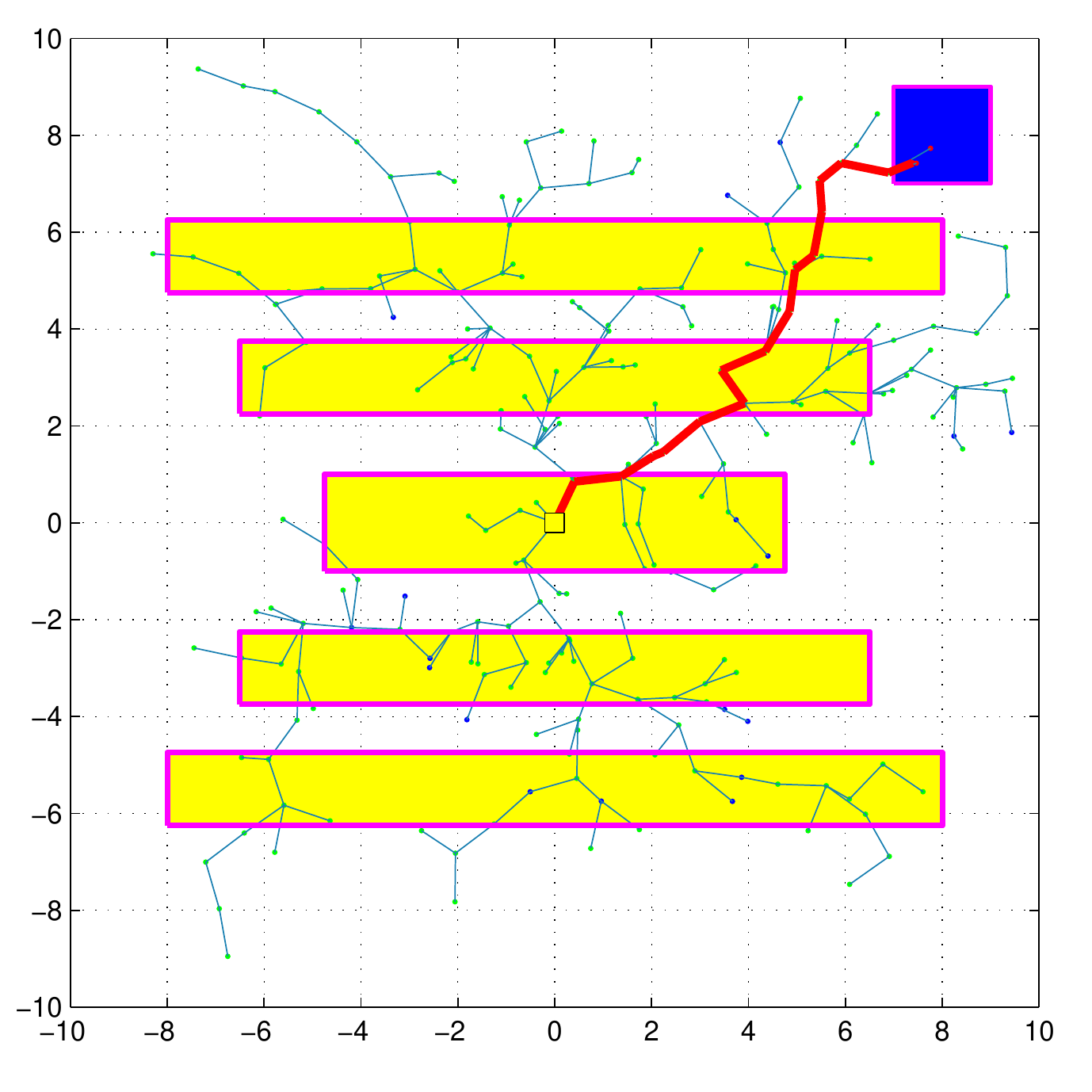}} \label{figure:pt4_rrtsharp_v3_it500}}
    \subfigure[]{\scalebox{0.35}{\includegraphics[trim = 4.0cm 6.937cm 3.587cm 7.0cm, clip =
          true]{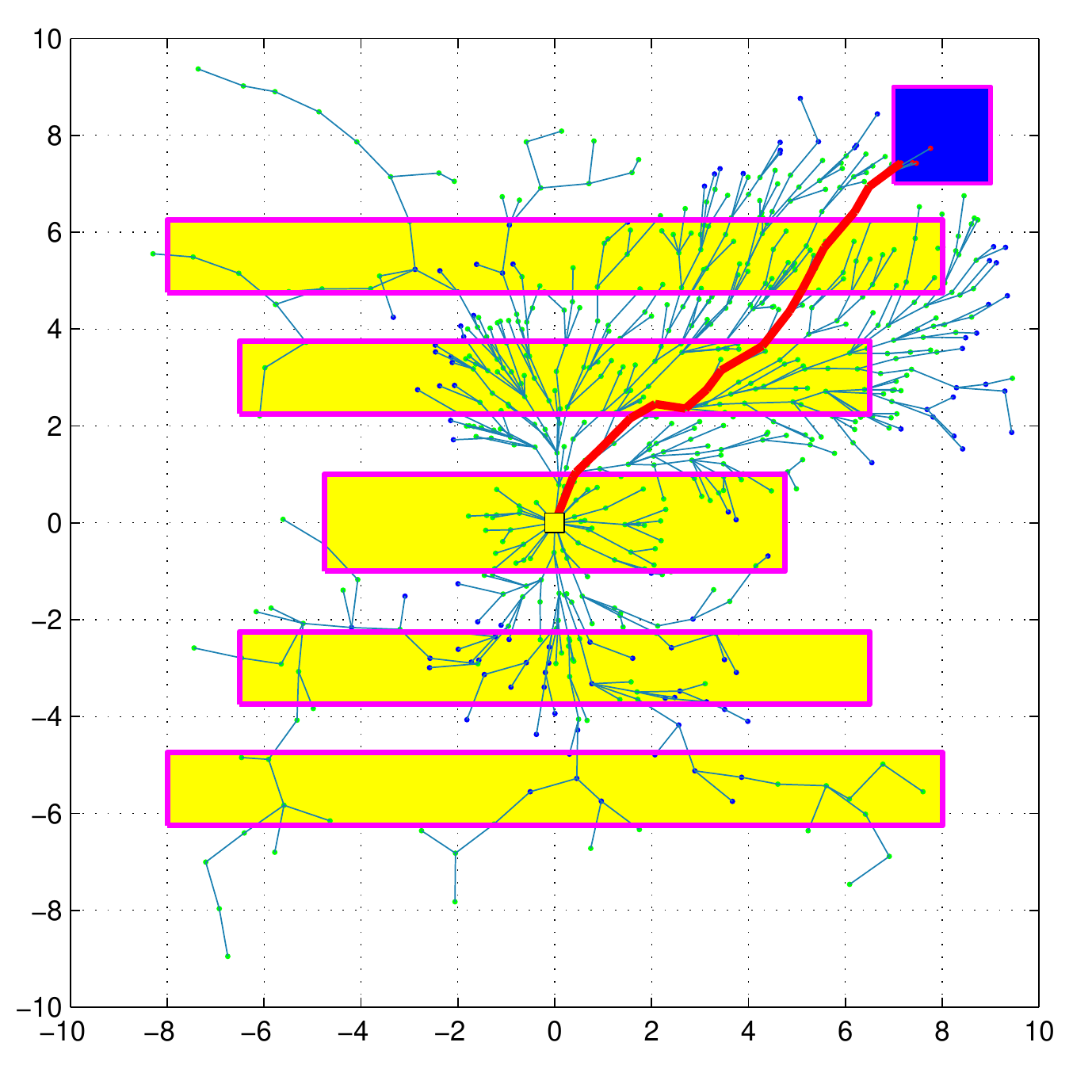}} \label{figure:pt4_rrtsharp_v3_it2500}}
   	}
   	
	\mbox{
	\subfigure[]{\scalebox{0.35}{\includegraphics[trim = 4.0cm 6.937cm 3.587cm 7.0cm, clip =
          true]{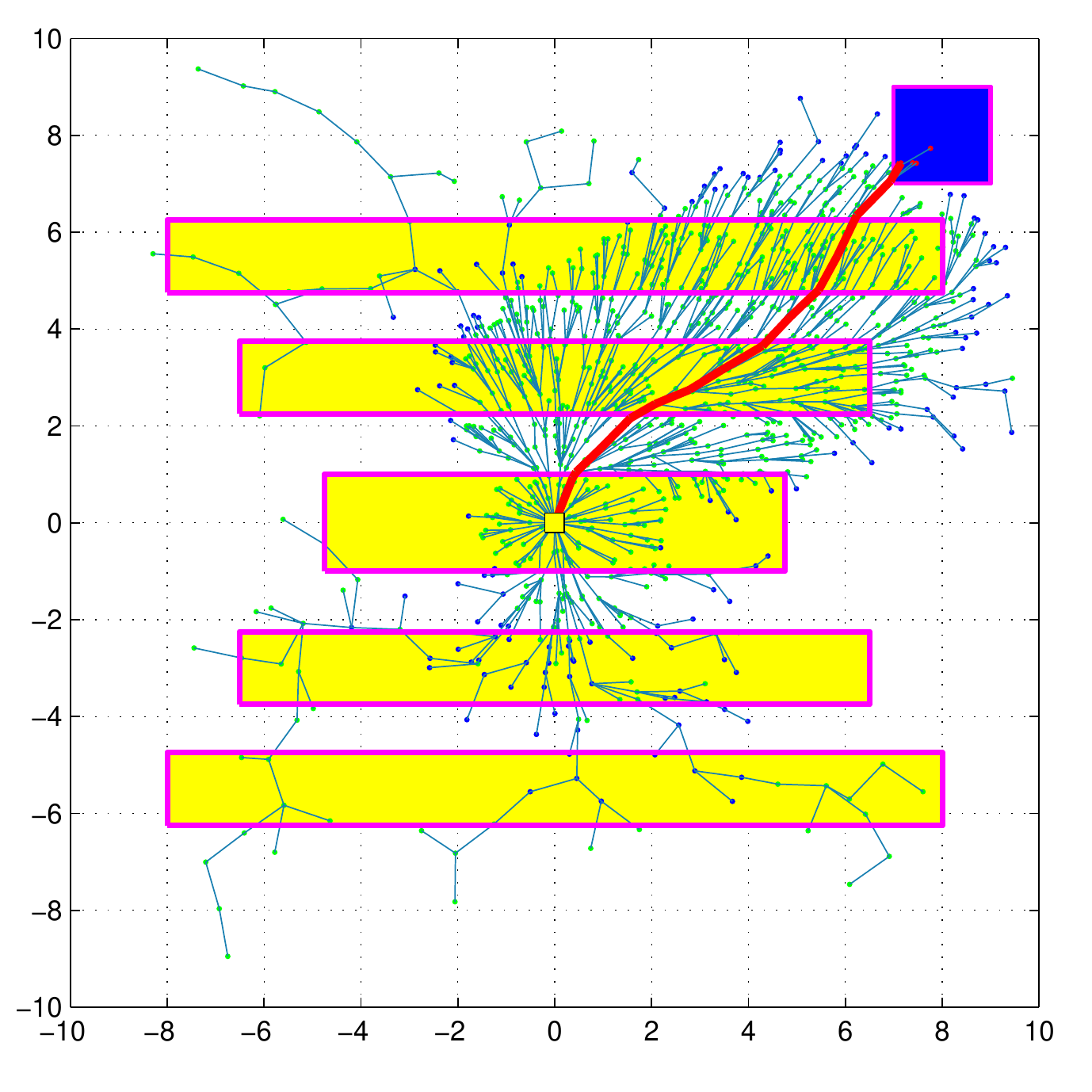}} \label{figure:pt4_rrtsharp_v3_it5000}}
    \subfigure[]{\scalebox{0.35}{\includegraphics[trim = 4.0cm 6.937cm 3.587cm 7.0cm, clip =
          true]{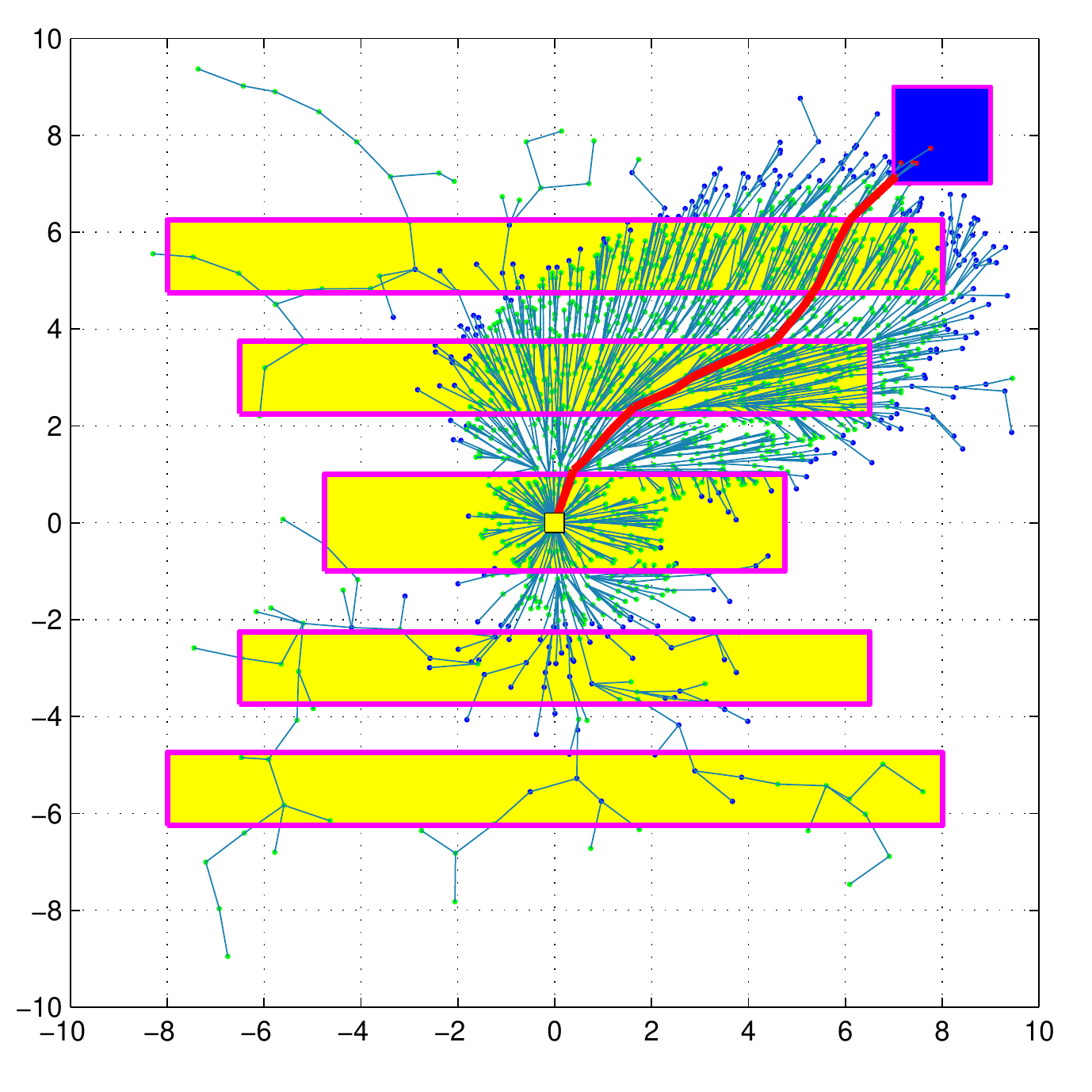}} \label{figure:pt4_rrtsharp_v3_it10000}}
    \subfigure[]{\scalebox{0.35}{\includegraphics[trim = 4.0cm 6.937cm 3.587cm 7.0cm, clip =
          true]{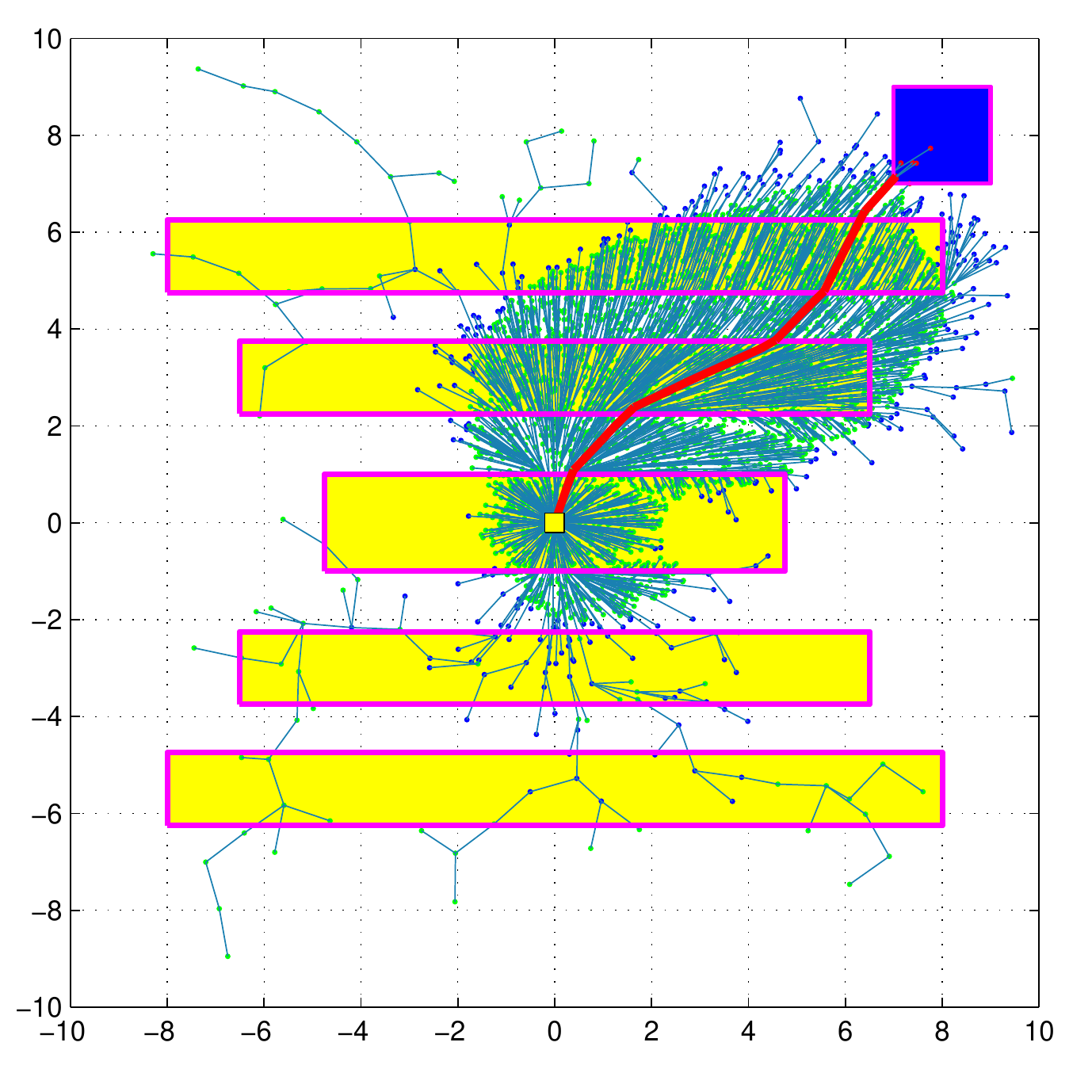}} \label{figure:pt4_rrtsharp_v3_it24999}}
    }

    \caption{The evolution of the tree computed by \AlgRRTsharpPromisingNewVertex{} algorithm is shown in   \subref{figure:pt4_rrtsharp_v3_it250}-\subref{figure:pt4_rrtsharp_v3_it24999}. The configuration of the trees in \subref{figure:pt4_rrtsharp_v3_it250} is at 250 iterations, in \subref{figure:pt4_rrtsharp_v3_it500} is at 500 iterations, in \subref{figure:pt4_rrtsharp_v3_it2500} is at 2500 iterations, in \subref{figure:pt4_rrtsharp_v3_it5000} is at 5000 iterations, in \subref{figure:pt4_rrtsharp_v3_it10000} is at 10000 iterations, and in \subref{figure:pt4_rrtsharp_v3_it24999} is at 25000 iterations.
    }

    \label{figure:sim_d2_pt4_rrtsharp_v3_iterations}
  \end{center}
\end{figure*}

\FloatBarrier

\begin{figure*}[htp]
  \begin{center}
	\mbox{
        \tikzmark{lr1st}\subfigure[]{\scalebox{0.26}{\includegraphics[trim = 4.0cm 3.0cm 4.0cm 3.0cm, clip =
          true]{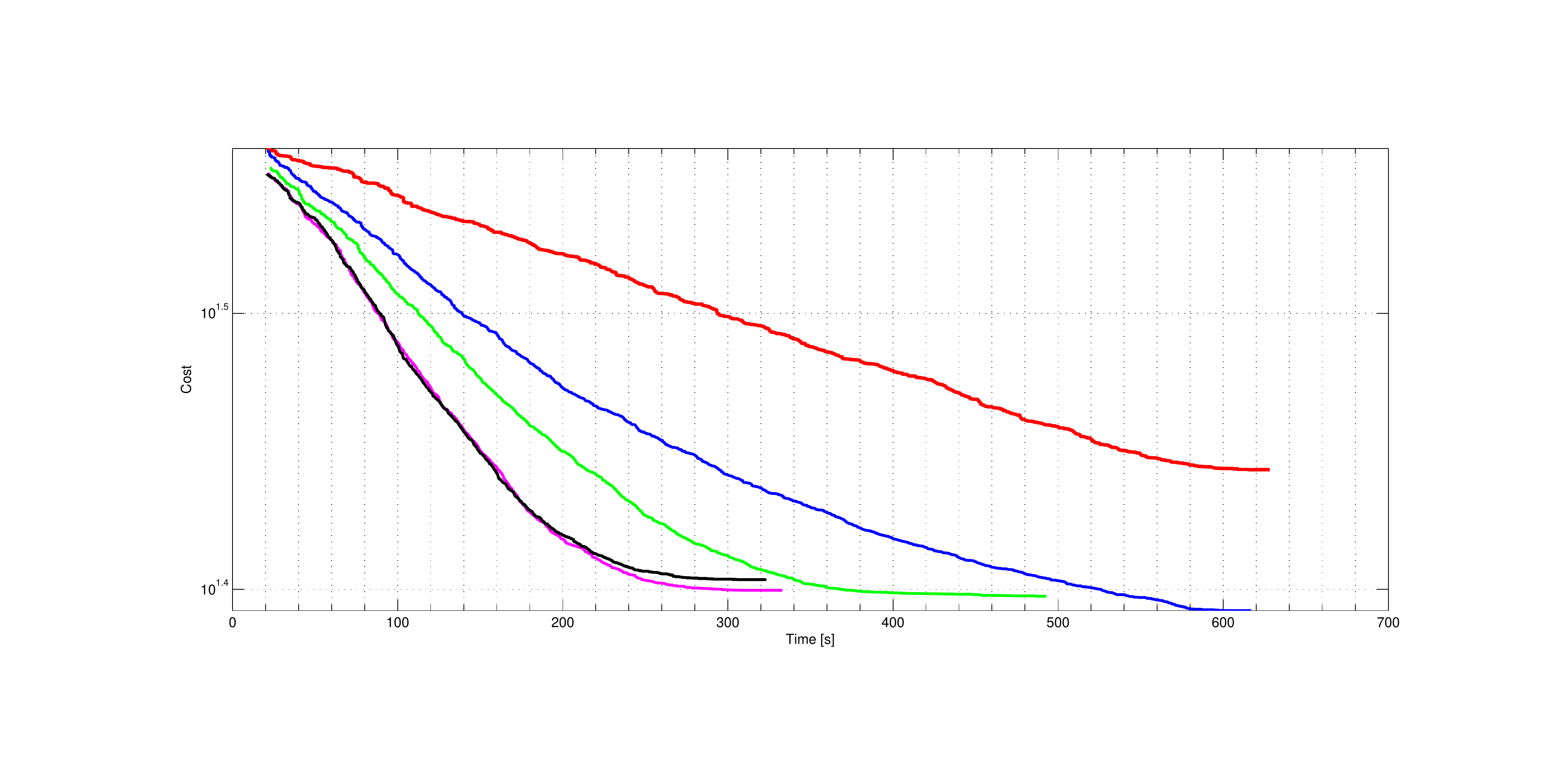}}\label{figure:time_cost_mean_d5_pt1_all}}\tikzmark{lr1en}
    	\tikzmark{lr2st}\subfigure[]{\scalebox{0.26}{\includegraphics[trim = 4.0cm 3.0cm 4.0cm 3.0cm, clip =
          true]{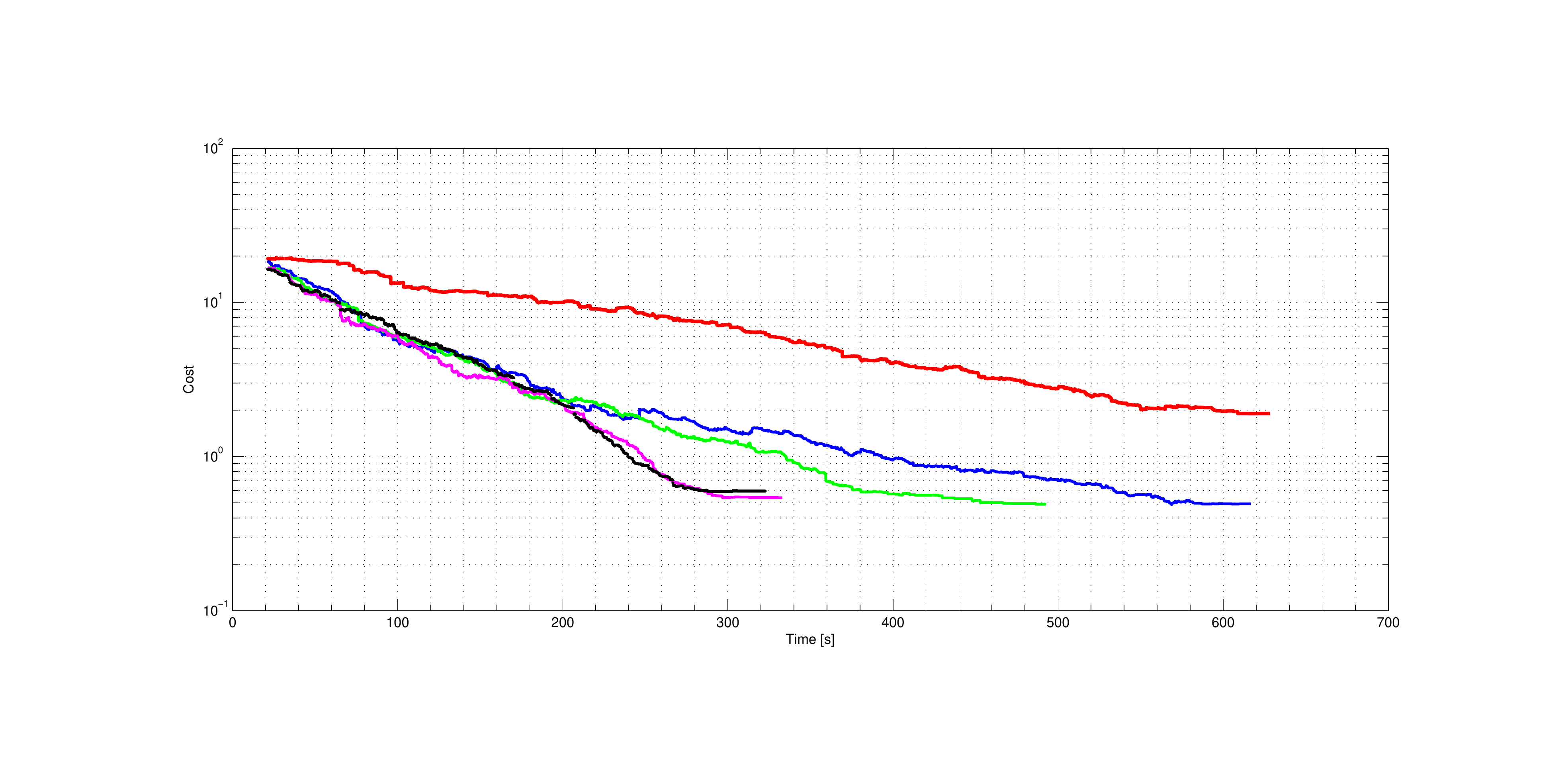}}\label{figure:time_cost_variance_d5_pt1_all}}\tikzmark{lr2en}
    }


    \caption{The change in the cost of the best paths computed by \AlgRRTstar{}, \AlgRRTsharp{}, and its variant algorithms and the variance of the trials is shown in \subref{figure:time_cost_mean_d5_pt1_all} and \subref{figure:time_cost_variance_d5_pt1_all}, respectively (problem type 1, 5D search space).}
    \label{figure:sim_d5_pt1_all_histories}

    \begin{tikzpicture}[
	remember picture,
	overlay,
	init/.style={inner sep=0pt}]
	\tiny
	\coordinate (lr1s) at ($(lr1st)+(6.7,1.95)$);
	\coordinate (lr1e) at ($(lr1en)+(-0.3,3.3)$);
	\coordinate (l1s) at ($(lr1st)+(6.7,3.25)$);
	\coordinate (l1e) at ($(lr1st)+(7.1,3.25)$);
	\coordinate (l2s) at ($(lr1st)+(6.7,2.95)$);
	\coordinate (l2e) at ($(lr1st)+(7.1,2.95)$);
	\coordinate (l3s) at ($(lr1st)+(6.7,2.65)$);
	\coordinate (l3e) at ($(lr1st)+(7.1,2.65)$);
	\coordinate (l4s) at ($(lr1st)+(6.7,2.35)$);
	\coordinate (l4e) at ($(lr1st)+(7.1,2.35)$);
	\coordinate (l5s) at ($(lr1st)+(6.7,2.05)$);
	\coordinate (l5e) at ($(lr1st)+(7.1,2.05)$);
	
	\node[draw=black,rectangle, fit=(lr1s) (lr1e)] {};
	\node[init] at (l1s) (n1s) {};
	\node[init] at (l1e) (n1e) [label=right:{\fontsize{0.5mm}{1mm}\selectfont\AlgRRTstar}]{};
	\node[init] at (l2s) (n2s) {};
	\node[init] at (l2e) (n2e) [label=right:{\fontsize{0.5mm}{1mm}\selectfont\AlgRRTsharp}]{};
	\node[init] at (l3s) (n3s) {};
	\node[init] at (l3e) (n3e) [label=right:{\fontsize{0.5mm}{1mm}\selectfont\AlgRRTsharpNoBlackVertex}] {};
	\node[init] at (l4s) (n4s) {};
	\node[init] at (l4e) (n4e) [label=right:{\fontsize{0.5mm}{1mm}\selectfont\AlgRRTsharpPromisingParent}]{};
	\node[init] at (l5s) (n5s) {};
	\node[init] at (l5e) (n5e) [label=right:{\fontsize{0.5mm}{1mm}\selectfont\AlgRRTsharpPromisingNewVertex}] {};
	\draw[red,thick] (n1s) -- (n1e);
	\draw[blue,thick] (n2s) -- (n2e);
	\draw[green,thick] (n3s) -- (n3e);
	\draw[magenta,thick] (n4s) -- (n4e);
	\draw[black,thick] (n5s) -- (n5e);
	
	\coordinate (lr2s) at ($(lr2st)+(6.7,1.95)$);
	\coordinate (lr2e) at ($(lr2en)+(-0.3,3.3)$);
	\coordinate (l6s) at ($(lr2st)+(6.7,3.25)$);
	\coordinate (l6e) at ($(lr2st)+(7.1,3.25)$);
	\coordinate (l7s) at ($(lr2st)+(6.7,2.95)$);
	\coordinate (l7e) at ($(lr2st)+(7.1,2.95)$);
	\coordinate (l8s) at ($(lr2st)+(6.7,2.65)$);
	\coordinate (l8e) at ($(lr2st)+(7.1,2.65)$);
	\coordinate (l9s) at ($(lr2st)+(6.7,2.35)$);
	\coordinate (l9e) at ($(lr2st)+(7.1,2.35)$);
	\coordinate (l10s) at ($(lr2st)+(6.7,2.05)$);
	\coordinate (l10e) at ($(lr2st)+(7.1,2.05)$);
	
	\node[draw=black,rectangle, fit=(lr2s) (lr2e)] {};
	\node[init] at (l6s) (n6s) {};
	\node[init] at (l6e) (n6e) [label=right:{\fontsize{0.5mm}{1mm}\selectfont\AlgRRTstar}]{};
	\node[init] at (l7s) (n7s) {};
	\node[init] at (l7e) (n7e) [label=right:{\fontsize{0.5mm}{1mm}\selectfont\AlgRRTsharp}]{};
	\node[init] at (l8s) (n8s) {};
	\node[init] at (l8e) (n8e) [label=right:{\fontsize{0.5mm}{1mm}\selectfont\AlgRRTsharpNoBlackVertex}] {};
	\node[init] at (l9s) (n9s) {};
	\node[init] at (l9e) (n9e) [label=right:{\fontsize{0.5mm}{1mm}\selectfont\AlgRRTsharpPromisingParent}]{};
	\node[init] at (l10s) (n10s) {};
	\node[init] at (l10e) (n10e) [label=right:{\fontsize{0.5mm}{1mm}\selectfont\AlgRRTsharpPromisingNewVertex}] {};
	\draw[red,thick] (n6s) -- (n6e);
	\draw[blue,thick] (n7s) -- (n7e);
	\draw[green,thick] (n8s) -- (n8e);
	\draw[magenta,thick] (n9s) -- (n9e);
	\draw[black,thick] (n10s) -- (n10e);	
\end{tikzpicture}
  \end{center}
\end{figure*}

\begin{figure*}[htp]
  \begin{center}
	\mbox{
        \tikzmark{lr1st}\subfigure[]{\scalebox{0.26}{\includegraphics[trim = 4.0cm 3.0cm 4.0cm 3.0cm, clip =
          true]{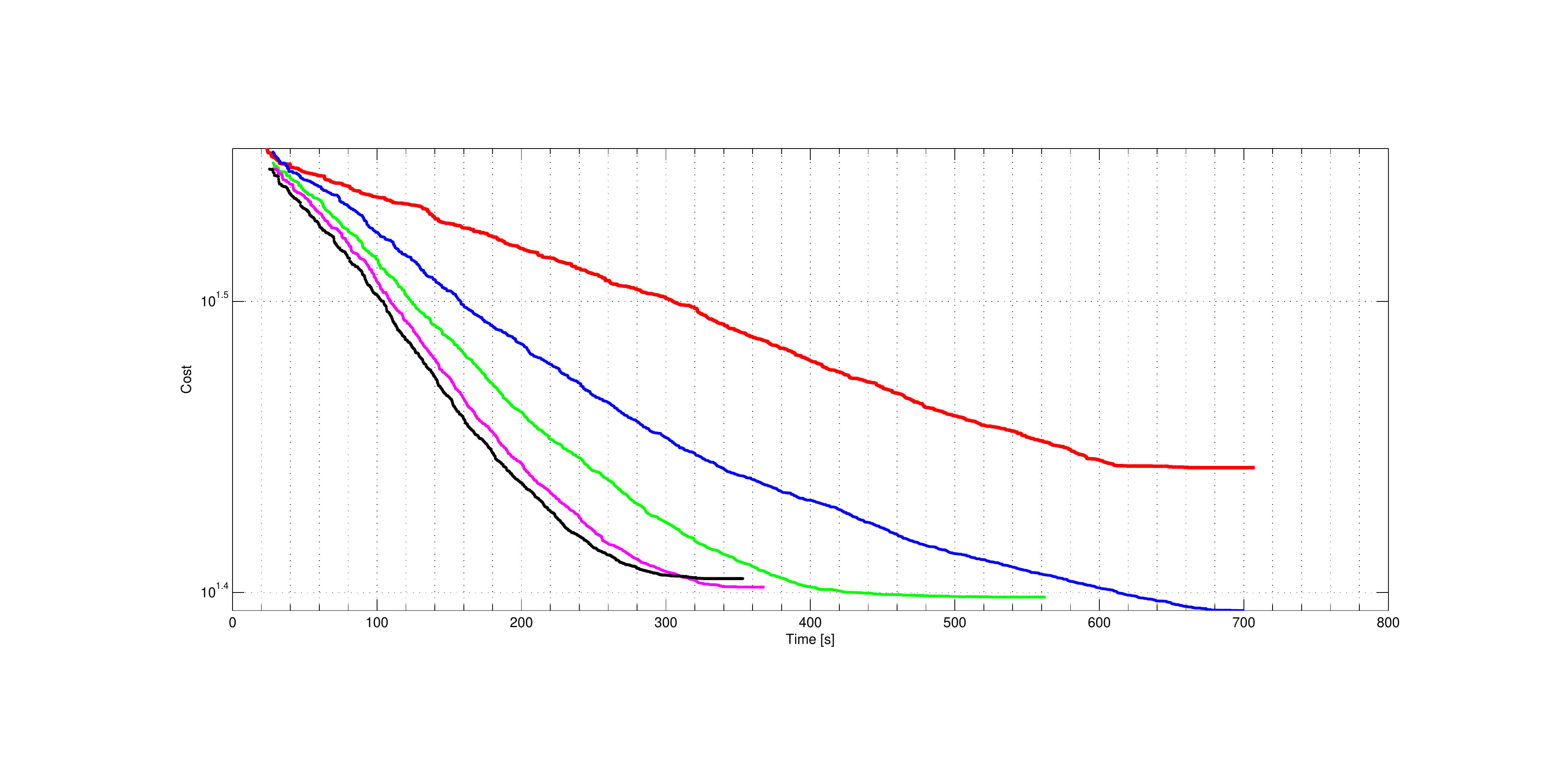}}\label{figure:time_cost_mean_d5_pt2_all}}\tikzmark{lr1en}
        \tikzmark{lr2st}\subfigure[]{\scalebox{0.26}{\includegraphics[trim = 4.0cm 3.0cm 4.0cm 3.0cm, clip =
          true]{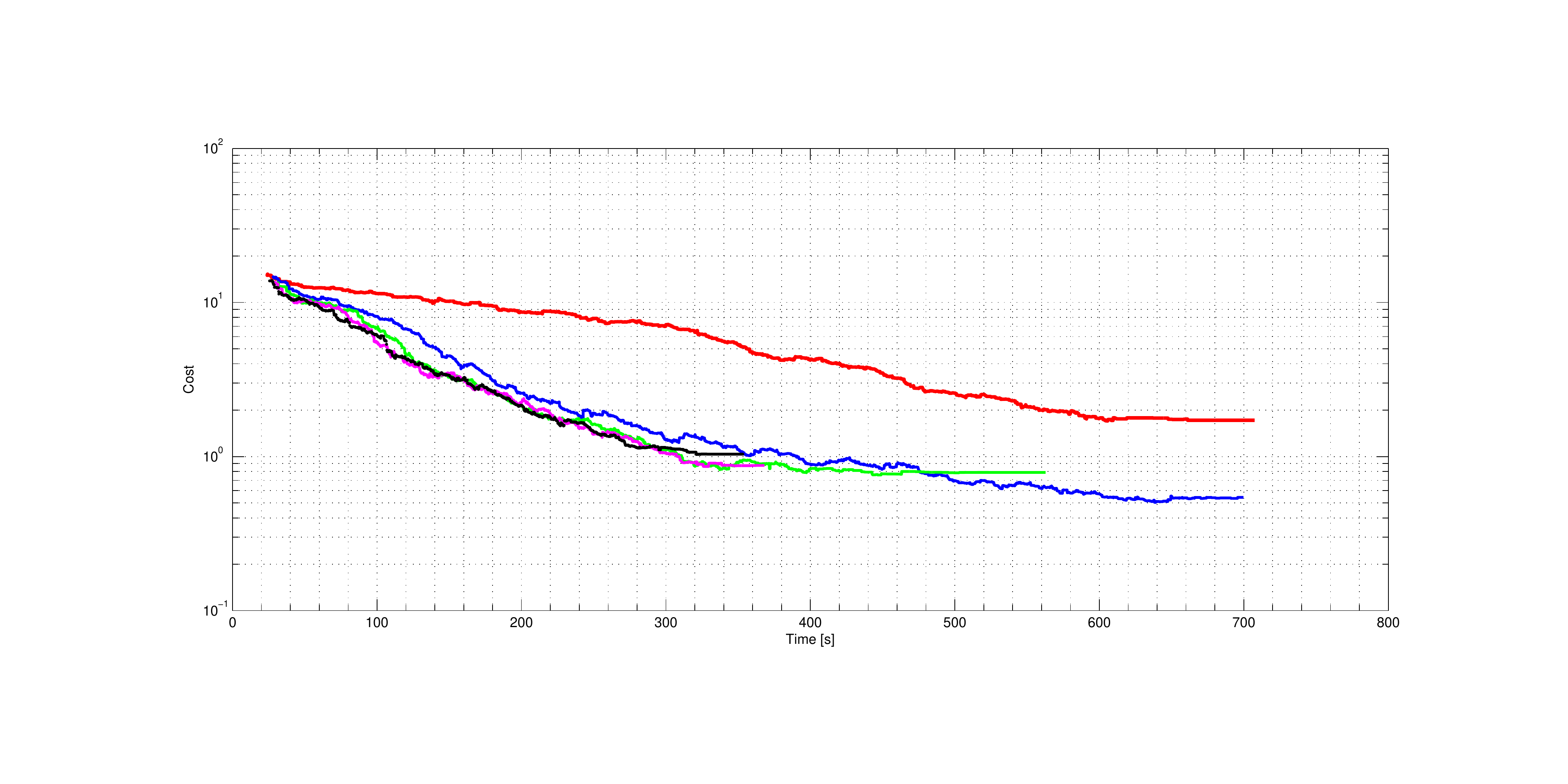}}\label{figure:time_cost_variance_d5_pt2_all}}\tikzmark{lr2en}
    }


    \caption{The change in the cost of the best paths computed by \AlgRRTstar{}, \AlgRRTsharp{}, and its variant algorithms and the variance of the trials are shown in \subref{figure:time_cost_mean_d5_pt2_all} and \subref{figure:time_cost_variance_d5_pt2_all}, respectively (problem type 2, 5D search space).}
    \label{figure:sim_d5_pt2_all_histories}

\begin{tikzpicture}[
	remember picture,
	overlay,
	init/.style={inner sep=0pt}]
	\tiny
	\coordinate (lr1s) at ($(lr1st)+(6.7,1.95)$);
	\coordinate (lr1e) at ($(lr1en)+(-0.3,3.3)$);
	\coordinate (l1s) at ($(lr1st)+(6.7,3.25)$);
	\coordinate (l1e) at ($(lr1st)+(7.1,3.25)$);
	\coordinate (l2s) at ($(lr1st)+(6.7,2.95)$);
	\coordinate (l2e) at ($(lr1st)+(7.1,2.95)$);
	\coordinate (l3s) at ($(lr1st)+(6.7,2.65)$);
	\coordinate (l3e) at ($(lr1st)+(7.1,2.65)$);
	\coordinate (l4s) at ($(lr1st)+(6.7,2.35)$);
	\coordinate (l4e) at ($(lr1st)+(7.1,2.35)$);
	\coordinate (l5s) at ($(lr1st)+(6.7,2.05)$);
	\coordinate (l5e) at ($(lr1st)+(7.1,2.05)$);
	
	\node[draw=black,rectangle, fit=(lr1s) (lr1e)] {};
	\node[init] at (l1s) (n1s) {};
	\node[init] at (l1e) (n1e) [label=right:{\fontsize{0.5mm}{1mm}\selectfont\AlgRRTstar}]{};
	\node[init] at (l2s) (n2s) {};
	\node[init] at (l2e) (n2e) [label=right:{\fontsize{0.5mm}{1mm}\selectfont\AlgRRTsharp}]{};
	\node[init] at (l3s) (n3s) {};
	\node[init] at (l3e) (n3e) [label=right:{\fontsize{0.5mm}{1mm}\selectfont\AlgRRTsharpNoBlackVertex}] {};
	\node[init] at (l4s) (n4s) {};
	\node[init] at (l4e) (n4e) [label=right:{\fontsize{0.5mm}{1mm}\selectfont\AlgRRTsharpPromisingParent}]{};
	\node[init] at (l5s) (n5s) {};
	\node[init] at (l5e) (n5e) [label=right:{\fontsize{0.5mm}{1mm}\selectfont\AlgRRTsharpPromisingNewVertex}] {};
	\draw[red,thick] (n1s) -- (n1e);
	\draw[blue,thick] (n2s) -- (n2e);
	\draw[green,thick] (n3s) -- (n3e);
	\draw[magenta,thick] (n4s) -- (n4e);
	\draw[black,thick] (n5s) -- (n5e);
	
	\coordinate (lr2s) at ($(lr2st)+(6.7,1.95)$);
	\coordinate (lr2e) at ($(lr2en)+(-0.3,3.3)$);
	\coordinate (l6s) at ($(lr2st)+(6.7,3.25)$);
	\coordinate (l6e) at ($(lr2st)+(7.1,3.25)$);
	\coordinate (l7s) at ($(lr2st)+(6.7,2.95)$);
	\coordinate (l7e) at ($(lr2st)+(7.1,2.95)$);
	\coordinate (l8s) at ($(lr2st)+(6.7,2.65)$);
	\coordinate (l8e) at ($(lr2st)+(7.1,2.65)$);
	\coordinate (l9s) at ($(lr2st)+(6.7,2.35)$);
	\coordinate (l9e) at ($(lr2st)+(7.1,2.35)$);
	\coordinate (l10s) at ($(lr2st)+(6.7,2.05)$);
	\coordinate (l10e) at ($(lr2st)+(7.1,2.05)$);
	
	\node[draw=black,rectangle, fit=(lr2s) (lr2e)] {};
	\node[init] at (l6s) (n6s) {};
	\node[init] at (l6e) (n6e) [label=right:{\fontsize{0.5mm}{1mm}\selectfont\AlgRRTstar}]{};
	\node[init] at (l7s) (n7s) {};
	\node[init] at (l7e) (n7e) [label=right:{\fontsize{0.5mm}{1mm}\selectfont\AlgRRTsharp}]{};
	\node[init] at (l8s) (n8s) {};
	\node[init] at (l8e) (n8e) [label=right:{\fontsize{0.5mm}{1mm}\selectfont\AlgRRTsharpNoBlackVertex}] {};
	\node[init] at (l9s) (n9s) {};
	\node[init] at (l9e) (n9e) [label=right:{\fontsize{0.5mm}{1mm}\selectfont\AlgRRTsharpPromisingParent}]{};
	\node[init] at (l10s) (n10s) {};
	\node[init] at (l10e) (n10e) [label=right:{\fontsize{0.5mm}{1mm}\selectfont\AlgRRTsharpPromisingNewVertex}] {};
	\draw[red,thick] (n6s) -- (n6e);
	\draw[blue,thick] (n7s) -- (n7e);
	\draw[green,thick] (n8s) -- (n8e);
	\draw[magenta,thick] (n9s) -- (n9e);
	\draw[black,thick] (n10s) -- (n10e);	
\end{tikzpicture}

  \end{center}

\end{figure*}

A Monte-Carlo study was performed in order to compare the convergence rate and variance in the trials of all algorithms in a high dimensional search space. All algorithms were run up until 4 million iterations 100 times in a 5-dimensional search space for Problem types 1 and 2. In the second problem type, several 5-dimensional hypercubes of different size were randomly placed in the environment in order to represent obstacles. As shown in Figures~\ref{figure:sim_d5_pt1_all_histories} and \ref{figure:sim_d5_pt2_all_histories}, the \AlgRRTsharpPromisingParent{} and \AlgRRTsharpPromisingNewVertex{} algorithms find the solution in a similar amount of time, and they are faster than the other algorithms. In addition, they compute solutions of lower cost than the other algorithms with smaller variance in the trials.

\FloatBarrier

\begin{figure*}[htp]
  \begin{center}
	\mbox{
        \tikzmark{l1st}\subfigure[]{\scalebox{0.26}{\includegraphics[trim = 4.0cm 3.0cm 4.0cm 3.0cm, clip =
          true]{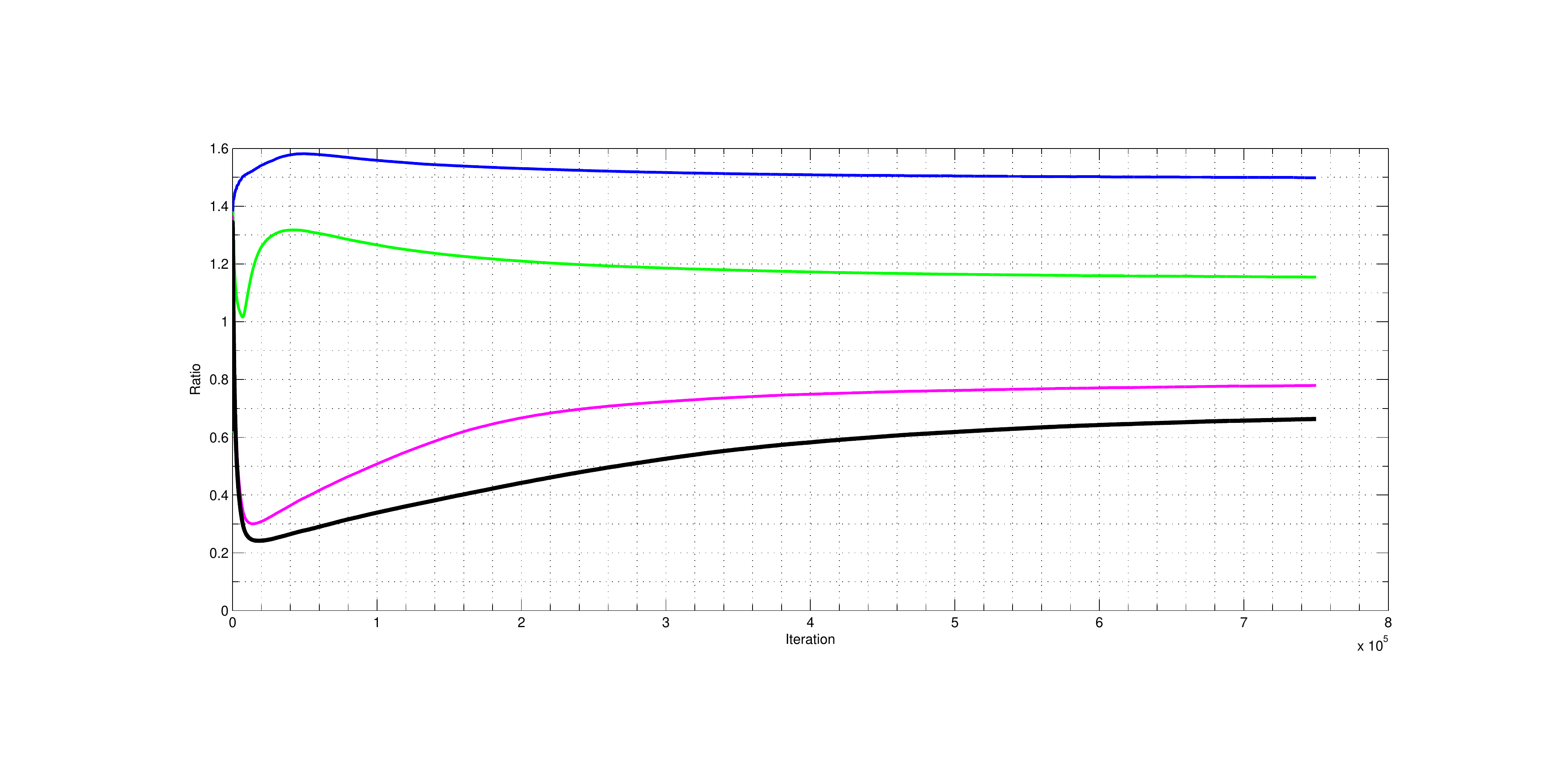}}\label{figure:time_ratio_d2_pt1_all}}\tikzmark{l1en}
        \tikzmark{l6st}\subfigure[]{\scalebox{0.26}{\includegraphics[trim = 4.0cm 3.0cm 4.0cm 3.0cm, clip =
          true]{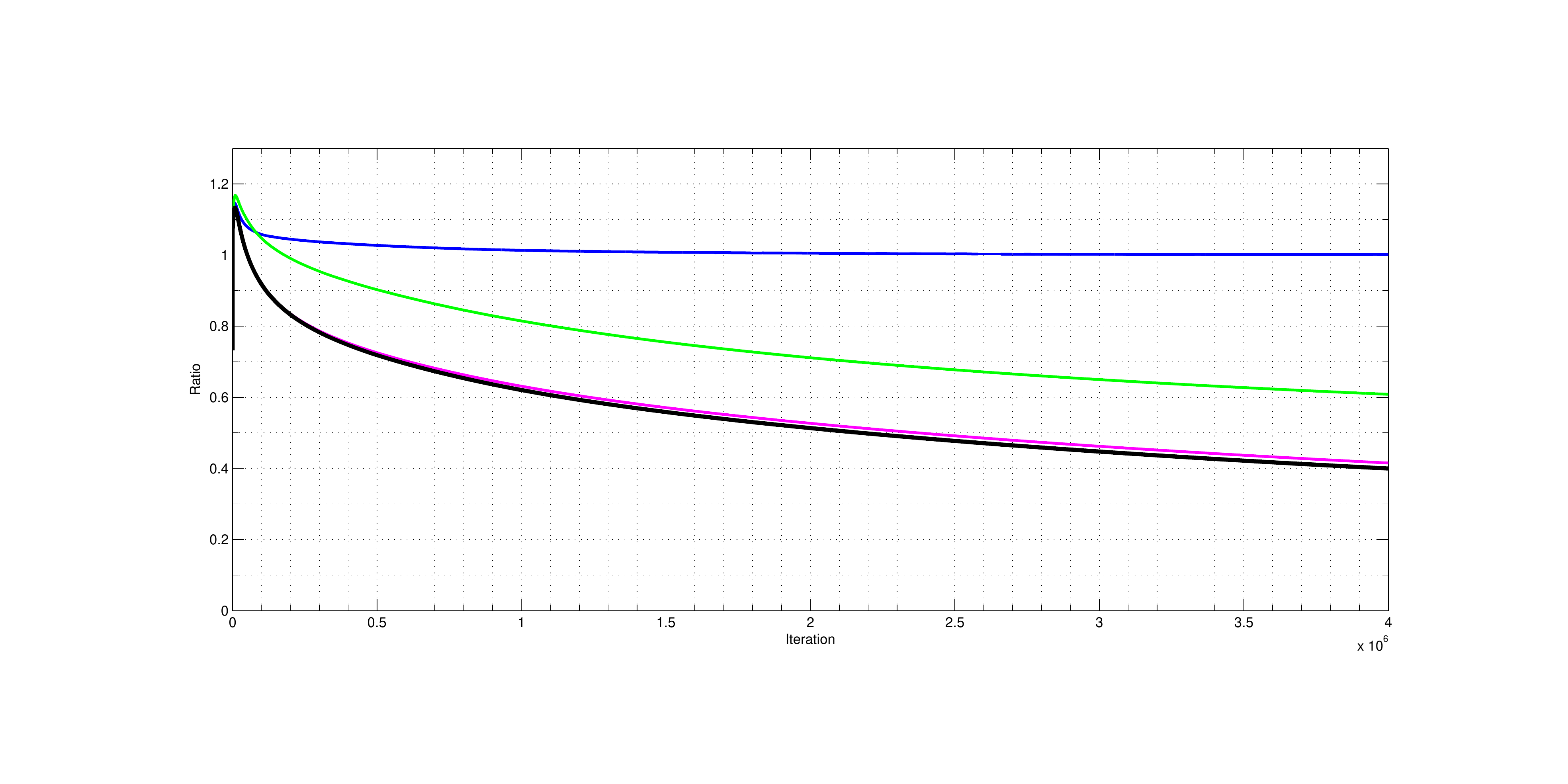}}\label{figure:time_ratio_d5_pt1_all}}\tikzmark{l6en}
    }


    \caption{Comparison of execution time of all algorithms (Problem type 1)}
    \label{figure:sim_d2_pt1_all_time_ratio}

    \begin{tikzpicture}[
	remember picture,
	overlay,
	init/.style={inner sep=0pt}]
	\tiny
	\coordinate (l1s) at ($(l1st)+(6.7,0.5)$);
	\coordinate (l1e) at ($(l1en)+(-0.3,1.55)$);
	\coordinate (l2s) at ($(l1st)+(6.7,1.5)$);
	\coordinate (l2e) at ($(l1st)+(7.1,1.5)$);
	\coordinate (l3s) at ($(l1st)+(6.7,1.2)$);
	\coordinate (l3e) at ($(l1st)+(7.1,1.2)$);
	\coordinate (l4s) at ($(l1st)+(6.7,0.9)$);
	\coordinate (l4e) at ($(l1st)+(7.1,0.9)$);
	\coordinate (l5s) at ($(l1st)+(6.7,0.6)$);
	\coordinate (l5e) at ($(l1st)+(7.1,0.6)$);
	
	\node[draw=black,rectangle, fit=(l1s) (l1e)] {};
	\node[init] at (l2s) (n2s) {};
	\node[init] at (l2e) (n2e) [label=right:{\fontsize{0.5mm}{1mm}\selectfont\AlgRRTsharp}]{};
	\node[init] at (l3s) (n3s) {};
	\node[init] at (l3e) (n3e) [label=right:{\fontsize{0.5mm}{1mm}\selectfont\AlgRRTsharpNoBlackVertex}] {};
	\node[init] at (l4s) (n4s) {};
	\node[init] at (l4e) (n4e) [label=right:{\fontsize{0.5mm}{1mm}\selectfont\AlgRRTsharpPromisingParent}]{};
	\node[init] at (l5s) (n5s) {};
	\node[init] at (l5e) (n5e) [label=right:{\fontsize{0.5mm}{1mm}\selectfont\AlgRRTsharpPromisingNewVertex}] {};
	
	\draw[blue,thick] (n2s) -- (n2e);
	\draw[green,thick] (n3s) -- (n3e);
	\draw[magenta,thick] (n4s) -- (n4e);
	\draw[black,thick] (n5s) -- (n5e);
	
	\coordinate (r0) at ($(-6.05,0)$);
	\coordinate (l6s) at ($(l6st)+(6.7,0.5) + (r0)$);
	\coordinate (l6e) at ($(l6en)+(-0.3,1.55) + (r0)$);
	\coordinate (l7s) at ($(l6st)+(6.7,1.5) + (r0)$);
	\coordinate (l7e) at ($(l6st)+(7.1,1.5) + (r0)$);
	\coordinate (l8s) at ($(l6st)+(6.7,1.2) + (r0)$);
	\coordinate (l8e) at ($(l6st)+(7.1,1.2) + (r0)$);
	\coordinate (l9s) at ($(l6st)+(6.7,0.9) + (r0)$);
	\coordinate (l9e) at ($(l6st)+(7.1,0.9) + (r0)$);
	\coordinate (l10s) at ($(l6st)+(6.7,0.6) + (r0)$);
	\coordinate (l10e) at ($(l6st)+(7.1,0.6) + (r0)$);
	
	\node[draw=black,rectangle, fit=(l6s) (l6e)] {};
	\node[init] at (l7s) (n7s) {};
	\node[init] at (l7e) (n7e) [label=right:{\fontsize{0.5mm}{1mm}\selectfont\AlgRRTsharp}]{};
	\node[init] at (l8s) (n8s) {};
	\node[init] at (l8e) (n8e) [label=right:{\fontsize{0.5mm}{1mm}\selectfont\AlgRRTsharpNoBlackVertex}] {};
	\node[init] at (l9s) (n9s) {};
	\node[init] at (l9e) (n9e) [label=right:{\fontsize{0.5mm}{1mm}\selectfont\AlgRRTsharpPromisingParent}]{};
	\node[init] at (l10s) (n10s) {};
	\node[init] at (l10e) (n10e) [label=right:{\fontsize{0.5mm}{1mm}\selectfont\AlgRRTsharpPromisingNewVertex}] {};
	
	\draw[blue,thick] (n7s) -- (n7e);
	\draw[green,thick] (n8s) -- (n8e);
	\draw[magenta,thick] (n9s) -- (n9e);
	\draw[black,thick] (n10s) -- (n10e);	
\end{tikzpicture}
  \end{center}
\end{figure*}
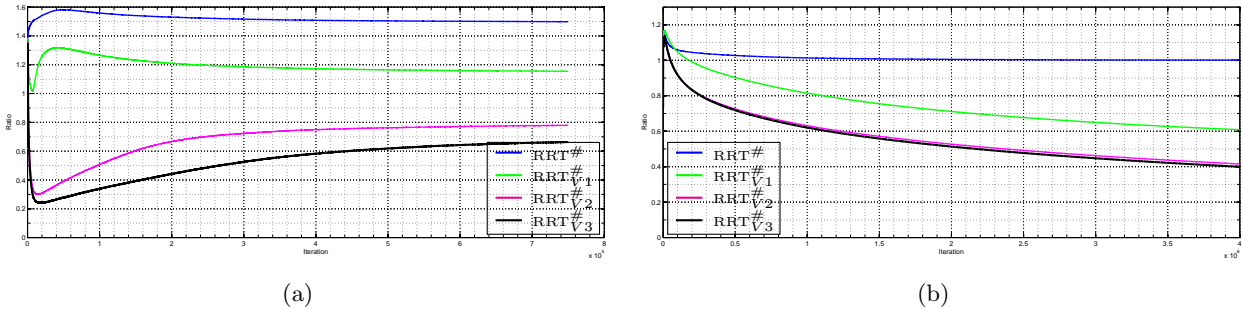

The execution times of all algorithms were also compared. Results of the \AlgRRTsharp{}, \AlgRRTsharpNoBlackVertex{}, \AlgRRTsharpPromisingParent{}, and \AlgRRTsharpPromisingNewVertex{} are plotted in blue, green, magenta, and black, respectively. All algorithms were run in a 2D and a 5D environment with no obstacles for up to 750,000 and 4,000,000 iterations, respectively. The execution time of the \AlgRRTsharp{} and its variant algorithms is normalized over that of the \AlgRRTstar algorithm and is plotted versus the number of iterations averaged over 50 trials for the 2D search space in Figure~\ref{figure:time_ratio_d2_pt1_all}. A similar plot is also created for 100 trials in the 5D search space and shown in Figure~\ref{figure:time_ratio_d5_pt1_all}.

\FloatBarrier

\section{Conclusion} \label{section:conclusion}

In this paper, a new incremental sampling-based algorithm, denoted by \AlgRRTsharp{} is presented, which offers asymptotically optimal solutions for solving motion planning problems. The \AlgRRTsharp{} algorithm relies heavily on the random geometric graph data structure and the \AlgRRG{} algorithm~\cite{penrose2003random}, which is also known to have asymptotic optimality properties. A bottleneck of optimal sampling-based algorithms is the slow convergence to the optimal solution, although sampling-based algorithms are capable of finding a feasible solution, often almost in real-time. By incorporating consistency information of all current vertices in the tree (essentially by comparing the current cost-to-come values of the vertices with the cost-to-come values via one of the neighboring vertices) we can have more informed estimates of the optimal values of the potential paths, thus speeding up convergence. Furthermore, once a feasible path has been found, vertex consistency can be used to estimate the region where the optimal solution should be found. This results in an initial convergence rate that is better than the one of the \AlgRRTstar{} algorithm.

We have also introduced three variants to improve the convergence rate of the baseline \AlgRRTsharp{} algorithm by implementing two key features: preventing the expansion of the tree towards unfavorable regions in search space, and propagating new information throughout the tree in an efficient way. The first feature allows us to limit the number of vertices in the tree, thus resulting to the algorithm running faster.
The second feature allows us to compute solutions with a less number of vertices in the tree since any new information is exploited to the highest degree.  As a result, the convergence rate of the baseline \AlgRRTsharp{} can be improved significantly.  Extensive numerical results have verified these observations in several simulation scenarios.

The work in this paper can be extended in several directions. First, a thorough theoretical analysis is warranted in order to provide strict bounds on the convergence rate of \AlgRRTsharp{}. Second, since \AlgRRTsharp{} decomposes the vertex set into ``promising'' and ``non-promising'' ones, smarter sampling strategies can be developed to exploit this information. It is also crucial for the algorithm to reach the target set as early as possible in order to converge to the optimal solution faster. In that respect, a bi-directional version of the \AlgRRTsharp{} (like the RRT-connect in \cite{kuffner2000rrt}) can be developed in order to shorten the first time-to-connect to the goal set. Also, a parallel version of the algorithm could be implemented by running the $\PrcExtend$ and $\PrcReduceInconsistency$ procedures as separate threads. A possible implementation would be to have multiple threads implementing the $\PrcExtend$ procedure and single thread implementing the $\PrcReduceInconsistency$. Finally, the algorithm can be modified to solve motion planning problems for vehicles with complex dynamics (ground vehicles, aircraft, helicopters etc) by implementing specific local steering functions.

%

\bibliography{arslan.tsiotras.arxiv.final}
\bibliographystyle{plain}

\end{document}

%% file: rrtsharp.tex
\IncMargin{1em}
\begin{algorithm}[h]
    \small
    \DontPrintSemicolon
    \SetKwInOut{Input}{input}
    \SetKwInOut{Output}{output}
    \SetKwBlock{NoBegin}{}{end}


    \SetKwFunction{fRRTsharp}{\AlgRRTsharp}
    \SetKwFunction{fSample}{Sample}
    \SetKwFunction{fExtend}{Extend}
    \SetKwFunction{fReduceInconsistency}{ReduceInconsistency}
    \SetKwFunction{fParent}{parent}


    \SetKwData{vV}{$V$}
    \SetKwData{vE}{$E$}
    \SetKwData{vEPrime}{$E^{\prime}$}
    \SetKwData{vXInit}{$x_{\mathrm{init}}$}
    \SetKwData{vI}{$i$}
    \SetKwData{vXRand}{$x_{\mathrm{rand}}$}
    \SetKwData{vT}{$\mathcal{T}$}
    \SetKwData{vG}{$\mathcal{G}$}
    \SetKwData{vCXGoal}{$\mathcal{X}_{\mathrm{goal}}$}
    \SetKwData{vCX}{$\mathcal{X}$}
    \SetKwData{vX}{$x$}


    \SetFuncSty{textbf}
    \fRRTsharp{\vXInit, \vCXGoal, \vCX}
    \SetFuncSty{texttt}
    \NoBegin
    {
        $\vV \leftarrow \{\vXInit\}$;
        $\vE \leftarrow \emptyset$;

        $\vG \leftarrow (\vV,\vE)$;

        \For{$\vI = 1$ to $N$ \label{line:rrtsharp_itbegin}}
        {
            $\vXRand \leftarrow \fSample(\vI)$;

            $\vG \leftarrow \fExtend(\vG, \vXRand)$;

            $\fReduceInconsistency(\vG, \vCXGoal)$; \label{line:rrtsharp_itend}
        }

        $(\vV,\vE) \leftarrow \vG$;
        $\vEPrime \leftarrow \emptyset$;

        \ForEach{$\vX \in \vV$}
        {
            $\vEPrime \leftarrow \vEPrime \cup \{(\fParent(\vX),\vX)\}$
        }

        \Return{$\vT = (\vV,\vEPrime)$}
    }

\caption{Body of the \AlgRRTsharp{} Algorithm\label{alg:rrtsharp}}
\end{algorithm}
\DecMargin{1em}

%% file: extend_rrtsharp.tex
\IncMargin{1em}
\begin{algorithm}[htb]
	\small

    \DontPrintSemicolon
    \SetKwInOut{Input}{input}
    \SetKwInOut{Output}{output}
    \SetKwBlock{NoBegin}{}{end}


	\SetKwFunction{fExtend}{Extend}      
    \SetKwFunction{fNearest}{Nearest}
    \SetKwFunction{fSteer}{Steer}
    \SetKwFunction{fObstacleFree}{ObstacleFree}
    \SetKwFunction{fUpdateState}{UpdateQueue}
    \SetKwFunction{fNear}{Near}
    \SetKwFunction{fG}{g}
    \SetKwFunction{fC}{c}
    \SetKwFunction{fLMC}{lmc}
    \SetKwFunction{fParent}{parent}
    \SetKwFunction{fInitState}{Initialize}



    \SetKwData{vT}{$\mathcal{T}$}
    \SetKwData{vG}{$\mathcal{G}$}
    \SetKwData{vTPrime}{$\mathcal{T}^{\prime}$}
    \SetKwData{vGPrime}{$\mathcal{G}^{\prime}$}
    \SetKwData{vX}{$x$}
    \SetKwData{vV}{$V$}
    \SetKwData{vE}{$E$}
    \SetKwData{vEPrime}{$E^{\prime}$}
    \SetKwData{vXNearest}{$x_{\mathrm{nearest}}$}
    \SetKwData{vXNew}{$x_{\mathrm{new}}$}
    \SetKwData{vCXNear}{$\mathcal{X}_{\mathrm{near}}$}
    \SetKwData{vXNear}{$x_{\mathrm{near}}$}



    \SetFuncSty{textbf}
    \fExtend{\vG,\vX}
    \SetFuncSty{texttt}
    \NoBegin
    {

        $(\vV,\vE) \leftarrow \vG$;
        $\vEPrime \leftarrow \emptyset$;

        $\vXNearest \leftarrow \fNearest(\vG,\vX)$;

        $\vXNew \leftarrow \fSteer(\vXNearest, \vX)$;

        \If{$\fObstacleFree(\vXNearest, \vXNew)$}
        {
			$\fG(\vXNew) \leftarrow \infty$;

			$\fLMC(\vXNew) = \fG(\vXNearest) + \fC(\vXNearest, \vXNew)$;
			
			$\fParent(\vXNew) = \vXNearest$;
            
            $\vCXNear \leftarrow \fNear(\vG,\vXNew,|\vV|)$;

            \ForEach{$\vXNear \in \vCXNear$}
            {
                \If{$\fObstacleFree(\vXNear, \vXNew)$}
                {
                    \If{ $\fLMC(\vXNew) > \fG(\vXNear) + \fC(\vXNear, \vXNew)$}
                    {
						$\fLMC(\vXNew) = \fG(\vXNear) + \fC(\vXNear, \vXNew)$;                        

                        $\fParent(\vXNew) = \vXNear$;
                    }

                    $\vEPrime \leftarrow \vEPrime \cup \{ (\vXNear, \vXNew),(\vXNew, \vXNear) \}$;
                }
            }

            $\vV \leftarrow \vV \cup \{ \vXNew \}$;

            $\vE \leftarrow \vE \cup \vEPrime$;


            $\fUpdateState(\vXNew)$;

        }

        \Return{$\vGPrime \leftarrow (\vV,\vE)$}
    }

%
%

\caption{${\tt Extend}$ Procedure for \AlgRRTsharp{} Algorithm\label{alg:extend_rrtsharp}} 
\end{algorithm}
\DecMargin{1em} 

%% file: reduce_inconsistency.tex
\IncMargin{1em}
\begin{algorithm}[htb]
    \small
    \DontPrintSemicolon
    \SetKwInOut{Input}{input}
    \SetKwInOut{Output}{output}
    \SetKwBlock{NoBegin}{}{end}


    \SetFuncSty{ptm}    
    \SetKwFunction{fReduceInconsistency}{ReduceInconsistency}
    \SetKwFunction{fKey}{Key}
    \SetKwFunction{fLMC}{lmc}

    \SetKwFunction{fG}{g}
    \SetKwFunction{fSucc}{succ}
    \SetKwFunction{fC}{c}
    \SetKwFunction{fParent}{parent}
    \SetKwFunction{fUpdateState}{UpdateQueue}



    \SetKwData{vV}{$V$}
    \SetKwData{vE}{$E$}
    \SetKwData{vFrontier}{$q$}
    \SetKwData{vXGoal}{$x_{\mathrm{goal}}$}
    \SetKwData{vT}{$\mathcal{T}$}
    \SetKwData{vG}{$\mathcal{G}$}
    \SetKwData{vTPrime}{$\mathcal{T}^{\prime}$}
    \SetKwData{vGPrime}{$\mathcal{G}^{\prime}$}
    \SetKwData{vCXGoal}{$\mathcal{X}_{\mathrm{goal}}$}
    \SetKwData{vXminGoal}{$x^{*}_{\mathrm{goal}}$}
    
    \SetKwData{vX}{$x$}
    \SetKwData{vS}{$s$}


    \SetFuncSty{textbf}
    \fReduceInconsistency{\vG,\vCXGoal}
    \SetFuncSty{texttt}
    \NoBegin
    {

        \While{$\vFrontier.findmin() \prec \fKey(\vXminGoal)$}
        {
            $\vX = \vFrontier.findmin()$;

              $\fG(\vX) = \fLMC(\vX)$;

              $\vFrontier.delete(\vX)$;

              \ForEach{$\vS \in \fSucc(\vG,\vX)$}
              {
                \If{$\fLMC(\vS) > \fG(\vX) + \fC(\vX,\vS)$}
                {


                    $\fParent(\vS) = \vX$;

                    $\fLMC(\vS) = \fG(\vX) + \fC(\vX,\vS)$;

                    $\fUpdateState(\vS)$;
                }
              }
        }

    }

%
%
%
%
%
%

\caption{${\tt ReduceInconsistency}$ Procedure\label{alg:reduce_inconsistency}}
\end{algorithm}
\DecMargin{1em} 

%% file: auxiliary_procedures.tex
\IncMargin{1em}
\begin{algorithm}[htb]
    \small
    \DontPrintSemicolon
    \SetKwInOut{Input}{input}
    \SetKwInOut{Output}{output}
    \SetKwBlock{NoBegin}{}{end}


    \SetFuncSty{ptm}    
    \SetKwFunction{fReduceInconsistency}{ReduceInconsistency}
    \SetKwFunction{fKey}{Key}
    \SetKwFunction{fLMC}{lmc}

    \SetKwFunction{fG}{g}
    \SetKwFunction{fSucc}{succ}
    \SetKwFunction{fC}{c}
    \SetKwFunction{fParent}{parent}
    \SetKwFunction{fUpdateState}{UpdateQueue}



    \SetKwData{vV}{$V$}
    \SetKwData{vE}{$E$}
    \SetKwData{vFrontier}{$q$}
    \SetKwData{vXGoal}{$x_{\mathrm{goal}}$}
    \SetKwData{vT}{$\mathcal{T}$}
    \SetKwData{vG}{$\mathcal{G}$}
    \SetKwData{vTPrime}{$\mathcal{T}^{\prime}$}
    \SetKwData{vGPrime}{$\mathcal{G}^{\prime}$}
    \SetKwData{vCXGoal}{$\mathcal{X}_{\mathrm{goal}}$}
    \SetKwData{vX}{$x$}
    \SetKwData{vS}{$x$}


    \SetFuncSty{textbf}
    \fInitState{\vX}
    \SetFuncSty{texttt}
    \NoBegin
    {
        $\fG(\vX) \leftarrow \infty$;

        $\fLMC(\vX) \leftarrow \infty$;

        $\fParent(\vX) \leftarrow \varnothing$;
    }

    \SetFuncSty{textbf}
    \fUpdateState{\vX}
    \SetFuncSty{texttt}
    \NoBegin
    {
         \If{$\fG(\vX) \neq \fLMC(\vX)$ and $\vX \in \vFrontier$}
         {
            $\vFrontier.update(\vX, \fKey(\vX))$;
         }
         \ElseIf{$\fG(\vX) \neq \fLMC(\vX)$ and $\vX \notin \vFrontier$}
         {
            $\vFrontier.insert(\vX, \fKey(\vX))$;
         }
         \ElseIf{$\fG(\vX) = \fLMC(\vX)$ and $\vX \in \vFrontier$}
         {
            $\vFrontier.delete(\vX)$;
         }
    }

    \SetKwFunction{fH}{h}

    \SetKwData{vGPrime}{$g^{\prime}$}
    \SetKwData{vGMin}{$g_{\min}$}
    \SetKwData{vF}{$f$}
    \SetKwData{vKey}{$key$}

    \SetFuncSty{textbf}
    \fKey{\vS}
    \SetFuncSty{texttt}
    \NoBegin
    {
        $\vGMin = \min(\fG(\vS),\fLMC(\vS))$;

        $\vF = \vGMin + \fH(\vS)$;

        \Return{$\vKey = (\vF,\vGMin)$};
    }

\caption{Auxiliary Procedures\label{alg:auxiliary_procedures}}
\end{algorithm}
\DecMargin{1em} 

%% file: extend_rrtsharp_v1.tex
\IncMargin{1em}
\begin{algorithm}[htb]
    \small

    \DontPrintSemicolon
    \SetKwInOut{Input}{input}
    \SetKwInOut{Output}{output}
    \SetKwBlock{NoBegin}{}{end}


    \SetKwFunction{fExtend}{Extend}
    \SetKwFunction{fNearest}{Nearest}
    \SetKwFunction{fSteer}{Steer}
    \SetKwFunction{fObstacleFree}{ObstacleFree}
    \SetKwFunction{fKey}{Key}
    \SetKwFunction{fUpdateState}{UpdateQueue}
    \SetKwFunction{fNear}{Near}
    \SetKwFunction{fG}{g}
    \SetKwFunction{fC}{c}
    \SetKwFunction{fLMC}{lmc}
    \SetKwFunction{fParent}{parent}
    \SetKwFunction{fInitState}{Initialize}



    \SetKwData{vT}{$\mathcal{T}$}
    \SetKwData{vG}{$\mathcal{G}$}
    \SetKwData{vTPrime}{$\mathcal{T}^{\prime}$}
    \SetKwData{vGPrime}{$\mathcal{G}^{\prime}$}
    \SetKwData{vX}{$x$}
    \SetKwData{vXGoal}{$x_{\mathrm{goal}}$}
    \SetKwData{vV}{$V$}
    \SetKwData{vE}{$E$}
    \SetKwData{vEPrime}{$E^{\prime}$}
    \SetKwData{vXNearest}{$x_{\mathrm{nearest}}$}
    \SetKwData{vXNew}{$x_{\mathrm{new}}$}
    \SetKwData{vCXNear}{$\mathcal{X}_{\mathrm{near}}$}
    \SetKwData{vXNear}{$x_{\mathrm{near}}$}



    \SetFuncSty{textbf}
    \fExtend{\vG,\vX}
    \SetFuncSty{texttt}
    \NoBegin
    {

        $(\vV,\vE) \leftarrow \vG$;
        $\vEPrime \leftarrow \emptyset$;

        $\vXNearest \leftarrow \fNearest(\vG,\vX)$;

        $\vXNew \leftarrow \fSteer(\vXNearest, \vX)$;

        \If{$\fObstacleFree(\vXNearest, \vXNew)$}
        {
            \fInitState{\vXNew};

            $\vCXNear \leftarrow \fNear(\vG,\vXNew,|\vV|)$;

            \ForEach{$\vXNear \in \vCXNear$}
            {
                \If{$\fObstacleFree(\vXNear, \vXNew)$}
                {
                    \If{$\fLMC(\vXNew) > \fG(\vXNear) + \fC(\vXNear, \vXNew)$}
                    {
						$\fLMC(\vXNew) = \fG(\vXNear) + \fC(\vXNear, \vXNew)$;
						                        
                        $\fParent(\vXNew) = \vXNear$;
                    }

                    $\vEPrime \leftarrow \vEPrime \cup \{ (\vXNear, \vXNew),(\vXNew, \vXNear) \}$;
                }
            }

            
            
            \If{$\fParent(\vXNew) \not= \emptyset$ \label{line:node_inclusion_condition}} 
            {
                $\vV \leftarrow \vV \cup \{ \vXNew \}$;

                $\vE \leftarrow \vE \cup \vEPrime$;


                $\fUpdateState(\vXNew)$;
            }
        }

        \Return{$\vGPrime \leftarrow (\vV,\vE)$}
    }

%
%

\caption{${\tt Extend}$ Procedure for \AlgRRTsharpNoBlackVertex Algorithm\label{alg:extend_rrtsharp_v1}} 

\end{algorithm}
\DecMargin{1em} 